\documentclass[
11pt, 
oneside, 
english, 
singlespacing, 
headsepline, 
]{MastersDoctoralThesis} 

\usepackage[utf8]{inputenc} 
\usepackage[T1]{fontenc} 

\usepackage{microtype}

\usepackage{palatino} 
\usepackage[style=authoryear,natbib=true]{biblatex} 

\usepackage{graphicx}

\usepackage{tabularx, colortbl, xcolor}
\definecolor{mygray}{gray}{0.95}
\definecolor{mycyan}{HTML}{005397}
\definecolor{myred}{HTML}{E13333}
\definecolor{mymagenta}{HTML}{BF3E87}
\definecolor{mypurple}{HTML}{1B2278}
\definecolor{p13}{HTML}{BFB5D7}
\definecolor{b14}{HTML}{BEA1A5}
\definecolor{y15}{HTML}{F0Cf61}
\newcolumntype{a}{>{\columncolor{p13}}l}

\usepackage{easyReview}

\usepackage{amsthm}
\usepackage{amsmath}
\usepackage{amssymb}
\usepackage{amsfonts}
\usepackage{mathtools}

\usepackage{enumitem}
\theoremstyle{definition}
\newtheorem{definition}{Definition}[section]
\theoremstyle{remark}
\newtheorem{example}{Example}[section]
\theoremstyle{remark}
\newtheorem{corpus}{Corpus Example}[section]

\usepackage{braket}
\newcommand{\norm}[1]{\left\lVert#1\right\rVert}
\newcommand{\abs}[1]{\left\lvert#1\right\rvert}
\newcommand{\concat}{\mathbin{{+}\mspace{-8mu}{+}}}
\newcommand{\inner}[1]{\left\langle#1\right\rangle}

\DeclareMathOperator*{\argmax}{arg\,max}
\DeclareMathOperator*{\argmin}{arg\,min}

\usepackage{tikz-dependency}
\usepackage{tikz}
\usetikzlibrary{trees}
\usepackage{tikz-3dplot}
\usetikzlibrary{chains,scopes}
\usetikzlibrary{arrows.meta,automata,positioning}
\usetikzlibrary{shapes.geometric, arrows}
\usepackage{pgfplots}
\usepackage{pifont}

\usepackage{algorithm}
\usepackage{algpseudocode}

\usepackage{times}
\usepackage{latexsym}

\usepackage{textcomp}
\usepackage{multirow}
\usepackage{lscape}
\usepackage{subcaption}

\usepackage{listings}
\newcommand{\RomanNumeralCaps}[1]{\MakeUppercase{\romannumeral #1}}
\usepackage{array}
\newcolumntype{H}{>{\setbox0=\hbox\bgroup}c<{\egroup}@{}}
\newcommand\T{\rule{0pt}{3.4mm}}
\newcommand\B{\rule[-1.5mm]{0pt}{0pt}}

\usepackage{setspace}
\usepackage[autostyle=true]{csquotes} 

\hypersetup{%
  colorlinks = true,
  linkcolor  = black
}

\thesistitle{Similarity between Units of Natural Language: The Transition from Coarse to Fine Estimation} 
\supervisor{Dr. \textsc{Lim} Kwan Hui} 
\examiner{} 
\degree{Doctor of Philosophy} 
\author{\textsc{Mu} Wenchuan} 
\addresses{} 
\university	{SUTD}
\pillar{\ } 

\hypersetup{pdftitle=\ttitle} 
\hypersetup{pdfauthor=\authorname} 
\hypersetup{pdfkeywords=\keywordnames} 

\begin{document}

\frontmatter 

\pagestyle{plain} 


\begin{titlepage}
\begin{center}

\begin{figure}
\centering
\includegraphics[width=0.5\textwidth]{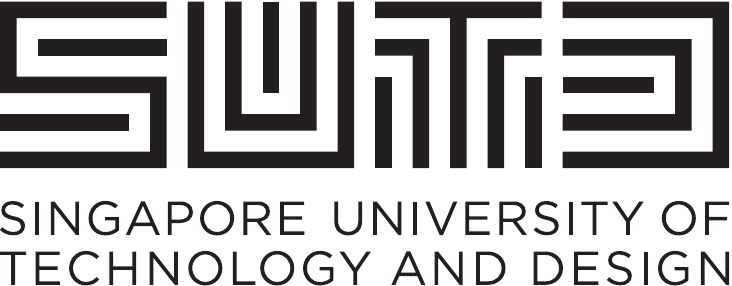}
\end{figure}

 \hfill\break\\[2.3cm]

{\huge \bfseries \ttitle}\\[3cm] 

Submitted by\\[1cm]
\authorname 

\vspace{4em}

Thesis Advisor\\[1cm]
\supname 

\vspace{4em}

\pillarname\\[1.5cm] 

\large{A thesis submitted to the Singapore University of Technology and Design in fulfillment of the requirement for the degree of \degreename}\\[1cm] 

{\large \the\year}\\[4cm] 

\vfill
\end{center}
\end{titlepage}


\begin{tec}
\addchaptertocentry{\tecname}
\begin{tabular}{ll}
	TEC Chair: & Prof. Yuen Chau \\
	Main Advisor: & Prof. Lim Kwan Hui \\
	Internal TEC member 1: & Prof. Mohan Rajesh Elara \\
	Internal TEC member 2: & Prof. Dorien Herremans\\
\end{tabular}
\end{tec}
\vfill\eject

\begin{abstract}
\addchaptertocentry{\abstractname} 

Capturing the similarities between human language units is crucial for explaining how humans associate different objects, and therefore its computation has received extensive attention, research, and applications. With the ever-increasing amount of information around us, calculating similarity becomes increasingly complex, especially in many cases, such as legal or medical affairs, measuring similarity requires extra care and precision, as small acts within a language unit can have significant real-world effects. My research goal in this thesis is to develop regression models that account for similarities between language units in a more refined way.

Computation of similarity has come a long way, but approaches to debugging the measures are often based on continually fitting human judgment values. To this end, my goal is to develop an algorithm that precisely catches loopholes in a similarity calculation. Furthermore, most methods have vague definitions of the similarities they compute and are often difficult to interpret. The proposed framework addresses both shortcomings. Itconstantly improves the model through catching different loopholes. In addition, every refinement of the model provides a reasonable explanation. The regression model introduced in this thesis is called progressively refined similarity computation, which combines attack testing with adversarial training. The similarity regression model of this thesis achieves state-of-the-art performance in handling edge cases. Chapter~\ref{Chapter2} is an introductory chapter on similarity computation in general. The main four chapters of the thesis explore the applications of general similarities, how to capture their omissions, and how to design a similarity model that can be refined over time.

The first practical work looks at applying similarity classify topics in online discussions on social networking services such as Twitter. The popularity of these services, and the large number of tweet, challenges the automatic topic detection models. To complicate matters, these topics need to be identified in the absence of prior knowledge about their type and number, and expertise is needed to tune numerous parameters. To address this challenge, I modified the cluster-based topic modelling algorithm that is originally based on word networks and n-grams co-occurrence. A more stable similarity scheme based on word embedding is used to construct networks, and the refined algorithm can better utilise community detection methods to determine topics.

Similarity so far takes a binary value, despite its regression value being calculated. Only a medium level of precision is required to determine whether similar or not. However, regression model requires higher precision. To test similarity schemes, I next construct a special task, "revising for concision", stemming from an academic writing need, which is to write concisely to convey meaning clearly. I curate a parallel dataset that describes concise revisions through 536 pairs of sentences before and after revision. These syntactically distinct but semantically identical sentences have a unique role in detecting the type of similarity that a scheme actually captures. I find that the precise semantics are still elusive, and lexical similarity is more frequently captured. Meanwhile, I use this dataset for language generation tasks to help writers struggling to revise drafts. I formulate revision as a constrained paraphrase generation task, where an algorithm is required to rewrite the sentence with only the necessary words, while preserving its meaning.

    
After stress testing these regression models using ready-made sentences, I define attacks to the regression models. I choose the scoring system of automatic summarization to be attacked. The automatic scoring of summaries guides the development of summarizers and involves aspects such as fluency, grammar, and even textual entailment. In this work, I perform evasion attacks to explore robustness of scoring systems. Attack systems predict a non-summary string from each input, and these non-summary strings achieve competitive similarity scores on representative metrics, even outperforming state-of-the-art summarization methods, indicating the low robustness of current similarity schemes. Backdoor attacks are also performed.
    
In a final work, I explore the indistinct nature of similarity.
I focus on semantic similarity, a well-known interpretation for which is natural language inference. But, there is little community consensus on its nature. Disagreements are reflected in deciding whether the inference is based on semantics or pragmatics, the strength and form of the inference, and the relevance of the inference to the real world. Differing views on the nature of inferences can further lead to differing views on whether a particular inference holds. Therefore, a system may face difficulties in evaluating reasonably under different perspectives. I propose an interpretation of the inference that takes advantage of predicate logic, natural deduction, and conditional probability to resolve seemingly inconsistencies between different viewpoints. 

I conclude this thesis by analysing the overall issues plaguing similarity research, the pros and cons of my approach, and challenges for future work. Computing similarity is a widely-applied and philosophically profound task, and I hope this thesis benefits future studies.

\end{abstract}


\begin{publications}
\addchaptertocentry{\publicationsname} 

Mu, Wenchuan, Kwan Hui Lim, Junhua Liu, Shanika Karunasekera, Lucia Falzon, and Aaron Harwood. "A clustering-based topic model using word networks and word embeddings." Journal of Big Data 9, no. 1 (2022): 1-38.

Mu, Wenchuan, and Kwan Hui Lim. "Revision for Concision: A Constrained Paraphrase Generation Task." In Proceedings of the Workshop on Text Simplification, Accessibility, and Readability (TSAR-2022), pp. 57-76. 2022.

Mu, Wenchuan, and Kwan Hui Lim. "Universal Evasion Attacks on Summarization Scoring." In Proceedings of the Fifth BlackboxNLP Workshop on Analyzing and Interpreting Neural Networks for NLP, pp. 104-118. 2022.

Mu, Wenchuan, and Kwan Hui Lim. "Explicitly stating assumptions reduces hallucinations in natural language inference." In The Second Tiny Papers Track at ICLR 2024. 2024.

Mu, Wenchuan, and Kwan Hui Lim. "Modelling Text Similarity: A Survey." In Proceedings of the International Conference on Advances in Social Networks Analysis and Mining, pp. 698-705. 2023.

\end{publications}


\begin{acknowledgements}
\addchaptertocentry{\acknowledgementname} 

First and foremost, I would like to thank my advisor Dr. Lim Kwan Hui that he gave me the chance to pursue my doctoral studies on this exciting topic. I am enormously grateful to him for all his guidance, patience, and support especially at the start of this work. I for his guidance and support throughout this work. I am also grateful to Dr. Yuen Chau, Dr. Mohan Elara, and Dr. Dorien Herremans for being on my committee and their helpful comments.

\end{acknowledgements}


\tableofcontents 

\listoffigures 

\listoftables 


\begin{abbreviations}{ll} 

\textbf{FTC} & The \textbf{F}ederal \textbf{T}rade \textbf{C}ommission\\
\textbf{SVD} & \textbf{S}ingular \textbf{v}alue \textbf{d}ecomposition)\\
\textbf{NLP} & \textbf{N}atural \textbf{L}anguage \textbf{P}rocessing\\
\textbf{NLI} & \textbf{N}atural \textbf{l}anguage \textbf{i}nference\\
\textbf{BOW} & \textbf{B}ag-\textbf{o}f-\textbf{W}ords\\
\textbf{TF-IDF} & The \textbf{T}erm \textbf{F}requency-\textbf{I}nverse \textbf{D}ocument \textbf{F}requency\\
\textbf{SVM} & \textbf{s}upport \textbf{V}ector \textbf{M}achines\\
\textbf{LCS} & \textbf{L}ongest \textbf{C}ommon \textbf{S}ubsequence\\
\textbf{LDA} & \textbf{L}atent \textbf{D}irichlet \textbf{a}llocation\\
\textbf{LSI} & \textbf{L}atent \textbf{S}emantic \textbf{I}ndexing\\

\textbf{CNN} & \textbf{C}onvolutional \textbf{n}eural \textbf{n}etworks\\
\textbf{RNN} & \textbf{R}ecurrent \textbf{n}eural \textbf{n}etworks\\
\textbf{LSTM} & \textbf{L}ong \textbf{s}hort \textbf{t}erm \textbf{m}emory\\
\textbf{BERT} & \textbf{B}idirectional \textbf{E}ncoder \textbf{R}epresentations from \textbf{T}ransformers\\
\textbf{WSD} & \textbf{W}ord \textbf{S}ense \textbf{D}isambiguation\\
\textbf{IC} & \textbf{I}nformation \textbf{C}ontent index\\
\textbf{TER} & \textbf{T}ranslation \textbf{E}dit \textbf{R}ate\\
\textbf{PER} & \textbf{P}osition-independent word \textbf{E}rror \textbf{R}ate\\
\textbf{WER} & \textbf{W}ord \textbf{E}rror \textbf{R}ate\\
\textbf{ITER} & \textbf{I}mproved version of TER\\
\textbf{EED} & \textbf{E}xtended \textbf{E}dit \textbf{D}istance\\
\textbf{HAL} & \textbf{H}yperspace \textbf{A}nalogue to \textbf{L}anguage\\
\textbf{TOT} & \textbf{T}opic \textbf{o}ver \textbf{T}ime\\
\textbf{ESA} & \textbf{E}xplicit \textbf{S}emantic \textbf{A}nalysis\\
\textbf{PPDB} & \textbf{P}ara\textbf{p}hrase \textbf{D}ata\textbf{b}ase\\
\textbf{NGD} & \textbf{N}ormalised \textbf{G}oogle \textbf{D}istance\\
\textbf{DAM} & \textbf{D}ecomposable \textbf{a}ttention \textbf{m}odel\\
\textbf{RUSE} & \textbf{R}egressor \textbf{u}sing \textbf{s}entence \textbf{e}mbeddings\\
\textbf{USE} & \textbf{U}niversal \textbf{s}entence \textbf{e}ncoder\\
\textbf{WMD} & The \textbf{W}ord \textbf{M}over \textbf{D}istance\\
\textbf{GMT} & \textbf{G}eneral \textbf{T}ext \textbf{M}atcher\\
\textbf{BEER} & \textbf{BE}tter \textbf{E}valuaion as \textbf{R}anking\\
\textbf{ROUGE} & The \textbf{R}ecall-\textbf{O}riented \textbf{U}nderstudy for \textbf{G}ist \textbf{E}valuation\\
\textbf{ESIM} & \textbf{E}nhanced \textbf{S}equential \textbf{I}nference \textbf{M}odel\\
\textbf{ClusTop} & \textbf{Clus}tering-based \textbf{Top}ic Modelling

\end{abbreviations}


%
%
%


%
%
%
%


\dedicatory{To my family}


\mainmatter 

\pagestyle{thesis} 



\chapter{Introduction} 

\label{Chapter1} 


\newcommand{\keyword}[1]{\textbf{#1}}
\newcommand{\tabhead}[1]{\textbf{#1}}
\newcommand{\code}[1]{\texttt{#1}}
\newcommand{\file}[1]{\texttt{\bfseries#1}}
\newcommand{\option}[1]{\texttt{\itshape#1}}


Representing similarities between human language units is important both for answering fundamental philosophical questions and for solving modern applications.

Since Plato 24 centuries ago, attention has been paid to the cognitive enigma, how the observation of a relatively small set of stimuli produces broadly adaptive knowledge. Philosophers and scientists have tackled this problem in a number of ways after Plato, one powerful approach came from Thomas K. Landauer, one of the most creative and innovative cognitive scientists. In his article titled "A solution to Plato's problem: The latent semantic analysis theory of acquisition, induction, and representation of knowledge", he states that the cognitive aspect of Plato's Problem can be cast as determining the similarity of objects, and learning about the appropriate similarity of words may greatly enhance knowledge and bridge the gap between the information~\citep{landauer1997solution}. It is also in this article that he presents the first computational method to represent objects and contexts based solely on general mathematical learning methods.

Computation scientists after \citeauthor{landauer1997solution} are no longer so obsessed with explaining the problem between similarity and cognition, and instead, study how to compute the similarity that the general public understands. In modern times, computing similarity has become increasingly important as a fundamental means of many natural language processing (NLP) tasks and real-world applications, including information retrieval, clustering, topic detection, question-answer (Q\&A) sessions, and financial document evaluation. Search engines need to find documents relevant to a query through modelling the correlation between documents, and Q\&A sites also need to determine whether duplicate questions have been asked before. Artificial intelligence service systems need to understand queries or commands formed by human language and find the most appropriate responses. In legal and financial matters, the similarity to existing contracts may allow an estimate of the risk level of a new contract~\citep{rawte2020comparative}.

Techniques for computing the similarity between human language units have come a long way. They can derive the similarity of word meanings from sophisticated ontology databases or rely on a large number of contexts to determine the relationship between sentences. These methods perform well for certain tasks, such as determining whether a document or block of text is similar to a particular search topic, in part because only a rough overall estimate of similarity is required here, i.e. a choice between similar and dissimilar.

However, simply judging similarity is not enough to solve more complex and specific problems in the real world. What we need is some way of telling us where and why the objects being compared are similar. Take the following two sentences for example:

\begin{enumerate}[label=(\alph*)]
\item  Activia and DanActive yoghurt\footnote{yogurt} are clinically proven to help regulate the digestive system.
\item Clinical studies show that Activia and DanActive help regulate the digestive system.
\end{enumerate}

At first glance, even judged by humans, the two statements seem to be similar. But in fact, these two sentences are very different in a certain sense. In 2010, FTC charged the Dannon Company, Inc. of deceptive advertising~\citep{ftc2010dannon}. Dannon claimed that Activia and DanActive yoghurt were "clinically proven" to help regulate the digestive system. Dannon was then prohibited from making such misleading claims, and agreed to pay \$21 million to resolve FTC investigations. At the same time, Dannon agreed to pay \$35 million to its customers. However, Dannon denied any wrong-doing and believed that softening "clinically proven" to "clinical studies show" might avoid breaches of the lawsuit settlement~\citep{ctvnews2012danone}.

There are sometimes large gaps between seemingly similar expressions. In this example, Dannon stands by its product, we may thus assume the two expressions are somehow "similar" to each other. However, how does the difference between these two expressions create the multi-million dollar gap? Dannon's revision of the wording is clearly a careful measure, and how is this measure reflected?

In this thesis, I study whether this similarity between units of natural language can be explained and quantified, and if so, how we would quantify it in the context of computational linguistics. In my research, I examine how existing lexical/semantic similarity schemes are computed, applicable, and in some cases misleading. I also propose a more refined similarity quantification scheme.

\section{Overview}

Public misinformation, from advertising to politics to science, is often about subtle differences between similar ways of covering it~\citep{west2021misinformation,levi-etal-2019-identifying,fact2019report}. I believe that computer programs that identify, quantify, and ultimately explain these differences can help people in their daily lives. My research goal in this thesis is to develop better quantification schemes for the similarity between units of natural language, especially English. I use "similarity" and "dissimilarity" as general terms to describe any estimate of the relationship between two English units, such as words, phrases, sentences, paragraphs, etc.

Answering how "similar" two sentences are is tricky. This is not just because the term is often used as an umbrella term to cover different phenomena~\citep{bar-etal-2011-reflective}. The scariest part is that the word "similar" itself has no objective definition. It is like someone pointing at a cheetah in the African savannah and asking another how agile she thinks it is. By contrast, if she was asked about speed or acceleration, the answer would be much clearer. There are attempts to formalise similarity through models such as geometry or set theory~\citep{bar-etal-2011-reflective,tversky1977features,widdows2004geometry,gardenfors2004conceptual}, but the overall agreement on the formalisation remains vague.

Despite lack of formalisation, researchers never stop to develop similarity estimation schemes. \citet{budanitsky-hirst-2006-evaluating} covers five till \citeyear{budanitsky-hirst-2006-evaluating}, \citet{bar2015composing} covers 31 calculating till \citeyear{bar2015composing}, and \citet{chandrasekaran2021evolution} report another 16 till \citeyear{chandrasekaran2021evolution}.

These similarity schemes are useful when the numerical values themselves are not important. They are very efficient at judging whether there is a similarity relationship between two units. The first practical work in this thesis also uses these similarity computations to model topics, and achieves topic coherence that surpasses traditional modelling methods.

Similarity, however, comes in two forms: overall similarity, which expresses the degree of resemblance between two objects all things considered, and respective similarity, which relates to one respect or feature. In this way, the first of two common shortcomings of similarity calculation methods is revealed: they do not specify the aspects they measure, so their scope of application is sometimes questionable.

Another downside is that the numbers are not enough to show how similar two objects are. The next three practical works in this thesis investigate this question. The first experimental step shows that these schemes do not give reliable scales when used to quantify the similarity between human language units. To overcome these shortcomings, we then developed our similarity measurement scheme with built-in grammar detection.

\section{Contributions and Outline of This Thesis}
The majority of human language similarity computing work has focused on how to obtain better correlations with human judgments. In contrast, the stress test and evasion attack method presented in this thesis can better discover details that previous methods ignore. This makes the similarity calculation more robust. It provides a framework to continuously refine similarity measures.

Determining similarity is known to be regression, and the similarity between two objects is generally a relative value of -1 to 1. I stress-tested these regression models and found that they have critical omissions. Since this corner-case testing showed great promise in uncovering the shortcomings of regression models, I decided to further investigate regression models with more sophisticated local and global attacks. More refined regression models are then developed according to the discovered deficiencies. The similarity regression model in this thesis achieves state-of-the-art performance in handling edge case similarity.

Chapter 2 surveys various language unit similarity regression models, categorizing them into set-theoretic, sequence-theoretic, real-vector-based, and fully-connected graph-based methods according to their mathematical formalisation of similarity. This survey focuses on the intrinsic connections between the independently proposed approaches, outlines the strengths and weaknesses of each approach, and provides insights into possible innovations for new researchers to address the problem of human language similarity. The final conclusion chapter distils the findings and discusses weaknesses, strengths, and potential future directions.

\subsection*{Summary Chapter 3: Topic modelling using similarity}

My initial exposure to natural language similarity is to study its application, sometimes called extrinsic evaluation~\citep{bar2015composing}, which is to measure what we can do when we get the similarity between two objects. Based on the work by \citet{lim2017clustop}, I decide to apply it to a topic modelling task. In the topic clustering of tweets, I use an existing word-embedding similarity model to obtain the similarity between two words and add them successively to the graph in the form of edge weights, where each node in the graph is a tweet. The next step is to use the Louvain method~\citep{blondel2008fast} to divide a graph into several subgraphs to achieve the effect of topic clustering. This not only surpasses the Latent Dirichlet Allocation~\citep{blei2003latent} baseline method, but also leverages a more stable similarity measure to achieve better topic coherence than the 2017 version of the ClusTop method, in terms of topic coherence, pointwise mutual information, precision, recall and F-score.

\subsection*{Summary Chapter 4: Distinct syntax but same semantics}

Starting from this chapter, I study the downsides of existing similarity schemes. I formulate a new task based on real academic writing needs, "revising for concision", a constrained paraphrase task. Like any other paraphrase task, it can be bifurcated into paraphrase identification and paraphrase generation. To the identification end, it is well suited for the goal of determining the dimensions that a similarity scheme really cares about, i.e. if the similarity scheme predicts a high score because the two units are semantically equivalent, or a low score because the two units are syntactically different. To the generation end, it clearly states the syntactical changes required from one unit to another, alleviating the ill-posedness of common paraphrase generation tasks~\citep{cao-etal-2020-unsupervised-dual,rus-etal-2014-paraphrase}.

\subsection*{Summary Chapter 5: Loopholes in the similarity schemes}

I use evasion attacks to detect vulnerabilities in similar schemes. Evasion attacks are a superset of adversarial attacks that aim to alter the input to make the system misjudge. When these schemes are used to evaluate other systems, such as automatic summarizers, vulnerabilities in similar schemes can be very harmful, as this can directly affect the evolution of generative algorithms. We select three of the most popular similarity schemes used in the evaluation of system-generated summaries, and show that poorer system outputs can receive more favourable scores from these similarity schemes than state-of-the-art summarizers. Furthermore, my research suggests that recent deep learning similarity schemes may not only be vulnerable to evasion attacks, but also leave backdoors open. Based on the vulnerabilities exposed during penetration testing, I developed a series of fine-tuned similarities to better handle these evasive examples. The attack-and-defence framework proposed in this chapter can be used to gradually refine the similarity schemes.

\subsection*{Summary Chapter 6: Interpret similarity using logic}

Finally, I am still trying to figure out the answer to this question: what makes two language units similar? In this chapter, I focus on one specific, and perhaps the most concerning dimension of similarity, semantic similarity. Different viewpoints have long existed to explain semantic similarity, a well-known explanation is through entailment~\citep{dagan2005pascal}. From the perspectives of entailment, semantic parsing, first-order logic, and conditional probability, I develop a systematic approach for human annotators to determine a meaning similarity between expressions, rather than relying on subjective guesswork.

\section{Similarity Computation from a Design Science and Application Viewpoint}

This thesis mainly studies language unit similarity calculation from the perspective of design science~\citep{simon2019sciences}. I focus on selecting possible and useful processes for improving and applying similar schemes, rather than currently existing options. From this point of view, the main purpose of this thesis is to gain knowledge and understanding of language unit similarity computation by constructing and applying artefacts~\citep{hevner2004design}, such as computer programs and datasets. These artefacts are mainly reflected in Chapter~\ref{Chapter3},\ref{Chapter4}, and \ref{Chapter5}. It is  promising that the deployment of these artefacts can help people in real-world work, such as checking the similarity of content in news articles, or rewriting texts as needed.

In contrast, the "explanatory science" aspect of similarity estimation between language units focuses on developing knowledge to \textit{describe, explain, and predict}~\citep{van2005management} the \textit{actually} similarity shared among language units. Many similar schemes investigated in Chapter~\ref{Chapter2} and those proposed in Chapter~\ref{Chapter6} do fall into this category.

\chapter{Background} 

\label{Chapter2} 

\begin{figure}[!htb]
\resizebox{\linewidth}{!}{
\tikzstyle{every node}=[draw=black,thick,anchor=west, minimum height=2.5em]

\begin{tikzpicture}[
criteria/.style={text centered, text width=2cm, fill=gray!50},
attribute/.style={%
    grow=down, xshift=0cm,
    text centered, text width=2cm,
    edge from parent path={(\tikzparentnode.225) |- (\tikzchildnode.west)}},
first/.style    ={level distance=8ex},
second/.style   ={level distance=16ex},
third/.style    ={level distance=24ex},
fourth/.style   ={level distance=32ex},
fifth/.style    ={level distance=40ex},
level 1/.style={sibling distance=10em}]
    \node[anchor=south]{Presupposition and terminology $\to$ Existing similarity schemes $\to$ Evaluation and criticism}
    [edge from parent fork down]

    child{node (crit1) [criteria] {Set / bag}
        child[attribute,first]  {node {$n$-gram~[\ref{sec:sim_in_set}]}}
        child[attribute,second] {node {Lesk~[\ref{sec:sim_in_set}]}}
        child[attribute,third]  {node {Wiki-feature~[\ref{sec:sim_in_set}]}}}
    child{node [criteria] {Sequence}
        child[attribute,first]  {node {ROUGE-L~[\ref{background_rouge}]}}
        child[attribute,second] {node {WER~[\ref{sec:symbolic_seq}]}}
        child[attribute,third]  {node {TER~[\ref{sec:symbolic_seq}]}}
        child[attribute,fourth]  {node {CDER~[\ref{sec:symbolic_seq}]}}
        child[attribute,fifth] {node {EED~[\ref{sec:symbolic_seq}]}}}
    child{node [criteria] {Real vector}
        child[attribute,first]  {node {TFIDF~[\ref{sec:sim_in_vec}]}}
        child[attribute,second] {node {GloVe~[\ref{sec:sim_in_vec}]}}     
        child[attribute,third]  {node {Word2Vec [\ref{sec:sim_in_vec}]}}
        child[attribute,fourth]  {node {LSI~[\ref{sec:sim_in_vec}]}}
        child[attribute,fifth]  {node {USE~[\ref{sec:sim_in_vec}]}}
        }
    %
    %
    child{node [criteria] {No representation}
        child[attribute,first]  {node {BLEU~[\ref{sec:complete_graph}]}}
        child[attribute,second] {node {IC~[\ref{sec:complete_graph}]}}
        child[attribute,third]  {node {BERTScore [\ref{sec:complete_graph}]}}
        child[attribute,fourth] {node {SIMILE~[\ref{sec:complete_graph}]}}
        child[attribute,fifth] {node {BLEURT [\ref{bleurt}]}}
        };
\end{tikzpicture}}
\caption{\label{fig:survey}
Survey architecture, containing the most frequently adopted schemes in each formalisation category. Besides, 55\% of the coverage scheme converts linguistic units to some form of representation, including sets (8\%), sequences (15\%), or real vectors (32\%). The remaining 45\% are just multivariate functions that return real numbers. The atomic tokens recognised by most coverage schemes are either words, including sub-words, or characters, with 90\% of the schemes being word-based. 70\% of the schemes came out after 2010, and the number of similarity schemes appears to be growing rapidly.
}
\end{figure}
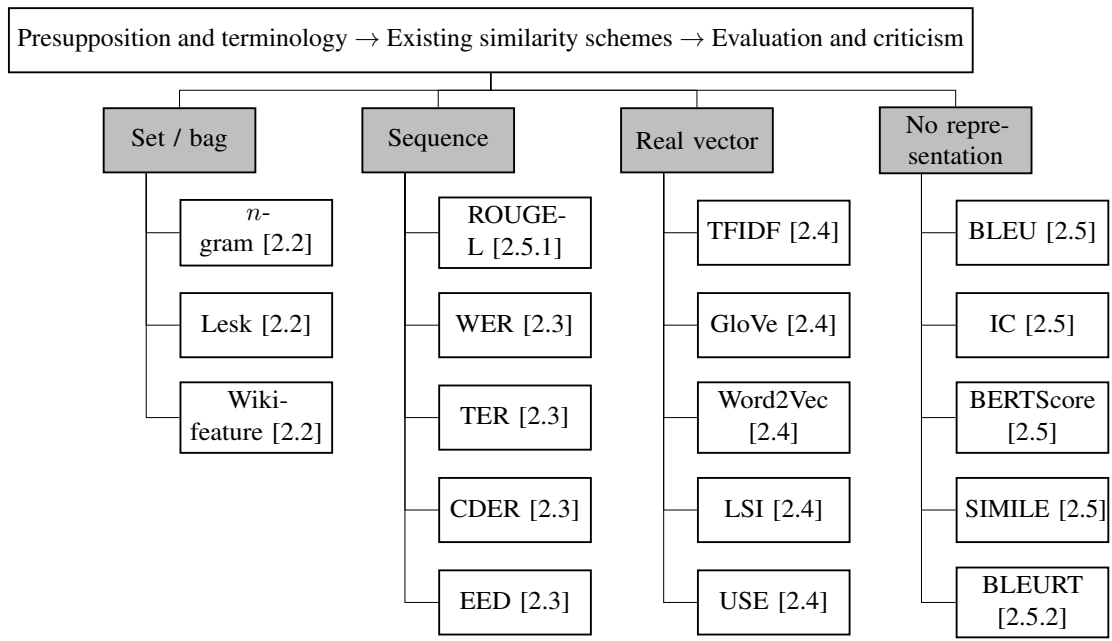

This chapter mainly covers the formalisation of various existing similarity schemes. Essentially, if we consider any unit of natural language as variable $u \in \mathcal{U}$, every similarity scheme can be shown as a multivariate function as below,

\begin{equation}
    s: \mathcal{U}, \mathcal{U}\ldots \mathcal{U} \to \mathbb{R},
\end{equation}

where the number of variables depends on the number of units under comparison. In most cases, a similarity scheme can be deemed as a bivariate function whose output range is a real number greater than -1 and less than 1. This chapter starts from grounding my entire study on clear and confine presuppositions and terminology, then I trace how similarity schemes evolve in each formalisation shown in Figure~\ref{fig:survey}, and at the end of this chapter I discuss how humans evaluate and criticise these similarity schemes.

\section{Presupposition and Terminology}
My dissertation begins by discussing the "similarity" between natural language units. But what exactly do we mean when it comes to "similarity"? How is "similarity" different from features like "relatedness", "proximity", "(reciprocal) distance", "resemblance", etc.? There may be some differences in the usage of these terms in some literature, while in others the terms are interchangeable~\citep{papineni-etal-2002-bleu,budanitsky-hirst-2006-evaluating,ferret-2010-testing,bar-etal-2011-reflective,bar-etal-2012-text,mallinson-etal-2017-paraphrasing}. While opinions on these characteristics are divided, it is certain that all of these terms lack a well-founded basis. Measuring these properties is like how people measure "agility", "dexterity", "fastness" of an object before they learn to calculate velocity and acceleration. Because of this vagueness, we refer to all these properties as "similarity" and vice versa as "dissimilarity". "Similar" is indeed the most widely used term in literature.

However, defining similarities between language units can be difficult, even if we try not to be literal. This is because similarity can be an irreducible psychological primitive~\citep{blough2001perception}, and might be too abstract for us to grasp. Here we make two presuppositions. First, there is such a property between two or more language units, possibly named for similarity, that we can attempt to understand and quantify. Second, every study of similarity seeks to understand one same property, which is the property in the first presupposition.

With these two presuppositions, we can get a clearer picture of what psychologists and linguists think about similarity, who try to formalise similarity using set theory, real vector spaces, etc. Unfortunately, a unified mathematical formalisation has not yet been achieved, and it is currently difficult to directly convert from one form to another, which makes the nature of "similarity" even more obscure.

As given by \citet{mw_similarity}, "similarity" is "the quality or state of being similar", which is again defined by "having characteristics in common"~\citep{mw_similar}. It seems that set theory captures the similarity between two units better. However, real vector space models have high popularity in recent studies. I will discuss the different formalisation of similarity between natural language units in the following three sections.

Language units are "one of the natural units into which linguistic messages can be analysed"~\citep{miller1995wordnet,miller1998wordnet,fellbaum2010wordnet}. Language units are also called linguistic units. This thesis only studies natural language units, to distinguish them from units in other languages such as programming languages. For simplicity, we occasionally omit the modifier "natural" and just use the term "linguistic unit" or "linguistic unit". In addition, we may use more informal terms such as "text," whose definition is primarily related to writing or authority~\citep{mw_text}, but is often extended in literature~\citep{bar-etal-2011-reflective,bar2015composing,shahmirzadi2019text}.

Subtypes of language units include syllables, morphemes, words, collocations, sentences~\citep{allerton1969sentence}, discourses (including paragraphs), etc.


\section{Similarity in Set Theory}
\label{sec:sim_in_set}
A set can been seen as a collection of objects, whose orders do not matter. Roster notation defines a set by listing its elements between curly brackets, separated by commas, as seen below.
\begin{equation}
\begin{aligned}
A &= \set{4, 3, 9, 33},\\
B &= \set{\text{blue}, \text{orange}, \text{cyan}}.
\end{aligned}
\end{equation}

Fundamental set operations can be applied to either construct new sets from given sets, calculate the size of a set, etc. A few operations are listed below.
\begin{itemize}
	\item Two sets $A$ and $B$ are equal if and only if they have the same elements. Formally, $A = B \equiv \forall x. x\in A \iff x \in B$.
	\item Set $A$ is a subset of $B$ if and only if every element of $A$ is an element of $B$. Formally, $(A \subseteq	B) \equiv \forall x. x \in A \to x \in B$.
	\item Union ($\cup$): $A\cup B = \set{x | x \in A \lor x\in B}$,
	\item intersection ($\cap$): $A\cap B = \set{x | x \in A \land x\in B}$, and 
	\item complement ($\complement$): $\complement A = \set{x | x \in U \land x\notin A}$,
\end{itemize}

where $U$ is a given universal set.

\subsection{Similarity between Sets}
\label{subsec:sim_btw_set}
In set theory, the two or more units under comparison are sets ($A$ and $B$, etc.). Usually, union ($A\cup B$), intersection ($A\cap B$), complement ($\complement A$), cardinality ($\abs{A}$), etc. are involved when estimating similarity.

For example, Jaccard index, developed by Paul Jaccard in 1901, takes the ratio of intersection over union and formally:

\begin{equation}
J(A, B) = \frac{\abs{A\cap B}}{\abs{A\cup B}}
\end{equation}

In bag algebra, extended from set algebra~\citep{bertossi2018datalog}, Jaccard index also applies to bags~\citep{rajaraman2011mining}, but the symbols mean bag intersection and bag sum.

\begin{equation}
J_\text{bag}(A, B) = \frac{\abs{A\cap B}}{\abs{A} + \abs{B}}
\end{equation}

Complementary to the Jaccard index measures dissimilarity, also called the Jaccard distance or Tanimoto distance, which is obtained by subtracting the Jaccard index from 1.

The S{\o}rensenDice index of two sets $A, B$ is defined as:

\begin{equation}
S(A, B) = \frac{2\abs{A\cap B}}{\abs{A} + \abs{B}}
\end{equation}


From the S{\o}rensenDice index $S$, one can calculate the respective Jaccard index $J$ and vice versa, using the following equations
\begin{equation}
\begin{aligned}
    J&=\frac{S}{2-S},\\
    S&=\frac{2J}{1 + J}.
\end{aligned}
\end{equation}

Otsuka-Ochiai coefficient~\citep{otsuka1936faunal}, also known as the Ochiai-Barkman coefficient~\citep{barkman1969phytosociology}, can be represented as:

\begin{equation}
K(A, B) = \frac{\abs{A\cap B}}{\sqrt{\abs{A}\abs{B}}}
\end{equation}

Moreover, if sets are represented as bit vectors, the Otsuka-Ochiai coefficient is equivalent to cosine.

Generally, similarity based on the set theory counts intersection portion against other parts and quantitative predictions are obtained. Larger intersection portion corresponds to higher similarity, and unique elements to a certain set decrease similarity.
\subsection{Applications}

In practice, the rationale for choosing a particular index is usually empirical rather than theoretical. For example, in psychology, which formalises similarity better than linguistics~\citep{bar-etal-2011-reflective}, earlier models of similarity measure common elements between sets of stimuli~\citep{estes1955statistical}, but \citet{tversky1977features}, when assuming that objects are represented by a set of features or attributes, gives an asymmetric similarity measure, the Tversky index ($S_\text{T}$). 

\begin{equation}
S_\text{T}(A, B) = \frac{\abs{A\cap B}}{\abs{A\cap B} + \alpha\abs{A\setminus B} + \beta\abs{B\setminus A}},
\end{equation}

where $A\setminus B$ denotes the relative complement of $B$ in $A$. Further, $\alpha, \beta \ge 0$  are parameters of the Tversky index. When $\alpha =\beta =1$, Tversky index equals Jaccard index, and when $\alpha =\beta =0.5$, Tversky index equals the S{\o}rensenDice index.

It can also be observed in linguistics that set theory is applied in various ways in measuring similarity. One of these categories can be called "feature-based methods", which transforms a given text into a set of various features, which can then be used with basic set theory. One of the most straightforward features is an $n$-gram, a contiguous sequence of $n$ items from a given text, such as phonemes, syllables, letters, words, etc.

In 2000s, simple coefficients like the Jaccard index are used to compare two sets of word $n$-grams~\citep{lyon-etal-2001-detecting,lyon2004theoretical} or two sets of character $n$-gram profiles~\citep{kevselj2003n}. ROUGE-N, the mostly used variant of ROUGE~\citep{lin-2004-rouge} whose details will be elaborated in Section , calculates the Jaccard index between two bags of word $n$-grams derived from texts, where a bag of word $n$-grams from text $\mathbf{s}$ is given as:

\begin{equation}\label{eq:bag}
    \set{x \mid x \text{ is an } n\text{-gram in } \mathbf{s}}_{\text{bag}}.
\end{equation}

In addition to getting $n$-grams directly, features can also be functions of word attributes, such as gloss or neighbour concepts~\citep{chandrasekaran2021evolution,sanchez2012ontology}. \citet{lesk1986automatic} introduces gloss overlaps for word sense disambiguation (WSD). A gloss is converted into a weighted bag-of-words, and the cardinality of intersection measures the semantic similarity. Lesk algorithm was later extended by \citet{banerjee2003extended}, where not only gloss but the WordNet synset of a word is converted into a bag. Similarly, \citet{jiang2015feature} measure semantic similarity using Wikipedia glosses. These features usually depend on the ontology, which may not contain many semantic features other than taxonomic relationships~\citep{sanchez2012ontology}. A rich feature set covering various textual features can be very promising, as observed by \citet{bar2015composing}.

In another category of exploits of set theory, contrary to the categories above, set theory is not used to compute the similarity in the final step. Conversely, similarity can be a parameterised score based on the union, intersection, or complement of sets. For example, \citet{amer2020set} formulate lexical similarity between two documents by treat each document as a set whose elements are the lexicon in the document, but do not choose the above coefficient calculation. While the authors prefer this approach to be based on set theory, we think this approach should be classified as one in Section~\ref{sec:complete_graph}. Despite flexibility, these parameters may lead to less interpretability of these formalisations.

\section{Similarity of Symbolic Sequences}
\label{sec:symbolic_seq}
A sequence, also called a string, is an enumerated collection of symbolised objects where order matters. Like a bag, it contains members (also called elements, or terms) and repetitions are allowed. The number of elements, possibly infinite, is called the length of the sequence. Usually, a sequence is defined as follows:
\begin{equation}
    s: \mathbb{N}_{+,\leq L}\to \mathcal{X},
\end{equation}

where $\mathcal{X}$ is the set of all possible members of the sequence, and $L$ is the length of the sequence. This function maps from positive natural numbers, the positions of elements in the sequence, to the elements at each position.

\subsection{Similarity between Sequences}

\paragraph{Similarity}

There a multiple approaches to calculating the similarity between sequences \citep{kozarzewski2011similarity,marteau2018sequence}. On intuitive way is to get the longest common subsequence (LCS, \citealp{allison1986bit}). The LCS problem is the problem of finding the longest subsequence common to all sequences in a set of sequences (often just two sequences). Subsequences may occupy inconsecutive positions within the original sequences. Formally, for two sequences $a$ and $b$, their LCS is recursively defined as follows:

\begin{equation}
    LCS(a, b) =  
    \begin{cases}
      LCS(a_{<-1}, b_{<-1})\concat [a(\abs{a})], & \abs{a} \cdot \abs{b} \ne 0, a(\abs{a}) = b(\abs{b})\\
      \max(LCS(a, b_{<-1}), LCS(a_{<-1}, b)), &\abs{a} \cdot \abs{b} \ne 0, a(\abs{a})\ne b(\abs{b}) \\
     \emptyset, &\abs{a} \cdot \abs{b} = 0,
    \end{cases}    
\end{equation}

where $a_{<-1}$ denotes a subsequence of $a$, obtained from deleting the last member of sequence $a$. $s_1\concat s_2$ denotes concatenation operation~\citep{wadler1989theorems}, and here it means appending/extending members of sequence $s_2$ to the end of sequence $s_1$, sequentially. Also, $\emptyset$ here denotes an empty sequence whose length is zero.

Not to be confused with LCS is the longest common substring problem ~\citep{wise1996yap3,gusfield1997algorithms}. A substring is a contiguous sequence of objects within a string.

\paragraph{Dissimilarity}

Another well-known fundamental operation on sequences is the Levenshtein distance~\citep{levenshtein1966binary}, also known as edit distance. It operates between two input sequences, returning a number equal to the minimum number of atomic operations required to transform the given input into another input. There are three basic atomic operations: substitution, insertion, and deletion.

The Levenshtein distance between two sequences $a, b$ (of length $\abs{a}$ and $\abs{b}$, respectively) is given as follows.

\begin{equation}
d(a, b) =
\begin{cases}
\abs{a}, & \abs{b} = 0,\\
\abs{b}, & \abs{a} = 0,\\
d(a_{<-1}, b_{<-1}), & a(\abs{a}) = b(\abs{b})\\
1 + \min \bigg(d(a_{<-1}, b), d(a, b_{<-1}, d(a_{<-1}, b_{<-1}))\bigg), & \text{otherwise},
\end{cases}
\end{equation}
where $s(i)$ means getting the $i$-th member in sequence $s$.

\subsection{Direct and Indirect Applications}

This subsection was supposed to be discussing how language units are represented in symbolic sequences, and how LCS or Levenshtein distance can be applied. However, current sequence representations of language units are actually simpler than set or real vector representations, since language units themselves are often sequences. Converting a language unit into a sequence often resorts to three approaches, namely a sequence of characters, a sequence of words (separated by spaces), or a sequence of sub-words~\citep{bojanowski-etal-2017-enriching}. To my best knowledge, no other approaches have been taken.

Existing sequence-based similarity scheme are either a direct or an indirect application of LCS or Levenshtein distance.

\citet{jaro1989advances} use LCS to match person or place names, as in different documents, full names of persons or places might occasionally not be used. ROUGE-L~\citep{lin-2004-rouge} is also a direct use of LCS. Greedy string tiling~\citep{wise1996yap3} is a direct use of the longest common substring problem. Jaro-Winkler distance~\citep{winkler1990string}, in contrast, cares more about some particular substrings, like prefixes.

Levenshtein distance have expanded to include phonetic, token, grammatical and character-based methods of statistical comparisons~\citep{han2016machine}.
\citet{su-etal-1992-new} introduce the metric named word error rate (WER), which is a direct use of Levenshtein distance on evaluating machine translated texts. The Levenshtein distance always needs to delete and re-insert a "misplaced" word. In order to give higher similarity to those texts with misplaced but still well-placed words, the position-independent word error rate (PER) \cite{tillmann1997accelerated} is designed to ignore word order when matching two texts. PER is adapted from Levenshtein distance in a way that two units are equivalent as long as they share the same bag-of-words. Another adaptation to the Levenshtein distance is to introduce a novel editing step that allows moving a substring~\cite{snover-etal-2006-study}, whose representatives are the translation edit rate (TER), and CDER~\citep{leusch-etal-2006-cder}. 
An independent evolution of WER is proposed by \citet{niessen-etal-2000-evaluation} to compare one text with multiple possible references. In an improved version of TER (ITER, \citealp{panja-naskar-2018-iter}), words with the same stemming forms are considered equivalent.

Levenshtein distance and its adapted variants have been applied on character sequences as well, where the variants are called characTER~\citep{wang-etal-2016-character} and Extended Edit Distance (EED, \citealp{stanchev-etal-2019-eed}, correponding to TER and CDER for word sequences~\citep{sai2022survey}. Nevertheless, EED is not a clean adaptation from Levenshtein distance and should be part of section ~\ref{sec:complete_graph} schemes. EED normalises the distance with the sentence length of \textit{each} unit individually, and thus impairs the symmetry of this similarity scheme.

\section{Similarity in Real Vector Spaces}
\label{sec:sim_in_vec}

This type of method has different names from different perspectives, but in fact the core is the same. If the similarity calculation based on set theory is to model a unit as a set, then here is to model a unit as a vector. This vector may sometimes called feature, embedding, high-dimension representation, etc. Similarity calculation between vectors can take advantage of numerous operations in real vector spaces.

In the following two sub-sections, we first discuss how similarity between vectors can be calculated, and then discuss how to represent a unit of natural language using a vector.

\subsection{Similarity Formulae between Vectors}

Converting something into a vector is not a secret skill in natural language processing (NLP), therefore computing the similarity between vectors is a more general problem with widespread applications in other fields such as computer vision. \citep{bar2015composing,chandrasekaran2021evolution,sai2022survey} looked at a number of formulae, some of which are standard operations in mathematics and others that only make sense when used to estimate "similarity". In this subsection, we select the most relevant content for presentation.

\paragraph{Similarity}
Given two vectors $\mathbf{a}, \mathbf{b}$, one of the most straightforward combinations is to compute their inner product $\inner{\mathbf{a}, \mathbf{b}}$. Then we can ignore the modulo of the vectors and just compute the angle cosine as follows,

\begin{equation}
\frac{\inner{\mathbf{a}, \mathbf{b}}}{\abs{\mathbf{a}}\abs{\mathbf{b}}}.
\end{equation}

Besides normalising the inner product with product of modulo, Jaccard, S{\o}rensen-Dice, and Overlap formulation for real vectors have also been widely used. There respective formulae are 
\begin{equation}
\begin{aligned}
    J(\mathbf{a}, \mathbf{b}) &= \frac{\inner{\mathbf{a}, \mathbf{b}}}{\abs{\mathbf{a}}^2 + \abs{\mathbf{b}}^2 - \inner{ \mathbf{a}, \mathbf{b} }}, \\
    S(\mathbf{a}, \mathbf{b}) &= \frac{2\inner{ \mathbf{a}, \mathbf{b} }}{\abs{\mathbf{a}}^2 + \abs{\mathbf{b}}^2}, \\
    O(\mathbf{a}, \mathbf{b}) &= \frac{\inner{ \mathbf{a}, \mathbf{b} }}{\min(\abs{\mathbf{a}}^2, \abs{\mathbf{b}}^2)}.
\end{aligned}
\end{equation}

Besides inner products, cross-covariance between vectors, from a statistics perspective, can act as "similarity". Usually, three types of bivariate analysis, also called correlations, are calculated: Pearson correlation ($r$), Kendall rank correlation ($\tau$), and Spearman correlation ($\rho$). Pearson correlation is equal to the cosine  of the angle between two vectors~\citep{rummel1976understanding}. Kendall correlation and Spearman correlation are nonparametric tests that measure the strength of dependence and the degree of association between two vectors, respectively~\citep{colwell1982spearman}.

\paragraph{Dissimilarity}

The flip side of measuring similarity are the formulae for measuring dissimilarity, where "distance" is usually calculated. A distance is a nonnegative number assigned to every pair of vectors, in accord with the following three conditions:

\begin{equation}
\begin{aligned}
	d(\mathbf{a}, \mathbf{b}) &> d(\mathbf{a}, \mathbf{a}) = 0, \forall \mathbf{a}, \mathbf{b} \in V, \mathbf{a} \neq \mathbf{b} &\text{Minimality}, \\
	d(\mathbf{a}, \mathbf{b}) &= d(\mathbf{b}, \mathbf{a}), \forall \mathbf{a}, \mathbf{b} \in V &\text{Symmetry}, \\
	d(\mathbf{a}, \mathbf{b}) &\leq d(\mathbf{a}, \mathbf{c}) + d(\mathbf{c}, \mathbf{b}), \forall \mathbf{a}, \mathbf{b}, \mathbf{c} \in V &\text{The triangle inequality}.
\end{aligned}
\end{equation}

The most popular distance function is the Euclidean distance (L2 norm, $\abs{\mathbf{a}-\mathbf{b}}$), which is differentiable and convex but not scale in-variant. Differences in measurement units can skew distances, and vector normalisation is often a remedy. Besides, if the vectors are normalised as $\inner{ \mathbf{a}, \mathbf{a} } = \inner{ \mathbf{b}, \mathbf{b} } = 1$, the Euclidean distance can be derived from the inner product as follows,

\begin{equation}
\frac{\abs{\mathbf{a}-\mathbf{b}}^2}{2} = 1 - \inner{ \mathbf{a}, \mathbf{b} }
\end{equation}

Although widely used, Euclidean distance is not very good at measuring vectors in high-dimensional spaces, which is called the concentration of norm or the curse of dimensionality~\citep{demartines1994analyse,beyer1999nearest}.




Often along with the L2 norm is the L1 norm ($\abs{\mathbf{a}-\mathbf{b}}_1$, City Block, Manhattan, or taxicab distance). Unlike the L2 norm, the L1 norm is not the shortest path.
Norms with other power $p$ are called p-norm or Minkowski distance.
As $p$ approaches infinity, the p-norm approaches the infinity norm (maximum norm, or Chebyshev distance).
The norm of each dimension can be multiplied by different weights, such as the Canberra distance for L1 norm and Chi-square distance for squared L2 norm~\citep{jurman2009canberra}. Moreover, if vectors are represented as bit vectors, the L1 norm can be seen to be the same as Hamming distance~\citep{phillips2013distances}.



Besides calculating norms, distance can also be obtained from similarity directly. Cosine can be used to calculated cosine distance $1 - \inner{ \mathbf{a}, \mathbf{b} } / (\abs{\mathbf{a}}\abs{\mathbf{b}})$. Like cosine, cosine distance computes the difference in terms of directions and not magnitude. This is often useful in comparing high-dimensional vectors~\citep{spruill2007asymptotic}. Cosine distance is sometimes called Pearson Correlation distance.

If the two vectors are regarded as two probability distributions, Kullback Leibler divergence (KL divergence, the relative entropy) measures how one probability distribution is different from a second. Kullback Leibler divergence is not symmetric, but averaging the KL divergence from the average of two vectors to each results in a symmetric divergence, called Jensen-Shannon divergence~\citep{osterreicher2003new}, whose square root is often referred to as Jensen-Shannon distance.



\subsection{Real Vector Representation of Expressions}
\label{sec:vector}
Now that we known the means of measuring "similarity" between two real vectors. The rest of the journey is to represent each unit of natural language in a real vector, namely how the following function is instantiated,
\begin{equation}
    v: \mathcal{U} \to \mathbb{R}^k,
\end{equation}

where $k$ is the dimension of the vector space, usually from tens to thousands. Different strategies of representation capture different information of a language unit. Usually, these aspects include lexicon, syntax, style, semantics, pragmatics, etc.

\citet{bar2015composing} find that the "similarity" between two texts indeed comes from different dimensions and believe a thorough definition of these dimensions is the basis for text similarity calculation.
They further argue that the three main feature dimensions associated with common tasks in NLP are content, structure, and style, and cite book of \citet{gardenfors2004conceptual} acknowledging that dimensions in conceptual space are not completely independent.

However, the definitions of these dimensions themselves are ambiguous. For example, \citeauthor{bar2015composing} believes that style refers to grammar, usage, mechanics, and lexical complexity, while \citet{hickey1993stylistics}, in his article titled "Stylistics, Pragmatics and Pragmastylistics", suggests that these (stylistic) features are not imposed by the grammar of the language. Given this ambiguity, we employ a simple dichotomy in dimension: semantic or non-semantic. Next, we introduce methods to convert semantic features or non-semantic features into vectors, respectively.

\subsubsection*{Non-semantic Representation\footnote{They are further away from measuring semantics than the other, but still may be used to do so occasionally.}}

\subparagraph{Function word frequency}
\citet{dinu2009ordinal} show that function word frequencies can well indicate authorship attribution. They compute feature vector for a text based on the frequencies of 70 function words identified by \citet{mosteller2007inference}.

\subparagraph{TF-IDF model}
Term frequency-inverse document frequency~\citep{luhn1957statistical,jones1972statistical,ramos2003using} is a widely used vectorisation method. In the space it defines, each term in the vocabulary is an orthonormal basis. The modulo length of an orthonormal basis is determined by the product of two values: how many times a word appears in a document, and the logarithm of inverse document frequency of the word across a set of documents. Some variations of TF-IDF are taking $n$-grams as terms ~\citep{shahmirzadi2019text}, or using time to increase the IDF~\citep{kelly2021measuring}, where a term has a low IDF and high discrimination when it is first introduced, and the IDF decays as the term's usage increases.

Consensus-based Image Description Evaluation (\citealp{vedantam2015cider}, CIDEr)  weighs each $n$-gram in a sentence based on its frequency in the corpus and in the reference set of the particular instance, using TF-IDF.

It was first proposed in the context of image captioning where each image is accompanied by multiple reference captions. It is based on the premise that $n$-grams that are relevant to an image would occur frequently in its set of reference captions. 

However, $n$-grams that appear frequently in the entire dataset (\textit{i.e.}, in the reference captions of different images) are less likely to be informative/relevant and hence they are assigned a lower weight using inverse-document-frequency (IDF) term.

\subparagraph{Word order}
\citet{hatzivassiloglou-etal-1999-detecting} create a feature vector for each text, where each vector element corresponds to the word order of word pairs shared by the two texts. However, although vector representations are involved, each text can be represented in a different way when computing similarity with different texts. Therefore, we think this method should be classified as one in Section ~\ref{sec:complete_graph}.

\subsubsection*{Semantic Representation}

\subparagraph{Bag of words}
As introduced in subsection~\ref{subsec:sim_btw_set}, bag-of-words (BOW) was originally formulated in bag algebra. BOW ignores the order of words in the text and only cares about the number of times each word occurs. A BOW can be converted to a fixed-size vector representing the probability distribution of each word appearing in this text.

\subparagraph{Latent Semantic Indexing (LSI)}
LSI~\citep{deerwester1990indexing} is also known as LSA~\citep{landauer1997solution}, which finds low-dimension vector representation of words or documents. In LSI, a document-term matrix is first built from a given set of documents and a set of vocabulary terms. The dimension of the matrix is number of documents times number of terms, and the matrix entries count occurrences of each term in each document. Then a reduced rank approximation, or truncated SVD (singular value decomposition) is found to reduce the matrix dimensions to the number of preset topics. Truncated SVD can be updated incrementally as new documents are added \citep{brand2002incremental,sarwar2002incremental}.

\subparagraph{Hyperspace Analogue to Language (HAL)}

HAL~\citep{lund1996producing} is a computationally and conceptually simpler method that does not require SVD. HAL builds a word matching matrix that has rows and columns representing the words in the dictionary, and the elements of the matrix are filled with association strength values. Co-occurrence, or "strength of association," is often determined by whether two words appear within a certain window distance, or simply by the number of other word being in between. Determined by how many there are, in general, the association between words decreases as the number of separating words increases.

\subparagraph{Latent Dirichlet allocation (LDA)} 

LDA~\citep{blei2003latent} is a popular topic model for determining the set of potential topics associated with a set of documents, where the topic of the document rather than each word in the document is represented as a vector, reducing the dimensionality of the vector\citep{sinoara2019knowledge}. Each document is usually represented as a BOW vector, which is then assigned a topic distribution through the generative process.

LDA can also be used with variants of BOW to improve the semantic coherence of topics, such as keeping only nouns, removing stop words, or restricting keywords to specific domains~\citep{martin-johnson-2015-efficient,yang2018text,nikolenko2017topic}. In the absence of word co-occurrences, LDA can be used with distributed word embeddings that capture semantic and syntactic correlations between words~\citep{gao2019incorporating}. It helps discover interpretable topics, even with large vocabularies containing rare words and stopwords~\citep{dieng-etal-2020-topic,dai2017social}. Domain-specific semantic relationships of words are useful in areas such as clinical predictive modelling~\citep{bagheri2020etm}. LDA can also be built on a conditional random field, two-layer bidirectional long short-term memory, or other neural network representations~\citep{gao2019incorporating,jansson-liu-2017-distributed,bhat2020deep}.


Although LDA is traditionally used for longer documents such as news articles and academic papers, LDA has also been applied to Twitter where each tweet is considered a document. To address the limitations caused by short texts such as tweets, researchers have used aggregation schemes where tweets by the same author or with the same terms, hashtags, posted time are combined as one document~\citep{hong2010empirical,mehrotra2013improving,steinskog-etal-2017-twitter, gao2019incorporating}. \citet{zhao2011comparing} have also used LDA to study the differences between Twitter and New York Times in terms of the discussed topics and content, while \citet{aiello2013sensing} applied LDA for the purpose of trending topics detection in sports and politics, using different textual pre-processing steps. Similarly, researchers have modified LDA to capture the temporal nature of documents, such as the Topic over Time (TOT) algorithm~\citep{wang2006topics} for detecting topical trends over continuous time, and Temporal-LDA~\citep{wang2012tm} for modelling topics and their transitions in streaming documents. LDA has also been applied in various domains, such as urban analytics~\citep{lansley2016geography,wang2018happiness}, advertising/marketting~\citep{chen2013emerging,barry2018alcohol}, diseases/medical~\citep{missier2016tracking,kwan2020understanding}, climate sentiment measurement~\citep{dahal2019topic}, communication research~\citep{maier2018applying}, and aspect-based product review~\citep{jeong2019social}.

\subparagraph{Explicit Semantic Analysis (ESA)}
ESA~\citep{gabrilovich2007computing} represents input text as weighted vectors of word concepts, called the "interpretation vectors." In ESA, each word gains its concept from the constantly updated Wikipedia which is applicable to various domains and languages. ESA uses TF-IDF to build the associated strength of a word with all words that appear in its Wikipedia page. Again, the words in the page have their own pages.

\subparagraph{Dependency-based models}

In general, these approaches takes inductive dependency parsing~\citep{nivre2006inductive} to convert the linear form language units in a corpus into trees.

Then the co-occurrence window~\citep{agirre-etal-2009-study} or bag-of-words~\citep{levy-goldberg-2014-dependency} can be applied according to syntactic context templates, considering preceding and succeeding nodes of a certain word in the dependency tree.

\subparagraph{Tree Kernels}

Machine learning algorithms such as support vector machines (SVMs) use kernels to fit the similarity between language units~\citep{shawe2004kernel}. Kernels find patterns in language units. \citet{moschitti-etal-2008-tree} propose a tree kernel for language units considering the grammar of the language. Tree kernels usually identify sentence structures based on constituencies~\citep{amir2017sentence} or dependencies.
Syntactic tree kernel~\citep{severyn-etal-2013-learning-semantic}, partial tree kernel~\citep{moschitti2006efficient,collins-duffy-2002-new} have been used encode a sentence. Multiple well-performing kernels can be further linearly combined~\citep{bar-etal-2012-ukp,saric-etal-2012-takelab}. The byproducts from parsing, such as named entities, part of speech tags, can also be formatted into vectors whose cosine can be then calculated.

\subparagraph{Word-Alignment models}

Word-Alignment models calculate the semantic similarity of sentences based on their alignment over a large corpus \citep{sultan-etal-2014-dls,kajiwara-komachi-2016-building,cer-etal-2017-semeval}.

The second, third, and fifth positions in SemEval tasks 2015 were secured by methods based on word alignment. The unsupervised method which was in the fifth place implemented the word alignment technique based on Paraphrase Database (PPDB)  \citep{ganitkevitch-etal-2013-ppdb}.

The system calculates the semantic similarity between two sentences as a proportion of the aligned context words in the sentences over the total words in both the sentences. The supervised methods which were at the second and third place used word2vec to obtain the alignment of the words.

In the first method, a sentence vector is formed by computing the "component-wise average" of the words in the sentence, and the cosine similarity between these sentence vectors is used as a measure of semantic similarity. The second supervised method takes into account only those words that have a contextual semantic similarity \citep{sultan-etal-2015-dls}.

\subparagraph{Word2Vec}
Vector representation for words gets a leap when \citet{mikolov2013efficient} propose Word2Vec (W2V), based on the assumptions that words with similar meanings tend to appear in similar contexts. A shallow neural network is trained to capture information from the context of other words surrounding a given word to predict a vector, which is believed to preserve latent linguistic relationships\citep{schnabel-etal-2015-evaluation} and semantic similarity \citep{mikolov-etal-2013-linguistic} between words.

Specifically, W2V is a three-layer shallow neural network. Given Context window size, the W2V algorithm has two forms, namely continuous bag-of-words and skip-gram. The former trains a neural model to predict the probability of a center word given each of its context word. The latter takes a reversed manner, training a model to predict context words given the center word. The model is trained with a large corpus, where probability prediction relying on softmax is computational expensive. Two efficient alternatives to softmax are hierarchical softmax and negative sampling~\citep{mikolov2013distributed}, although it may fall short in polysemy~\citep{camacho2018word}.
\subparagraph{GloVe} 

While also taking distributional semantics, GloVe globally builds a global word co-occurrence matrix on a corpus~\citep{pennington-etal-2014-glove}. It also builds context window co-occurrence matrices such that the GloVe loss function minimises the least-square distance between the context window co-occurrence values and the global co-occurrence values \citep{lastra2019reproducible}. The vector that represents the global co-occurrence matrix is thus representation the word.
\subparagraph{FastText}

Texts in corpus are tokenised in characters, and the character representation training takes skip-gram approach~\citep{bojanowski-etal-2017-enriching}. Each word can then be represented as a collection of character n-grams, and its embedding is the average of its character embeddings. FastText learns to account for the morphological structure of the word and efficient in handling out-of-the-vocabulary words with their characters or subunits.

\subparagraph{Document embedding}

The vectors at sentence or document level can be obtained from word vectors~\citep{sai2022survey,landauer1997solution}, including taking an average of the word vectors or the dimension-wise max/min (Vector Extrema, \citealp{forgues2014bootstrapping}). Simple averaging gives equal weights to important and unimportant words; taking the extreme values along each dimension may help ignore the common words and prioritise informative words; more information may be preserved by assigning TF-IDF as weights in averaging.

Many researchers have extended the word2vec model to propose a context vector \citep{melamud-etal-2016-context2vec}, a dictionary vector \citep{tissier-etal-2017-dict2vec}, a sentence vector \citep{pagliardini-etal-2018-unsupervised} a paragraph vector \citep{le2014distributed}, and a document vector (D2V, \citealp{mikolov2013distributed}).

\section{Similarity as Weighted Edges of A Complete Graph}
\label{sec:complete_graph}
While set theory and real vector spaces provide a reasonable basis for similarity formalisation, some similarity measures are not designed to represent linguistic units in some form, such as sets, bags, or vectors. Instead, these methods focus on bivariate mapping from two units to real-valued scales.

For any two linguistic units, there is one of two similarity values that connect them. If this value is symmetric, we assume that these units form the vertices of a  complete graph, while the values are the edges in between. If this value is asymmetric, we assume that these units form vertices of a complete digraph, while the values are the edges in between.

\subsection{Undirected Graph}

\begin{figure}
\centering
\resizebox{0.6\linewidth}{!} {
\begin{tikzpicture}[
    text width=5em,text centered,
    > = stealth, 
    shorten > = 1pt, 
    auto,
    node distance = 4cm, 
    semithick 
    ]

    \tikzset{every state}=[
    draw = black,
    thick,
    fill = white,
    minimum size = 1mm
    ]

    \node[state] (A) {A jaguar kills a crocodile};
    \node[state] (B) [right=of A]{A crocodile is not murdered by a jaguar};
    \node[state] (C) [below=of A] {Sonny kills Dr. Lanning};
    \node[state] (D) [below=of B] {Dr. Lanning is not murdered by Sonny};

    \path[-] (A) edge  node[] {0.947} (B);
    \path[-] (A) edge  node[left] {0.868} (C);
    \path[-] (A) edge  node[pos=0.05, right] {0.857} (D);
    \path[-] (B) edge  node[pos=0.15, below] {0.859} (C);
    \path[-] (B) edge  node[] {0.891} (D);
    \path[-] (C) edge  node[] {0.938} (D);

    \end{tikzpicture}
}
\caption{The edge between each two vertices has a weight, indicating the similarity score between the two vertices. The graph is undirected, such that this bivariate function is symmetric.}
\label{fig:complete_graph}
\end{figure}
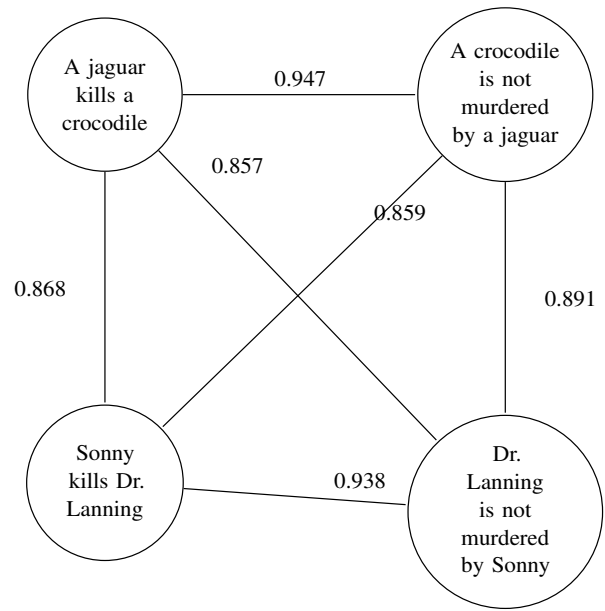


Measuring similarity between two language units has been challenging for long and various methods emerge over the years. In this (sub)section, we describe the representative one in a rough chronological order to understand how purely bivariate similarity schemes evolve over years. Early works focus more on analysing specific knowledge sources of concepts while more recent methods take advantage of distributional semantics assumption, stronger computing resources, and pretrained contextualised embeddings. 

\subsubsection{Ontology Measures}

Knowledge bases are very useful in early days in terms of delivering the human annotated meaning of terms, structure semantic relationships in between \citep{sanchez2012ontology}.

They compute the similarity between two terms based on information derived from at least one underlying knowledge source such as ontology/lexical databases, thesauri, dictionaries, taxonomies, etc. A few widely used lexical databases are WordNet~\citep{miller1995wordnet,zhu2016computing}
Wiktionary\footnote{\url{https://en.wiktionary.org}},
Wikipedia\footnote{\url{http://www.wikipedia.org}},
and BabelNet~\citep{navigli2012babelnet}.
DBPedia \citep{bizer2009dbpedia}
These methods are also known as edge counting~\citep{martinez2014overview} or knowledge-based methods~\citep{chandrasekaran2021evolution}. 

These methods are often accompanied by information content indices (IC). Words with higher IC values are more specific \citep{zhu2016computing}.

For any two terms $A, B$ in ontology, the general form of existing similarity formulae using ontology and IC can be described as

\begin{equation}
    s(A,B) = \frac{1}{1+ IC_1(A, B) \cdot k^{IC_2}(A, B)}
\end{equation}

where $k$ is a hyperparameter to be tuned for different IC calculations~\citep{hoffart2013yago2}.

One way to calculate IC is using the intrinsic structure of the ontology~\citep{sanchez2011ontology}. The assumption is that the ontology is constructed in a meaningful way. \citet{gao2015wordnet} estimate the semantic similarity between two terms by using the depth of their subsumer in WordNet synset tree. A deeper postition in the tree indicates more specific meaning, and therefore higher IC. 

Apart from intrinsic IC values, corpus IC has also been explored and believed to correlated better with similarity in gold standard datasets. Usually IDF, as described in section ~\ref{sec:complete_graph} is used to measure IC~\citep{sanchez2013semantic}. For example, semantic similarity between two terms can be directly estimated through the IC of their shared subsumer (least common subsumer, or hypernym) \citep{resnik1995using}, the ratio of the IC of their shared subsumer and the average IC of two terms~\citep{lin1998information}, or subtracting the IC of their shared subsumer from the average IC of two terms\footnote{This is a dissimilarity measure.}\citep{jiang-conrath-1997-semantic}. IC based on corpora tends to be language and domain independent \citep{altinel2018semantic}. However, these values consider less the human-annotated meaning of the words, and an ideal clean corpus is still hard to be clearly built~\citep{raffel2019exploring}

Multiple ontologies are sometimes combined to calculate the relationship among terms that are scattered in several ontologies \citep{rodriguez2003determining}. Some resources like Wikipedia can be both used as a structured taxonomy as well as a corpus to provide IC~\citep{jiang2017wikipedia}.

Underlying knowledge well handles the most common ambiguities such as synonyms, idioms, and phrases. Ontology methods are hardly applied to measure similarity/dissimilarity between longer units like sentences or documents, as ontologies are usually for words or tokens. But, they can be extended with aggregation rules to compute sentence similarity~\citep{lee2011novel,lastra2017hesml}.

\subsubsection{Normalised Google Distance (NGD)}

Frequent patterns in text fragments such as those indexed by web search engines also contain information for finding similarities

NGD measures the probability of co-occurrence of two terms on the Web to estimate the similarity between the two terms~\citealp{cilibrasi2007google}. The probability of a particular term is determined by the ratio of the number of pages associated with that term to the total number of pages, both numbers returned by a search on the Google engine. Therefore, the joint probability of two terms is derived as the ratio between the number of pages containing both terms and the total number of web pages. It is based on the assumption that if two words are more related, they appear together more frequently in web pages. For two terms $A, B$, NGD calculates as follows.

\begin{equation}
    NGD(A, B) = \frac{\max(\log  g(A),\log  g(B)-\log  g(A,B)}{\log  G - \min (\log   g(A), \log   g(B))}
\end{equation}

where function $g(\cdot)$ returns the number of returned related pages by searching Google engine, and $G$ represent the total number of web pages. NGD is sometimes believed to measure semantic relatedness rather than semantic similarity because occurring words in web pages may have opposite meaning. But, as antonyms may contain similarity as well we include NGD in this section~\citep{bollegala2007measuring,martinez2014overview}.

Apart from the ratio between numbers of pages, joint probability has been computed by pointwise mutual information (PMI), Dice, Overlap, or Jaccard index~\citep{bollegala2007measuring,martinez2014overview}.

\subsubsection{Lexico-syntactic patterns}

\citet{bollegala2007measuring} use regular expressions between two terms to estimate their semantic similarity. In the websites, two terms can co-occur with certain regular expressions. Such expressions are considered to indicate semantic similarity between the two terms. A support vector machine (SVM) is used to determine if the connection between two words are of synonymous or not. This is trained using gold synonyms.

\subsubsection{Early Composite methods}
As opposed to non-composite methods, composite methods explicitly describes how the atomic similarity among tokens or pre-defined heuristic-based features contributes to the overall similarity of longer units.

A large portion of composite methods in early days come from the context-free metrics to evaluate "how close two sentences are" to each other~\citep{papineni-etal-2002-bleu}. The units being compared are usually first represented as bags of words or $n$-grams. The overlap is then used to compute a (series of) similarity scores between units. Unlike the methods presented in Section~\ref{sec:symbolic_seq}, composite methods typically use more NLP-specific operations than basic operations in set theory when computing similarity. Notably, although these context-free metrics are believed to "check the similarity" between the units~\citep{sai2022survey}, they are sometimes overlooked by the textual similarity community~\citep{chandrasekaran2021evolution}.

\subparagraph{General Text Matcher (GTM)}

GTM tries to match two string of words. Unlike the LCS, GTM does not require a subsequence to be found in both units under comparison~\citep{turian-etal-2003-evaluation} . Instead,  it takes the concept of "maximum matching" from graph theory~\citep{cormen2022introduction}. A matching is a sub-bag between two units, but they are clustered to form longer substrings in the original sequences. The maximum match size (MMS) is the size of the sub-bag shared by the two units, equivalent to the sum of all substrings. F-measure, harmonic mean of precision and recall, between two language units. The calculation has been so far both symmetric, and can be seen as with a \textit{bag} representation. However, GTM further uses the term below to reward longer matched substrings.

\begin{equation}
    \bigg(\sum_{s \in \text{all substrings}} \abs{s}^p\bigg)^{1/p}, p > 0
\end{equation}

Specifically when $p = 1$, GTM is equivalent to Jaccard index on bags, and equivalent to ROUGE-1 described below.

\subparagraph{Recall-Oriented Understudy for Gisting Evaluation (ROUGE)}
\label{background_rouge}
It is a combination of variants used to measure relatively long language units formed with unigrams or words. It measures the number of overlapping $n$-grams (ROUGE-N), LCS between two string of words (ROUGE-L), etc~\citep{lin-2004-rouge}. Particularly, when F-measures of ROUGE-N are taken~\citep{see-etal-2017-get}, ROUGE is a symmetric metric.

ROUGE-N between two language units equals to the Jaccard index on the $n$-gram bags from the two units. ROUGE-L measures the ratio between the LCS between a pair of units over the length sum of the pair~\citep{cormen2022introduction}. Thus, ROUGE-L can be considered as a kind of F-measure as well.

There are more sub-variant of ROUGE, such as ROUGE-W and ROUGE-S. ROUGE-W adds gap-penalty weights to LCS matching to reward longer substrings. ROUGE-S uses unigram order co-occurrence matching, where S stands for skip-bigram, a pair of words in the same sentence order, with arbitrary words in between. ROUGE-S also takes a F-measure.

ROUGE variants were originally proposed for comparing the similarity between outputs and references of automatic summarization, but have been adopted for other NLG tasks. Due to the number of sub-variants, ROUGE overall does not take any representation among set, sequence, or real vector. It is essentially a lexical similarity measure~\citep{han2016machine}. When every sub-variants takes F-measure, ROUGE is symmetric, although it was initially proposed as \textit{recall oriented}, which is asymmetric.

\subparagraph{Word Order}

As already seen in ROUGE-S, the correct word order is very important. However, the diversity of languages also allows for different appearances or structures of sentences. Its essence is how penalty is \textit{only} given to wrong word order that misstructure sentences. Many word-order involved schemes exist, ATEC features the explicit assessment of word order and word choice in terms of characteristic forms at various linguistic levels, including surface form, stem, sound and sense, and further by weighing the informativeness of each word~\citep{wong2009atec} . 

Some word order schemes are often accompanied with measuring $n$-gram matching or sentence length as well, such as PORT~\citep{chen-etal-2012-port} and LEPOR \cite{han-etal-2012-lepor,han2014unsupervised}. They are usually considered as symmetric lexical similarity measures.
\subparagraph{MEANT}
MEANT considers both the structure and semantics of the sentences~\citep{lo-wu-2011-meant,lo-etal-2012-fully,lo-wu-2012-unsupervised,lo-etal-2014-xmeant}. It leverages semantic role labelling (shallow semantic parsing) to assign the agent/act/patient labels to words or phrases in a sentence. MEANT then calculated a weighted F-score computed according to the semantic roles of each token. It starts from matching maximum weighted tokens, usually predicates, based on lexical similarities. The lexical similarity can be obtained from word vectors \citep{dagan-etal-1993-contextual,dagan1999contextual}, or any other lexical similarity listed in sections above.

The 2.0 version of MEANT further weighs the amount of content in each word by IDF~\citep{lo-2017-meant}. The aggregation on $n$-gram lexical similarities also considers word order instead of only relying on BOW of phrases. Other variants, like YiSi~\citep{lo-2019-yisi} replace (non-contextualised) word embeddings like word2vec with contextualised word embeddings from BERT.

\subparagraph{Greedy Embedding Matching}
When the language units are taken as sequences of vectors, the similarity between token vectors can accumulate to compute the similarity of two units. Greedy matching finds the closest vectors in two units based on their cosine similarity~\citep{rus-lintean-2012-comparison}. Then the token similarity can be average to get an overall similarity from one to the other. So far greedy matching is not symmetric, therefore the process is usually repeated in the reverse to atain symmetry. Formally, for language units $A$ and $B$

\begin{equation}
    s(A, B) = \frac{1}{2} \bigg(\frac{1}{\abs{A}}\sum_{\mathbf{a}\in A}\max_{\mathbf{b}\in B} \inner{\mathbf{a}, \mathbf{b}} + \frac{1}{\abs{B}}\sum_{\mathbf{b}\in B}\max_{\mathbf{a}\in A} \inner{\mathbf{a}, \mathbf{b}}\bigg)
\end{equation}

\subparagraph{Word Mover-Distance (WMD)}
WMD measures dissimilarity between two documents by computing the minimum cumulative distance between the embeddings of their words~\citep{kusner2015word}. WMD uses optimal matching rather than greedy matching, based on the l2-norm of the embedding difference between two words.

In WMD, each language unit is first converted into a normalised BOW vector with length $\abs{\mathcal{V}}$, where $\mathcal{V}$ stands for the vocabulary. With two normalised BOW vectors, WMD measures the total distance of "moving" one to the other. A constrained optimisation problem is thus formulated as

\begin{equation}
\begin{aligned}
    d(\mathbf{a}, \mathbf{b}) = \min_T \quad & \sum_{i=1}^{\abs{\mathcal{V}}}\sum_{j=1}^{\abs{\mathcal{V}}} T_{ij}\cdot\abs{\mathbf{e}_i - \mathbf{e}_j}_2\\
    \text{s.t.} \quad &\sum_{j=1}^{\abs{\mathcal{V}}} T_{ij}= a_i, \forall i = 1, 2, \ldots, \abs{\mathcal{V}}\\
    &\sum_{i=1}^{\abs{\mathcal{V}}} T_{ij} = b_j, \forall j = 1, 2, \ldots, \abs{\mathcal{V}}\\
\end{aligned}
\end{equation}

where $a_i$ stands for the normalised occurrence of $i^\text{th}$ word in the vocabulary in language unit $\mathbf{a}$, and the same is applied to $\mathbf{b}$. $\mathbf{e}_i$ stands for the word embedding of $i^\text{th}$ word in the vocabulary.

WMD is a symmetric scheme and often used to measure the similarity between language generation outputs and references \citep{kilickaya-etal-2017-evaluating}.

WMD can be modified to account for word orders by either explicitly add a fragmentation penalty (WMD-O, \citealp{chow-etal-2019-wmdo}) or replace the l2-norm with a customised symmetric function (WE-WPI, \citep{echizenya-etal-2019-word}).

\subsubsection{Composite methods with Contextualised Embedding}
The positioning of contextualised embeddings is somewhat tricky to pin down. It is indeed an upgraded version of word embeddings, but a word is not always mapped to a vector. A longer unit does have a fixed mapping, but its form is not a vector, but a sequence of vectors. A vector can be obtained by simply taking the maximum or averaging, but this is sometimes not done. On the other hand, some methods of similarity measure are to directly use the sequence formed by this vector.

In constructing contextualised word embeddings, the embedding of a word depends on the context in which it is used. This approach aims to solve the disambiguation in polysemy words and in effect enhance the tasks in NLP that use embeddings. These contextualised embeddings are either based on recurrent neural networks (RNN) or transformers~\citep{vaswani2017attention}. Transformers models and RNN take two routes in processing sequential data. Particularly, transformers combine multi-level feed-forward neural networks with attention components.

Some popular examples of such contextualised embeddings include ElMo \citep{peters-etal-2018-deep} , BERT \citep{devlin-etal-2019-bert}, ERNIE~\citep{sun2020ernie} and XLNet \citep{yang2019xlnet}. A number of variations of BERT models were proposed based on the corpus used to train the model and by optimising the computational resources~\citet{lan2019albert,jiao-etal-2020-tinybert,liu2019roberta,sanh2019distilbert}. Another variation of BERT is to focus on domain-specific corpus training~\citep{beltagy-etal-2019-scibert}. \citet{raffel2019exploring}.

Pretrained contextualised embeddings show great potential in various NLP tasks, and many similarity schemes take advantage of contextualised embeddings. Here, we only introduce schemes that utilise off-the-shelf embeddings, not those that fine-tune RNNs or transformers. These schemes are sometimes called "untrained"~\citep{sai2022survey}.

\subparagraph{BERTscore}

\citep{zhang2019bertscore} measures soft overlap between two token-aligned texts, by selecting alignments, BERTScore returns the maximum cosine similarity between contextual BERT~\citep{devlin-etal-2019-bert} embeddings. Cosine is obtained between each token in one language unit and each token in the other unit. Then greedy matching is taken and finally an F1-score is calculated.

\begin{equation}
    \begin{aligned}
        Recall (A, B) &= \frac{1}{\abs{A}}\sum_{\mathbf{a}\in A}\max_{\mathbf{b}\in B} \inner{\mathbf{a}, \mathbf{b}}\\
        Precision (A, B) &= \frac{1}{\abs{B}}\sum_{\mathbf{b}\in B}\max_{\mathbf{a}\in A} \inner{\mathbf{a}, \mathbf{b}}\\
        s(A, B) &= 2\cdot\frac{Recall (A, B)\cdot Precision (A, B)}{Recall (A, B) + Precision (A, B)}
    \end{aligned}
\end{equation}

\citet{zhang2019bertscore} believe that BERTScore correlates better with human judgements in evaluating outputs from the NLG tasks like image captioning and machine translation. BERTr, only the recall part of BERTScore, is sometimes also believed to work well~\citet{mathur-etal-2019-putting}. BERTr is not symmetric, only recall is considered.

\subparagraph{MoverScore}
MoverScore can be seen as taking the complicated part from both BERTScore and WMD~\citep{zhao-etal-2019-moverscore}. It uses contextualised embeddings to compute the minimum cumulative distance from two language units. Compared with BERTscore, MoverScore allows partial matching and formulates a constrained optimisation to find the best matching. Compared with WMD, MoverScore uses contextualised word embeddings.

\subsubsection{Tuned Regression Models}
The similarity estimate is a regression task and it may rely on human annotated similarity scores to help improve the model. With the human annotated similarity between language units, various machine learning techniques such as linear regression, SVM, deep neural networks can be used.

I believe this category is the most flexible similarity model framework. First, it does not require any prior knowledge about why two units are similar, which is often required in other classes of schemes. Instead, the only requirement is a parallel dataset containing language unit pairs and similarity scores. Second, there are rich existing features that can be used as input, such as WMD, NGD, etc., and the permutation of these inputs can breed a large number of regression model variants. Third, the flexibility of neural networks allows learning of similarities directly from tokenised language units, which accelerates the growth of scheme diversity. Therefore, we will only be able to cover representative ones, rather than going through all variants that may not differ significantly.

\subparagraph{Grid search with bagging}
Q-Metrics~\citep{nema-khapra-2018-towards} focus on improving existing $n$-gram measures like ROUGE to better correlate with human annotated scores. It is believed that some words in a language unit carry more information than the other and therefore $n$-grams can be grouped separately. For each group the $n$-gram precision and recall are computed and they are averaged with weights to obtain overall precision and recall, whose harmonic mean yields an F-score $s_F(A, B)$. The overall Q-Metric is given as follows
\begin{equation}
    Q\text{-}Metric = k \cdot s_F(A, B) + (1 - k)\cdot m
\end{equation}

where $m$ is one metric of BLEU~\citep{papineni-etal-2002-bleu}, NIST~\citep{doddington2002automatic}, METEOR~\citep{banerjee-lavie-2005-meteor}, or ROUGE~\citep{lin-2004-rouge}. The weights in $s_F$ and $k$ are tuned using grid search and bagging to find the best values that maximise correlation with human scores.

\subparagraph{Better Evaluation as Ranking (BEER)}

The most straightforward option is to find a linear regression model that combines the heuristic-based features.
BEER is a linear regression model of several features derived from two language units~\citep{stanojevic-simaan-2014-beer,stanojevic-simaan-2014-fitting}. Here, features are functions or properties of two language units. These features could be statistical measures such as $n$-gram precision, recall or even other schemes ROUGE. BEER's input features include precision, recall and F1-score on word unigrams and character $n$-grams, as well as word order information based on permutation trees~\citep{zhang-gildea-2007-factorization}. Unigram statistics are computed separately for function words and content words and for the entire word set.

For any (phrase or longer) language units $A, B$, the BEER model is a linear combination of the input features as follows.

\begin{equation}
    s(A, B) = \sum_{i\in \text{all features}} W_i \cdot s_i(A, B)
\end{equation}

where $W_i$ is the learnt model weight for each feature $s_i(A, B)$.

\subparagraph{Enhanced Sequential Inference Model (ESIM)}

In ESIM~\citep{chen-etal-2017-enhanced,mathur-etal-2019-putting}, sentence embedding with Bi-LSTM is obtained for each language unit, where the sentence embedding is obtained from a concatenation of average and maximal value of each token embedding. Then a two-layer neural network is used to obtain the similarity value, as seen below:
\begin{equation}\label{eq:esim}
    s(A, B) = \sigma(\mathbf{u}\cdot ReLU\big(\mathbf{W}(\mathbf{e}_A \concat \mathbf{e}_B) + \mathbf{b}\big) + c)
\end{equation}

where $\mathbf{u}, \mathbf{W}, \mathbf{b}$ and $c$ are learnt parameters, and sigmoid ($\sigma$) and rectified linear unit (ReLU) functions are element-wise functions, given as below,

\begin{equation}
    \begin{aligned}
        \sigma(x) &= \frac{1}{1 + e^{-x}}\\
        ReLU(\mathbf{x}) &= [\ldots, \max(0, x_i), \ldots]^T, \forall i = 1, 2, \ldots N; \mathbf{x}\in\mathbb{R}^N.
    \end{aligned}
\end{equation}

\subparagraph{Neural Network Methods in General}
Semantic similarity methods have exploited the recent developments in neural networks to enhance performance. A neural network is a combination of linear combinations and non-linear activation functions such as the hyperbolic tangent function. Fitting a neural network to dataset requires gradient descent algorithm and automatic gradient back propagation as an numerical approximation.

The most widely used techniques include feed-forward neural network~\citep{sharif-etal-2018-learning,sharif2018nneval}, convolutional neural networks (CNN, \citealp{wang-etal-2016-sentence,shao-2017-hcti}), long short term memory (LSTM, \citealp{tien2019sentence, tai-etal-2015-improved}), Bidirectional long short term memory (Bi-LSTM, \citealp{he-lin-2016-pairwise,bahdanau2014neural,rush-etal-2015-neural}), decomposable attention model (DAM, \citealp{parikh-etal-2016-decomposable,lopez2019word}).

Recent progress with transformer models has led to improvements in semantic similarity measurement~\citep{vaswani2017attention,jiao-etal-2020-tinybert,shimanaka2019machine,kane-etal-2020-nubia}. The performance of the models is improved by the use of a larger corpus, which again stresses the importance of building an ideal corpus.

Most deep learning models are black-box models, and it is difficult to ascertain the features for which performance is achieved, making them difficult to interpret, unlike corpus-based methods with strong mathematical underpinnings will be. This also leads to the fact that these neural networks schemes are not symmetric measures at birth, because the output regression can be unpredictable. some schemes such as DAM was even originally proposed for entailment determination (introduced below). However, simple symmetric modification can be applied as follows,

\begin{equation}
    s(A, B) = \frac{1}{2}\big(s_\text{neural network}(A, B) + s_\text{neural network}(B, A)\big)
\end{equation}

Depending on whether a symmetric approach is taken, these neural-network method can be symmetric or asymmetric.

Many fields that deal with sensitive data are hesitant to use deep neural network-based methods because they are not easily understood~\citep{chandrasekaran2021evolution}.

\subparagraph{Regressor Using Sentence Embeddings (RUSE)}

RUSE is a regression model that considers three pre-trained sentence embeddings~\citep{shimanaka-etal-2018-ruse}. The three sentence embeddings, InferSent~\citep{conneau-etal-2017-supervised}, Quick-Thought~\citep{logeswaran2018efficient}, and Universal Sentence Encoder~\citep{cer2018universal,cer-etal-2018-universal}, are concatenated to form a longer sentence embedding for each sentence. A multi-layer perceptron regressor (MLP, \citealp{scikit-learn}) similar to the one in equation~\ref{eq:esim} predicts the RUSE score by taking the sentence embeddings as input, as seen below.

\begin{equation}
    s(\mathbf{a}, \mathbf{b}) = MLP\bigg(\mathbf{a} \concat \mathbf{b} \concat [\abs{\mathbf{a} - \mathbf{b}}_2, \inner{\mathbf{a}, \mathbf{b}}]^T\bigg).
\end{equation}

\subsubsection{SIMILE}

This measure has two parts: the cosine similarity between sentence embeddings to better compare two sentences, and a length penalty (LP) to prevent word repetitions~\citep{wieting-etal-2019-beyond}. For two sentence $A$ and $B$, LP is given as follows,

\begin{equation}
    LP(A, B)= \exp{\bigg(1-\frac{\max(\abs{A}, \abs{B})}{\min(\abs{A}, \abs{B})}\bigg)}
\end{equation}

Together with the sentence embedding $\mathbf{a}, \mathbf{b}$ for $A, B$ respectively, SIMILE is given as,

\begin{equation}
    s(A, B) = LP(A, B)^k \inner{\mathbf{a}, \mathbf{b}}
\end{equation}

where $k$ is a coefficient to balance the two parts.


\subsection{Directed Graph}

\begin{figure}
\centering
\resizebox{0.6\linewidth}{!} {
\begin{tikzpicture}[
    text width=5em,text centered,
    > = stealth, 
    shorten > = 1pt, 
    auto,
    node distance = 4cm, 
    semithick 
    ]

    \tikzset{every state}=[
    draw = black,
    thick,
    fill = white,
    minimum size = 1mm
    ]

    \node[state] (A) {A jaguar kills a crocodile};
    \node[state] (B) [right=of A]{A crocodile is not murdered by a jaguar};
    \node[state] (C) [below=of A] {Sonny kills Dr. Lanning};
    \node[state] (D) [below=of B] {Dr. Lanning is not murdered by Sonny};

    \path[<->] (A) edge  node[] {0.947, 0.923} (B);
    \path[<->] (A) edge  node[left] {0.868, 0.878} (C);
    \path[<->] (A) edge  node[pos=0.05, right] {0.857, 0.890} (D);
    \path[<->] (B) edge  node[pos=0.15, below] {0.859, 0.903} (C);
    \path[<->] (B) edge  node[] {0.891, 0.766} (D);
    \path[<->] (C) edge  node[] {0.938, 0.899} (D);

    \end{tikzpicture}
}
\caption{In a complete digraph, between each two vertices there are two edges. Both edges have directions, with possibly different weights. The bivariate similarity function is not symmetric.}
\label{fig:complete_digraph}
\end{figure}
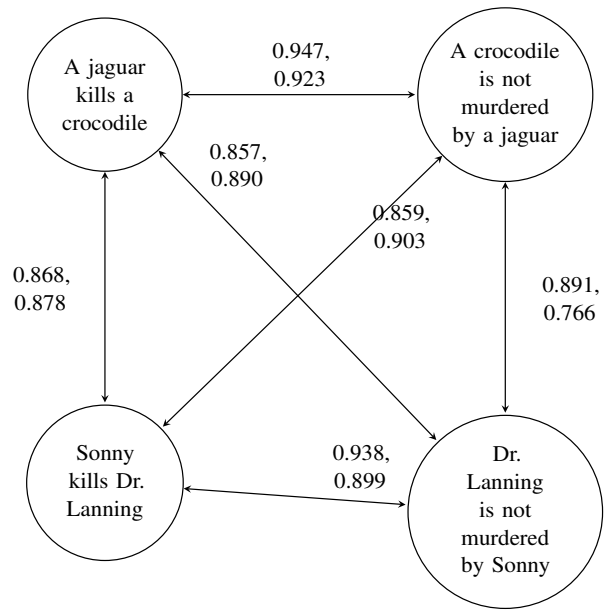


Some bivariate similarity functions are not symmetric, and the order or permutation of the two arguments to the function makes a difference. Such relations can be represented in a directed complete graph, also known as complete digraph, where all edges are directed, as seen in figure~\ref{fig:complete_digraph}. We list these asymmetric measures below.

\subparagraph{Bilingual Evaluation Understudy (BLEU)}
BLEU consists of two parts, a \textit{precision}-based bag semantic measure that computes the $n$-gram overlap between two language units, and a brevity penalty term that penalises very short sentences~\citep{papineni-etal-2002-bleu}. This is because BLEU is used to evaluate the output of machine translation, and short but meaningless sentences need to be avoided. In a stricter sense, BLEU is not a similarity measure, since the brevity penalty is imposed unilaterally on one of the two units. However, \citet{papineni-etal-2002-bleu} explicitly mentioned that this metric estimates the "closeness of two sentences", which is why I added it here.

\subsubsection{Metric for Evaluation of Translation with Explicit Ordering (METEOR)}
METEOR~\citep{banerjee-lavie-2005-meteor} measures overlapping unigrams, equating a unigram with its stemmed form, synonyms, and paraphrases. This is believed to mitigate two drawbacks of BLEU~\citep{papineni-etal-2002-bleu}, namely recall not being considered and exact $n$-gram matching only. Therefore, METEOR is designed as an F-measure scheme and relaxes matching criteria, where it first tries to match exact uni-grams, then stemmed-words, and eventually the synonyms. This is an asymmetrically parameterised F-score, as follows.

\begin{equation}
    F = \frac{Precision \cdot Recall}{0.1\cdot Precision + 0.9\cdot Recall}
\end{equation}

Complementary to the property that METEOR only matches uni-grams, as opposed to $n$-gram matches, is a penalising term to reward longer consecutive matches. Despite the effectiveness of this term, it excludes METEOR from the scope of schemes with abstract representation.

\subsubsection{ChrF}

Language can be tokenised not only in words but in sub-words or even characters as well. Character-based metrics is believed to better capture morphological information among words~\citep{wang-etal-2016-character,popovic-2015-chrf}. \citet{popovic-2015-chrf} propose chrf, which happens to be an asymmetric measure. Essentially, chrf follows bag semantics on character level, but gives different weights to precision and recall.

The final chrF score is then computed as:
\begin{equation}
    chrF_\beta = (1+\beta^2) \frac{Precision\cdot Recall}{\beta^2\cdot Precision + Recall}
\end{equation}

where the value of $\beta$ indicates that recall is given $\beta$ times more weightage than precision. Character-level measures are not necessarily asymmetric.

\subsubsection{BLEURT}
\label{bleurt}
A linear regression model, when helped by a large number of signals, achieves the state-of-the-art performance in measuring similarity between outputs and references in machine translation~\citep{sellam-etal-2020-bleurt}.

Essentially, BLEURT is a BERT model, retrained using several flags and signals. First sentence pairs are synthesised by perturbing Wikipedia sentences via mask-filling with BERT, back-translation or randomly dropping words. Next, the signals are provided, including several schemes discussed above, namely BLEU, ROUGE, and BERTscore. The signals also include a flag indicating how the synthesised sentence has been perturbed, and textual entailment signal predicted by a off-the-shelf entailment predicting model. Then, a BERT model is fine-tuned  using least square error loss as follows.

\begin{equation}
    loss= \frac{1}{N}\sum_{n=1}^N \abs{y_i-\hat{y}}^2
\end{equation}

The fine-tuned BERT model, known as BLEURT now, takes sentence pairs as input and outputs contextualised token embeddings as other BERT models do. A linear layer ($(\mathbf{u}, c)$) is used to predict the regression number, as seen below.

\begin{equation}
    s(A, B) = \hat{y} = \mathbf{u}\cdot\mathbf{e}_{A\concat B} + c.
\end{equation}

\subsubsection{Textual Inference}

As seen above, entailment or inference information can help estimate similarity. Uncertain entailment, as proposed by \citet{chen-etal-2020-uncertain}, essentially works as conditional probability between two statement $A$ and $B$, one-directional similarity can thus be:
\begin{equation}
    s(A, B) = P(A|B)
\end{equation}

where $P(A|B)$ is the likelihood of statement $A$ holds, given that statement $B$ holds.

\section{Evaluation of Similarity Schemes}

In this section, we discuss how humans evaluate various similarity computation methods. For this purpose, one of the most widely used methods is to calculate the correlation between the numerical value given by a program and the subjective judgment of humans, which is also called intrinsic evaluation~\citep{bar2015composing}. Commonly used correlation measures are Pearson correlation ($r$), Spearman's correlation ($\rho$), and Kendall's $\tau$ index~\citep{sai2022survey}. Each of the three correlation indices plays two roles: for example the Pearson correlation itself can be used to measure the similarity between real-vector formed language units, and the numerical distribution given by a similarity scheme can be again compared with the human judgment scale to calculate a similarity.

Because of this recursive property, sometimes we find poor correlation of the scales reported by a similarity scheme with direct human judgements~\citep{stent2005evaluating}, though it is uncertain which part of "similarity measurement" goes wrong.

In addition to correlation, other aspects like interpretability~\citep{zhang-etal-2004-interpreting,callison-burch-etal-2006-evaluating} or adaptability~\citep{reiter-2018-structured,nema-khapra-2018-towards,liu-etal-2016-evaluate,kilickaya-etal-2017-evaluating} have been studied in various previous works and found to be in need of improvement. 

The complexity of languages is challenging to model with all its nuances. One similarity scheme is difficult to account for all required or related nuances~\citep{kryscinski-etal-2019-neural,wiseman-etal-2017-challenges,liu-etal-2016-evaluate,sai2019re}, for example, consistency of fact and style can be very different characteristics. The purpose of this thesis is indeed to work towards a scheme that captures more nuanced similarities.



\chapter{Topic Modelling with Word Vector Representations} 

\label{Chapter3} 

Expression similarity can be used to cluster topics. "Similar" expressions are grouped in the same topic cluster, while "disimilar" expressions are grouped in different topic clusters. The study of similarity schemes in this chapter focuses on one of their applications, topic modelling, and it turns out that these similarities are fairly useful.


\section{Introduction}
\label{sectIntro}

Twitter is a popular microblogging service with a high volume of 500 million tweets posted each day~\citep{stats_twitter}. Microblogging services such as Twitter offer a range of topics from basic and popular (e.g., movies, television, music, entertainment) to niche and specialised topics such as politics, religion and current affairs. The ability to detect and understand discussions on these topics can be used for a variety of purposes, such as understanding general sentiment and trends on these topics, as well as recommending accurate and up-to-date content. However, due to the high volume and high frequency of tweets, it is difficult for users to understand the topics discussed in these tweets~\citep{kumar2014twitter,liao2012mining}.

A popular approach is to utilise topic modelling algorithms to automatically detect the topics discussed in a set of traditional text-based documents, such as news articles, academic papers, etc. In such algorithms, the output is a set of keywords denoting the topics that are relevant to each document. Examples of topic modelling algorithms are the original Latent Semantic Analysis~\citep{deerwester1990indexing,landauer1997solution}, Probabilistic Latent Semantic Analysis~\citep{hofmann1999probabilistic} and latent Dirichlet allocation~\citep{blei2003latent}. These algorithms were developed mainly for topic modelling on traditional and large documents such as news articles or papers~\citep{de2009cross,jacobi2016quantitative}.

The advent of microblogging services has led to the widespread use of short documents (\emph{i.e.}, tweets) in social media, which traditional topic modelling algorithms do not work well on. In response, various researchers have proposed variants of these traditional topic models, based on various types of aggregation schemes to combine a set of tweets as larger documents~\citep{hong2010empirical,mehrotra2013improving}. While latent Dirichlet allocation and its variant have been shown to model topics well for traditional documents, the number of topics needs to be defined in advance and they do not account for the syntactic structure of sentences. To overcome these limitations, \cite{lim2017clustop} introduce a topic modelling algorithm that is able to automatically determine the appropriate number of topics.

Today, we are looking into whether newer similarity schemes can improve the performance of a community detection method for topic modelling. In this study, I use embeddings to calculate the similarity between words. Multiple variants of different word embedding techniques are used to create the edges weights.

\section{Preliminary}

\citet{lim2017clustop} propose a clustering-based topic modelling (ClusTop) algorithm that leverage community detection approaches for modelling topics on Twitter using a word network graph. In the word network graph, nodes represent different definitions of words and phrases and edges represent either word or phrase co-occurrences  Unlike more traditional topic models, ClusTop automatically determines the appropriate number of topics by maximising a modularity score among words in the network.

In addition to using a traditional co-word usage network, \citet{lim2017clustop} experiment with different variants of ClusTop algorithm based on numerous definitions of words (unigrams, bigrams, trigrams, hashtags, nouns from part-of-speech tagging), types of relations (word co-occurrence frequency and word embedding similarity distance) and different aggregation schemes (individual tweets, hashtags and mentions). 

Using three Twitter datasets with labelled topics, \citet{lim2017clustop} evaluate ClusTop and its variants against various LDA baselines based on measures of topic coherence, pointwise mutual information, precision, recall and F-score. Experimental results show that ClusTop offers superior performance based on these evaluation metrics, compared to the various baselines.

I am hoping to improve the accuracy of community detection algorithms by expanding their options of edge weights. In a network graph where the vertices are still words and the edges are newer proposed relationships between words, I explore whether the advanced algorithm is also able to better reflect syntactic nature. Empirical research is needed to better understand the different types of network graphs based on node types, edges, and embedding methods, and their impact on the accuracy and quality of discovered topics.

\subsection{Structure and Organization}
The rest of this chapter is structured as follows. Section~\ref{sectRelatedWork} discusses key literature on studying topics on microblogs and topic modelling algorithms. Section~\ref{sectAlgorithm} describes our ClusTop algorithm. Section~\ref{sectDataEval} outlines our experimental methodology in terms of the dataset used, baseline algorithms and evaluation metrics.
Section~\ref{sectExperiments} highlights the results from our evaluation and discusses our main findings.
Section~\ref{sectConclusion} concludes this paper and highlights possible future directions for this work.

\section{Refined Algorithm}
\label{sectAlgorithm}

\begin{figure} [t!]
  	\centering
    \includegraphics[width=0.9\columnwidth, keepaspectratio=true, angle=0, trim={0mm 0mm 0mm 0mm}]{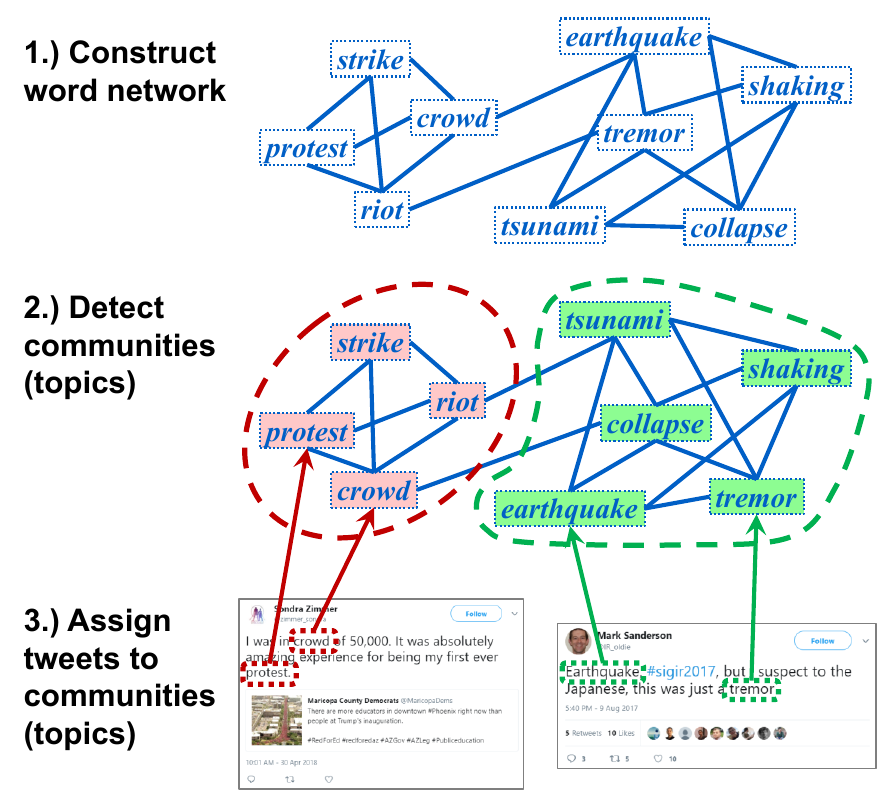}
    \caption{Overview of the ClusTop Algorithm}
    \label{ClusTopOverview}
\end{figure}

We now describe our extending to ClusTop algorithm by first defining the basic notations and preliminaries used in this algorithm. Using standard network theory notations, we denote $V$ and $E$ to represent the set of vertices and edges, respectively. Following this, an undirected graph $G = (V, E)$ is represented as a collection of vertices $V$ that are connected by a set of edges $E$. In turn, each edge $e \in E$ is denoted by
\begin{equation}
    e = (\{v_i, v_j\}, w)
\end{equation}

where $w$ represents the weight of the link between vertices $v_i$ and $v_j$. In our application of community detection algorithms to topic modelling, \citet{lim2017clustop} explore the use of an undirected graph as $G = (U, R)$, where $U$ is the set of unigrams (vertices) and $R$ is the set of relations (edges) between the unigrams. They further examine the effects of different definitions of vertices, such as bi-grams, tri-grams, hashtags, etc, as well as different types of edge weights, such as frequency counts. Figure~\ref{ClusTopOverview} provides an overview of our ClusTop algorithm, with the following three main steps of network construction, community (topic) detection and topic assignment. 

\paragraph{Network Construction}

The first step of this algorithm involves constructing a unigram network, \emph{i.e.}, an undirected graph $G = (U, R)$, based on a particular definition of vertices (unigrams) and edges (relations). This step will be elaborated further in Section~\ref{networkConstruct}, where we will describe the various types of vertices (unigrams) and edges (relations) modelled in this work.

\paragraph{Community Detection}

Using the network graph obtained from Step 1, we will next apply community detection approaches to identify the main communities (topics), where sets of vertices (unigrams) will be grouped into different communities that represent different topics. This step will be further described in Section~\ref{sectCommDetect}. 

\paragraph{Topic Assignment}

Based on the detected community from Step 2 that corresponds to a specific topic, this step examines individual tweets and aims to assign this tweet to a specific community. In short, this step aims to label each tweet with an appropriate topic. More details about this step are provided later in Section~\ref{sectTopicAssign}.

\subsection{Network Construction \citep{lim2017clustop}}
\label{networkConstruct}
\begin{algorithm}
\caption{Network Construction}\label{alg:networkConstruction}
\small
\begin{algorithmic}
\Require $T$: Collection of tweets in corpus
\Return $G = (U, R)$: Network graph of unigrams (vertices) and relations (edges).
\State Initialise an empty graph $G$
\For{each tweet $t \in T$}
\For{each word-pair $(p_1,p_2) \in t$}
\State $e \gets (\{p_1,p_2\}, 1)$
\If{edge $e$ exists in graph $G$}
\State increment edge $e$ in graph $G$ by 1
\Else
\State add edge $e$ to graph $G$
\EndIf
\EndFor
\EndFor

\end{algorithmic}
\end{algorithm}

\citet{lim2017clustop} construct a network graph based on the combination of two aspects: different vertices and the edges in between, and different types of document aggregation, \emph{i.e.}, individual tweets, aggregated by hashtag or mentions. 

The first stage of our algorithm involves constructing a network graph of word usage, as shown in algorithm~\ref{alg:networkConstruction}. This algorithm involves the following: (i) examining all tweets and tokenizing all words in each tweet based on whitespaces; (ii) for each word-pair in each tweet, build a weighted edge $e$ linking the two words; and (iii) repeating Steps 1 and 2 for all tweets, until we obtain a network graph, where the vertices represent uni-grams and edges represent a relation between two unigrams. The choice of vertices and edges will lead to a different type of network graph being constructed. To better examine the effect of these vertices and edges on the graph type, we experiment with a variety of relations types between different types of uni-gram, including a combination of the two aspects: 
\begin{itemize}
    \item $\{$ uni-gram (\textbf{Word}), bi-gram (\textbf{BiG}), tri-gram (\textbf{TriG}), noun uni-gram (\textbf{Noun}), hashtag uni-gram (\textbf{Hash}), a combination of bigram occurrence and co-hashtag usage (\textbf{BiHa})$\}$, and
    \item $\{$no aggregation (\textbf{NA}), aggregate by hashtags (\textbf{AH}), aggregate by mentions (\textbf{AM})$\}$.
\end{itemize}

Edge weights have been determined based on the co-occurrence frequency of  terms observed in a set of tweets, \emph{i.e.}, the training set. We here make use of word embedding to model edge weights as the cosine similarity between a pair of words, \emph{i.e.}, more similar words will be linked with a higher edge weight. For this purpose, we first use the GloVe algorithm~\citep{pennington-etal-2014-glove} for generating the word vector based on hashtags used, then construct a network with vertices based on hashtags and edge weights based on the cosine similarity scores between hashtags. In addition to GloVe, we also generalise this variant using other popular word embedding algorithms such as Word2Vec\citep{mikolov2013efficient} and FastText\citep{joulin-etal-2017-bag} to better examine the effects of different word embedding techniques on our approach.

We denote the three variants of these word embedding based networks with its similarity based edge weights as:

\paragraph{Hash2Vec-Glove (H2VG)}

A network based on co-usage of hashtags in the same tweet, where the edge weights are based on cosine similarity scores of a word vector trained using GloVe~\citep{pennington-etal-2014-glove}.
\paragraph{Hash2Vec-Word2Vec (H2VW)}

A network based on co-usage of hashtags in the same tweet, where the edge weights are based on cosine similarity scores of a word vector trained using Word2Vec~\citep{mikolov2013efficient}.
\paragraph{Hash2Vec-FastText (H2VF)}

A network based on co-usage of hashtags in the same tweet, where the edge weights are based on cosine similarity scores of a word vector trained using FastText~\citep{joulin-etal-2017-bag}.

\subsection{Community Detection}
\label{sectCommDetect}

\begin{algorithm}[t]
\begin{algorithmic}
\Require $G = (U, R)$: Network graph of unigrams (vertices) and relations (edges)
\State Assign all unigrams $u$ into their own community

\For{Community structure stabilises and modularity score is maximised}
\For{\upshape{each unigram $u \in U$}}
\State	$MaxModularity \leftarrow -1$
\State	$MaxModNeighbour \leftarrow NULL$
\For{\upshape{each neighbour $u_n$ of unigram $u$}}
\State		$ShiftMod \leftarrow$ Modularity score of shifting unigram $u$ to neighbour $u_n$'s community
\If{$ShiftMod > MaxModularity$}
\State			$MaxModularity \leftarrow ShiftMod$
\State			$MaxModNeighbour \leftarrow u_n$
		\EndIf	
	\EndFor
\State	$OriginalMod \leftarrow$ Modularity score of unigram $u$ in its original community
\If{$OriginalMod > MaxModularity$}
	Shift unigram $u$ to the community of MaxModNeighbour
\Else{
	Keep unigram $u$ in its original community
}
\EndIf
\EndFor
\State	$U_n \leftarrow$ New unigrams (vertices) $U_n$ based on the newly-formed communities
\State	$R_n \leftarrow$ New relations (edges) $R_n$ based on edge weights between nodes in two communities
\State	$G_n \leftarrow$ New network graph $G_n = (U_n, R_n)$
\State	The algorithm iterates again (Lines 4 to 19) with network graph $G_n$ as input
\EndFor

\State \Return $a_n$
\end{algorithmic}

\caption{Topic Modelling using Louvain \citep{lim2017clustop}}
\label{louvainTopicAlgo}
\end{algorithm}

After constructing the network graph in the previous section, we now describe the approach to modelling the topics in this graph using community detection approaches. The main example in this work is following the adaptation of the Louvain algorithm~\citep{blondel2008fast} by \citet{lim2017clustop}. The Louvain algorithm has been shown to be one of the best performing algorithm in a comprehensive survey of community detection algorithms~\citep{fortunato2010community}.

The adaptation of the Louvain algorithm~\citep{blondel2008fast} for the purpose of topic modelling is described by the pseudo-code in Algorithm~\ref{louvainTopicAlgo}, which comprises the following steps:

\begin{enumerate}

\item Initially, each unigram is placed in its own community/cluster (Line 2).
\item Following which, for each unigram, we examine each neighbour of this unigram and combine two unigrams into the same community/cluster if their modularity gain is the greatest among all of the neighbours (Lines 4 to 16).
\item Next, we build a new network graph where unigrams in the same community/cluster are combined as a single vertex (unigram), and Step 2 is repeated until the modularity score is maximised (Lines 17 to 20).
\end{enumerate}

One of the reasons for the Louvain algorithm's good performance is due to its local adjustment of unigrams (vertices) into communities/clusters, by maximizing the gain in the following modularity function~\citep{blondel2008fast}:

\begin{equation}
Q = \Bigg[ \frac{\sum_{in} + k_{i,in}}{2m} - \bigg( \frac{\sum_{tot} + k_{i}}{2m} \bigg)^2 \Bigg] - 
\Bigg[ \frac{\sum_{in}}{2m} - \bigg( \frac{\sum_{tot}}{2m} \bigg)^2 - \bigg( \frac{k_{i}}{2m} \bigg)^2 \Bigg]
\label{louvainModularity}
\end{equation} 
\vspace{3mm}

{\noindent where $\sum_{in}$ and $\sum_{tot}$ represents the total weight of all links inside a community/cluster and total weight of all links to a community/cluster, respectively. Similarly, the terms $k_{i}$ and $k_{i,in}$ denote the total weight of all links to $i$ and total weight of links to $i$ within the community/cluster. Lastly, $m$ denotes the total weight of all links in the network graph.}

At the end of this step, we will obtain a set of communities/clusters based on the provided network graph. Each community/cluster will represent a particular topic, where the members of each community/cluster serve as the representative words of each topic. For each topic, we also rank the keywords (\emph{i.e.}, members of each community) based on the total weight of all links to a unigram/vertex.

\subsection{Topic Assignment}
\label{sectTopicAssign}

Given the detected communities/topics $C$ from Section~\ref{sectCommDetect} and a tweet $t = \{u_1,...u_n\}$, we define the most likely topic for this tweet as:

\begin{equation}
\argmax_{c \in C} \sum_{u \in c} k_u \delta(u=u_t), ~~~~~\forall~u_t \in t
\label{topicAssign}
\end{equation} 

{\noindent where $\delta(u=u_t) = 1$ if a unigram $u$ of a community/topic $c \in C$ is the same as a unigram $u_t$ of a tweet $t$ and $\delta(p=c) = 0$ otherwise, and $k_u$ denotes the total weight of links to unigram $u$ (as previously described in Section~\ref{sectCommDetect}).}

In short, we assign a tweet $t$ to a community/topic $c$ that has the highest co-occurrence of unigrams in both the tweet and community/topic, where the unigram in the community/topic is weighted based on its co-occurrence to other unigrams.

\section{Dataset and Evaluation Methodology}
\label{sectDataEval}

In this section, we give an overview of our experimental dataset and describe our evaluation methodology in terms of the ClusTop algorithm variants, baseline algorithms and evaluation metrics.

\subsection{Dataset}

For our experimental evaluation, we utilise the same three Twitter datasets with labelled topics~\citep{olteanu2014crisislex,olteanu2015expect,zubiaga2016analysing} as those used by \citet{lim2017clustop}, which enables us to fairly evaluate our algorithm and baselines against the ground truth topics compared to an unlabelled dataset.

\begin{table} [h]
\label{datasetDesp}
\small
\centering
\begin{tabular}{lllll} \hline
Dataset & Paper	Reference &	Number	of Topics &	Total Tweets & Labelled	Tweets \\
\hline
A	&	\citep{olteanu2014crisislex}	&	6	&	7.67mil	&	60k	 \\ 
B	&	\citep{olteanu2015expect}	&	26	&	0.28mil	&	27.9k	 \\ 
C	&	\citep{zubiaga2016analysing}	&	8	&	4.8k	&	3.6k	 \\
\hline 
\end{tabular}
\caption{Description of Dataset}

\end{table}

We split each dataset into four partitions and perform a 4-fold cross validation~\citep{kohavi1995study}. At each evaluation iteration, we use three partitions as our training set and the last partition as our testing set. After completing all evaluations, we compute and report the mean results for each algorithm based on the metrics of topic coherence, pointwise mutual information, precision, recall and f-score, which we elaborate further in the rest of the paper.

\subsection{Topic Quality Metrics}

For determining the quality of the detected topics, we measure the topic quality based on the topic coherence and pointwise mutual information metrics. These two metrics have also been widely used by many topic modelling researchers~\citep{mimno-etal-2011-optimizing,yao2015incorporating,mehrotra2013improving,lim2017clustop}. For both evaluation metrics, we denote a detected topic $t$ that comprises a set of $n$ representative unigrams/keywords $U^{(t)} = (u^{(t)}_1,...,u^{(t)}_n)$ for each topic.

\paragraph{Topic Coherence (TC)}
Given that $D(u_i, u_j)$ denotes the number of times both unigrams $u_i$ and $u_j$ appeared in the same document/tweet, and similarly, $D(u_i)$ for a single unigram $u_i$, topic coherence is defined as:

\begin{equation}
TC(t, U^{(t)}) = \sum_{u_i \in U^{(t)}}\sum_{u_j \in U^{(t)}, u_i \ne u_j} log~\frac{D(u_i, u_j)}{D(u_j)}
\label{topicCoherence}
\end{equation} 

\paragraph{Pointwise Mutual Information (PMI)}
Given that $P(u_i, u_j)$ denotes the probability of a unigram pair $u_i$ and $u_j$ appearing in the same document/tweet, and $P(u_i)$ for the probability of a single unigram $u_i$, pointwise mutual information is defined as:

\begin{equation}
PMI(t, U^{(t)}) = \sum_{u_i \in U^{(t)}}\sum_{u_j \in U^{(t)}, u_i \ne u_j} log~\frac{P(u_i, u_j)}{P(u_i) P(u_j)}
\label{topicPMI}
\end{equation} 

In both the TC and PMI metrics, it is possible for a division by 0 or taking the log of 0 when the appropriate numerator or denominator is 0, \emph{i.e.}, when a particular word or word pair has not been previously observed. As such, we adopt a similar strategy as~\citep{mimno-etal-2011-optimizing,mehrotra2013improving} by adding a small value $\epsilon=1$ to both components to avoid the situation of a division by 0 or log of 0.

\subsection{Topic Relevance Metrics}

Precision, recall and f-score are popular metrics used in Information Retrieval and other related fields, such as in topic modelling~\citep{aiello2013sensing,ritter2012open}, tour recommendation~\citep{halder2021transformer,brilhante2015planning,zhou2020semi}, location prediction and tagging~\citep{zheng2015towards,cao2015inferring,zheng2018survey}, event detection~\citep{george2021real}, among others. In contrast to the previous topic quality metrics (TC and PMI), these metrics allow us to evaluate how relevant and accurate the detected topics are, compared to the ground truth topics. In topic modelling, researchers typically manually curate a set of ground truth keywords to describe a specific topic, then evaluate how well the detected keywords from their topic models match these ground truth keywords~\citep{aiello2013sensing}. For our evaluation, we adopt a similar methodology except that we automatically determine the ground truth keywords from the respective Wikipedia article for each topic.


Given that $U^{D} = (u^{D}_1,...,u^{D}_n)$ and $U^{G} = (u^{G}_1,...,u^{G}_n)$ denotes the set of detected unigrams and ground truth unigrams for a specific topic, the metrics we use are as follows:

\paragraph{Precision}
The proportion of unigrams for the detected topic $U^{D}$ that also appears in the ground truth unigrams $U^{G}$. For a topic $t$, precision is defined as:
\begin{equation}
P(t) = \frac{|U^{D} \cap U^{G}|}{|U^{D}|}
\label{precision}
\end{equation} 

\paragraph{Recall}
The proportion of ground truth unigrams $U^{G}$ that also appears in the unigrams for the detected topic $U^{D}$. For a topic $t$, recall is defined as:
\begin{equation}
R(t) = \frac{|U^{D} \cap U^{G}|}{|U^{G}|}
\label{recall}
\end{equation} 

\paragraph{F-score}
The harmonic mean of precision $P(t)$ and recall $R(t)$, which was introduced in Equations~\ref{precision} and ~\ref{recall}, respectively. For a topic $t$, F-score is defined as:
\begin{equation}
F(t) = \frac{2 \times P(t) \times R(t)} {P(t) + R(t)}
\label{fscore}
\end{equation} 

In our experiments, we compute the precision, recall and F-score derived from the testing set, in terms of the top 5 and 10 keywords of each topic modelled.

\subsection{Summary Rank Metrics} 

As our experiments involve five evaluation metrics, three datasets and 18 algorithms, we develop an intuitive approach to represent the performance of each algorithm. This approach first ranks an algorithm's performance from 1 to 18 for each evaluation metric and dataset, with the lowest rank being the best performing one. For each combination of topic quality metrics (topic coherence and pointwise mutual information) and topic relevance metrics (precision, recall and f-score), we take the average of each metrics group across all three datasets for an average rank. For example, if an algorithm ranked 1st, 1st and 2nd in terms of topic coherence and 2nd, 1st, 2nd in terms of pointwise mutual information for datasets A, B, C, respectively, this algorithm will be assigned an overall rank of 1.5 for the topic quality metric. 


Note that we did not use the ClusTop variants based on bigrams and trigrams combined with the hashtag and mention aggregation schemes, as these variants provide minimal improvements compared to their original non-aggregated variants. Consider a simple example of three tweets with a common hashtag, the hashtag aggregation scheme with bigrams will only produce an additional two bigrams resulting from the first and second tweet as well as the second and third tweet. Moreover, these two additional bigrams will be generated from the last word of the first tweet and the first word of the second tweet, which will not be syntactically meaningful in most cases.

\subsection{Baseline Algorithms}

LDA is a popular topic modelling algorithm that was used for traditional documents (such as news articles), and more recently for social media (such as tweets on Twitter). Given the popularity of LDA for topic modelling, we compare our ClusTop algorithm and its variants against the following LDA-based algorithms, namely:

\paragraph{LDA-Orig}
The original version of LDA introduced by~\citep{blei2003latent}, where each document corresponds to a single tweet.  
\paragraph{LDA-Hash}
A variant of LDA applied on Twitter, where each document is aggregated from multiple tweets with the same hashtag~\citep{mehrotra2013improving}. 

\paragraph{LDA-Ment}
An adaptation of the Twitter-based LDA variant proposed by~\citep{weng2010twitterrank}, where we aggregate tweets with the same mention into a single document. 

\paragraph{Basic ClusTop Series}

Basic variants of ClusTop \citep{lim2017clustop} utilises token co-occurrence as graph edges, which already demonstrate its better modelling capability compared with the LDA baselines.

\section{Experimental Results and Discussion}
\label{sectExperiments}

In this section, we report on the results of our experiments and discuss some implications of these findings.

\begin{table*} [!t]
\caption{Comparison of ClusTop algorithm against various baselines, in terms of Topic Coherence (TC) and Pointwise Mutual Information (PMI) for the top 5, 10, 15 and 20 keywords. The rank of an algorithm's performance for each metric are provided in brackets.}
\label{resultsTCPMI-summary}
\centering
\begin{tabular}{r H H H H H H c H H H H H H c H H H H H H H c H H H H H H c} \hline	
{\bf Algorithm}& {TC} & {PMI} & {TC} & {PMI} & {TC} & {PMI} & {\bf Rank@5} & {TC} & {PMI} & {TC} & {PMI} & {TC} & {PMI} & {\bf Rank@10}  & 	{\bf Algorithm}& {TC} & {PMI} & {TC} & {PMI} & {TC} & {PMI} & {\bf Rank@15} & {TC} & {PMI} & {TC} & {PMI} & {TC} & {PMI} & {\bf Rank@20} \B \T \B \\ \hline
ClusTop-Word-NA   & -37.6 (21) & -5.5 (15)  & -34.1 (15) & -7.7 (14)  & -37.9 (18) & -14.4 (17) & (16.7) & -171.0 (21) & -49.2 (16)  & -160.8 (17) & -39.5 (15)  & -173.4 (18) & -67.5 (19)  & (17.7) & 	ClusTop-Word-NA   & -409.7 (21) & -157.4 (17) & -382.6 (19) & -115.5 (16) & -403.5 (18) & -147.6 (21) & (18.7) & -768.2 (21)  & -318.1 (19)  & -695.1 (19)  & -234.8 (17)  & -732.7 (18)  & -247.8 (21) & (19.2) \T \B \\
ClusTop-BiG-NA    & -36.6 (20) & 7.3 (8)    & -35.9 (17) & 1.2 (9)    & -42.5 (22) & -16.4 (18) & (15.7) & -153.4 (20) & -29.6 (13)  & -158.2 (16) & -25.8 (13)  & -194.8 (21) & -63.4 (18)  & (16.8) & 	ClusTop-BiG-NA    & -357.2 (20) & -108.3 (15) & -360.3 (17) & -77.2 (12)  & -447.9 (20) & -116.5 (18) & (17.0) & -645.5 (20)  & -212.3 (16)  & -633.7 (17)  & -147.8 (13)  & -794.7 (20)  & -165.4 (18) & (17.3) \T \B \\
ClusTop-TriG-NA   & -30.9 (17) & 10.7 (5)   & -35.8 (16) & -2.6 (13)  & -42.0 (21) & -18.2 (19) & (15.2) & -122.6 (16) & -16.1 (12)  & -166.5 (19) & -25.1 (12)  & -194.2 (20) & -73.5 (21)  & (16.7) & 	ClusTop-TriG-NA   & -289.2 (16) & -83.1 (13)  & -382.0 (18) & -86.2 (15)  & -451.6 (21) & -145.4 (20) & (17.2) & -530.2 (16)  & -170.0 (14)  & -674.8 (18)  & -183.1 (14)  & -804.0 (21)  & -222.8 (19) & (17.0) \T \B \\
ClusTop-BiHa-NA   & -23.3 (12) & 19.6 (1)   & -32.3 (14) & 4.7 (7)    & -37.9 (18) & -11.2 (14) & (11.0) & -81.4 (12)  & 7.1 (4)     & -140.8 (15) & -14.9 (11)  & -169.9 (17) & -50.7 (15)  & (12.3) & 	ClusTop-BiHa-NA   & -179.9 (13) & -24.2 (10)  & -324.4 (15) & -82.8 (14)  & -389.5 (17) & -98.6 (17)  & (14.3) & -319.3 (13)  & -64.1 (10)   & -569.2 (15)  & -189.6 (15)  & -692.0 (17)  & -141.3 (17) & (14.5) \T \B \\
ClusTop-Hash-NA   & -7.1 (2)   & 5.8 (9)    & -14.8 (4)  & 0.3 (11)   & -14.1 (4)  & 2.6 (6)    & (6.0)  & -19.4 (2)   & 2.3 (8)     & -54.9 (5)   & -6.9 (8)    & -47.8 (4)   & 4.4 (5)     & (5.3)  & 	ClusTop-Hash-NA   & -38.4 (2)   & -4.0 (5)    & -116.4 (5)  & -11.3 (7)   & -98.6 (6)   & 13.1 (4)    & (4.8)  & -63.7 (2)    & -11.0 (6)    & -195.2 (7)   & -9.7 (6)     & -148.8 (9)   & 28.2 (4)    & (5.7)  \T \B \\
ClusTop-Noun-NA   & -17.1 (6)  & 10.6 (6)   & -21.4 (8)  & 6.9 (5)    & -22.8 (10) & -0.3 (10)  & (7.5)  & -64.7 (8)   & 2.9 (6)     & -90.5 (10)  & -3.6 (7)    & -97.8 (14)  & -14.1 (10)  & (9.2)  & 	ClusTop-Noun-NA   & -139.1 (8)  & -23.3 (9)   & -208.5 (11) & -31.4 (9)   & -219.9 (15) & -29.6 (11)  & (10.5) & -242.1 (8)   & -63.7 (9)    & -363.5 (11)  & -86.9 (9)    & -384.7 (15)  & -36.4 (10)  & (10.3) \T \B \\
ClusTop-H2VG-NA    & -17.9 (8)  & -7.8 (16)  & -22.8 (10) & -9.3 (15)  & -22.5 (8)  & -7.9 (12)  & (11.5) & -69.3 (9)   & -35.7 (14)  & -87.2 (9)   & -35.5 (14)  & -62.1 (7)   & -20.9 (11)  & (10.7) & 	ClusTop-H2VG-NA    & -160.2 (10) & -77.7 (12)  & -183.4 (9)  & -70.6 (11)  & -75.6 (3)   & -24.8 (10)  & (9.2)  & -292.1 (11)  & -127.5 (11)  & -295.8 (10)  & -107.8 (11)  & -75.6 (3)    & -24.8 (8)   & (9.0)  \T \B \\
ClusTop-H2VW-NA & -9.4 (3)   & 16.6 (3)   & -11.0 (2)  & 18.2 (1)   & -8.9 (2)   & 21.2 (1)   & (2.0)  & -28.0 (3)   & 40.3 (2)    & -32.5 (2)   & 48.4 (1)    & -18.4 (1)   & 48.6 (2)    & (1.8)  & 	ClusTop-H2VW-NA & -49.9 (3)   & 72.7 (2)    & -61.4 (2)   & 96.8 (1)    & -19.3 (1)   & 52.5 (2)    & (1.8)  & -73.4 (3)    & 111.3 (2)    & -95.7 (2)    & 161.4 (1)    & -19.3 (1)    & 52.5 (3)    & (2.0)  \T \B \\
ClusTop-H2VF-NA & -10.5 (4)  & 18.8 (2)   & -10.4 (1)  & 18.0 (2)   & -8.6 (1)   & 20.3 (2)   & (2.0)  & -31.8 (5)   & 45.7 (1)    & -30.7 (1)   & 45.5 (2)    & -19.0 (2)   & 48.7 (1)    & (2.0)  & 	ClusTop-H2VF-NA & -59.5 (4)   & 87.5 (1)    & -60.0 (1)   & 94.2 (2)    & -20.7 (2)   & 54.8 (1)    & (1.8)  & -92.4 (4)    & 145.3 (1)    & -95.0 (1)    & 157.7 (2)    & -20.7 (2)    & 54.8 (2)    & (2.0)  \T \B \\
ClusTop-Word-AH   & -30.9 (17) & -1.6 (14)  & -40.2 (19) & -27.6 (19) & -24.2 (11) & 10.3 (3)   & (13.8) & -137.6 (18) & -57.7 (17)  & -198.3 (21) & -131.1 (20) & -88.5 (12)  & 9.1 (4)     & (15.3) & 	ClusTop-Word-AH   & -311.9 (18) & -172.7 (19) & -480.2 (21) & -306.9 (21) & -177.0 (14) & -14.8 (8)   & (16.8) & -567.7 (17)  & -336.0 (21)  & -903.7 (21)  & -562.7 (21)  & -298.8 (14)  & -50.2 (13)  & (17.8) \T \B \\
ClusTop-Hash-AH   & -6.3 (1)   & 5.4 (10)   & -12.8 (3)  & 0.9 (10)   & -13.1 (3)  & 2.6 (6)    & (5.5)  & -16.2 (1)   & 1.5 (9)     & -47.3 (3)   & -7.3 (9)    & -43.7 (3)   & 2.2 (6)     & (5.2)  & 	ClusTop-Hash-AH   & -32.3 (1)   & -4.2 (6)    & -99.8 (3)   & -16.3 (8)   & -90.7 (5)   & 8.7 (5)     & (4.7)  & -54.8 (1)    & -9.5 (5)     & -169.0 (3)   & -22.2 (7)    & -143.6 (6)   & 21.9 (5)    & (4.5)  \T \B \\
ClusTop-Noun-AH   & -28.7 (14) & -11.4 (18) & -41.8 (20) & -19.8 (18) & -17.6 (5)  & 4.6 (4)    & (13.2) & -132.4 (17) & -72.3 (19)  & -185.6 (20) & -102.5 (18) & -63.8 (8)   & -2.5 (9)    & (15.2) & 	ClusTop-Noun-AH   & -309.9 (17) & -173.6 (20) & -444.0 (20) & -242.1 (19) & -131.8 (9)  & -17.5 (9)   & (15.7) & -575.1 (18)  & -316.4 (18)  & -811.4 (20)  & -425.6 (19)  & -221.7 (12)  & -35.0 (9)   & (16.0) \T \B \\
ClusTop-H2VG-AH    & -17.6 (7)  & -9.1 (17)  & -32.0 (13) & -15.6 (17) & -29.5 (14) & -11.3 (15) & (13.8) & -71.2 (10)  & -41.0 (15)  & -136.3 (14) & -64.7 (17)  & -97.9 (15)  & -33.7 (14)  & (14.2) & 	ClusTop-H2VG-AH    & -166.4 (11) & -96.1 (14)  & -304.6 (14) & -127.6 (17) & -143.0 (12) & -45.3 (13)  & (13.5) & -303.3 (12)  & -166.6 (13)  & -530.9 (14)  & -195.1 (16)  & -147.6 (8)   & -46.9 (12)  & (12.5) \T \B \\
ClusTop-H2VW-AH & -27.2 (13) & -23.8 (20) & -38.7 (18) & -32.1 (20) & -26.6 (13) & -19.2 (21) & (17.5) & -84.0 (13)  & -83.3 (20)  & -133.6 (13) & -113.9 (19) & -87.7 (11)  & -60.7 (16)  & (15.3) & 	ClusTop-H2VW-AH & -167.4 (12) & -166.8 (18) & -281.9 (13) & -237.7 (18) & -116.9 (8)  & -79.1 (15)  & (14.0) & -274.1 (10)  & -268.2 (17)  & -476.9 (12)  & -390.7 (18)  & -122.6 (5)   & -82.5 (15)  & (12.8) \T \B \\
ClusTop-H2VF-AH & -29.0 (15) & -23.8 (21) & -45.1 (21) & -38.4 (21) & -25.7 (12) & -18.3 (20) & (18.3) & -97.3 (14)  & -93.3 (21)  & -166.1 (18) & -144.0 (21) & -85.1 (10)  & -62.9 (17)  & (16.8) & 	ClusTop-H2VF-AH & -201.1 (14) & -197.3 (21) & -350.3 (16) & -299.5 (20) & -134.5 (10) & -95.8 (16)  & (16.2) & -335.0 (14)  & -329.3 (20)  & -591.2 (16)  & -494.6 (20)  & -160.1 (10)  & -115.3 (16) & (16.0) \T \B \\
ClusTop-Word-AM   & -34.4 (19) & 8.3 (7)    & -30.9 (12) & 11.0 (4)   & -37.7 (17) & -14.3 (16) & (12.5) & -146.1 (19) & -5.4 (11)   & -126.8 (12) & 8.1 (4)     & -179.9 (19) & -69.0 (20)  & (14.2) & 	ClusTop-Word-AM   & -326.6 (19) & -57.3 (11)  & -278.7 (12) & -42.5 (10)  & -422.5 (19) & -143.4 (19) & (15.0) & -599.9 (19)  & -155.4 (12)  & -492.7 (13)  & -122.3 (12)  & -766.5 (19)  & -239.6 (20) & (15.8) \T \B \\
ClusTop-Hash-AM   & -19.7 (11) & 11.4 (4)   & -18.5 (5)  & 16.3 (3)   & -33.7 (16) & -7.3 (11)  & (8.3)  & -73.9 (11)  & 8.5 (3)     & -52.9 (4)   & 14.3 (3)    & -153.6 (16) & -30.4 (12)  & (8.2)  & 	ClusTop-Hash-AM   & -152.7 (9)  & -17.5 (8)   & -102.1 (4)  & -1.8 (5)    & -350.0 (16) & -50.1 (14)  & (9.3)  & -262.2 (9)   & -56.7 (8)    & -169.8 (4)   & -26.5 (8)    & -615.2 (16)  & -67.0 (14)  & (9.8)  \T \B \\
ClusTop-Noun-AM   & -11.2 (5)  & 4.8 (12)   & -19.2 (6)  & 0.3 (11)   & -22.6 (9)  & 4.0 (5)    & (8.0)  & -29.9 (4)   & -0.3 (10)   & -70.1 (8)   & -9.0 (10)   & -70.3 (9)   & 11.9 (3)    & (7.3)  & 	ClusTop-Noun-AM   & -60.7 (5)   & -6.0 (7)    & -150.6 (8)  & -10.5 (6)   & -138.9 (11) & 29.7 (3)    & (6.7)  & -99.3 (5)    & -11.4 (7)    & -257.4 (8)   & -2.4 (5)     & -227.4 (13)  & 58.0 (1)    & (6.5)  \T \B \\
ClusTop-H2VG-AM    & -30.9 (17) & -13.9 (19) & -26.7 (11) & -12.0 (16) & -29.5 (14) & -10.9 (13) & (15.0) & -119.1 (15) & -59.3 (18)  & -99.4 (11)  & -43.6 (16)  & -95.7 (13)  & -31.9 (13)  & (14.3) & 	ClusTop-H2VG-AM    & -273.6 (15) & -123.9 (16) & -193.0 (10) & -78.1 (13)  & -152.4 (13) & -43.1 (12)  & (13.2) & -480.9 (15)  & -202.2 (15)  & -289.8 (9)   & -103.9 (10)  & -167.3 (11)  & -46.4 (11)  & (11.8) \T \B \\
ClusTop-H2VW-AM & -18.7 (9)  & 3.7 (13)   & -21.6 (9)  & 3.1 (8)    & -19.9 (7)  & 0.7 (9)    & (9.2)  & -53.3 (6)   & 2.7 (7)     & -69.6 (7)   & 6.9 (5)     & -58.8 (6)   & 0.3 (8)     & (6.5)  & 	ClusTop-H2VW-AM & -95.4 (6)   & 4.1 (4)     & -128.1 (7)  & 25.2 (3)    & -89.3 (4)   & 5.5 (6)     & (5.0)  & -146.2 (6)   & 9.5 (4)      & -192.2 (6)   & 53.3 (3)     & -102.6 (4)   & 12.2 (7)    & (5.0)  \T \B \\
ClusTop-H2VF-AM & -19.6 (10) & 5.2 (11)   & -20.2 (7)  & 5.0 (6)    & -19.6 (6)  & 1.9 (8)    & (8.0)  & -54.0 (7)   & 3.4 (5)     & -65.2 (6)   & 5.4 (6)     & -58.2 (5)   & 0.4 (7)     & (6.0)  & 	ClusTop-H2VF-AM & -96.2 (7)   & 4.3 (3)     & -122.6 (6)  & 14.9 (4)    & -108.3 (7)  & 4.1 (7)     & (5.7)  & -149.5 (7)   & 11.3 (3)     & -185.6 (5)   & 30.2 (4)     & -146.6 (7)   & 13.0 (6)    & (5.3)  \T \B \\
LDA-Orig          & -74.7 (24) & -74.4 (24) & -66.9 (24) & -62.2 (24) & -54.2 (24) & -43.1 (24) & (24.0) & -323.5 (24) & -307.5 (24) & -297.1 (24) & -269.2 (24) & -251.3 (24) & -191.9 (24) & (24.0) & 	LDA-Orig          & -722.5 (24) & -659.1 (24) & -695.0 (24) & -619.0 (24) & -593.5 (24) & -442.7 (24) & (24.0) & -1279.6 (24) & -1133.6 (24) & -1262.1 (24) & -1102.0 (24) & -1083.2 (24) & -790.0 (24) & (24.0) \T \B \\
LDA-Hash          & -51.2 (22) & -43.8 (22) & -55.1 (23) & -42.9 (22) & -41.5 (20) & -23.4 (22) & (21.8) & -247.1 (22) & -185.4 (22) & -256.8 (22) & -199.2 (22) & -206.6 (22) & -112.5 (22) & (22.0) & 	LDA-Hash          & -607.9 (22) & -435.4 (22) & -611.4 (22) & -449.2 (22) & -500.7 (22) & -271.5 (22) & (22.0) & -1134.2 (22) & -794.2 (22)  & -1120.9 (22) & -805.6 (22)  & -930.7 (22)  & -463.1 (22) & (22.0) \T \B \\
LDA-Ment          & -52.8 (23) & -45.9 (23) & -54.1 (22) & -45.3 (23) & -47.3 (23) & -27.7 (23) & (22.8) & -250.2 (23) & -198.8 (23) & -258.6 (23) & -206.5 (23) & -225.5 (23) & -136.4 (23) & (23.0) & LDA-Ment          & -611.6 (23) & -465.7 (23) & -613.9 (23) & -477.7 (23) & -540.9 (23) & -316.9 (23) & (23.0) & -1141.5 (23) & -850.2 (23)  & -1130.6 (23) & -855.9 (23)  & -981.9 (23)  & -566.6 (23) & (23.0) \T \B  \\ \hline
\end{tabular}	
\end{table*}

\subsection{Topic Coherence and Pointwise Mutual Information}

Table~\ref{resultsTCPMI-summary} shows a summary of the performance of the refined ClusTop algorithm and its variants against the various LDA baselines, in terms of average rank based on Topic Coherence and Pointwise Mutual Information scores on the top 5, 10, 15 and 20 keywords in the detected topics.
For a more detailed breakdown, Tables~\ref{resultsTCPMI-top5n10} and~\ref{resultsTCPMI-top15n20} show the performance of the ClusTop algorithm and its variants against the various LDA baselines, in terms of Topic Coherence and Pointwise Mutual Information, based on the top 5, 10, 15 and 20 keywords in the detected topics.

The results generally show that all variants of our ClusTop algorithm outperform the various LDA baselines, in terms of the average rank metrics. All ClusTop variants also out-perform the LDA baselines in terms of the individual evaluation metrics of Topic Coherence and Pointwise Mutual Information across all datasets.

\begin{table*} [!t]
\caption{Comparison of ClusTop algorithm against various baselines, in terms of Precision (Pre), Recall (Rec) and F-score (FS) for the top 5 keywords/unigrams of each topic. The rank of an algorithm's performance for each metric are provided in brackets.}
\label{resultsPreRecF1-summary}
\centering
\begin{tabular}{r H H H H H H H H H c H H H H H H H H H H c H H H H H H H H H H c H H H H H H H H H H c} \hline																								
{\bf Algorithm} & {Precision} & {Recall} & {F-score} & {Precision} & {Recall} & {F-score} & {Precision} & {Recall} & {F-score} & {\bf Rank@5}   & 	{\bf Algorithm} & {Precision} & {Recall} & {F-score} & {Precision} & {Recall} & {F-score} & {Precision} & {Recall} & {F-score} & {\bf Rank@10}   & 	{\bf Algorithm} & {Precision} & {Recall} & {F-score} & {Precision} & {Recall} & {F-score} & {Precision} & {Recall} & {F-score} & {\bf Rank@15}   & 	{\bf Algorithm} & {Precision} & {Recall} & {F-score} & {Precision} & {Recall} & {F-score} & {Precision} & {Recall} & {F-score} & {\bf Rank@20}  \B \T \B \\
\hline																					
\end{tabular}
\end{table*}

\subsection{Precision, Recall and F-score}

Table~\ref{resultsPreRecF1-summary} shows the average ranks based on the Precision, Recall and F-score scores of the ClusTop algorithm and variants, and the various LDA baselines based on the top 5, 10, 15 and 20 keywords of detected topics. 
For a more detailed breakdown of the results, Table~\ref{resultsPreRecF1Top5} shows the Precision, Recall and F-score of our ClusTop algorithm and variants, and the various LDA baselines based on the top 5 keywords of detected topics, while Tables~\ref{resultsPreRecF1Top10},~\ref{resultsPreRecF1Top15}, and ~\ref{resultsPreRecF1Top20} show the same results based on top 10, 15 and 20 keywords of detected topics, respectively. 

\subsection{Discussion}

We observe that the word embedding variants of ClusTop outperform the LDA baselines and the basic co-occurrence variants of ClusTop, in terms of average rank of topic coherence and pointwise mutual information.


The Hash2Vec variants consistently fosters the best performers (except one case in Table~\ref{resultsPreRecF1Top5} where LDA-Orig is ranked 2nd, beating all  Hash2vec variants except ClusTop-H2VW-AH). The next two best performers are ClusTop-Hash-NA and ClusTop-Noun-AM, except for the aforementioned case where they are beaten by LDA-Orig Table~\ref{resultsPreRecF1Top5}.
ClusTop-H2VW-* and ClusTop-H2VF-* have slightly higher precision than ClusTop-H2VG-*, observed from that ClusTop-H2VG-* never scores top precision in Table~\ref{resultsPreRecF1Top5}. The existence of this slight difference is due to each word embedding algorithm being trained using its own vocabulary set. While GloVe (twitter.27B) provides embeddings for some hashtags that are not standard English, the other two algorithms do not. Missing these hashtags possibly increases precision of ClusTop-H2VW-* and ClusTop-H2VF-* in Table~\ref{resultsPreRecF1Top5}, as vast majority of ground truth words are in standard English vocabulary.

Our observations show that ClusTop is an adaptive algorithm that can be improved with different similarity measures, and this information can be helpful when performing downstream tasks, such as topic modelling.

I here answer the question raised in section~\ref{sectIntro} that a more proper similarity scheme can indeed improve the performance of a community detection method in topic modelling. It is likely that further improving the similarity schemes may be helpful in many other applications as well. 



\section{Related Work}
\label{sectRelatedWork}

In this section, we discuss two main areas of research related to our work, namely the study of topics on microblogs and general topic modelling algorithms.

\subsection{Studying Topics using Communities}

The most closely related works to our proposed approach are those that make use of community detection techniques for understanding and studying topics on microblogs. As such, we discuss main works that utilise such techniques. Towards the effort to better understand research themes in the Human Computer Interaction domain, Liu et al.~\citep{liu2014chi} used hierarchical clustering on co-keywords usage in academic papers to identify the main research clusters across two different time periods. Researchers have also proposed approaches for identifying communities that frequently interact about common interest topics using various types of community detection algorithms. These approaches are based on topological links such as friendship networks among users and celebrities~\citep{lim2012topological} and interaction links in the form of explicit mentions of other users~\citep{lim2016interaction}. Researchers like~\citep{paranyushkin2011identifying} and~\citep{barbon2017artificial} have also used community detection algorithms on word networks to identify topics with a focus on network analysis and visualization, and detection of spammer topics, respectively. Fried et al.~\citep{fried2014analyzing} used topic modelling on a series of food-related tweets to understand health information such as overweight rate and diabetes rate. Others like Surian et al.~\citep{surian2016characterizing} and \citep{amati2021topic} combined the use of topic modelling algorithms with community detection algorithms to characterise discussions relating to vaccines on Twitter, study discussion topics of Italian users, respectively.

\subsection{Social Media Analytics}

\section{Detailed Results on Topic Coherence and Pointwise Mutual Information}
\begin{landscape}
\begin{table*} [!t]
\caption{Comparison of ClusTop algorithm against various baselines, in terms of Topic Coherence (TC) and Pointwise Mutual Information (PMI) for the top 5 and 10 keywords. The rank of an algorithm's performance for each metric are provided in brackets.}
\label{resultsTCPMI-top5n10}
\centering
\resizebox{\columnwidth}{!} {
\begin{tabular}{r c c c c c c c c c c c c c c} \hline
{\bf }	&	\multicolumn{7}{c}{\bf Top 5 Keywords / Unigrams}	&	\multicolumn{7}{c}{\bf Top 10 Keywords / Unigrams}	\T \B \T \B \\
{\bf }	&	\multicolumn{2}{c}{\bf Dataset A}	&	\multicolumn{2}{c}{\bf Dataset B}	&	\multicolumn{2}{c}{\bf Dataset C} & {\bf Average}	&	\multicolumn{2}{c}{\bf Dataset A}	&	\multicolumn{2}{c}{\bf Dataset B}	&	\multicolumn{2}{c}{\bf Dataset C} & {\bf Average}	\T \B \T \B \\
{\bf Algorithm}&	{TC}	&	{PMI}	&	{TC}	&	{PMI}	&	{TC}	&	{PMI} & {\bf Rank@5}	&	{TC}	&	{PMI}	&	{TC}	&	{PMI}	&	{TC}	&	{PMI} & {\bf Rank@10}	\B \T \B \\ \hline
ClusTop-Word-NA   & -37.6 (21) & -5.5 (15)  & -34.1 (15) & -7.7 (14)  & -37.9 (18) & -14.4 (17) & (16.7) & -171.0 (21) & -49.2 (16)  & -160.8 (17) & -39.5 (15)  & -173.4 (18) & -67.5 (19)  & (17.7) \T \B \\
ClusTop-BiG-NA    & -36.6 (20) & 7.3 (8)    & -35.9 (17) & 1.2 (9)    & -42.5 (22) & -16.4 (18) & (15.7) & -153.4 (20) & -29.6 (13)  & -158.2 (16) & -25.8 (13)  & -194.8 (21) & -63.4 (18)  & (16.8) \T \B \\
ClusTop-TriG-NA   & -30.9 (17) & 10.7 (5)   & -35.8 (16) & -2.6 (13)  & -42.0 (21) & -18.2 (19) & (15.2) & -122.6 (16) & -16.1 (12)  & -166.5 (19) & -25.1 (12)  & -194.2 (20) & -73.5 (21)  & (16.7) \T \B \\
ClusTop-BiHa-NA   & -23.3 (12) & 19.6 (1)   & -32.3 (14) & 4.7 (7)    & -37.9 (18) & -11.2 (14) & (11.0) & -81.4 (12)  & 7.1 (4)     & -140.8 (15) & -14.9 (11)  & -169.9 (17) & -50.7 (15)  & (12.3) \T \B \\
ClusTop-Hash-NA   & -7.1 (2)   & 5.8 (9)    & -14.8 (4)  & 0.3 (11)   & -14.1 (4)  & 2.6 (6)    & (6.0)  & -19.4 (2)   & 2.3 (8)     & -54.9 (5)   & -6.9 (8)    & -47.8 (4)   & 4.4 (5)     & (5.3)  \T \B \\
\hline
\end{tabular}
}
\end{table*}
\end{landscape}

\begin{landscape}
\begin{table*} [!t]
\caption{Comparison of ClusTop algorithm against various baselines, in terms of Topic Coherence (TC) and Pointwise Mutual Information (PMI) for the top 15 and 20 keywords. The rank of an algorithm's performance for each metric are provided in brackets.}
\label{resultsTCPMI-top15n20}
\centering
\resizebox{\columnwidth}{!} {
\begin{tabular}{r c c c c c c c c c c c c c c} \hline
{\bf }	&	\multicolumn{7}{c}{\bf Top 15 Keywords / Unigrams}	&	\multicolumn{7}{c}{\bf Top 20 Keywords / Unigrams}	\T \B \T \B \\
{\bf }	&	\multicolumn{2}{c}{\bf Dataset A}	&	\multicolumn{2}{c}{\bf Dataset B}	&	\multicolumn{2}{c}{\bf Dataset C} & {\bf Average}	&	\multicolumn{2}{c}{\bf Dataset A}	&	\multicolumn{2}{c}{\bf Dataset B}	&	\multicolumn{2}{c}{\bf Dataset C} & {\bf Average}	\T \B \T \B \\
{\bf Algorithm}&	{TC}	&	{PMI}	&	{TC}	&	{PMI}	&	{TC}	&	{PMI} & {\bf Rank@15}	&	{TC}	&	{PMI}	&	{TC}	&	{PMI}	&	{TC}	&	{PMI} & {\bf Rank@20}	\B \T \B \\ \hline
ClusTop-Word-NA   & -409.7 (21) & -157.4 (17) & -382.6 (19) & -115.5 (16) & -403.5 (18) & -147.6 (21) & (18.7) & -768.2 (21)  & -318.1 (19)  & -695.1 (19)  & -234.8 (17)  & -732.7 (18)  & -247.8 (21) & (19.2) \T \B \\
ClusTop-BiG-NA    & -357.2 (20) & -108.3 (15) & -360.3 (17) & -77.2 (12)  & -447.9 (20) & -116.5 (18) & (17.0) & -645.5 (20)  & -212.3 (16)  & -633.7 (17)  & -147.8 (13)  & -794.7 (20)  & -165.4 (18) & (17.3) \T \B \\
ClusTop-TriG-NA   & -289.2 (16) & -83.1 (13)  & -382.0 (18) & -86.2 (15)  & -451.6 (21) & -145.4 (20) & (17.2) & -530.2 (16)  & -170.0 (14)  & -674.8 (18)  & -183.1 (14)  & -804.0 (21)  & -222.8 (19) & (17.0) \T \B \\
ClusTop-BiHa-NA   & -179.9 (13) & -24.2 (10)  & -324.4 (15) & -82.8 (14)  & -389.5 (17) & -98.6 (17)  & (14.3) & -319.3 (13)  & -64.1 (10)   & -569.2 (15)  & -189.6 (15)  & -692.0 (17)  & -141.3 (17) & (14.5) \T \B \\
ClusTop-Hash-NA   & -38.4 (2)   & -4.0 (5)    & -116.4 (5)  & -11.3 (7)   & -98.6 (6)   & 13.1 (4)    & (4.8)  & -63.7 (2)    & -11.0 (6)    & -195.2 (7)   & -9.7 (6)     & -148.8 (9)   & 28.2 (4)    & (5.7)  \T \B \\
\hline
\end{tabular}
}
\end{table*}
\end{landscape}

\section{Detailed Results on Precision, Recall and F-score}

\begin{landscape}
\begin{table*} [!t]
\caption{Comparison of ClusTop algorithm against various baselines, in terms of Precision (Pre), Recall (Rec) and F-score (FS) for the top 5 keywords/unigrams of each topic. The rank of an algorithm's performance for each metric are provided in brackets.}
\label{resultsPreRecF1Top5}
\centering
\resizebox{\columnwidth}{!} {
\begin{tabular}{r c c c c c c c c c c} \hline
{\bf }	&	\multicolumn{10}{c}{\bf Top 5 Keywords / Unigrams}	\T \B \T \B \\
{\bf }	&	\multicolumn{3}{c}{\bf Dataset A}	&	\multicolumn{3}{c}{\bf Dataset B}	&	\multicolumn{3}{c}{\bf Dataset C}	&	{\bf Average}	\T \B \T \B \\
{\bf Algorithm}	&	{Precision}	&	{Recall}	&	{F-score}	&	{Precision}	&	{Recall}	&	{F-score}	&	{Precision}	&	{Recall}	&	{F-score}	&	{\bf Rank@5} 	\B \T \B \\ \hline
ClusTop-Word-NA   & .754$\pm$.002	(22) & .031$\pm$.000	(6)  & .059$\pm$.000	(6)  & .866$\pm$.003	(21) & .043$\pm$.000	(5)  & .082$\pm$.001	(5)  & .840$\pm$.009	(18) & .027$\pm$.001	(7)  & .052$\pm$.001	(7)  & (10.8) \T \B \\
ClusTop-BiG-NA    & .786$\pm$.002	(18) & .034$\pm$.000	(2)  & .064$\pm$.000	(2)  & .857$\pm$.003	(22) & .046$\pm$.000	(2)  & .086$\pm$.001	(2)  & .833$\pm$.010	(20) & .029$\pm$.001	(4)  & .056$\pm$.001	(3)  & (8.3)  \T \B \\
ClusTop-TriG-NA   & .791$\pm$.002	(17) & .034$\pm$.000	(2)  & .064$\pm$.000	(2)  & .871$\pm$.003	(20) & .046$\pm$.000	(2)  & .087$\pm$.001	(1)  & .822$\pm$.009	(21) & .031$\pm$.001	(2)  & .058$\pm$.001	(2)  & (7.7)  \T \B \\
ClusTop-BiHa-NA   & .784$\pm$.002	(19) & .032$\pm$.000	(4)  & .060$\pm$.000	(4)  & .886$\pm$.003	(18) & .045$\pm$.000	(4)  & .084$\pm$.001	(4)  & .820$\pm$.009	(22) & .029$\pm$.001	(4)  & .055$\pm$.001	(5)  & (9.3)  \T \B \\
ClusTop-Hash-NA   & .898$\pm$.004	(10) & .023$\pm$.000	(21) & .044$\pm$.000	(18) & .916$\pm$.007	(13) & .032$\pm$.001	(17) & .062$\pm$.001	(17) & .936$\pm$.011	(8)  & .022$\pm$.000	(17) & .042$\pm$.001	(18) & (15.4) \T \B \\
\hline
\end{tabular}
}
\end{table*}
\end{landscape}

\begin{landscape}
\begin{table*} [!t]
\caption{Comparison of ClusTop algorithm against various baselines, in terms of Precision (Pre), Recall (Rec) and F-score (FS) for the top 10 keywords/unigrams of each topic. The rank of an algorithm's performance for each metric are provided in brackets.}
\label{resultsPreRecF1Top10}
\centering
\resizebox{\columnwidth}{!} {
\begin{tabular}{r c c c c c c c c c c} \hline
{\bf }	&	\multicolumn{10}{c}{\bf Top 10 Keywords / Unigrams}	\T \B \T \B \\
{\bf }	&	\multicolumn{3}{c}{\bf Dataset A}	&	\multicolumn{3}{c}{\bf Dataset B}	&	\multicolumn{3}{c}{\bf Dataset C}	&	{\bf Average}	\T \B \T \B \\
{\bf Algorithm}	&	{Precision}	&	{Recall}	&	{F-score}	&	{Precision}	&	{Recall}	&	{F-score}	&	{Precision}	&	{Recall}	&	{F-score}	&	{\bf Rank@10} 	\B \T \B \\ \hline
ClusTop-Word-NA   & .690$\pm$.002	(24) & .033$\pm$.000	(7)  & .062$\pm$.000	(7)  & .804$\pm$.003	(20) & .051$\pm$.000	(3)  & .095$\pm$.001	(3)  & .764$\pm$.010	(22) & .033$\pm$.001	(6)  & .062$\pm$.001	(6)  & (10.9) \T \B \\
ClusTop-BiG-NA    & .707$\pm$.002	(22) & .035$\pm$.000	(2)  & .065$\pm$.000	(2)  & .789$\pm$.004	(22) & .052$\pm$.000	(2)  & .096$\pm$.001	(2)  & .765$\pm$.010	(20) & .035$\pm$.001	(3)  & .067$\pm$.002	(2)  & (8.6)  \T \B \\
ClusTop-TriG-NA   & .717$\pm$.002	(21) & .034$\pm$.000	(4)  & .064$\pm$.000	(4)  & .811$\pm$.003	(19) & .053$\pm$.000	(1)  & .098$\pm$.001	(1)  & .746$\pm$.010	(23) & .036$\pm$.001	(1)  & .068$\pm$.001	(1)  & (8.3)  \T \B \\
ClusTop-BiHa-NA   & .719$\pm$.002	(20) & .034$\pm$.000	(4)  & .064$\pm$.000	(4)  & .782$\pm$.004	(24) & .051$\pm$.000	(3)  & .095$\pm$.001	(3)  & .765$\pm$.010	(20) & .035$\pm$.001	(3)  & .065$\pm$.001	(3)  & (9.3)  \T \B \\
ClusTop-Hash-NA   & .860$\pm$.004	(8)  & .023$\pm$.000	(24) & .045$\pm$.000	(17) & .896$\pm$.008	(9)  & .032$\pm$.001	(17) & .061$\pm$.001	(17) & .925$\pm$.012	(8)  & .022$\pm$.000	(18) & .043$\pm$.001	(18) & (15.1) \T \B \\
\hline
\end{tabular}
}
\end{table*}
\end{landscape}

\begin{landscape}
\begin{table*} [!t]
\caption{Comparison of ClusTop algorithm against various baselines, in terms of Precision (Pre), Recall (Rec) and F-score (FS) for the top 15 keywords/unigrams of each topic. The rank of an algorithm's performance for each metric are provided in brackets.}
\label{resultsPreRecF1Top15}
\centering
\resizebox{\columnwidth}{!} {
\begin{tabular}{r c c c c c c c c c c} \hline
{\bf }	&	\multicolumn{10}{c}{\bf Top 15 Keywords / Unigrams}	\T \B \T \B \\
{\bf }	&	\multicolumn{3}{c}{\bf Dataset A}	&	\multicolumn{3}{c}{\bf Dataset B}	&	\multicolumn{3}{c}{\bf Dataset C}	&	{\bf Average}	\T \B \T \B \\
{\bf Algorithm}	&	{Precision}	&	{Recall}	&	{F-score}	&	{Precision}	&	{Recall}	&	{F-score}	&	{Precision}	&	{Recall}	&	{F-score}	&	{\bf Rank@15} 	\B \T \B \\ \hline
ClusTop-Word-NA   & .648$\pm$.002	(24) & .034$\pm$.000	(5)  & .063$\pm$.000	(7)  & .743$\pm$.004	(23) & .053$\pm$.000	(3)  & .098$\pm$.001	(3)  & .717$\pm$.010	(22) & .038$\pm$.001	(4)  & .070$\pm$.002	(4)  & (10.6) \T \B \\
ClusTop-BiG-NA    & .671$\pm$.002	(21) & .035$\pm$.000	(2)  & .065$\pm$.000	(2)  & .747$\pm$.004	(21) & .055$\pm$.001	(1)  & .100$\pm$.001	(1)  & .719$\pm$.010	(21) & .040$\pm$.001	(2)  & .073$\pm$.002	(2)  & (8.1)  \T \B \\
ClusTop-TriG-NA   & .667$\pm$.002	(22) & .035$\pm$.000	(2)  & .065$\pm$.000	(2)  & .757$\pm$.004	(20) & .055$\pm$.000	(1)  & .100$\pm$.001	(1)  & .677$\pm$.010	(24) & .041$\pm$.001	(1)  & .075$\pm$.002	(1)  & (8.2)  \T \B \\
ClusTop-BiHa-NA   & .667$\pm$.002	(22) & .034$\pm$.000	(5)  & .064$\pm$.000	(5)  & .729$\pm$.004	(24) & .052$\pm$.000	(4)  & .096$\pm$.001	(4)  & .722$\pm$.009	(20) & .040$\pm$.001	(2)  & .073$\pm$.002	(2)  & (9.8)  \T \B \\
ClusTop-Hash-NA   & .848$\pm$.004	(9)  & .023$\pm$.000	(24) & .045$\pm$.000	(18) & .880$\pm$.008	(9)  & .031$\pm$.001	(17) & .060$\pm$.001	(17) & .914$\pm$.012	(8)  & .023$\pm$.001	(17) & .044$\pm$.001	(17) & (15.1) \T \B \\
\hline
\end{tabular}
}
\end{table*}
\end{landscape}

\begin{landscape}
\begin{table*} [!t]
\caption{Comparison of ClusTop algorithm against various baselines, in terms of Precision (Pre), Recall (Rec) and F-score (FS) for the top 20 keywords/unigrams of each topic. The rank of an algorithm's performance for each metric are provided in brackets.}
\label{resultsPreRecF1Top20}
\centering
\resizebox{\columnwidth}{!} {
\begin{tabular}{r c c c c c c c c c c} \hline
{\bf }	&	\multicolumn{10}{c}{\bf Top 20 Keywords / Unigrams}	\T \B \T \B \\
{\bf }	&	\multicolumn{3}{c}{\bf Dataset A}	&	\multicolumn{3}{c}{\bf Dataset B}	&	\multicolumn{3}{c}{\bf Dataset C}	&	{\bf Average}	\T \B \T \B \\
{\bf Algorithm}	&	{Precision}	&	{Recall}	&	{F-score}	&	{Precision}	&	{Recall}	&	{F-score}	&	{Precision}	&	{Recall}	&	{F-score}	&	{\bf Rank@20} 	\B \T \B \\ \hline
ClusTop-Word-NA   & .617$\pm$.002	(24) & .034$\pm$.000	(6)  & .064$\pm$.000	(6)  & .709$\pm$.003	(23) & .055$\pm$.000	(2)  & .100$\pm$.001	(3)  & .669$\pm$.009	(22) & .038$\pm$.001	(4)  & .070$\pm$.002	(4)  & (10.4) \T \B \\
ClusTop-BiG-NA    & .646$\pm$.002	(21) & .035$\pm$.000	(3)  & .065$\pm$.000	(3)  & .722$\pm$.004	(22) & .056$\pm$.001	(1)  & .102$\pm$.001	(1)  & .679$\pm$.010	(20) & .040$\pm$.001	(2)  & .073$\pm$.002	(2)  & (8.3)  \T \B \\
ClusTop-TriG-NA   & .636$\pm$.002	(22) & .035$\pm$.000	(3)  & .065$\pm$.000	(3)  & .726$\pm$.004	(20) & .055$\pm$.001	(2)  & .101$\pm$.001	(2)  & .662$\pm$.009	(24) & .041$\pm$.001	(1)  & .075$\pm$.002	(1)  & (8.7)  \T \B \\
ClusTop-BiHa-NA   & .635$\pm$.002	(23) & .035$\pm$.000	(3)  & .065$\pm$.000	(3)  & .702$\pm$.004	(24) & .054$\pm$.000	(4)  & .099$\pm$.001	(4)  & .679$\pm$.009	(20) & .040$\pm$.001	(2)  & .073$\pm$.002	(2)  & (9.4)  \T \B \\
ClusTop-Hash-NA   & .831$\pm$.004	(9)  & .024$\pm$.000	(23) & .045$\pm$.000	(18) & .873$\pm$.008	(9)  & .031$\pm$.001	(17) & .060$\pm$.001	(17) & .911$\pm$.012	(8)  & .023$\pm$.001	(17) & .044$\pm$.001	(17) & (15.0) \T \B \\
\hline
\end{tabular}
}
\end{table*}
\end{landscape}

\section{Applications}

Topic models are relatively new, and their application to analyzing large repositories of information is currently being studied. Examples of their practical application have been found in cases such as investigating historical topics in large volumes of text~\citep{blevins_2010} or conducting root cause analysis on the New York Times Opinion page on the Civil War~\citep{nelson_2011}. The proposed topic modelling algorithm has shown promising result in detecting major themes and events in a given text.

Moreover, one of the most recent and popular application for topic modelling is Recommendation System. A user would tend to view or click content under the same or similar topic. For example, an effective topic modelling system can help Netflix in recommending movies that users are interested in based on users' browsing history, so as to increase revenue and users' online time to occupy the market.

\section{Conclusion}
\label{sectConclusion}

In this chapter, I refine the ClusTop algorithm for topic modelling on Twitter, using more coherent edge weights in the graph containing tweets as nodes. The basic ClusTop algorithm addresses the problem that traditional topic modelling algorithms require the tuning and setting of numerous parameter. ClusTop does not require this parameter tuning and is able to automatically determine the appropriate number of topics using a local maximization of modularity among the word network graph, where the edge weights are usually word co-occurrence frequency. With edges weight adapted into word embedding similarity scores, the refined variants of our ClusTop algorithm, which we use to compare among the variants as well as against various LDA baselines, exhibits better  topic quality. The experimental results based on three Twitter datasets with labelled topics (crises and events) show that our ClusTop algorithm can be refined by newer and more appropriate similarity schems to outperform the various LDA baselines and its basic variants.

\chapter{Being Similar Versus Paraphrasing} 
\label{Chapter4} 

\section{Paraphrasing, Special Requirements for Semantic Similarity}

Paraphrase is an important part of language since the Renaissance, which helps to gain understanding. A proper paraphrase may exist in different forms, such as changing the original wording in styles, tightening up overly verbose part, speaking more gently, using auspicious words, or even using ominous words~\citep{liddell1894greek}. More precisely, a paraphrase of something written or spoken is the same thing expressed in a different way\footnote{\url{https://www.collinsdictionary.com/dictionary/english/paraphrase}}.


The first question studied in this chapter is what kind of relationship exists between a pair of paraphrases and a pair of similar expressions. In general, we can assume that if the similarity between two units in a certain dimension is infinitely close to the upper limit of similarity, we can say that the two units are the same in this dimension. Linguistically, units with the same meaning are called paraphrases (or synonyms in the case of words)

Rigorous quantification of expression similarity is a regression task. Recognising paraphrases, in contrast, is usually regarded as a binary classification task, usually defined as follows:

\begin{equation}\label{eq:paraphrase_recognition}
f: \mathcal{U}, \mathcal{U} \to \set{0, 1}
\end{equation}

where $\mathcal{U}$ is a set containing all possible linguistic units, and the output is whether a pair of units are paraphrasing each other. Because of the difference in problem definition between similarity quantification and paraphrase, the two tasks are sometimes considered to be independent tasks~\citep{bar2015composing}. However, a closer look at the formulations will lead us to the conclusion that paraphrase recognition can be considered as applying a threshold function on the output of a similarity function, as shown below:

\begin{equation}
    f(u_1, u_2) = 
    \begin{cases}
    0, &s(u_1, u_2) < T,\\
    1, &s(u_1, u_2) \geq T,
    \end{cases}
\end{equation}

where $T$ is a real-value constant. Given a threshold $T$, paraphrase recognition may be a suitable task to verify the effectiveness of similarity schemes. For example, if a pair of units that humans consider to be paraphrases end up getting a similarity score from a scheme below a threshold, we shall know there is something wrong with that scheme.

For this, I use a set of paraphrase pairs with different level of syntactic similarity. The experiment is to check whether syntactic similarity interferes with the calculation of semantic similarity. This set of paraphrasing pairs is called Concise-536, details of which are given in section~\ref{sectDataset}.

This experiment can be thought of as a stress test for similar schemes. The results of the stress test show that for most schemes, syntactic or lexical similarity strongly affects the similarity calculation, even for schemes proposed for semantic similarity, such as BERTScore~\citep{zhang2019bertscore}. A pair of paraphrases that use very different words tends to receive a lower score than a pair of paraphrases that use the same words. The result is shown in Table~\ref{tab:results}.

\section{Paraphrase Recognition and Paraphrase Generation}

Currently, there are two NLP tasks closely related to paraphrasing, one is the paraphrase recognition task defined above, and the other is paraphrase generation, which is a conditional natural language generation task. In addition to testing similarity, I extend this research slightly to paraphrase generation through a task called "revising for concision".

Paraphrase generation is unofficially regarded as the task to generate an output sentence which is semantically identical to the input sentence but contains variations in lexicon or syntax\footnote{\url{http://nlpprogress.com/english/paraphrase-generation.html}}. Formally,

\begin{equation}
    g: \mathcal{U} \to \mathcal{U}
\end{equation}

Paraphrase generation is closely related to paraphrase recognition, as the efficacy of function $g$ is usually determined by the function value from $f$. In other words, these two tasks can only be solved together, and solving one is equivalent to solving the other~\citep{iyyer-etal-2018-adversarial}

However, paraphrase generation is essentially not a function, or at least not a well-posed function. Its input can expecting multiple equally plausible outputs. For units $u_1, u_2$, and $u_3$ that are paraphrases to each other, one may expect that either $u_2$ or $u_3$ to be a valid output if $u_1$ is an input. Analogous to that a circle cannot be defined through a explicit function, paraphrase generation is a function $g$ that satisfies the following condition:

\begin{equation}
\begin{aligned}
    f\big(u, g(u)\big) &= 1, \text{ zor}\\
    s\big(u, g(u)\big) &\geq T.
\end{aligned}
\end{equation}

\section{Revision for Concision: A Constrained Paraphrase Generation Task}

Academic writing should be concise as concise sentences better keep the readers' attention and convey meaning clearly. Writing concisely is challenging, for writers often struggle to revise their drafts. We introduce and formulate revising for concision as a natural language processing task at the sentence level. Revising for concision requires algorithms to use only necessary words to rewrite a sentence while preserving its meaning. The revised sentence should be evaluated according to its word choice, sentence structure, and organization. The revised sentence also needs to fulfil semantic retention and syntactic soundness. To aide these efforts, we curate and make available a benchmark parallel dataset that can depict revising for concision. The dataset contains 536 pairs of sentences before and after revising, and all pairs are collected from college writing centres. We also present and evaluate methods that address this task, which may assist researchers in this area.

Concision and clarity\footnote{We treat \textit{concision} and \textit{conciseness} as equivalent, and \textit{clarity} as part of \textit{concision}} are important in academic writing as wordy sentences will obscure good ideas (Figure~\ref{fig:obscure}). Concise writing encourages writers to choose words deliberately and precisely, construct sentences carefully to eliminate deadword, and use grammar properly~\citep{stanford2021}, which often requires experience and time. A first draft often contains far more words than necessary, and achieving concise writing requires revisions~\citep{monash2020}. As far as we know, currently this revision process can only be done manually, or semi-manually with the help of some rule-based wordiness detectors~\citep{long_2013}. We therefore introduce and formulate revising for concision as a natural language processing (NLP) task and address it. In this study, we make the following contributions:

\begin{figure}[t!]
	\centering
	\includegraphics[width=0.7\textwidth]{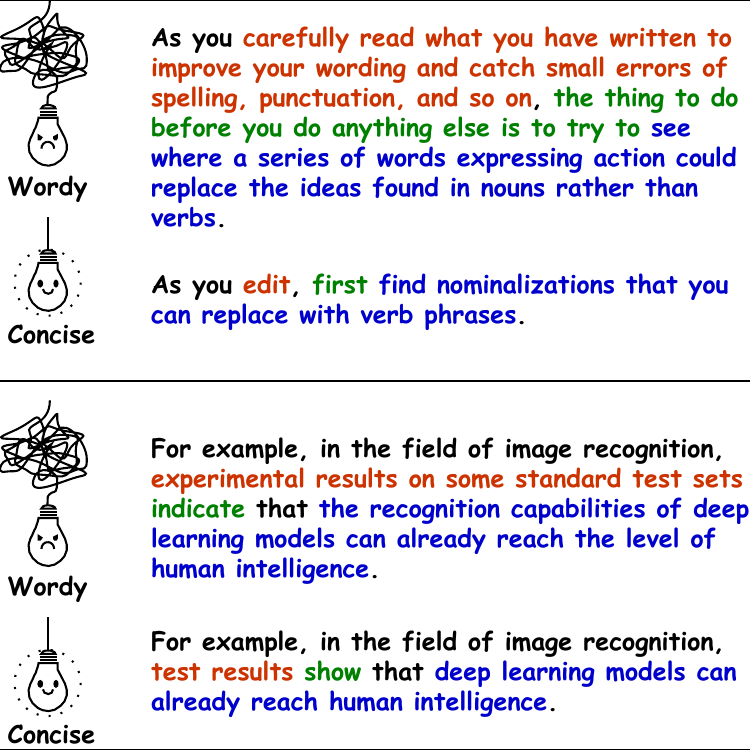}
	\caption{Wordy sentences are more boring to read than concise sentences. But how do we turn lengthy sentences into concise ones? We show two examples. The above sentence pair is taken from the Purdue Writing Lab, which suggests how college students should succinctly revise their writing~\citep{purdue_writing_lab2021}. In the other example, the wordy sentence comes from a scientific paper~\citep{chen2020survey}, and its concise counterpart is predicted from the concise revisioner we developed (Section~\ref{sec:baseline}). In each pair, text with the same colour delivers the same information.}
	\label{fig:obscure}
\end{figure}

\begin{enumerate}
	\item We formulate the revising for concision NLP task at the sentence level, which reflects the revising task in academic writing. We also survey the differences between this task and sentence compression, paraphrasing, etc.
	\item We release a corpus of 536 sentence pairs, curated from 72 writing centres and additionally coded with the various linguistic rules for concise sentence revision.
	\item We propose baseline methods by fine-tuning Seq2Seq models for this task, and conduct automatic and human evaluations. We observed promising preliminary results and we believe that our findings will be useful for researchers working in this area.
\end{enumerate}

\section{Problem Statement}
\label{sec:problem}
\subsection{Revision as an English Writing Task}

Concise writing itself is a lesson that is often emphasized in colleges, and revision is crucial in writing. The following definitions are helpful when we set out to formulate the task.

\begin{definition}[Concise]
Marked by brevity of expression or statement: free from all elaboration and superfluous detail~\citep{merriam_webster_concise}.
\end{definition}

\begin{definition}[Concise writing, English]
Writing that is clear and does not include unnecessary or vague/unclear words or language~\citep{auckland2021}.
\end{definition}

Revising for concision at paragraph level, or even article level, may be the best practice. However, sentence-level revising usually suffices. We focus on revising for concision at the sentence level now. Indeed, in many college academic writing tutorials, revisions for concision are for individual sentences, and this process is defined as follows.

\begin{definition}[Revise for concision at the sentence level, English\footnote{Adapted from notes of~\citet{purdue_writing_lab2021} Writing Lab and~\citet{rambo2019}}]\label{def:eng}
Study a sentence in draft, use specific strategies\footnote{Presented in Appendix (Table~\ref{tab:strategies}) as a periphery of this study.}  to edit the sentence concisely without losing meaning.
\end{definition}

If someone, such as a college student, wants to concisely modify a sentence, specific strategies (\emph{e.g.}, \emph{delete} weak modifiers, \emph{replace} phrasal verbs with single verbs, or \emph{rewrite} in active voice, etc.) tell us how to locate wordiness and how to edit it~\citep{purdue_writing_lab2021,waldenu2021,ualr2021,massey2021,monash2020}. The rule is to repeatedly detect wordiness and revise it until no wordiness is detected or it cannot be removed without adding new wordiness. The final product serves as a concise version of the original sentence, if it does not lose its meaning.

\subsection{Task Definition in NLP}
Now that we know how humans can revise a sentence, what about programs? Each strategy is clear to a trained college student, but not clear enough to program in code. On the one hand, existing verbosity detectors may suggest which part of a sentence is too "dense"~\citep{long_2013}, but fail to expose fine-grained wordiness details. On the other hand, how programs can edit sentences without losing their meaning remains challenging. In short, no existing program can generate well-modified sentences in terms of concision.

Eager for a program that revises sentences nicely and concisely, we set out to formulate this modification process as a sequence-to-sequence (Seq2Seq) NLP task. In this task, the input is any English sentence and the output should be its concise version. We define it as follows.

\begin{definition}[Revise for concision at the sentence level, NLP]\label{def:nlp}
Produce a sentence where minimum wordiness can be identified. (And,) the produced sentence delivers the same information as input does. (And,) the produced sentence is syntactically correct.

\end{definition}

As many other NLP tasks, \emph{e.g.}, machine translation, named-entity recognition, etc., Definition\,\ref{def:nlp} describes the product (text) of a process, not the process itself, \emph{i.e.}, how the text is produced. This perspective is different from that of Definition\,\ref{def:eng}

Among three components in Definition\,\ref{def:nlp}, both the first and the third are clear and self-contained. They are related to syntax; hence, at least human experts would think it straightforward to determine the soundness of a sentence on both.  For example, the syntax correctness of an English sentence will not be judged differently by different experts, unless the syntax itself changes. Unfortunately, the second component is neither clear nor self-contained. This component asks for information retention, which is a rule inherited from Definition\,\ref{def:eng}. Determining the semantic similarity between texts has long been challenging, even for human experts~\citep{rus-etal-2014-paraphrase}. 

We then clarify the definition by assuming that combining the second and third components in Definition\,\ref{def:nlp} meet the definition of the paraphrase generation task~\citep{rus-etal-2014-paraphrase}. Henceforth, Definition\,\ref{def:nlp} can be simplified to Definition\,\ref{def:confined}.

\begin{definition}[Revise for concision at the sentence level, NLP, simplified]\label{def:confined}
Produce a \textit{paraphrase} where minimum wordiness can be identified.
\end{definition}

The revising\footnote{stands for \textit{(machine) revising for concision} if not otherwise specified, so does \textit{revision}} task is well-defined, as long as "paraphrase generation" is well-defined. It is a paraphrase generation task with a syntactic constraint.

\subsection{Task Performance Indicator}\label{sec:indicator}
How does one approximately measure revision performance? In principle, Definition~\ref{def:nlp} should be used as a checklist. A good sample requires correct grammar ($\gamma$), complete information ($\rho$) and reduced wordiness ($1-\omega$), assuming each component as a float number between 0 and 1. The overall assessment ($\chi$) of the three components is as follows,
\begin{equation}
	\chi = \alpha^2\cdot (\gamma - 1) + \alpha\cdot (\rho - 1) + (1-\omega),
\end{equation}
where $\alpha\in\mathbb{R}_{>1}$ is a large enough number, as we believe that $\gamma$ and $\rho$ overweigh $1-\omega$. Intuitively, if a revised sentence does not paraphrase the original one, assessing the reduction of wordiness makes little sense. Concision $\chi$ would always be negative if $\gamma<1$ or $\rho<1$. For a balance among syntax, information, and wordiness, the coefficient $\alpha$ tells how much syntax overweighs information, or information overweighs reduced wordiness. Empirically, minimum of $\alpha$ can be around the word count in a standard sentence. In other words, even if a single key word is missing, the decrease in $\rho$ is bigger than the increase in $1 - \omega$. 

Corresponding to the three componets is a mix of three tasks, including grammatical error correction for $g$, textual semantic similarity for $r$, and wordiness detection for $w$. Unfortunately, both a reference-free metric good enough to characterize the paraphrase and a robust wordiness detector are rare. Therefore, such assessment of concision is now only feasible through human evaluation.

To enable automatic evaluation for faster feedback, we currently follow Papineni's viewpoint~\citep{papineni-etal-2002-bleu}. The closer a machine revision is to a professional human revision, the better it is. To judge the quality of a machine revision, one measures its closeness to one or more reference human revisions according to a numerical metric. Thus, our revising evaluation system requires two main components:
\begin{enumerate}
	\item A numerical "revision closeness" metric.
	\item A corpus of good quality human reference revisions.
\end{enumerate}

Different from days when Papineni needed to propose a closeness metric, we can adopt various metrics from machine translation and summarization community~\citep{lin-2004-rouge,banerjee-lavie-2005-meteor}. Since it is certain which criterion correlates best, we take multiple relevant and reasonable metrics into account to estimate quality of revision. These metrics include those measuring higher order n-grams precision (BLEU,~\citealp{papineni-etal-2002-bleu}), explicit Word-matching, stem-matching, or synonym-matching (METEOR,~\citealp{banerjee-lavie-2005-meteor}), surface bigram units overlapping (ROUGE-2-F1,~\citealp{lin-2004-rouge}), cosine similarity between matched contextual words embeddings (BERTScore-F1,~\citealp{zhang2019bertscore}), edit distance with single-word insertion, deletion, or replacement (word error rate,~\citealp{su1992new}), edit distance with block insertion, deletion, or replacement (translation edit rate,~\citealp{snover-etal-2006-study}), and explicit goodness of words editing against reference and source (SARI,~\citealp{xu-etal-2016-optimizing}). In short, BLEU, METEOR, ROUGE-2-F1, SARI, word error rate and translation edit rate estimate sentence wellformedness lexically; METEOR and BERTScore-F1 consider semantic equivalence. Comparing grammatical relations found in prediction with those found in references can also measure semantic similarity~\citep{clarke-lapata-2006-models,riezler-etal-2003-statistical,toutanova-etal-2016-dataset}. Grammatical relations are extracted from dependency parsing, and F1 scores can then be used to measure overlap. 

In contrast, the lack of good parallel corpus impedes (machine) revising for concision. To address this limitation, we curate and make available such a corpus as benchmark. Each sample in the corpus contains a wordy sentence, and at least one sentence revised for concision. Samples are from English writing centres of 57 universities, ten colleges, four community colleges, and a postgraduate school.

\section{Benchmark Corpus}
\label{sectDataset}

The collated corpus, named \textbf{Concise-536}, contains 536 pairs of sentences. This is a fair starting size, comparable with 385 of RST-DT (semantic parsing,~\citealp{carlson2003building}), 500 of DUC 2004 (summarization\footnote{\url{https://duc.nist.gov/duc2004/}}), or 575 by~\citet{cohn-lapata-2008-sentence} (sentence compression). Each concise sentence is revised from its wordy counterpart by English specialists from the 72 universities, colleges, or community colleges. Sentence ID, category and original link are available for each data point\footnote{\url{https://huggingface.co/datasets}}, and a 120-point validation split from other sources is attached.


\begin{table*}[t]
\scriptsize
\centering
\begin{tabularx}{\textwidth}{lllXXX}
\hline
\textbf{Category} & \textbf{Action} & \textbf{\# sents.} & \textbf{Mean words wordy sent.} & \textbf{Mean words concise sent.} & \textbf{Translation Edit Rate} \\ \hline
\RomanNumeralCaps{1} & Delete & 169 & 13.16 & 9.17 & 4.72 \\
\RomanNumeralCaps{2} & Replace & 116 & 12.37 & 9.02 & 5.1 \\
\RomanNumeralCaps{3} & Rewrite & 153 & 14.43 & 9.73 & 9.54 \\
\RomanNumeralCaps{4} & Delete + Replace & 42 & 23.81 & 11.57 & 15.16 \\
\RomanNumeralCaps{5} & Replace + Rewrite & 33 & 21.52 & 12.85 & 14.88 \\
\RomanNumeralCaps{6} & Delete + Rewrite & 14 & 24.5 & 11.36 & 17.71 \\
\RomanNumeralCaps{7} & Delete + Replace + Rewrite & 9 & 32.56 & 14.56 & 25.56 \\
All & - & 536 & 15.32 & 9.86 & 8.31\\ \hline
\end{tabularx}
\caption{\label{tab:category}
Revising a sentence can involve either one of the three strategies (category \RomanNumeralCaps{1}, \RomanNumeralCaps{2}, \RomanNumeralCaps{3}), or a combination of them (category \RomanNumeralCaps{4}, \RomanNumeralCaps{5}, \RomanNumeralCaps{6}, \RomanNumeralCaps{7}). Sample sizes, average word counts before and after revisions, and average edit distance (translate edit rate, TER) for revision are listed.
}
\end{table*}

Revising different sentences can go through a completely different process. As seen below, simply crossing out a few words revises Example~\ref{ex:delete}; a new word is needed in revising Example~\ref{ex:replace}; and even the sentence structure needs changing in Example~\ref{ex:rewrite}.

\begin{corpus}[Delete]\label{ex:delete}
Any \remove{particular type of} dessert is fine with me.~\citep{purdue_writing_lab2021}
\end{corpus}

\begin{corpus}[Replace]\label{ex:replace}
She \replace{has the ability to}{can} influence the outcome.~\citep{purdue_writing_lab2021}
\end{corpus}

\begin{corpus}[Rewrite]\label{ex:rewrite}
\replace{The 1780 constitution of Massachusetts was written by John Adams.}{John Adams wrote the 1780 Massachusetts Constitution.}~\citep{north_carolina2021}
\end{corpus}

In Concise-536, we do not identify fine-grained wordiness because a phrase can have more than one type of verbosity at the same time. For instance, we can revise "\textit{Her poverty also helped in the formation of her character.}" to "\textit{Her poverty also helped form her character.}"~\citep{gmu2021}, treating "\textit{in the formation of}" as either a wordy prepositional phrase, or nominalization. Rather, we focus on editing. 

In editing, the three actions are not complementary, and instead have varying degrees of power. Deleting can be covered by replacing (See Section\,\ref{sec:sentcomp}), which could be again covered by rewriting, \emph{i.e.}, rewriting is the most flexible. However, Occam's razor pushes us to prioritize the actions requiring lower effort to complete, \emph{i.e.}, delete $<$ replace $<$ rewrite. Supposedly, the difficulty for implement each action with programs shares the same trend. In addition, some sentences contain multiple wordiness occurrences, each of which may need a different action, \emph{e.g.,} delete + replace.

Interested in how well a revising algorithm resembles each action, we label revisions in each sentence pair and divide them into seven categories. Revisions that require the same set of actions will be assigned to the same category. Each revision is assigned to one of seven categories in Table~\ref{tab:category}. 

For convenience, we standardize categorizing rules as follows where each sentence is a word-level sequence. For each pair, we have a wordy sequence ($w$) and a concise sequence ($c$).

\begin{enumerate}
	\item If $c$ is a (not necessarily consecutive) subsequence of $w$, we consider revision only requires deletion (category \RomanNumeralCaps{1}).
	\item If not, we only delete redundancy from $w$ to get $w'$, \emph{i.e.}, $w'$ paraphrases $w$, and $w'$ is a subsequence of $w$. Then, we make local\footnote{Empirically, in a sentence or clause, we do not replace the subject and predicate verb together.} replacement(s) to $w'$ to get $w^*$, and every individual state from $w'$ to $w^*$ (\emph{i.e.}, after each local replacement) paraphrases $w'$. If $w^* = c$ and $w' = w$, we consider revision only requires replacement (category \RomanNumeralCaps{2}). If $w^* = c$ and $w' \neq w$, we consider revision only requires deletion and replacement (category \RomanNumeralCaps{4}).
	\item If $w^* = w$, we consider revision relies solely on rewriting (category \RomanNumeralCaps{3}).
\end{enumerate}

\begin{corpus}[category \RomanNumeralCaps{1}]\label{exp:there}
\remove{There are} four rules \remove{that} should be observed.~\citep{purdue_writing_lab2021}
\end{corpus}

\begin{corpus}[category \RomanNumeralCaps{3}]\label{exp:regular}
\remove{Regular reviews of} online content should be \replace{scheduled}{reviewed regularly}.~\citep{monash2020}
\end{corpus}

\begin{corpus}[category \RomanNumeralCaps{4}]\label{exp:she}
She fell \remove{down} \replace{due to the fact that}{because} she hurried.
~\citep{purdue_writing_lab2021}
\end{corpus}

Example\,\ref{exp:there} used to be wordy in the running start, but deleting suffices in revision. Therefore, although counter intuitive, it belongs to category \RomanNumeralCaps{1}. An adjective-noun pair is the wordiness in Example\,\ref{exp:regular}, yet its revision is more complex than replacing a verb. Usually, revision involves multiple strategies, as seen in Example\,\ref{exp:she} (delete "\textit{down}" + replace "\textit{due to the fact that with}" with "\textit{because}").

Human annotators implement the rules, as we need to check whether the meaning is still the same at each step.

Usually, category \RomanNumeralCaps{3} sentences are the hardest to revise as the easier strategies of deleting and replacing are not applicable. In fact, revising category \RomanNumeralCaps{5}, \RomanNumeralCaps{6}, and \RomanNumeralCaps{7} sentences are more challenging, as these sentences are longer, more complex, and more deliberate than category \RomanNumeralCaps{3} sentences (Figure~\ref{fig:difficulty}, Table~\ref{tab:good},\,\ref{tab:bad}), which is a bias in this corpus.

\section{Experiments}
\label{sec:baseline}

We also offer a baseline solution in this study. Solutions to machine revision for concision can be diverse. Neural model solutions include tree-to-tree transduction models~\citep{cohn-lapata-2008-sentence}, or general Seq2Seq models. We present a Seq2Seq baseline method, for it is flexible and straightforward. The model architecture is BART~\citep{lewis-etal-2020-bart}.

Ideally, training corpora tune statistical models or neural models, such that we can test tuned models on the benchmark corpus. However, lacking authoritative revisions prompted us to let models fit relevant task data. We also use public external knowledge, \emph{e.g.}, WordNet~\citep{fellbaum2010wordnet}. This section describes how we build an ad hoc training corpus to initiate this task.

The BART base model (124,058,116 parameters) is then used to fit \emph{each} training set in this section. Training settings are fixed (batch size at 32, PyTorch Adam optimizer~\citep{NEURIPS2019_9015,kingma2014adam} with initial learning rate at $5\times10^{-5}$, validated every 5,000 iterations). We then evaluate trained models on Concise-536.

\subsection{Pre-baselines}
We prepare training samples by adjusting data from paraphrase generation (ParaNMT,~\citealp{wieting-gimpel-2018-paranmt}), sentence simplification (WikiSmall,~\citealp{zhang-lapata-2017-sentence}), or sentence compression (Gigaword,~\citealp{rush-etal-2015-neural}; Google News datasets,~\citealp{filippova-altun-2013-overcoming}; MSR Abstractive Text Compression Dataset,~\citealp{toutanova-etal-2016-dataset}).

\subsection{WordNet as Booster}

Pre-baselines are useful, but they are not developed for revision tasks after all. To replace a verb or noun phrase with a single word, we leverage word glosses in public dictionaries, \emph{i.e.}, WordNet~\citep{fellbaum2010wordnet}. Word semantics are close to semantics in their glosses. This feature is usually used to improve word embedding~\citep{bosc-vincent-2018-auto} or evaluate analogy of word embedding~\citep{mikolov2013efficient}. We use this feature to replace a verb or noun phrase with a single word.

We create data samples using WordNet and a language modelling corpus. For each sentence $s$ in the corpus, we use WordNet vocabulary glosses to inflate it and obtain $s'$. Resulted parallel data approximate phrase replacement in sentence revision.

We first pick a unigram $u$, one of nouns, verbs, adjectives, or adverbs in $s$. At the same time, we avoid common words, \emph{e.g.}, "\textit{old}", or collocations and compounds, \emph{e.g.}, "\textit{united}" in "\textit{United Kingdom}". Next, we apply Lesk's dictionary-based word sense disambiguation (WSD) algorithm~\citep{lesk1986automatic} on $u$ and $s$ to get gloss $g$. Then, we parse $s$ and $g$ to obtain respective dependency trees $T_s$ and $T_g$; $r_g$ denotes root node in $T_g$. Usually, if $u$ is a noun, $r_g$ is a noun, and if $u$ is an adjective, $r_g$ is a verb. Eight $u \rightarrow r_g$ patterns account for over 90\% of the WordNet vocabulary (Table~\ref{tab:pos}). In Algorithm~\ref{alg:tree}, We modify dependency trees ($T_s$ and $T_g$) according to the eight patterns. The remaining six patterns are NOUN$\rightarrow$VERB, ADJ(-S) $\rightarrow$ ADJ, ADJ $\rightarrow$ ADP, ADJ $\rightarrow$ VERB, and ADV $\rightarrow$ ADP.

\begin{algorithm}[t!]
\begin{algorithmic}
\Require $T_s, T_g$
\Return $s'$
\State Copy-children(from $u$, to $r_g$)
\State Locate $h_u$ \Comment{head node of $u$}
\State Delete ($u$ with children, from $T_s$)
\If{$u \in$  NOUN}
\State Insert-child-node ($r_g$ with children, to $h_u$)
\If {$r_g \in$  VERB}
\State $u \gets$ Gerund($u$)
\EndIf
\State Correct inflections (singular and plural forms)
\State Remove duplicate determiners
\ElsIf{$u \in$  VERB}
\State Insert-child-node ($r_g$ with children, to $h_u$)
\State Correct inflections (person and tense)
\State Add/Remove prepositions according to verb transitivity
\Else
\State Insert-right-child-node ($r_g$ with children, to $h_u$) \Comment{Post attributive}
\EndIf
\State $s'\gets$ Linearize($T_s$)
\end{algorithmic}
\caption{Rule-based Gloss Substitution}\label{alg:tree}
\end{algorithm}

Finally, we filter and post-process synthesized sentences. We parse $s'$ again and compare it with the dependency tree from which $s'$ is linearized. We drop those with more than three mismatches, or with accuracy lower than 0.9. We "smooth" synthesized sentences with parroting\footnote{\url{https://huggingface.co/prithivida/parrot_paraphraser_on_T5}}, to mitigate overfitting. We also drop those sharing low semantic similarity (BERTScore $\leq$ 0.82) with original $s$.

We take the first 0.2 million sentences from WikiText-103 corpus~\citep{merity2016pointer} and around 71 thousand data points after filtration are available to train the BART base model.

\begin{table*}[t!]
\tiny
\centering{
\begin{tabularx}{\textwidth}{Xlllllllllll}
\hline
\textbf{Methods} & \textbf{\RomanNumeralCaps{1}}& \textbf{\RomanNumeralCaps{2}}& \textbf{\RomanNumeralCaps{3}}& \textbf{\RomanNumeralCaps{4}}& \textbf{\RomanNumeralCaps{5}}& \textbf{\RomanNumeralCaps{6}}& \textbf{\RomanNumeralCaps{7}} & \textbf{All}   & \textbf{H} & $\rho$ & $\omega$  \\
\hline
ParaNMT~\citep{wieting-gimpel-2018-paranmt}   & 0.46          & \textbf{0.62} & 0.46          & \textit{0.53} & 0.44          & 0.45          & \textit{0.38} & 0.55   &    3.40  & &  \\
MSR~\citep{toutanova-etal-2016-dataset}       & \textit{0.74} & 0.58          & 0.44          & 0.51          & 0.41          & 0.44          & 0.37          & 0.57     &   \textit{2.79} & 0.78 & \textbf{0.40} \\
G. News~\citep{filippova-altun-2013-overcoming}      & 0.61          & 0.46          & 0.39          & 0.40          & 0.35          & 0.39          & 0.33          & 0.48      &  5.74 & & \\
Gigaword~\citep{rush-etal-2015-neural}  & 0.30          & 0.29          & 0.28          & 0.31          & 0.25          & 0.29          & 0.23          & 0.29    &  6.74  & &  \\
WikiSmall~\citep{zhang-lapata-2017-sentence} & 0.70          & 0.59          & \textit{0.48} & 0.52          & 0.44          & \textit{0.46} & \textit{0.38} & 0.57      &   3.31 & &\\
WordNet~\citep{fellbaum2010wordnet}   & 0.70          & 0.60          & 0.47          & 0.52          & \textbf{0.46} & 0.45          & 0.37          & 0.58    &    \textit{2.79}&\textbf{0.99}&0.47\\
Baseline  & \textbf{0.75} & 0.60          & 0.47          & \textbf{0.55} & 0.40          & 0.45          & \textbf{0.41} & \textbf{0.59} &  \textbf{2.62} &\textit{0.82}&\textit{0.41}\\
Input (control group)     & 0.73          & \textit{0.61} & \textbf{0.49} & \textit{0.53} & \textbf{0.46} & \textbf{0.47} & \textit{0.38} & \textbf{0.59}  & -& - &0.52\\
\hline
\end{tabularx}
}
\caption{We average BLEU, METEOR, ROUGE-2-F1, SARI, Parsed relation F1, BERTScore-F1, \textit{and} (negative) translation edit rate of (pre-)baseline methods. The most favorable score in each column is in bold, the second most favorable in italics. This table estimates the strengths and weaknesses of each variants.
System ranking from human evaluation (\textbf{H}), information retention ($\rho$), and wordiness ($\omega$) are presented in the right-most columns.  }\label{tab:brief_results}
\end{table*}

\subsection{Multi-Task Learning}\label{sec:mix}

Each dataset in pre-baselines, including WordNet, handles part of task. However, sentence compression or simplification does not emphasize complete information retention; paraphrase generation hardly encourages deletion; synthetic data limit editing scope because word glosses are limited. We hypothesize that mixing the good samples among these datasets could more closely approximate the revision task. Therefore, we adjust datasets again. We keep every sample in MSR as it is small (21,145, see Appendix). Semantic similarity lower bound for sentence compression and simplification datasets is set at BERTScore = 0.9.  For ParaNMT, we discard samples with less than 10 words. As a result, ablation of mixed and shuffled data samples shows that a mixture of MSR, filtered ParaNMT, and synthetic WordNet dataset leads to the strongest baseline. This baseline method uses transfer learning from multiple datasets to learn revising strategies such as deletion and phrase replacement.

\subsection{Baseline Result}
\begin{table*}[!t]
\centering
\small
\begin{tabularx}{\textwidth}{lXX}
\hline
\textbf{Category} & \textbf{Reference(s)} & \textbf{Prediction} \\\hline
5th percentile & Bob \replace{provided an explanation of}{explained} the computer to his grandmother. & Bob \replace{provided an explanation of}{explained} the computer to his grandmother. \\
95th percentile & \remove{Rather than taking the bull by the horns,} she  \replace{was quiet as a church mouse}{avoided confrontation by remaining silent}. & \remove{Rather than taking the bull by the horns,} she was quiet as a church mouse. \\
\hline
\end{tabularx}
\caption{\label{tab:illustrate}
Well/poorly revised samples in the corpus. Shorter sentences that require simpler actions are perfectly revised. Rewriting clich\'es is difficult, in which case the baseline approach tends to use deletion.
}
\end{table*}
Table~\ref{tab:brief_results} shows test results in each category. The baseline model has the highest overall score and is more robust than pre-baseline models on category \RomanNumeralCaps{1}, \RomanNumeralCaps{4}, and \RomanNumeralCaps{7}. The same architecture trained only on MSR outperforms any other pre-baseline for deletion (category \RomanNumeralCaps{1}) and ranks second for replacement (category \RomanNumeralCaps{2}). The top-ranked pre-baseline for replacement (category \RomanNumeralCaps{2}) is trained on ParaNMT. In category \RomanNumeralCaps{5}, the model trained on WordNet scores highest, slightly outperforming other pre-baselines. Trends in category \RomanNumeralCaps{3}, \RomanNumeralCaps{4}, \RomanNumeralCaps{6}, \RomanNumeralCaps{7} are less clear. Datasets Gigaword, Google News, and WikiSmall may be quite different from the benchmark corpus, and thus models trained on these datasets do not score well.

The baseline model suffers from two shortcomings common to all pre-baselines. First, the model relies on transfer learning from MSR and ParaNMT and struggle to rewrite (category \RomanNumeralCaps{3}) or to handle composite wordiness (category \RomanNumeralCaps{5}, \RomanNumeralCaps{6}, \RomanNumeralCaps{7}). Second, the baseline outputs score worse than the input text on many metrics in many categories, especially on category \RomanNumeralCaps{3}. These shortcomings suggest challenges in revision. We take the 5th and 95th percentile from all 536 samples to qualitatively illustrate the baseline in Table~\ref{tab:illustrate}. Apart from samples in Concise-536, Figure~\ref{fig:obscure} shows an arbitrary sentence by non-English native speakers~\citep{chen2020survey}. Our baseline revisioner removes repetition and unnecessary prepositional phrases, illustrating its potential in academic writing.

For human evaluation, we adopt an approach similar to \citet{hsu-etal-2018-unified,zhang2020pegasus,ravaut2022summaReranker}. We (1) rank the samples by overall automatic evaluation on the \textit{baseline} model in descending order; (2) divide the examples in \textit{each category} into two buckets; (3) randomly pick one example from each bucket. For each picked sample, we ask three graduate students (IELTS 7.0 or equivalent) to rank the predictions of seven systems, and the average ranking of each system is shown in \textbf{H} column in Table~\ref{tab:brief_results}.

For top three systems, human evaluators then assess information retention ($\rho$) and wordiness ($\omega$), since system outputs are in good syntax. Particularly, human assessment on wordiness engages the Paramedic Method~\citep{lanham1992revising} to highlight the wordy part and $\omega = $ (\# wordy words) / (\# all words). The model trained adapted WordNet data preserves information better, which also accounts for its good human ranking.

We observe general correlation between automatic score ranking and human evaluation ranking. However, information retention is not sufficiently represented by semantic similarity scores like BERTScore. These findings suggest further investigation on the evaluation scheme of this task.


\section{Related Work}
\label{sec:relatedWork}

\subsection{Deleting as in Sentence Compression}\label{sec:sentcomp}
When revising, deleting redundant words is common. For example, we can revise "\textit{research is increasing in the field of nutrition and food science}" to "\textit{research is increasing in nutrition and food science}"~\citep{uri2019}, simply by deleting "\textit{the field of}". Deleting is canonical in sentence compression, a task aiming to reduce sentence length from source sentences while retaining basic meaning~\citep{jing-2000-sentence,knight2000statistics,mcdonald-2006-discriminative}. For example, the compression task has been formulated as integer linear programming optimisation using syntactic trees~\citep{clarke-lapata-2006-constraint}, or as a sequence labelling optimisation problem using the recurrent neural networks (RNN)~\citep{filippova-etal-2015-sentence,klerke-etal-2016-improving,kamigaito-etal-2018-higher}. They explicitly or implicitly use dependency grammar. Pre-trained language models such as ELMo~\citep{peters-etal-2018-deep} and BERT~\citep{devlin-etal-2019-bert} can encode features apart from dependency parsing~\citep{kamigaito2020syntactically}, bringing prediction and reference sentences closer.

All methods rely on parallel datasets labelling parts to be deleted. However, the deleting part in sentence compression differs from that in revision.~\citep{filippova-altun-2013-overcoming} created Google dataset from titles and first sentence of news articles. The information retained in the first sentence depends on the title. While this creation is useful for reducing excessive information, the deleted part is probably not wordiness.

Deleting does not solve everything in revision.  We can revise "\textit{in this report I will conduct a study of ants and the setup of their colonies}" to "\textit{in this report I will study ants and their colonies}", taking advantage of noun-and-verb homograph. However, a more concise version "\textit{this report studies ants}"~\citep{commnet2021} requires changing "\textit{study}" to third-person singular. 

\subsection{Replacing as in Paraphrase Generation}\label{subsec:paraphrase}
Word choice matters as well, thus we revise by paraphrasing to stronger words. Paraphrase generation changes a sentence grammatically and re-selects words, while retaining meaning. Paraphrasing matters in academic writing, for it helps avoid plagiarism. Rule-based or statistical machine paraphrasing substitutes words by finding synonyms from lexical databases, and decodes syntax according to template sentences. This rigid method may undermine creativity~\citep{bui2021generative}. Pre-trained neural language models like GPT~\citep{radford2019language} or  BART~\citep{lewis-etal-2020-bart} paraphrase more accurately~\citep{hegde2020unsupervised}. Through paraphrasing, we can replace verb phrase "\textit{conduct a study}" to verb "\textit{study}" in the example above, rather than delete and rely on noun-and-verb homographs to keep the sentence syntactically correct.

Machine revision is a kind of paraphrase generation, and vice versa is not true. Current paraphrase generation does not require concision in generated sentences. Automatically annotated datasets for paraphrasing include ParaNMT~\citep{wieting-gimpel-2018-paranmt}, Twitter~\citep{lan-etal-2017-continuously}, or re-purposed noisy datasets such as MSCOCO~\citep{lin2014microsoft} and WikiAnswers~\citep{fader-etal-2013-paraphrase}. We may adapt paraphrase parallel datasets to train revising models, as investigated in Section~\ref{sec:baseline}.

\subsection{Other related tasks}

Summarization produces a shorter text of one or several documents, while retaining most of meaning~\citep{paulus2017deep}. This is similar to sentence compression. In practice, summarization welcomes novel words, allows specifying output length~\citep{kikuchi-etal-2016-controlling}, and removes much more information than sentence compression does. Datasets include XSum~\citep{narayan-etal-2018-dont}
, CNN/DM~\citep{hermann2015teaching}, WikiHow~\citep{koupaee2018wikihow}, NYT~\citep{sandhaus2008new}, DUC-2004~\citep{over2007duc}, and Gigaword~\citep{rush-etal-2015-neural}, where summaries are generally shorter than one-tenth of documents. On the other hand, sentence summarization~\citep{chopra-etal-2016-abstractive} uses summarization methods on sentence compression datasets, retaining more information and possibly generating new words.

Summarization compresses long document(s) into a short, fluent, and coherent text that delivers relevant, consistent, and as much information as possible from the source document~\cite{fabbri-etal-2021-summeval}. However, most works on machine summarization do not perform a direct evaluation based on these qualities of a good summary, which is challenging to quantify. Instead, a common approach is to compare the machine summary against that of a gold-standard human summary. Hence, contemporary summarization usually depends on three main aspects: (i) having benchmark datasets to provide source and references texts for comparison; (ii) using automatic metrics to quantify this closeness; and (iii) developing good summarizing approach, which is the main aim of this task.
The third aspect on developing good summarizing approaches is highly dependent on the earlier two, but current metrics miscalculate factual consistency~\cite{liu-liu-2008-correlation,cohan2016revisiting,chaganty-etal-2018-price,narayan-etal-2018-dont,owczarzak-etal-2012-assessing}. The rest of this section lists the frequently used metrics, and summarizing approaches.



Pointer Generator~\cite{see-etal-2017-get}, an encoder-decoder model built with Long short-term memory (LSTM). A copy mechanism helps generate out-of-vocabulary words in inputs. A coverage mechanism penalizes word repeating 

Fast-Abs-RL~\cite{chen-bansal-2018-fast}, built upon a Pointer Generator, rewrites selected sentences from inputs. ROUGE-L gives feedback in reinforcement learning of sentence selection.

Bottom-Up~\cite{gehrmann-etal-2018-bottom}, built upon a Pointer Generator, restricts the copy attention distribution during inference.

Pretrained encoding transformers such as BERT~\cite{devlin-etal-2019-bert} extract features from input texts. MatchSum uses features to select source sentences as summaries. BertSum-Abs~\cite{liu-lapata-2019-text} decode features to generate summaries. Feature extracting backbone shifts from recurrent neural networks (RNN) to transformers.

Pretrained conditional generative transformers addresses general text generations tasks. GPT-2~\cite{radford2019language,ziegler2019fine}, UniLM~\cite{dong2019unified}, or BART~\cite{lewis-etal-2020-bart} can write summaries if they are fine-tuned on correponding datasets. Pegasus~\cite{zhang2020pegasus} is a transformer pretrained on summarization datasets.

Recent summarizing approaches mix and match pretrained transformers and various training methods~\cite{dou-etal-2021-gsum,liu-liu-2021-simcls} and generate summaries with more favourable scores.

Although often evaluated by ROUGE, most approaches optimize maximum likelihood estimation (MLE) between outputs and references.

Text simplification modifies vocabulary and syntax for easier reading, while retaining approximate meaning~\citep{omelianchuk-etal-2021-text}. Hand-crafted syntactic rules~\citep{siddharthan2006syntactic,carroll-etal-1999-simplifying,chandrasekar-etal-1996-motivations} and aligned
sentences-driven simplification~\citep{yatskar-etal-2010-sake} have been explored. Corpora such as Turk~\citep{xu-etal-2016-optimizing} and PWKP~\citep{zhu-etal-2010-monolingual} are compiled from Wikipedia and Simple English Wikipedia~\citep{coster-kauchak-2011-simple}. Rules for simplification may deviate from that for revision, \emph{e.g.}, text simplification sometimes encourages prepositional phrases~\citep{xu-etal-2016-optimizing}. Still, adapting these approaches may benefit academic revising for concision.

Fluency editing~\citep{napoles-etal-2017-jfleg} not only corrects grammatical errors but paraphrases text to be more native sounding as well. Its paraphrasing section is constrained such that outputs represent a higher level of English proficiency than inputs. As a constrained paraphrase task, fluency editing may alleviate ill-posed problems in paraphrase generation~\citep{cao-etal-2020-unsupervised-dual,rus-etal-2014-paraphrase}. However, such constraints may not be consistent with those required for concision.

In general, machine revision for academic writing requires new methods. Rules for revision can be adapted from these related tasks, so do training strategies.


\section{Discussion}
\label{sec:discussion}

Comparing baseline revisioner's effectiveness for different categories, we understand deleting and replacing are much easier sub-tasks than rewriting is. The former two actions, especially deletion, are less ill-posed, while rewriting is open. Still, revision for concision requires an algorithm that is able to use all three actions in combination. Its goal is to resolve all seven categories of cases, marking distinction between revision and other tasks such as sentence compression.

We use seven metrics to estimate a revisioner's effectiveness, since each metric has its shortcomings. For example, METEOR does not adequately penalize nominalization, and thus wordy input texts typically score higher on METEOR than algorithm outputs. More targeted metrics for this task, including reference-free structural metrics~\citep{sulem-etal-2018-semantic}, might help. We do not include word counts. Although concision is marked by brevity and wordiness often correlates to high word count, concise writing does not always require the fewest words~\citep{purdue_writing_lab2021}. Optimizing a lower word count may be misleading even if it is constrained to zero information loss~\citep{siddharthan2006syntactic}. For example, abusing pronouns and ellipses can result in shorter sentences that are harder to read. 

Transferring knowledge from other tasks to approximate revising is a stopgap measure. Specialized revising methods exist, \emph{e.g.}, the Paramedic Method~\citep{lanham1992revising}. Automated specialized methods may be more efficient.

\subsection{Linguistic Rules in Revising for Concision}
We collate and present a set of practical linguistic rules for concise sentence revision, which we synthesize based on guidelines from writing centres at numerous major universities and educational institutes. Table~\ref{tab:strategies} illustrates how wordiness can be fine-grained, and what action is required once a wordiness is identified~\citep{north_carolina2021,purdue_writing_lab2021,monash2020}.

\begin{table}[!t]
\centering
\begin{tabularx}{0.7\textwidth}{Xl}
\hline
\textbf{Wordiness identified} &   \textbf{Action}        \\ \hline
           Weak modifiers (qualifiers / intensifiers)   &Delete\\
                 Redundant pairs                        &\\
                 Grouped synonyms                       &\\
                 Stock phrases                          &\\
                 Unnecessary hedging                    &\\
                 Implied information                    &\\
                 Yourself                               &\\ \hline
          Informal language                      &Replace\\
                 Vague pronoun references               &\\
                 Possessive constructions using "of"    \\
                 Prepositional phrases                  &\\
                 All-purpose nouns                      &\\
                 Vague Swamp                            &\\
                 Fancy words                            &\\
                 Helping verbs ("to be" verbs, "be" + adjective) &\\
                 Adjective-noun pairs                   &\\
                 Phrasal verbs                          &\\
                 Verb-adverb pairs                      &\\
                 Nominalisation / noun strings          \\
                 Cliches and Euphemisms                 &\\
                 Empty phrases                          &\\
                 Expletive constructions                &\\ \hline
          long sentences (\textgreater 25 words) &Rewrite\\
                 Running starts (with "there / it" + "be") &\\
                 Long opening phrases / clauses         \\
                 Needless transitions                   &\\
                 Interrupted subjects and verbs         \\
                 Interrupted verbs and objects          \\
                 Negatives (opposite to affirmatives)   \\
                                                        &\\
                 \textit{and anything violating:}       \\
                 A blend of active and passive verbs    \\
                 Elliptical constructions / parallelism \\
                 Only one main idea per sentence        \\ \hline
\end{tabularx}

\caption{\label{tab:strategies}
Revising rules collated from college writing centres. Three actions are available. Redundancy can be deleted; short, specific, concrete and stronger expressions shall replace vague ones; sentences should be rewritten if neither deleting nor replacing helps.
}
\end{table}

\subsection{Technical Difficulties in Reference-free Revision Evaluation}

Had we chosen not to follow Papineni's viewpoint, reference-free evaluation is the way to go. However, it is technically not trivial to use programs to detect wordiness or syntax errors these days (See Section~\ref{sec:indicator}), let alone detect semantic similarity. Progress in sentence embedding~\citep{lin2017structured} and semantic textual similarity~\citep{yang2019xlnet} enables meaning comparison between sentences, but relying on one developing system to evaluate another is risky. Moreover, information delivered by a sentence is sometimes beyond its textual meaning. Concise writing can suggest eliminating  first-person narratives; \emph{e.g.}, "\textit{I feel that the study is significant}" is revised to "\textit{The study is significant}"~\citep{waldenu2021}. Here, the first-person statement used to be the main clause, and removing it will shift sentence embedding. Nevertheless, in academic writing, these two sentences deliver identical information.

\subsection{Explaining Categories in the Corpus}

\begin{figure}[h!]
    \centering
    \small
    \begin{tikzpicture}
        \begin{scope}[blend group=soft light]
        \fill[p13!70!white]   ( 209.98945070661557:1.1982899710040975) circle (1.4911131108882707);
		
        \fill[b14!70!white] ( 95.9701298441318:1.020383701219179) circle (1.3785345586728925);
        		
        \fill[y15!70!white]  ( -13.379609620139817:1.2988177887329182) circle (1.409210281782305);
        		
        \end{scope}
        \node at ( 209.98945070661557:1.6776059594057364)    {\RomanNumeralCaps{1} };
        		
        \node at ( 95.9701298441318:1.4285371817068504)    {\RomanNumeralCaps{2} };
        		
        \node at ( -13.379609620139817:1.8183449042260853)    {\RomanNumeralCaps{3} };
        		
        \node at ( 152.9797902753737:0.679315988401639)    {\RomanNumeralCaps{4} };
        		
        \node at ( 41.295260111995994:0.679315988401639)     {\RomanNumeralCaps{5} };
        		
        \node at ( -82.75:0.679315988401639)     {\RomanNumeralCaps{6} };

        \node at ( 209.98945070661557:3.275895930409834)    {Delete};
        		
        \node at ( 95.9701298441318:2.9489208829260294)    {Replace};
        		
        \node at ( -13.379609620139817:3.3171626929590037)    {Rewrite};
        		
        \node at (0:0)   {\RomanNumeralCaps{7} };
    \end{tikzpicture}
    \caption{Revising a sentence can involve either one of the three strategies (category \RomanNumeralCaps{1}, \RomanNumeralCaps{2}, \RomanNumeralCaps{3}), or a combination of them (category \RomanNumeralCaps{4}, \RomanNumeralCaps{5}, \RomanNumeralCaps{6}, \RomanNumeralCaps{7}).}
    \label{fig:categories}
\end{figure}
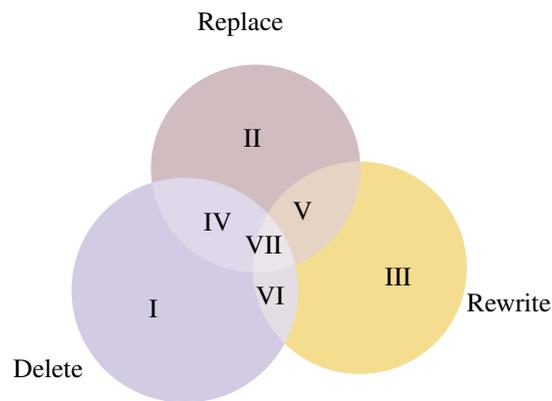

There are seven categories, as seen in Figure~\ref{fig:categories}. Note that although three actions (delete, replace, rewrite) are put side-by-side, they are with different levels of flexibility. In fact, every revision made with deleting can be done through replace, \emph{e.g.}, the fourth example in Table~\ref{tab:good}, "\textit{fell down}" could be replaced to "\textit{fell}", but we would simply consider the cheapest revision, which is to delete "\textit{down}"\footnote{"\textit{fell}" means "\textit{fell down}", as one never "\textit{fell up}".}. Similarly, rewriting is even more expensive and ambiguous. Therefore, our rule of Occam's razor is that only when a cheaper revision fails, will we use a more expensive one.

Here, we give examples of which category a revision corresponds to. Indeed, many sentence revisions are categorized in original websites. See the two examples from Purdue Writing Lab~\citep{purdue_writing_lab2021} below. The strategies applied are to "eliminate words that explain the obvious or provide excessive detail" (category \RomanNumeralCaps{1}) and to "replace several vague words with more powerful and specific words" (category \RomanNumeralCaps{2}), respectively.

\begin{corpus}[category \RomanNumeralCaps{1}]
Imagine \remove{a mental picture of} someone \remove{engaged in the intellectual activity of} trying to learn \remove{what} the rules \remove{are for how to play the game} of chess.

\end{corpus}

\begin{corpus}[category \RomanNumeralCaps{2}]
The politician \replace{talked about several of the merits of}{touted} after-school programs in his speech
\end{corpus}

For revisions not categorized in sources, we first align the segments of a pair of sentences by their meaning, as seen in Figure~\ref{fig:obscure}. This is intuitively straightforward when the revised sentence is given\footnote{If the revised sentences were not from a trustworthy site, this process could have been less intuitive.}.

Then, we determine the actions to revise. For example, in the fourth example (category \RomanNumeralCaps{4}) in Table~\ref{tab:good}, we find that we cannot delete any words in "\textit{due to the fact that}" without violating the second and third components in Definition\,\ref{def:nlp}. Thus, we have to put some more concise conjunction to take its place, \emph{i.e.}, "\textit{because}".

Another example is the sixth one (category \RomanNumeralCaps{4}) in Table~\ref{tab:bad}. Though it looks that the entire wordy sentence can only be written to reach the concise form, a cheaper revision is actually to first delete some redundancy, \emph{e.g.}, "\textit{sent \remove{to you} by us}", and then rewrite the necessary part.

Whether the subject and predicate of a sentence (clause) is changed together determines the border between replacing and rewriting. In the fifth example (category \RomanNumeralCaps{4}) in Table~\ref{tab:good}, "\textit{it was necessary}" is aligned to "\textit{had to}", and "\textit{us}" to "\textit{we}". However, we cannot change either of them individually without violating the third component in Definition\,\ref{def:nlp}. Therefore, when two or more replacements intertwine, we rewrite.

\subsection{Explaining Rule-based Gloss Substitution}
\label{sec:graft}

\begin{figure*}[!t]
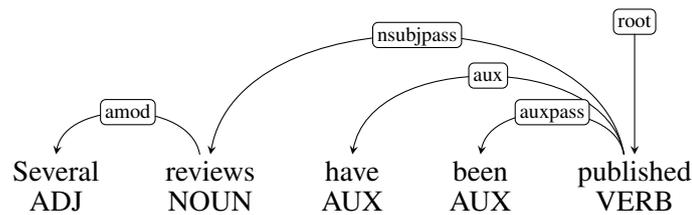
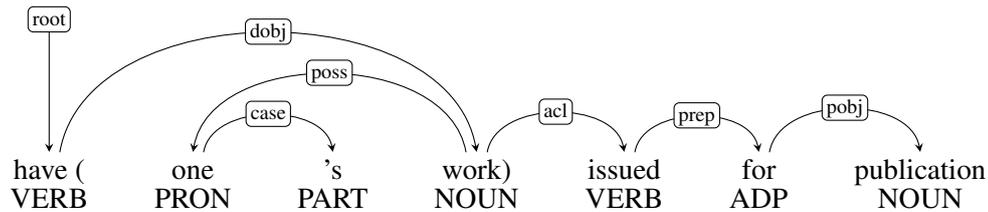
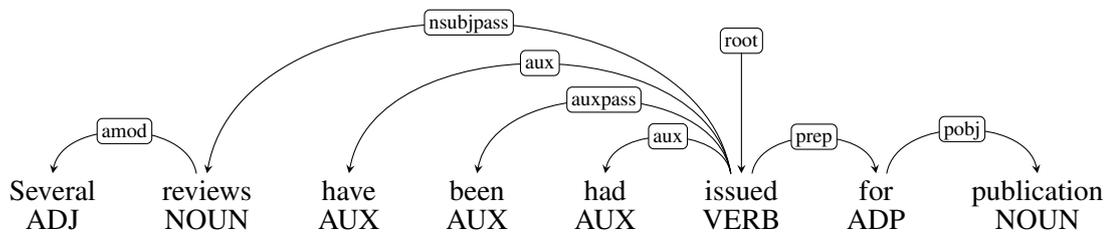

	\centering
	\begin{subfigure}{\textwidth}
	\centering
	\begin{dependency}[arc edge, arc angle=80]
	\begin{deptext}[column sep=.7cm]
	Several \& reviews \& have \& been \& published \\
	ADJ \& NOUN \& AUX \& AUX \& VERB \\
	\end{deptext}
	\deproot{5}{root}
	\depedge{2}{1}{amod}
	\depedge{5}{2}{nsubjpass}
	\depedge{5}{3}{aux}
	\depedge{5}{4}{auxpass}
	\end{dependency}
	\caption{Sentence "\textit{Several reviews have been published}" and its dependency tree. We expand the word "\textit{publish}" below.}
	\label{fig:s1}
	\end{subfigure}\\
	\begin{subfigure}{\textwidth}
	\centering
	\begin{dependency}[arc edge, arc angle=80]
	\begin{deptext}[column sep=.7cm]
	have ( \& one \& 's \& work) \& issued \& for \& publication \\
	VERB \& PRON \& PART \& NOUN \& VERB \& ADP \& NOUN \\
	\end{deptext}
	\deproot{1}{root}
	\depedge{1}{4}{dobj}
	\depedge{4}{2}{poss}
	\depedge{2}{3}{case}
	\depedge{4}{5}{acl}
	\depedge{5}{6}{prep}
	\depedge{6}{7}{pobj}
	\end{dependency}
	\caption{Gloss of "\textit{publish}" from WordNet\citep{fellbaum2010wordnet}; the root node is the verb "\textit{have}".}
	\label{fig:g1}
	\end{subfigure}\\
	\begin{subfigure}{\textwidth}
	\centering
	\begin{dependency}[arc edge, arc angle=80]
	\begin{deptext}[column sep=.7cm]
	Several \& reviews \& have \& been \& had \& issued \& for \& publication \\
	ADJ \& NOUN \& AUX \& AUX \& AUX \& VERB \& ADP \& NOUN \\
	\end{deptext}
	\deproot{6}{root}
	\depedge{2}{1}{amod}
	\depedge{6}{2}{nsubjpass}
	\depedge{6}{3}{aux}
	\depedge{6}{4}{auxpass}
	\depedge{6}{5}{aux}
	\depedge{6}{7}{prep}
	\depedge{7}{8}{pobj}
	\end{dependency}
	\caption{Synthesized sentence with the first stage of our approach; $r_g$ node "\textit{have/had}" is grafted onto the original sentence in (a).}
	\label{fig:ss1}
	\end{subfigure}\\

	\caption{Demonstration of dependency tree grafting in sentence synthesis. The dependency in (c) is obtained by re-parsing the synthesized sentence. As we can see, the POS tag of "\textit{have/had}" has changed from a verb to an auxiliary word, and the synthesized sentence is still syntactically and semantically correct, which shows that dependency changes may be unavoidable in the process of sentence synthesis. We also dealt with inflections to reduce grammatical errors.}
	\label{fig:graft}
\end{figure*}

A demonstration of Algorithm~\ref{alg:tree} is shown in Fig.\ref{fig:graft}, where a verb that appears in the past participle is replaced. By running this rule-based gloss replacement multiple times, we can recursively expand a sentence because the words used in a gloss have their associated glosses~\citep{bosc-vincent-2018-auto}. Table~\ref{tab:pos} describes $u \rightarrow r_g$ in the WordNet vocabulary.

\begin{table}[h!]
\centering
\small
\begin{tabular}{llllll}
\hline
\textbf{POS} & \textbf{ADJ} & \textbf{ADJS} & \textbf{ADV} & \textbf{NOUN} & \textbf{VERB} \\ \hline
VERB & \textbf{3627} & \textbf{6354} & 221 & \textbf{3349} & \textbf{11586} \\
DET & 1 & 6 & 4 & 1594 & 0 \\
ADJ & 1053 & \textbf{2825} & 50 & 544 & 316 \\
NOUN & 155 & 405 & 57 & \textbf{73527} & 1739 \\
CCONJ & 0 & 0 & 0 & 24 & 0 \\
PUNCT & 0 & 1 & 0 & 6 & 0 \\
PART & 0 & 5 & 4 & 4 & 6 \\
ADV & 10 & 84 & 235 & 37 & 52 \\
ADP & \textbf{2615} & 972 & \textbf{3019} & 222 & 29 \\
AUX & 5 & 5 & 4 & 108 & 1 \\
PRON & 0 & 3 & 2 & 1516 & 1 \\
SCONJ & 4 & 10 & 14 & 3 & 10 \\
PROPN & 0 & 6 & 0 & 534 & 15 \\
X & 2 & 2 & 2 & 16 & 19 \\
NUM & 1 & 16 & 8 & 658 & 0 \\
INTJ & 1 & 1 & 3 & 13 & 12 \\
SYM & 0 & 0 & 0 & 0 & 0 \\ \hline
\end{tabular}
\caption{Part-of-speech (POS) tags for a word $w$ and its corresponding $r_g$. Representation of POS tags follows the Stanford typed dependencies manual~\citep{de2008stanford} (except for ADJ-S, which stands for 'adjective satellite' in WordNet~\citep{fellbaum2010wordnet}). POS tags of $r_g$ are closely related to the POS tags of $w$, and we bold the pairs that appear frequently. In particular,  among nearly 117,000 word-gloss ($w \rightarrow r_g$) pairs, NOUN $\rightarrow$ NOUN is most frequent, accounting for more than three fifths. We have now studied the eight most frequently occurring pairs.}
\label{tab:pos}
\end{table}

\subsection{Details in Datasets Used to Train Pre-baselines}

We prepare training samples by adjusting data from paraphrase generation, sentence simplification, or sentence compression. ParaNMT~\citep{wieting-gimpel-2018-paranmt} contains over five million paraphrase pairs annotated from machine translation tasks; we sort each pair by sentence length. This is a rough approximation, since shorter sentences are not necessarily more concise. Google News datasets (News,~\citealp{filippova-altun-2013-overcoming}) contains 0.2 million pairs of sentences, where the longer one is the leading sentence of each article, and the shorter one is a subsequence of the longer one. Gigaword~\citep{rush-etal-2015-neural} contains four million pairs of article headline and the first sentence. Although these datasets are mainly for generating news headlines~\citep{p-v-s-meyer-2019-data}, they approximate the deletion aspect of sentence revision. MSR Abstractive Text Compression Dataset~\citep{toutanova-etal-2016-dataset} contains six thousand sentence pairs from business letters, newswire, journals, and technical documents sampled from the Open American National Corpus\footnote{\url{https://www.anc.org/data/oanc}}; humans rewrite sentences at a fixed compression ratio.  WikiSmall~\citep{zhang-lapata-2017-sentence} contains sentence pairs from Wikipedia articles and corresponding Simple English Wikipedia. We adopt training splits of these datasets, and Table~\ref{tab:trainsize} lists their sizes.


\begin{table}[h!]
\small
\centering
\begin{tabular}{ll}
\hline
\textbf{Dataset} & \textbf{Size}\\
\hline
MSR (~\citeyear{toutanova-etal-2016-dataset}) & 21,145 \\
ParaNMT (~\citeyear{wieting-gimpel-2018-paranmt}) & 5,306,522 \\
Google News (G. News~\citeyear{filippova-altun-2013-overcoming}) & 200,000 \\ 
Gigaword (~\citeyear{rush-etal-2015-neural}) & 3,803,957 \\ 
WikiSmall (~\citeyear{zhang-lapata-2017-sentence}) & 89,042 \\\hline
Baseline fine-tuning set (Section~\ref{sec:mix}) & 182,330 \\\hline
\end{tabular}
\caption{Sample numbers of training sets. MSR dataset has multiple references, we take each reference as a sample point. The mixed fine-tuning set in Section~\ref{sec:mix} is composed of 89,712 samples from ParaNMT, 21,145 from MSR, and 71,473 from our synthesized dataset from WordNet.}
\label{tab:trainsize}
\end{table}

\subsection{Random Sample Selection in Human Evaluation}
\begin{lstlisting}
import random

random.seed(0)
for k in [169, 116, 153, 42, 33, 14, 9]:
	print(random.randint(0, k//2))
	print(random.randint(k//2, k))
\end{lstlisting}

\subsection{Evaluation on Individual Metrics}
\label{sec:appendix}





\begin{table*}[t!]
\scriptsize
\begin{subtable}{.55\linewidth}\centering{
\begin{tabular}{llllllll}
\hline
          &\textbf{BL}   & \textbf{M}   & \textbf{R}   & \textbf{S}   & \textbf{P}  & \textbf{BS}   & \textbf{T}  \\\hline
ParaNMT   & .40          & .78          & .56          & .49          & .55          & .96          & .53          \\
MSR       & .55          & .86          & .69          & .62          & .73          & \textbf{.97} & .39          \\
News      & .36          & .67          & .56          & .52          & .62          & .95          & .51          \\
Gigaword  & .04          & .27          & .19          & .25          & .16          & .89          & .92          \\
WikiSmall & .47          & .91          & .65          & .59          & .63          & .95          & .82          \\
WordNet   & .48          & .90          & .65          & .59          & .66          & .94          & .82          \\
Input     & .48          & \textbf{.93} & .66          & \textbf{.68} & .68          & .96          & .56          \\
Baseline  & \textbf{.57} & .87          & \textbf{.71} & .66          & \textbf{.74} & \textbf{.97} & \textbf{.37} \\ \hline
\end{tabular}
}
\caption{Category \RomanNumeralCaps{1}, 169 / 536, Delete}
\end{subtable}
\begin{subtable}{.42\linewidth}\centering{
\begin{tabular}{lllllll}
\hline
\textbf{BL}  & \textbf{M}   & \textbf{R}   & \textbf{S}   & \textbf{P}  & \textbf{BS}   & \textbf{T}  \\
\hline
\textbf{.38} & .75          & \textbf{.54} & \textbf{.54} & \textbf{.51} & \textbf{.97} & \textbf{.50} \\
.33          & .73          & .51          & .45          & .47          & .96          & .56          \\
.17          & .56          & .39          & .35          & .37          & .94          & .64          \\
.01          & .26          & .17          & .28          & .15          & .89          & .92          \\
.33          & .80          & .51          & .48          & .44          & .95          & .92          \\
.36          & .80          & \textbf{.54} & .51          & .46          & .95          & .91          \\
.36          & \textbf{.82} & .53          & .52          & .47          & .96          & .65          \\
.36          & .76          & .53          & .50          & .50          & .96          & .53          \\ \hline
\end{tabular}
}
\caption{Category \RomanNumeralCaps{2}, 116 / 536, Replace}
\end{subtable}

\vspace{3mm}
\begin{subtable}{.55\linewidth}\centering{
\begin{tabular}{llllllll}
\hline
ParaNMT   & .16          & .60          & .33          & .44          & .28          & \textbf{.94} & .94          \\
MSR       & .15          & .56          & .31          & .38          & .28          & .93          & .92          \\
News      & .10          & .44          & .26          & .37          & .25          & .92          & .85          \\
Gigaword  & .03          & .20          & .12          & .38          & .09          & .88          & 1.00          \\
WikiSmall & \textbf{.19} & .66          & .35          & .45          & .29          & .93          & 1.26          \\
WordNet   & .17          & .65          & .36          & .43          & .29          & .93          & 1.27          \\
Input     & \textbf{.19} & \textbf{.67} & \textbf{.37} & \textbf{.47} & \textbf{.31} & \textbf{.94} & 1.09          \\
Baseline  & .18          & .60          & .35          & .42          & .30          & \textbf{.94} & \textbf{.91} \\ \hline
\end{tabular}
}
\caption{Category \RomanNumeralCaps{3}, 153 / 536, Rewrite}
\end{subtable}
\begin{subtable}{.42\linewidth}\centering{
\begin{tabular}{lllllll}
\hline
.27          & .68          & .40          & \textbf{.48} & .41          & .94          & 1.22          \\
.26          & .63          & .39          & .44          & .39          & .94          & 1.09          \\
.12          & .41          & .29          & .37          & .31          & .92          & \textbf{.77} \\
.03          & .24          & .16          & .35          & .16          & .89          & .86          \\
.24          & .71          & .39          & .46          & .37          & .93          & 1.69          \\
.23          & \textbf{.72} & .39          & .44          & .39          & .92          & 1.70          \\
.24          & .72          & .39          & .48          & .39          & .94          & 1.58          \\
\textbf{.32} & .67          & \textbf{.44} & .47          & \textbf{.43} & \textbf{.95} & .94          \\ \hline
\end{tabular}
}
\caption{Category \RomanNumeralCaps{4}, 42 / 536, Delete + Replace}
\end{subtable}

\vspace{3mm}
\begin{subtable}{.55\linewidth}\centering{
\begin{tabular}{llllllll}
\hline
ParaNMT   & .16          & .55          & .27          & .44          & .26          & \textbf{.94} & 1.18          \\
MSR       & .14          & .49          & .28          & .37          & .24          & .93          & 1.05          \\
News      & .08          & .34          & .21          & .35          & .22          & .92          & \textbf{.85} \\
Gigaword  & .01          & .15          & .09          & .34          & .06          & .87          & .90          \\
WikiSmall & .17          & .58          & .30          & .40          & .27          & .93          & 1.40          \\
WordNet   & .20          & \textbf{.59} & \textbf{.31} & .43          & .29          & .93          & 1.37          \\
Input     & \textbf{.19} & \textbf{.59} & \textbf{.31} & \textbf{.45} & \textbf{.30} & \textbf{.94} & 1.31          \\
Baseline  & .14          & .49          & .26          & .35          & .25          & .93          & 1.04          \\ \hline
\end{tabular}
}
\caption{Category \RomanNumeralCaps{5}, 33 / 536, Replace + Rewrite}
\end{subtable}
\begin{subtable}{.42\linewidth}\centering{
\begin{tabular}{lllllll}
\hline
.21          & .56          & .30          & .43          & .27          & \textbf{.93} & 1.35          \\
.19          & .53          & .30          & .40          & .30          & \textbf{.93} & 1.23          \\
.12          & .39          & .26          & .37          & .25          & .92          & \textbf{.81} \\
.04          & .20          & .12          & .38          & .10          & .87          & .87          \\
.17          & .62          & \textbf{.31} & .43          & .30          & .92          & 1.92          \\
.16          & .62          & \textbf{.31} & .41          & .29          & .92          & 1.96          \\
.18          & \textbf{.63} & \textbf{.31} & \textbf{.45} & \textbf{.32} & \textbf{.93} & 1.74          \\
\textbf{.22} & .55          & \textbf{.31} & .41          & .30          & \textbf{.93} & 1.23          \\ \hline
\end{tabular}
}
\caption{Category \RomanNumeralCaps{6}, 14 / 536, Delete + Rewrite}
\end{subtable}

\vspace{3mm}
\begin{subtable}{.55\linewidth}\centering{
\begin{tabular}{llllllll}
\hline
ParaNMT   & .06          & .50          & .17          & .40          & .20          & .92          & 1.61          \\
MSR       & .06          & .49          & .17          & .39          & .17          & .92          & 1.34          \\
News      & .03          & .28          & .15          & .38          & .21          & .91          & \textbf{.79} \\
Gigaword  & .00          & .11          & .04          & .36          & .02          & .86          & .95          \\
WikiSmall & \textbf{.08} & .54          & .20          & .38          & .18          & .91          & 2.02          \\
WordNet   & .04          & .53          & .18          & .39          & .17          & .91          & 1.96          \\
Input     & .06          & .54          & .18          & .41          & .17          & .92          & 1.84          \\
Baseline  & \textbf{.08} & \textbf{.55} & \textbf{.23} & \textbf{.43} & \textbf{.24} & \textbf{.93} & 1.18          \\ \hline
\end{tabular}
}
\caption{Category \RomanNumeralCaps{7}, 9 / 536, Delete + Replace + Rewrite}
\end{subtable}
\begin{subtable}{.42\linewidth}\centering{
\begin{tabular}{lllllll}
\hline
.29          & .69          & .44          & .48          & .42          & \textbf{.95} & .77          \\
.32          & .69          & .48          & .48          & .47          & \textbf{.95} & .71          \\
.20          & .53          & .38          & .41          & .40          & .93          & .69          \\
.03          & .23          & .15          & .31          & .13          & .88          & .94          \\
.31          & .76          & .48          & .49          & .43          & .94          & 1.12          \\
.31          & .76          & .49          & .50          & .45          & .94          & 1.12          \\
.32          & \textbf{.78} & .49          & \textbf{.54} & .47          & \textbf{.95} & .91          \\
\textbf{.35} & .72          & \textbf{.50} & .51          & \textbf{.49} & \textbf{.95} & \textbf{.68} \\ \hline
\end{tabular}
}
\caption{Overall}
\end{subtable}
\caption{BLEU (\textbf{BL}), METEOR (\textbf{M}) , ROUGE-2-F1 (\textbf{R}), SARI (\textbf{S}), Parsed relation F1 (\textbf{P}), BERTScore-F1 (\textbf{BS}), \textit{and} translation edit rate (\textbf{T}) of pre-baselines and baseline method. Numbers are shown in categories. Smaller edit distance is more favorable. The most favorable score(s) in each column is bold. In category \RomanNumeralCaps{5}, the model trained on WordNet has the highest scores on three metrics, slightly outperforming other pre-baselines. In category \RomanNumeralCaps{3},\RomanNumeralCaps{4}, \RomanNumeralCaps{6}, \RomanNumeralCaps{7}, no particular pre-baseline scores well on all metrics.}\label{tab:results}
\end{table*}

\begin{table*}[t!]
\begin{subtable}{.43\linewidth}\centering{
\begin{tabular}{llll}
\hline
          &\textbf{W}   & \textbf{R1}   & \textbf{RL}\\
\hline
ParaNMT   & .65          & .75          & .71          \\
MSR       & .43          & .85          & .8           \\
News      & .54          & .73          & .7           \\
Gigaword  & .97          & .39          & .37          \\
WikiSmall & .89          & .82          & .77          \\
WordNet   & .96          & .82          & .77          \\
Input     & .63          & .83          & .78          \\
Baseline  & \textbf{.42} & \textbf{.86} & \textbf{.81} \\ \hline
\end{tabular}
}
\caption{Category \RomanNumeralCaps{1}, 169 / 536, Delete}
\end{subtable}
\begin{subtable}{.41\linewidth}\centering{
\begin{tabular}{lll}
\hline
\textbf{W}   & \textbf{R1}   & \textbf{RL}  \\
\hline
.61          & \textbf{.73} & \textbf{.72} \\
.58          & .71          & .7           \\
.66          & .6           & .59          \\
.98          & .37          & .35          \\
.93          & .71          & .7           \\
1.03          & \textbf{.73} & \textbf{.72} \\
.65          & .72          & .71          \\
\textbf{.54} & \textbf{.73} & \textbf{.72} \\ \hline
\end{tabular}
}
\caption{Category \RomanNumeralCaps{2}, 116 / 536, Replace}
\end{subtable}

\vspace{3mm}
\begin{subtable}{.43\linewidth}\centering{
\begin{tabular}{llll}
\hline
ParaNMT   & 1.04          & .61          & .50          \\
MSR       & .99          & .60          & .48          \\
News      & .88          & .52          & .43          \\
Gigaword  & 1.03          & .32          & .29          \\
WikiSmall & 1.35          & .63          & .50          \\
WordNet   & 1.40          & .63          & .50          \\
Input     & 1.18          & \textbf{.64} & \textbf{.51} \\
Baseline  & \textbf{.97} & .63          & \textbf{.51} \\ \hline
\end{tabular}
}
\caption{Category \RomanNumeralCaps{3}, 153 / 536, Rewrite}
\end{subtable}
\begin{subtable}{.41\linewidth}\centering{
\begin{tabular}{lll}
\hline
1.28          & .59          & .58          \\
1.14          & .59          & .57          \\
\textbf{.81} & .48          & .46          \\
.90          & .37          & .32          \\
1.74          & .58          & .56          \\
1.79          & .59          & .56          \\
1.62          & .57          & .56          \\
.97          & \textbf{.63} & \textbf{.62} \\ \hline
\end{tabular}
}
\caption{Category \RomanNumeralCaps{4}, 42 / 536, Delete + Replace}
\end{subtable}

\vspace{3mm}
\begin{subtable}{.43\linewidth}\centering{
\begin{tabular}{llll}
\hline
ParaNMT   & 1.36          & .52          & .45          \\
MSR       & 1.17          & .51          & .43          \\
News      & \textbf{.91} & .42          & .36          \\
Gigaword  & .96          & .28          & .24          \\
WikiSmall & 1.58          & .53          & .44          \\
WordNet   & 1.59          & \textbf{.55} & \textbf{.46} \\
Input     & 1.48          & .54          & .45          \\
Baseline  & 1.15          & .52          & .43          \\ \hline
\end{tabular}
}
\caption{Category \RomanNumeralCaps{5}, 33 / 536, Replace + Rewrite}
\end{subtable}
\begin{subtable}{.41\linewidth}\centering{
\begin{tabular}{lll}
\hline
1.54          & .49          & .41          \\
1.37          & .51          & .41          \\
\textbf{.87} & .45          & .38          \\
.89          & .30          & .30          \\
2.04          & .51          & .43          \\
2.10          & .51          & .43          \\
1.89          & .51          & .44          \\
1.39          & \textbf{.54} & \textbf{.45} \\ \hline
\end{tabular}
}
\caption{Category \RomanNumeralCaps{6}, 14 / 536, Delete + Rewrite}
\end{subtable}

\vspace{3mm}
\begin{subtable}{.43\linewidth}\centering{
\begin{tabular}{llll}
\hline
ParaNMT   & 1.64          & .44          & .35          \\
MSR       & 1.38          & .47          & .35          \\
News      & \textbf{.81} & .38          & .34          \\
Gigaword  & .97          & .25          & .20          \\
WikiSmall & 2.04          & .44          & .33          \\
WordNet   & 1.98          & .45          & .34          \\
Input     & 1.88          & .44          & .33          \\
Baseline  & 1.24          & \textbf{.53} & \textbf{.43} \\ \hline
\end{tabular}
}
\caption{Category \RomanNumeralCaps{7}, 9 / 536, Delete + Replace + Rewrite}
\end{subtable}
\begin{subtable}{.41\linewidth}\centering{
\begin{tabular}{lll}
\hline
.88          & .66          & .61          \\
.76          & .69          & .63          \\
\textbf{.72} & .59          & .55          \\
.98          & .35          & .33          \\
1.19          & .69          & .63          \\
1.25          & .70          & .63          \\
.98          & .70          & .63          \\
.73          & \textbf{.71} & \textbf{.65} \\ \hline
\end{tabular}
}
\caption{Overall}
\end{subtable}
\caption{Other metrics include word error rate (\textbf{W}), ROUGE-1-F1 (\textbf{R1}), \textit{and} ROUGE-L-F1 (\textbf{RL}) . Numbers are shown in categories. Smaller edit distance is more favorable. The most favorable score(s) in each column is bold.}\label{tab:results_appendix}
\end{table*}

For each sample in the benchmark corpus, we compute individual metric score for its baseline-revised sentence and average the corpus ranking of its individual metric scores to obtain the final ranking for that sample. Table~\ref{tab:good} lists the well-treated samples (at the third percentile) in each category. Table~\ref{tab:bad} lists the cases that were not well resolved (at the 97th percentile). 

\begin{table*}[!t]
\centering
\begin{tabularx}{\textwidth}{lXX}
\hline
\textbf{Category} & \textbf{Reference(s)} & \textbf{Prediction} \\\hline
\RomanNumeralCaps{1} & \remove{The reason} she moved \remove{is} because she was offered a better position. & \remove{The reason} she moved \remove{is} because she was offered a better position. \\
\rowcolor{mygray}
\RomanNumeralCaps{2} & We outlined the letters for the banner \replace{in a careful way}{carefully}. & We outlined the letters for the banner \replace{in a careful way}{carefully}. \\
\RomanNumeralCaps{3} & \replace{The current focus of the medical profession is}{The medical profession currently focuses on} disease prevention. & \replace{The current focus of the medical profession is}{The medical profession focuses on} disease prevention.\\
\rowcolor{mygray}
\RomanNumeralCaps{4} & She fell \remove{down} \replace{due to the fact that}{because}  she hurried. & She fell \remove{down} \replace{due to the fact that}{because}  she hurried. \\
\RomanNumeralCaps{5} & We didn't renew the contract with our supplier \replace{on account of the fact that}{because} \replace{it was necessary for us to}{we had to} find a cheaper \replace{supplier}{one}. & We didn't renew the contract \remove{with our supplier} \replace{on account of the fact that}{because} it was necessary for us to find a cheaper supplier. \\
\rowcolor{mygray}
\RomanNumeralCaps{6} & Research is increasing in \remove{the field of} nutrition and food science. \emph{OR} Research \add{within nutrition and food science} is increasing \remove{the field of nutrition and food science}. & Research is increasing in \remove{the field of} nutrition and food science. \\
\RomanNumeralCaps{7} & Joe found \remove{that} \remove{the fictional novel by} Alcott\add{'s book}, Little Women, \replace{conveyed a sense of emotion and feeling}{both delightful and tragic}. & Joe found that \remove{the fictional novel by} Alcott\add{'s novel}, Little Women, conveyed \remove{a sense of} emotion and feeling. \\
\hline
\end{tabularx}
\caption{\label{tab:good}
Well-revised samples in each category (about the third percentile). Shorter sentences that require simpler actions are perfectly revised. When sentence length increases, not all wordiness can be resolved. The baseline method tends to delete and replace, even if sentences need to be rewritten.
}
\end{table*}

\begin{table*}[t!]
\centering
\small
\begin{tabularx}{\textwidth}{lXX}
\hline
\textbf{Category} & \textbf{Reference(s)} & \textbf{Prediction} \\\hline
\RomanNumeralCaps{1} & \remove{I would call your attention to the fact that} our President, \remove{who was} formerly the Governor of Arkansas, is \remove{basically} a Southerner. & I would call your attention to the fact that our President \remove{, who was formerly the Governor of Arkansas,} is a \remove{basically} Southerner. \\
\rowcolor{mygray}
\RomanNumeralCaps{2} & We have \replace{conducted an investigation}{investigated} and \replace{arrived at the conclusion}{concluded}. & We \remove{have} conducted an investigation and arrived at the conclusion. \\
\RomanNumeralCaps{3} & \replace{As far as my professor is concerned, the problem of wordiness is the thing she'd really like to see us involved with actually eliminating.}{My professor wants us to focus on eliminating wordiness.} & \remove{As far as my professor is concerned,} the problem of wordiness is the thing she'd \remove{really} like to see us involved with \remove{actually} eliminating.\\
\rowcolor{mygray}
\RomanNumeralCaps{4} & \remove{The existence of} computers \remove{and computer technology} \replace{has}{have} {improved} \replace{commercial enterprise}{business} \remove{and} \replace{information exchange}{communication}. & The existence of computers and \remove{computer} technology has \remove{greatly} influenced commercial enterprise and information exchange. \\
\RomanNumeralCaps{5} & \replace{It has come to my attention that}{I have noticed} \replace{there is a vast proliferation of undesirable vegetation}{many weeds growing} \replace{surrounding the periphery of}{around} \replace{this facility}{the building} & It has come to my attention that there is a vast proliferation of undesirable vegetation \remove{surrounding the periphery of this facility}. \\
\rowcolor{mygray}
\RomanNumeralCaps{6} & \replace{Your statement in your letter that has the contention that the information submitted and sent to you by us contained certain inaccuracies and errors has prompted us to embark on a careful and thorough reevaluation of the information submitted, with the result that the original informative data has been determined to be accurate and correct in all instances and aspects of the information.}{As you suggested, we have checked our information and confirmed its accuracy.} & Your statement in your letter that \remove{has the contention that} the information submitted and sent to you by us contained \replace{certain}{some} inaccuracies and errors has prompted us to embark on a \remove{careful and} thorough reevaluation of the information \remove{submitted}, with the result that the original informative data has been determined to be accurate and correct \remove{in all instances and aspects of the information}.
 \\
\RomanNumeralCaps{7} & \replace{In the event that}{If} \add{you get} \replace{some}{any} information \replace{concerning}{about} Mr. Smith \remove{should be brought to your attention}, \replace{it should be forwarded via mail or courier or telephone to us}{please contact us} \replace{in view of the possibility that}{in case} \replace{the information may reveal any attempt on the part of Mr. Smith to depart from the United States}{he tries to leave the country}. &	\replace{In the event that}{If} \replace{some}{any} information concerning Mr. Smith should be brought to your attention, it should be forwarded via mail or courier or telephone to us in view of the possibility that the information may reveal any attempt \remove{on the part of Mr. Smith} to depart from the United States.
 \\
\hline
\end{tabularx}
\caption{\label{tab:bad}
Badly-revised samples in each category (about the 97th percentile). These sentences are longer than sentences in Table~\ref{tab:good}. Informative part may be trimmed. Replacing nominalizations with verbs is hard. For severely wordy sentences (category \RomanNumeralCaps{6}, \RomanNumeralCaps{7}), the model fails to rewrite, and resorts to deletion. A lot of improvement is needed.
}
\end{table*}

\begin{figure}[t]
	\centering
	\includegraphics[width=0.5\textwidth]{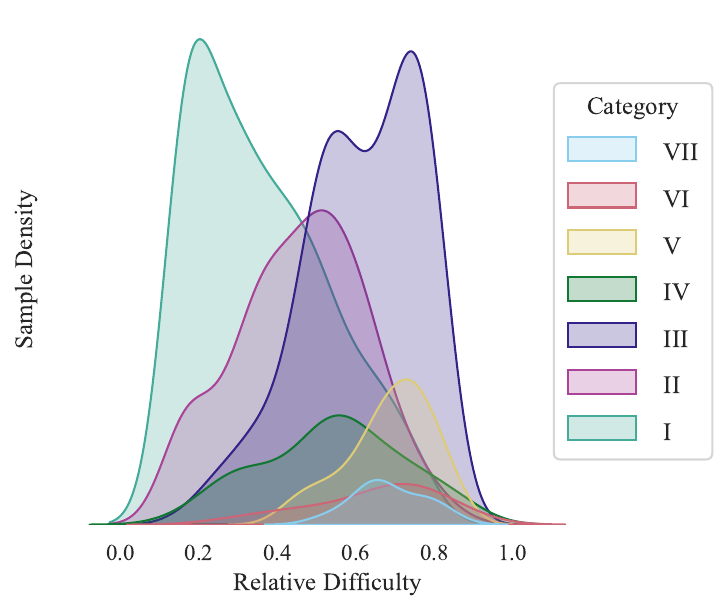}
	\caption{Difficulty faced by baseline models when dealing with sentences from different categories. This difficulty is relative to other samples in the corpus of 536 sentences. Deletion (category \RomanNumeralCaps{1}) is the least challenging. The most challenging samples are most likely from category \RomanNumeralCaps{3}. Handling sentences requiring more than one revising strategies (category \RomanNumeralCaps{4}-\RomanNumeralCaps{7}) is usually more challenging.}
	\label{fig:difficulty}
\end{figure}

Figure~\ref{fig:difficulty} shows the difficulty of revising sentences for each category. The data in Figure~\ref{fig:difficulty}, while demonstrating strengths and weaknesses of baseline, can also serve as an approximation of the difficulty of the corpus itself. The baseline model is better at deleting and replacing than rewriting due to heavy reliance on transfer learning.

\section{Applications}
A large number of realistic cases involve the recognition and generation of paraphrases. On the one hand, Question answering sites like Quora~\footnote{\url{www.quora.com}}, stackoverflow~\footnote{\url{www.stackoverflow.com}} or github issues~\footnote{\url{www.github.com/}} regularly check for duplicate questions. For example, stackoverflow has the problem that one duplicate question can account for up to 3\%~\citep{silva2018duplicate} of the total questions. Users from these websites have different expression habits and writing styles. An effective and efficient paraphrase  tool would greatly help in detecting the duplicates.

Itenticate~\footnote{\url{www.ithenticate.com}} or Grammarly~\footnote{\url{www.grammarly.com}} check paraphrases to detect text reuse or plagiarism. Some "clever" students may copy through rewriting some of the sentences, which cannot be detected at present. With this technology, these "smart copy" strategies would also be uncovered.

On the other hand, rewriting tools like Hemingway Editor~\citep{long_2013} or paraphrasetool~\footnote{\url{www.paraphrasetool.com}} can help people change the syntax of the text. The artefacts presented in this chapter will be reasonably applied to these tasks.
For example, data in this article is collected from institutional writing centres. The teacher-student ratio in the writing centre is relatively small. If there is a model that focuses on concise writing to help students get rid of most of the redundancy, it will be very helpful to improve the quality of guidance in writing.

\section{Conclusion}
We formulate sentence-level revision for concision as a constrained paraphrase generation task. The revision task not only requires semantics preservation as in usual paraphrasing tasks, but also specifies syntactic changes. A revised sentence is free of wordiness and as informative. Revising sentences is challenging and requires coordinated use of delete, replace, and rewrite. To benchmark revising algorithms, we collect 536 sentence pairs before and after revising from 72 college writing centres. We then propose a baseline Seq2Seq revising model and evaluate it on this benchmark. Despite scarcity of training data, our baseline model offers promising results for revising academic texts. We believe this corpus will drive specialized revision algorithms that benefit both authors and readers.

At the same time, the fact that existing similarity schemes cannot give very instructive scores in evaluating generated text also indicates that more detailed and comprehensive similarity schemes need to be proposed.


\chapter{Weaknesses in Existing Similarity Measures} 

\label{Chapter5} 

A long-standing paradox has plagued the task of automatic summarization. On the one hand, for about 20 years, there has not been any automatic scoring available as a sufficient or necessary condition to demonstrate summary quality, such as adequacy, grammaticality, cohesion, fidelity, etc. On the other hand, contemporaneous research more often uses one or several automatic scores to endorse a summarizer as state-of-the-art. More than 90\% of works on language generation neural models choose automatic scoring as the main basis, and about half of them rely on automatic scoring only~\citep{van2021human}. However, these scoring methods have been found to be insufficient~\citep{novikova-etal-2017-need}, oversimplified~\citep{van2021human}, difficult to interpret~\citep{sai2022survey}, inconsistent with the way humans assess summaries~\citep{rankel-etal-2013-decade,bohm-etal-2019-better}, or even contradict each other~\citep{gehrmann-etal-2021-gem,bhandari-etal-2020-metrics}.

\begin{figure}[t]
    \centering
    \includegraphics[width=\textwidth]{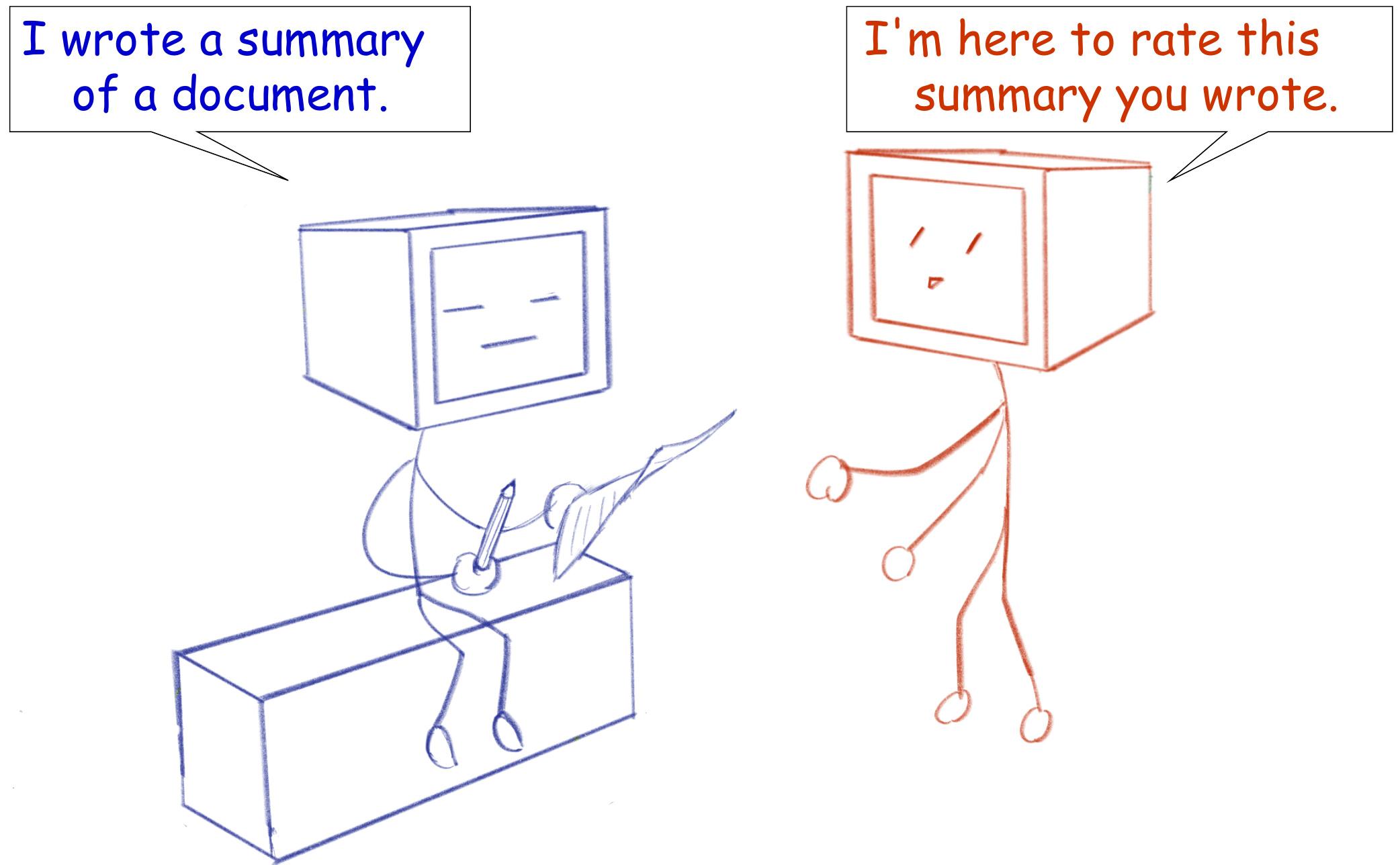}
    \caption{Automatic summarization (left) and automatic scoring (right) should be considered as two systems of the same rank, representing conditional language generation and natural language understanding, respectively. As a stand-alone system, the accuracy and robustness of automatic scoring are also important. In this study, we create systems that use bad summaries to fool existing scoring systems.}
    \label{fig:robots}
\end{figure}

Why do we have to deal with this paradox? The current work may not have suggested that summarizers assessed by automatic scoring are de facto ineffective. However, optimizing for flawed evaluations~\citep{gehrmann-etal-2021-gem,peyrard-etal-2017-learning}, directly or indirectly, ultimately harms the development of automatic summarization~\citep{narayan-etal-2018-ranking,kryscinski-etal-2019-neural,paulus2017deep}. One of the most likely drawbacks is shortcut learning (surface learning, \citealp{geirhos2020shortcut}), where summarizing models may fail to generate text with more widely accepted qualities such as adequacy and authenticity, but instead pleasing scores. Here, we quote and adapt\footnote{We underline adaptations.} this hypothetical story by \citeauthor{geirhos2020shortcut}.

\textit{"Alice loves \underline{literature}. Always has, probably always will. At this very moment, however, she is cursing the subject: After spending weeks immersing herself in the world of \underline{Shakespeare's The Tempest}, she is now faced with a number of exam questions that are (in her opinion) to equal parts dull and difficult. 'How many \underline{times is Duke} \underline{of Milan addressed}'... Alice notices that Bob, sitting in front of her, seems to be doing very well. Bob of all people, who had just boasted how he had learned the whole book chapter by rote last night ..."}

\begin{table*}[!t]
\tiny
\begin{tabularx}{\textwidth}{p{0.065\textwidth}Xp{0.28\textwidth}}
\hline
System &
  Summary &
  Document \\
\hline
Gold &
  Kevin Pietersen was sacked by   England 14 months ago after Ashes defeat. Batsman scored 170 on his county   cricket return for Surrey last week. Pietersen wants to make a sensational   return to the England side this year. But Andrew Flintoff thinks time is running   out for him to resurrect career. \hfill (ROUGE-1, ROUGE-2, ROUGE-L, METEOR, BERTScore)&
  \multirow{6}{16\baselineskip }{Andrew Flintoff fears   Kevin Pietersen is 'running out of time' to resurrect his England career. The   dual Ashes-winning all-rounder is less convinced, however, about Pietersen's   prospects of forcing his way back into Test contention. Kevin Pietersen scored   170 for Surrey in The Parks as he bids to earn a recall to the England   squad... ... Flintoff senses he no longer has age on his side. Pietersen has   not featured for England since he was unceremoniously sacked 14 months ago.   ... ... Flintoff said ... 'If he'd started the season last year with Surrey, and   scored run after run and put himself in the position... whereas now I think   he's looking at the Ashes ... ... you get the sense everyone within the England   set-up wants him as captain,' he said.' ... The former England star is hoping   to win back his Test place with a return to red ball cricket. ... ...  'this stands up as a competition.'} \\
\cellcolor{mygray}
Good \citep{liu-liu-2021-simcls}&
\cellcolor{mygray}
  Kevin   pietersen scored 170 for surrey against mccu oxford. Former england star   andrew flintoff fears pietersen is 'running out of time' to resurrect his   england career. Pietersen has been surplus to requirements since being sacked   14 months ago. Flintoff sees a bright future for 'probably the premier   tournament' in this country. \hfill(55.45, 18.18, 41.58, 40.03, 85.56) &
   \\
Broken&
  \textcolor{mypurple}{Andrew Flintoff fears Kevin Pietersen is running out of time to resurrect his England career Flintoff. Pietersen scored 170 for   Surrey in The.  Former England star   Andrew.  batsman has been .  since being sacked 14 months ago   after.  three in the.  the Ashes and he s.}  \hspace{100pt}\textcolor{white}{>}\,\hfill (\textbf{56.84}, \textbf{21.51}, \textbf{44.21}, \textbf{47.26}, 85.95) &
   \\
\cellcolor{mygray}
A dot&
\cellcolor{mygray}
  \textbf{\color{mymagenta}.} \hfill(0, 0, 0, 0, \textbf{88.47}) &
   \\
Scrambled code&
\texttt{\color{mymagenta}\textbackslash{}x03\textbackslash{}x18\$\textbackslash{}x18...\textbackslash{}x03\$\textbackslash{}x03|...\textbackslash{}x0f\textbackslash{}x01\textless{}\textless{}\$\$\textbackslash{}x04...\textbackslash{}x0e \textbackslash{}x04\# \$...\textbackslash{}x0f\textbackslash{}x0f\textbackslash{}x0f...\textbackslash{}x0e...\textbackslash{}x0f...\textbackslash{}x0f\textbackslash{}x0f\$\textbackslash{}x0f \textbackslash{}x04\textbackslash{}x0f\textbackslash{}x0f} (many tokens omitted) \hfill(0, 0, 0, 0, 87.00) &
   \\
\cellcolor{mygray}
Scrambled code + broken&
\cellcolor{mygray}
  \texttt{\color{mymagenta}\textbackslash{}x03\textbackslash{}x18\$\textbackslash{}x18...\textbackslash{}x03\$\textbackslash{}x03|...\textbackslash{}x0f\textbackslash{}x01\textless{}\textless{}\$\$\textbackslash{}x04...\textbackslash{}x0e \textbackslash{}x04\# \$...\textbackslash{}x0f\textbackslash{}x0f\textbackslash{}x0f...\textbackslash{}x0e...\textbackslash{}x0f...\textbackslash{}x0f\textbackslash{}x0f\$\textbackslash{}x0f \textbackslash{}x04\textbackslash{}x0f\textbackslash{}x0f}...    \textcolor{mypurple}{Andrew Flintoff fears Kevin Pietersen is running out of time to   resurrect his England career Flintoff.    Pietersen scored 170 for Surrey in The.  Former England star Andrew.  batsman has been .  since being sacked 14 months ago   after.  three in the.  the Ashes and he s.} (many tokens omitted)\hspace{100pt} \hfill(\textbf{56.84}, \textbf{21.51}, \textbf{44.21}, \textbf{47.26}, 87.00) &\\
\hline
  
\end{tabularx}

\caption{We created non-summarizing systems, each of which produces bad text when processing any document. Broken sentences get higher lexical scores; non-alphanumeric characters outperform good summaries on BERTScore. Concatenating two strings produces equally bad text, but scores high on both. The example is from CNN/DailyMail (for
visualization, document is abridged to keep content most consistent with the corresponding gold summary).
}
\label{tab:example}
\end{table*}

According to \citeauthor{geirhos2020shortcut}, Bob might get better grades and consequently be considered a better student than Alice, which is an example of surface learning. The same could be the case with automatic summarization, where we might end up with significant differences between expected and actual learning outcomes~\citep{paulus2017deep}. To avoid going astray, it is important to ensure that the objective is correct.

In addition to understanding the importance of correct justification, we also need to know what caused the fallacy of the justification process for these potentially useful summarizers. There are three mainstream speculations that are not mutually exclusive. (1) The transition from extractive summarization to abstractive summarization~\citep{kryscinski-etal-2019-neural} could have been underestimated. For example, the popular score ROUGE~\citep{lin-2004-rouge} was originally used to judge the ranking of sentences selected from documents. Due to constraints on sentence integrity, the generated summaries can always be fluent and undistorted, except sometimes when anaphora is involved. However, when it comes to free-form language generation, sentence integrity is no longer guaranteed, but the metric continues to be used. (2) Many metrics, while flawed in judging individual summaries, often make sense at the system level~\citep{reiter-2018-structured,gehrmann-etal-2021-gem,bohm-etal-2019-better}. In other words, it might have been believed that few summarization systems can \emph{consistently} output poor-quality but high-scoring strings. (3) Researchers have not figured out how humans interpret or understand texts~\citep{van2021human,gehrmann-etal-2021-gem,schluter-2017-limits}, thus the decision about how good a summary really is varies from person to person, let alone automated scoring. In fact, automatic scoring is more of a natural language understanding (NLU) task, a task that is far from solved. From this viewpoint, automatic scoring itself is fairly challenging.

Nevertheless, the current work is not to advocate (and certainly does not disparage) human evaluation. Instead, we argue that automatic scoring itself is not just a sub-module of automatic summarization, and that automatic scoring is a stand-alone system that needs to be studied for its own accuracy and robustness. The primary reason is that NLU is clearly required to characterize summary quality, \emph{e.g.}, semantic similarity to determine adequacy~\citep{morris-2020-second}, or textual entailment~\citep{dagan2005pascal} to determine fidelity. Besides, summary scoring is similar to automated essay scoring (AES), which is a 50-year-old task measuring grammaticality, cohesion, relevance etc. of written texts~\citep{ke2019automated}. Moreover, recent advances in automatic scoring also support this argument well. Automatic scoring is gradually transitioning from well-established metrics measuring N-gram overlap (BLEU~\citep{papineni-etal-2002-bleu}, ROUGE~\citep{lin-2004-rouge}, METEOR~\citep{banerjee-lavie-2005-meteor}, etc.) to emerging metrics aimed at computing semantic similarity through pre-trained neural models (BERTScore~\citep{zhang2019bertscore}, MoverScore~\citep{zhao-etal-2019-moverscore}, BLEURT~\citep{sellam-etal-2020-bleurt}, etc.) These emerging scores exhibit two characteristics that stand-alone machine learning systems typically have: one is that some \emph{can be fine-tuned} for human cognition; the other is that they \emph{still have room to improve} and still have to learn how to match human ratings.

Machine learning systems can be attacked. Attacks can help improve their generality, robustness, and interpretability. In particular, evasion attacks are an intuitive way to further expose the weaknesses of current automatic scoring systems. Evasion attack is the parent task of adversarial attack, which aims to make the system fail to correctly identify the input, and thus requires defence against certain exposed vulnerabilities.

In this work, we try to answer the question: do current representative automatic scoring systems really work well at the system level? How hard is it to say they do not work well at the system level? In summary, we make the following major contributions in this study:
\begin{itemize}
    \item We are the first to treat automatic summarization scoring as an NLU regression task and perform evasion attacks.
    \item We are the first to perform a \emph{universal}, \emph{targeted} attack on NLP \emph{regression} models.
    \item Our evasion attacks support that it is not difficult to deceive the three most popular automatic scoring systems simultaneously.
    \item The proposed attacks can be directly applied to test emerging scoring systems.

\end{itemize}

We develop universal evasion attacks for individual scoring system, and make sure that the combined attacker can fool ROUGE, METEOR, and BERTScore at the same time. It incorporates two parts, a white-box attacker on ROUGE, and a black-box universal trigger search algorithm for BERTScore, based on genetic algorithms. METEOR can be attacked directly by the one designed for ROUGE. Concatenating output strings from black-box and white-box attackers leads to a sole universal evasion attacking string.

\subsection{Problem formulation}

Summarization is conditional generation. A system $\sigma$ that performs this conditional generation takes an input text ($\mathbf{a}$) and outputs a text ($\hat{\mathbf{s}}$), \emph{i.e.}, $\hat{\mathbf{s}} = \sigma(\mathbf{a})$. In single-reference scenario, there is a gold reference sequence $\mathbf{s}_\text{ref}$. A summary scoring system $\gamma$ calculates the "closeness" between sequence $\hat{\mathbf{s}}$ and $\mathbf{s}_\text{ref}$. In order for a scoring system to be sufficient to justify a good summarizer, the following condition should always be avoided:
\begin{equation}\label{eq:cond}
    \gamma(\sigma_\text{far worse}(\mathbf{a}), \mathbf{s}_\text{ref}) > \gamma(\sigma_\text{better}(\mathbf{a}), \mathbf{s}_\text{ref}).
\end{equation}

Indeed, to satisfy the condition above is our attacking task. In this section, we detail how we find a suitable $\sigma_\text{far worse}$.

All (existing) automatic summary scoring are monotonic regression models. Most scoring requires at least one gold-standard text to be compared to the output from summarizers. One can opt to combine multiple available systems in one super system~\citep{lamontagne2006combining}. We will focus on the three most frequently used systems, including rule-based systems and neural systems.

\section{ROUGE and METEOR: Limitations of Lexicon Matching}
\section{Weaknesses of ROUGE}\label{sec:weakness}

Although initially developed for summaries, ROUGE can also evaluate machine translation~\citep{lin-och-2004-automatic}. In fact, we can compare every pair of sequences ($\mathbf{s}_1, \mathbf{s}_2$) using ROUGE. Moreover, each member in a sequence is an index, and an index can only be either equal or unequal when compared with another index. Two popular variants of ROUGE are ROUGE-N ($R_{\text{N}}(n, \mathbf{s}_1, \mathbf{s}_2), n \in\mathbb{Z}^+$) and ROUGE-L($R_{\text{L}}(\mathbf{s}_1, \mathbf{s}_2)$). We use bag algebra extended from set algebra~\citep{bertossi2018datalog} and calculations are as follows:

\begin{equation}\label{eq:rougebag}
    b(n, \mathbf{s}) = \set{x\mid x \text{ is an } n\text{-gram in } \mathbf{s}}_{\text{bag}},
\end{equation}

\begin{align} 
    R_{\text{N}}(n, \mathbf{s}_1, \mathbf{s}_2) &= \frac{2\cdot\abs{b(n, \mathbf{s}_1) \cap b(n, \mathbf{s}_2)}}{\abs{b(n, \mathbf{s}_1)} + \abs{b(n, \mathbf{s}_2)}}, \\
    R_{\text{L}}(\mathbf{s}_1, \mathbf{s}_2) &=
    \frac{2\cdot \abs{b(1, \text{LCS}(\mathbf{s}_1, \mathbf{s}_1))}}{\abs{b(1, \mathbf{s}_1)} + \abs{b(1, \mathbf{s}_2)}}.
\end{align}

Here, $\abs{\cdot}$ denotes the size of a bag and $\cap$ denotes  \emph{bag} intersection. An example is below.

\begin{example}[Index sequences]\label{case:numbers}
\begin{align*}
    &(83, 67, 79, \textcolor{myred}{85}, 82, 73, 78, 71), \text{and} \\
    &(83, 67, 79, 82, 73, 78, 71).
\end{align*}
\end{example}

From a bag or sequence perspective, it is fairly reasonable that two sequences in Case~\ref{case:numbers} looks similar, and ROUGE-1/2/L scores are as high as $14/15$, $10/13$, and $14/15$, respectively. However, not all sequences of this form will appear as similar. Changing each integer to its ASCII character leads to two following sequences.

\begin{example}[Letter sequences]\label{case:scouring}
\begin{align*}
    &(S, C, O, \textcolor{myred}{U}, R, I, N, G),  \text{and} \\
    &(S, C, O, R, I, N, G).
\end{align*}
\end{example}

Considering the linguistic information delivered by two sequences from Case~\ref{case:scouring}, we are not able to easily conclude similarity between them, although ROUGE scores here are the same as scores in  Case~\ref{case:numbers}. This observation highlights that linguistic information matters in sentences as well. Missing a word in a sentence may not have much impact in certain cases but at times it may shift its meaning dramatically. Here is an example of a sentence version~\citep{utah} 

\begin{example}\label{case:eat}
We could eat \remove{on} every couple of hours.
\end{example}

\begin{example}\label{case:eatchange}
We could eat \remove{on} the bench this afternoon.
\end{example}

Four cases above share the same ROUGE, but the difference within each pair varies. Linguistic features often contain sophisticated and important information, and ROUGE is unable to capture this aspect. While ROUGE misses some differences, it does not always characterize equality, either.

\begin{example}\label{case:tom}
\begin{align*}
    &\text{\textit{\replace{Tom}{Jerry} and \replace{Jerry}{Tom}} was created in 1940.}\\
    &\text{\replace{Tom}{Jerry} and \replace{Jerry}{Tom} are left homeless.}
\end{align*}
\end{example}

Swapping names is critical in the first sentence of Case~\ref{case:tom}, but not in the second sentence. Nevertheless, ROUGE is unable to highlight this difference.

These examples show that considering a sentence or summary as a sequence, a high ROUGE score ($<1$) is neither necessary nor sufficient to indicate the text similarity as understood by humans. In addition to the above, there are still unlisted complex features, such as negation, synonyms, antonyms, co-referencing, numerical approximation, etc that cannot be measured by ROGUE.

\section{ROUGE on Denser Sequence Distribution}\label{sec:distribution}
ROGUE scores can be measured by comparing a text sequence (e.g., a generated summary) with a reference sequence (e.g., a gold standard summary). However, the correlation between professional judgements and ROUGE depends on the sampled texts. Extensive sampling shows that similarities between text and references are often different from what researchers have inferred from ROUGE in the past.

We denote $\mathcal{S}$, from which we sample sequences. For each sequence $\mathbf{s} \in\mathcal{S}$, we have it member $\mathbf{s}_i\in \mathcal{V}, i \in \mathbb{N}, i < \abs{\mathbf{s}}$, where $\mathcal{V}$ is a finite vocabulary set, and $\abs{\mathbf{s}}$ denotes length of $\mathbf{s}$. Given a reference sequence $\mathbf{s}_\text{ref}$, we can calculate $R_{\text{N}}(n, \mathbf{s}_\text{ref}, \cdot)$ or $R_{\text{L}}(\mathbf{s}_\text{ref}, \cdot)$ as in Section~\ref{sec:weakness}. A discrete variable problem is thus formed.

We run genetic algorithms (GA) to search samples \citealp{holland1992genetic}. GA imitates the mechanics of natural selection and natural genetics and operates on strings, including biological structures. Strings evolve through random but structured exchanges of information to suit (survival) fitness. Each generation creates a new set of strings from reproduction, crossover, and mutation between elements that have survived the old set.

We have three objective functions, and the population of $\mathbf{s}$ is randomly initialized.

\begin{align}\label{eq:obj}
     &\min_\mathbf{s} -R_{\text{N}}(1, \mathbf{s}_\text{ref}, \mathbf{s}), \\
     &\min_\mathbf{s} -R_{\text{N}}(2, \mathbf{s}_\text{ref}, \mathbf{s}), \\
     &\min_\mathbf{s} -R_{\text{L}}(\mathbf{s}_\text{ref}, \mathbf{s}).
\end{align}

In practice, we use the library from \citet{pymoo} for multi-objective optimizing, set population size at 200, and terminate at 1000 generations. $\mathcal{V}$ is a set of byte-pair encoded words and subwords~\citep{sennrich-etal-2016-neural}. For a reference $\mathbf{s}_\text{ref}$ from CNNDM, we presume that most words required by $\mathbf{s}_\text{ref}$ can be found in its corresponding news article. This helps to reduce the size of $\mathcal{V}$ to about one tenth of its original size and accelerates searching/optimization. The length of $\mathbf{s}$ is \emph{fixed}, which is twice the average length of all $\mathbf{s}_\text{ref}$ in CNNDM, but empty tokens are allowed. Therefore, text sequences with different word counts are available. This searching is conducted for individual $\mathbf{s}_\text{ref}$.

\begin{figure}[t]
    \centering
    \includegraphics[width=0.5\textwidth]{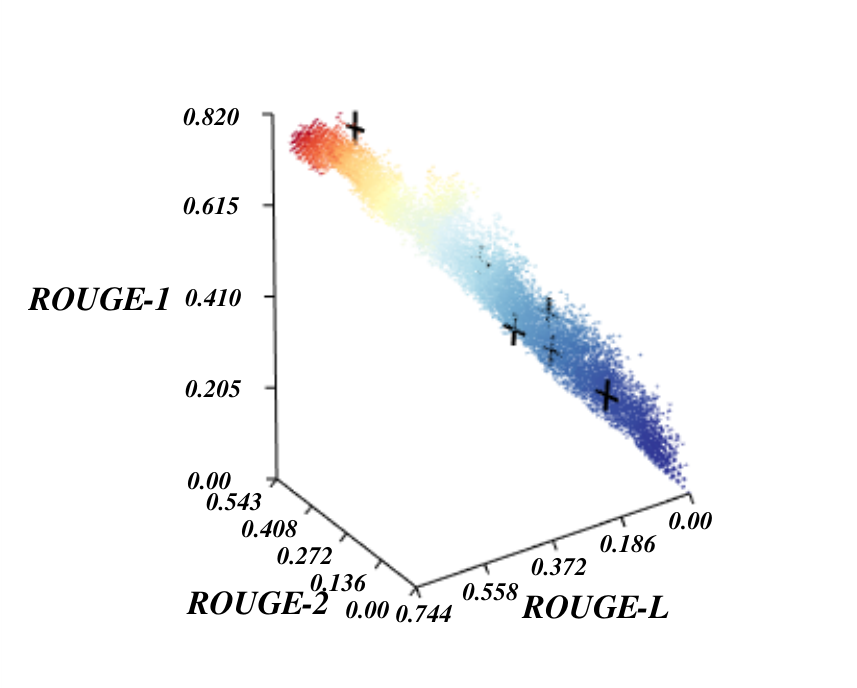}
    \caption{GA searched population of texts, black crosses mark the algorithm-output cases.}
    \label{fig:population}
\end{figure}

The genetic algorithm balances the sample size of each ROUGE score interval better than random sampling. Multi-objective optimization ensures that all ROUGE variants infer text quality consistently. Therefore, text sequences densely populate a large interval in each dimension of ROUGE-1/2/L. As seen in Figure~\ref{fig:population}, 200,000 samples are distributed within $[0, 0.82]$ along ROUGE-1, $[0, 0.55)$ along ROUGE-2, $[0, 0.75)$ along ROUGE-L. The scores of summaries predicted by previous works also appear in this interval.

In this three-dimensional space (Figure~\ref{fig:population}), the location of a text point represents its similarity with $\mathbf{s}_\text{ref}$ in terms of ROUGE. If ROUGE appropriately measures summary similarity, we can observe:
\begin{enumerate}
    \item Texts with lower scores look less "similar" to $\mathbf{s}_\text{ref}$,
    \item Two texts with close scores look "similar" to each other.
\end{enumerate}

"Similarity" here refers to human judgment. Although the determination of this similarity without extensive manual evaluation is difficult, Table~\ref{tab:neighbours}  shows intuitive examples (not edge cases) that probably violate this "similarity". First we run GA to obtain the aforementioned 200,000 samples for a $\mathbf{s}_\text{ref}$, and show their distribution in Figure~\ref{fig:population}. Then, we collect machine summaries specific to this $\mathbf{s}_\text{ref}$ and label them in Figure~\ref{fig:population} based on their ROUGE coordinates. Next, we find the nearest neighbour of each machine summary from 200,000 GA sampling points based on the Euclidean distance. The collected summaries and their respective nearest neighbours are shown in Table~\ref{tab:neighbours}.

\begin{table*}[t!]
\centering
\small
 \resizebox{\textwidth}{!}{
\begin{tabularx}{1.03\textwidth}{lX}
\hline
\rowcolor{mygray}
1  & \textit{The ch-53e copter is twice the size of the humpback whales that sometimes wash up on surrounding shores. It crouched on the sand below a bluff and only a few yards from the surf line, dwarfing surfboard-topped lifeguard vehicles parked nearby. Bing bush and his wife, julie, who live in a cliffside complex overlooking the beach, were in for a surprise.} (28.15 / 3.01 / 25.19, \citealp{see-etal-2017-get})                                                     \\
2  & \textit{Come was, super Ary a any San sun  \textcolor{lightgray}{\_} routine Station low back to  \textcolor{lightgray}{\_} Julie Old Amar  \textcolor{lightgray}{\_} civilians  \textcolor{lightgray}{\_}  \textcolor{lightgray}{\_}  \textcolor{lightgray}{\_} light Super and shot  \textcolor{lightgray}{\_}  \textcolor{lightgray}{\_} crew in The size  \textcolor{lightgray}{\_}  \textcolor{lightgray}{\_} set helicopter which video precaution injured the.} (28.07 / 3.57 / 17.54)                                                                                                                                                                                                                            \\
\rowcolor{mygray}
3  & \textit{The ch-53e landed on the shore of this northern san diego county town. The 100-foot copter is twice the size of the humpback whales that sometimes wash up on surrounding shores. Ch-53e super stallion 's crew landed on the beach in northern san diego county. Marine corps helicopter forced to make emergency landing on southern california beach.} (40.91 / 16.92 / 39.39, \citealp{chen-bansal-2018-fast})                                                                  \\
4  & \textit{No for  \textcolor{lightgray}{\_}  \textcolor{lightgray}{\_}  \textcolor{lightgray}{\_}  \textcolor{lightgray}{\_}  \textcolor{lightgray}{\_} back to Makes  \textcolor{lightgray}{\_} back for team or  \textcolor{lightgray}{\_}  \textcolor{lightgray}{\_} TO a  \textcolor{lightgray}{\_} County damage and heaviest in The landed largest  \textcolor{lightgray}{\_} landing came make  \textcolor{lightgray}{\_} Wednesday morning the.} (40.74 / 16.98 / 35.19)                                                                                                                                                                                                                                                                   \\
\rowcolor{mygray}
5  & \textit{The ch-53e super stallion landed on the shore of northern san diego county shortly after 11:30 am after a low oil-pressure indicator light went on in the cockpit. The 100-foot copter is twice the size of the humpback whales that sometimes wash up on surrounding shores. It crouched on the sand below a bluff and only a few yards from the surf line, dwarfing surfboard-topped lifeguard vehicles parked nearby.} (41.89 / 19.18 / 28.38, \citealp{liu-lapata-2019-text})    \\
6  & \textit{County beach  \textcolor{lightgray}{\_}': largest oil Pressure  \textcolor{lightgray}{\_}  \textcolor{lightgray}{\_} back to Makes  \textcolor{lightgray}{\_} landing 11 injured  \textcolor{lightgray}{\_} in awesome or down oil  \textcolor{lightgray}{\_}  \textcolor{lightgray}{\_}  \textcolor{lightgray}{\_}  \textcolor{lightgray}{\_}, TO a CH  \textcolor{lightgray}{\_} and heaviest in The landed  \textcolor{lightgray}{\_} parked  \textcolor{lightgray}{\_}  \textcolor{lightgray}{\_} 53  \textcolor{lightgray}{\_} Wednesday morning the.} (41.74 / 19.47 / 27.83)                                                                                                                                                                                                                              \\
\rowcolor{mygray}
7  & \textit{The ch-53e super stallion landed on the shore of northern san diego county shortly after 11:30 am after a low oil-pressure indicator light went on in the cockpit. The 100-foot copter is twice the size of the humpback whales that sometimes wash up on surrounding shores. The helicopter was fine to later take off and return to miramar air station.} (52.17 / 29.41 / 39.13, \citealp{liu-lapata-2019-text}))                                                              \\
8  & \textit{No one was: return the helicopter low  \textcolor{lightgray}{\_} parked back to Makes  \textcolor{lightgray}{\_} largest for Gun The helicopter town  \textcolor{lightgray}{\_} landing on: a CH  \textcolor{lightgray}{\_} and heaviest in The landed  \textcolor{lightgray}{\_} emergency landing  \textcolor{lightgray}{\_}  \textcolor{lightgray}{\_} an Wednesday morning the.} (52.25 / 29.36 / 37.84)                                                                                                                                                                                                                             \\
\rowcolor{mygray}
9  & \textit{Marine Corps helicopter made an emergency landing on a Southern California beach on Wednesday. The. Copter is twice. Humpback whales. Light: The CH-53E Super Stallion's crew landed on the beach in northern San Diego County after a low oil-pressure indicator light came on the in the cockpit. To Miramar Air Station. Was no injuries, no. } (61.53 / 34.38 / 49.23*)                                                                                      \\
10 & \textit{The ch-53e super stallion made an emergency landing on a beach in northern san diego county on wednesday morning. The crew was forced to land after a low oil-pressure indicator light went on in the cockpit. The helicopter is the largest and heaviest in the u.s. military and can carry a crew of four. No one was injured in the landing and the helicopter was fine to later take off and return to miramar air station.} (78.15 / 51.01 / 56.95, \citealp{liu-liu-2021-simcls}) \\
\rowcolor{mygray}
11 & \textit{No one was: in the helicopter San injured or  \textcolor{lightgray}{\_} .  \textcolor{lightgray}{\_} back to Makes an emergency landing The military damages in helicopter The helicopter back ,' Air Station  \textcolor{lightgray}{\_} parked on: a CH forced  \textcolor{lightgray}{\_} and heaviest in The landed the The cockpit  \textcolor{lightgray}{\_} landing came on largest beach Wednesday morning the.} (78.05 / 51.24 / 63.41)                                                                                                                                           \\\hline
12 & \textit{A CH-53E Super Stallion helicopter was forced to make an emergency landing on a north San Diego County beach Wednesday morning. The crew landed the helicopter - the largest and heaviest in the military- after a low oil pressure light came on in the cockpit. No one was injured or anything damaged in the landing. The helicopter parked on the back for four hours before taking off back to Miramar Air Station.} (Reference)    \\ \hline                  

\end{tabularx}
}
\caption{\label{tab:neighbours}
Machine summaries and the nearest neighbours of each, arranged in ascending order of ROUGE score (in brackets). Line 5 \& 7 is the extractive and abstractive summary of the same work. Line 9 is from the shortcut method
. For reading, misspelt words are replaced by "\textit{\_}", and sentences' first letters (if detectable) are capitalized.
}
\end{table*}

From the table, we observe that a generated text can obtain a high score on ROUGE-1/2/L but otherwise still provide unfaithful or corrupted information. Texts generated by the algorithms are good, and they stand out from their neighbours. Switching to other $\mathbf{s}_\text{ref}$ also showed the same trend. Therefore, ROUGE-1/2/L alone cannot reflect the sentence similarity understood by humans. Similarly, it does not accurately characterize the quality of summaries, either.

\subsection{White-box Input-agnostic Attack on ROUGE and METEOR}
\section{Methods}
We add bag algebra to formally describe both approaches because in both approaches the prediction is a bag of words $\hat{W}$, and the metric ROUGE (ROUGE-1) measures the following value:

\begin{equation}\label{intersection}
\text{ROUGE}=\frac{2\abs{\hat{W}\cap W}}{\abs{\hat{W}} + \abs{W}},
\end{equation}

where $\abs{\cdot}$ denote the size of a bag and $\hat{W}\cap W$ denotes the bag intersection between $\hat{W}$ and the ground truth bag of words $W$. Intersection of bags is defined differently than that for sets. First suppose that $b$ occurs $m$ times in bag $B_1$ and $b$ occurs $n$ times in bag $B_2$. Then $b$ occurs $\min(m,n)$ times in $B_1 \cap B_2$. For convenience, we also create a notation here for "left intersection" $\cap_\leftarrow$, such that $b$ occurs $m \times (n > 0)$ times in $B_1 \cap_\leftarrow B_2$, where $(n>0)$ is a Boolean value.

The difference of the two approaches becomes more obvious here. When we directly predict a bag of words ($\hat{W}$) from a sequence (e.g. an article) $\mathbf{x}$, $\hat{W}$ is not necessarily a subset of $A$, the bag of words of $\mathbf{x}$. Thus, it is always possible to obtain $\hat{W} = W$.

In contrast, when we apply token classification to the input sequence $\mathbf{x}$, then we have an extra constraint that $\hat{W} \subseteq A$. Thus, if $W \nsubseteq X$, we will inevitably miss some elements in the predicted $\hat{W}$. Still, we bear with this limitations for now, since if $\hat{W}$ is very close to $A\cap W$, we can obtain decent ROUGE scores.

\subsection{Direct Bag-of-words Prediction}
We can formulate abstractive summarization as a sequence-to-bag (Seq2Bag) task. In a Seq2Bag task, the input is a sequence $\mathbf{x}$ and the target is a bag of words $W$. The direct approach for Seq2Bag tasks looks for a mapping $\phi: \mathcal{X} \rightarrow \mathcal{B}$, where $\mathbf{x}\in \mathcal{X}$ and $A, W \in \mathcal{B}$. Fitting the model $\phi$ on a sample point $(\mathbf{x}, W)$ is equivalent to:

\begin{equation}\label{direct}
\begin{split}
\phi & =\argmin_{\phi} D_{\text{KL}}(P_{\hat{W}}\parallel P_{W}) \\
 & = \argmin_{\phi} \sum _{w\in  W}P_{\hat{W}}(w)\log \left({\frac {P_{\hat{W}}(w)}{P_{W}(w)}}\right) \\
\end{split}
\end{equation}

where $P_W$ is the probability distribution derived from bag of words $W$. Here we optimize $\phi$ using Kullback–Leibler divergence ($D_\text{KL}$).

\subsection{Binary Token Classification}
We create a binary tagging sequence $\mathbf{t} = \tau_{W}(\mathbf{x})$ for input sequence $\mathbf{x}$. All positively tagged tokens $\mathbf{t}_i$ form a bag $A_+$, such that $\norm{A \cap W} \subseteq A_+$ and $A_+ \subseteq (A\cap_\leftarrow W)$, where $\norm{\cdot}$ here returns a bag with distinct entries from the given bag. In other words, the positive token bag $A_+$ shares the same distinct entries with $A \cap W$, but counts of each entry may be different. This also suggests that the predicted $\hat{W}$ from binary token classification likely requires token revaluation (or multi-class classification) to determine the occurrence of each token in the bag, otherwise, even if $\hat{W}$ shares the same distinct tokens with $A \cap W$, they could vary greatly on the counts of each entry.

Binary token classification is a Seq2Seq task, though it avoids sequence writing (decoding). Formally, it looks for a mapping $\phi: \mathcal{X} \rightarrow \mathcal{T}$, where $\mathbf{x}\in \mathcal{X}$ and $\mathbf{t} \in \mathcal{T}$. The bag of positively tagged tokens is then formally define as $A_+\coloneqq \{\mathbf{x}_i | \forall \mathbf{x}_i \in \mathbf{x},  \tau_{W}(\mathbf{x})_i = 1\}$. Fitting the model $\phi$ on a sample point $(\mathbf{x}, \mathbf{t})$ is equivalent to fitting the rule-based labelling function $\tau_{W}$.

Choosing labelling function $\tau_{W}$ and determining the position of positive labels is crucial, for different $\tau_{W}$ will result in different $\phi$. We list in Table a few labelling schemes that has been explored. 

Multi-class may not be possible to predict every entry in $W$, which is similar to binary token classification. However, there are chances that we can skip the revaluation part. The benefit from this approach is also similar to that of binary token classification, which is to avoid the bottle neck.
In general, attacking ROUGE or METEOR can only be done with a white-box setup, since even the most novice attacker (developer) will understand how these two formulae calculate the overlap between two strings. We choose to game ROUGE with the most obvious bad system output (broken sentences) such that no additional human evaluation is required. In contrast, for other gaming methods, such as reinforcement learning~\citep{paulus2017deep}, even if a high score is achieved, human evaluation is still needed to measure how bad the quality of the text is.

We utilize a hybrid approach (we refer it as $\sigma_\text{ROUGE}$) of token classification neural models and simple rule-based ordering, since we know that ROUGE compares each pair of sequences ($\mathbf{s}_1, \mathbf{s}_2$) via hard N-gram overlapping. In bag algebra, extended from set algebra~\citep{bertossi2018datalog}, two trendy variants of ROUGE: ROUGE-N ($R_{\text{N}}(n, \mathbf{s}_1, \mathbf{s}_2), n \in\mathbb{Z}^+$) and ROUGE-L($R_{\text{L}}(\mathbf{s}_1, \mathbf{s}_2)$) calculate as follows:

\begin{align} 
    R_{\text{N}}(n, \mathbf{s}_1, \mathbf{s}_2) &= \frac{2\cdot\abs{b(n, \mathbf{s}_1) \cap b(n, \mathbf{s}_2)}}{\abs{b(n, \mathbf{s}_1)} + \abs{b(n, \mathbf{s}_2)}}, \\
    R_{\text{L}}(\mathbf{s}_1, \mathbf{s}_2) &=
    \frac{2\cdot \abs{b(1, \text{LCS}(\mathbf{s}_1, \mathbf{s}_1))}}{\abs{b(1, \mathbf{s}_1)} + \abs{b(1, \mathbf{s}_2)}},
\end{align}

where $\abs{\cdot}$ denotes the size of a bag, $\cap$ denotes  \emph{bag} intersection, and bag of N-grams is calculated as follows:
\begin{equation}\label{eq:rougebag2}
    b(n, \mathbf{s}) = \set{x\mid x \text{ is an } n\text{-gram in } \mathbf{s}}_{\text{bag}}.
\end{equation}


In our hybrid approach, the first step is that the neural model tries to predict the target's bag of words $b(1, \mathbf{s}_\text{ref})$, given any input $\mathbf{a}$ and corresponding target $\mathbf{s}_\text{ref}$. Then, words in the predicted bag are ordered according to their occurrence in the input $\mathbf{a}$. Formally, training of the neural model ($\phi$) is:

\begin{equation}\label{eq:train}
\min_\phi \frac{1}{\abs{\mathcal{A}}}\sum _{\mathbf{a}\in \mathcal{A}} \sum _{w\in  \mathbf{a}} H(P_\text{ref}(\cdot\mid w), P(\cdot\mid w, \phi)),
\end{equation}

where $H$ is the cross-entropy between the probability distribution of the reference word count and the predicted word count. An approximation is that the model tries to predict $b(1, \mathbf{s}_\text{ref})\cap b(1, \mathbf{a})$. Empirically, three-quarters of words in reference summaries can be found in their corresponding input texts.

Referencing the input text ($\mathbf{a}$) and predicted bag of words ($\hat{W}$) to construct a sequence is straightforward, as seen in Algorithm~\ref{alg:b2s}.

\begin{algorithm}
\caption{From bag of words to sequence}\label{alg:b2s}
\begin{algorithmic}
\Require $\mathbf{a}, \hat{W}$
\Return $\hat{\mathbf{s}}$
\State $\hat{\mathbf{s}} \gets ()$
\While{$\abs{\hat{W}} > 0$}
\State Salient Sequence $\mathbf{l} \gets (x \mid $ for $x \in \mathbf{a}$ if $[x \in \hat{W}])$
\State $\mathbf{c} \gets$ Longest Consecutive Salient Subsequence of $l$
\If{$\abs{\mathbf{c}} < C$} \Comment{Constant about 3}
\State break 
\EndIf
\State $\hat{\mathbf{s}} \gets \hat{\mathbf{s}} + \mathbf{c}$ \Comment{Concatenate $\mathbf{c}$ to $\hat{\mathbf{s}}$}
\State $\hat{W} \gets \hat{W} - \mathbf{c}$ \Comment{Remove used words}

\EndWhile
\end{algorithmic}
\end{algorithm}

Algorithm~\ref{alg:b2s} uses salient words to highlight the longest consecutive salient subsequences in $\mathbf{a}$, until the words in $\hat{W}$ are exhausted, or when each consecutive salient sequence is less than three words ($C=3$).




\section{BERTScore: A Metric with Weaknesses Common to Other Neural Models}
\subsection{Unintentional Backdoor in the Model}
Finding a $\sigma_\text{far worse}$ for BERTScore alone to satisfy condition\ref{eq:cond} is easy. A single dot (\emph{"."}) is an imitator of \emph{all} strings, as if it is a "backdoor" left by developers. We notice that, on default setting of BERTScore\footnote{\url{https://huggingface.co/metrics/bertscore}}, using a single dot can achieve around 0.892 in average when compared with any natural sentences. This figure "outperforms" all existing summarizers, making \textit{outputing a dot} a good enough $\sigma_\text{far worse}$ instance.

\subsection{Black-box Search for Universal Trigger}

This example is very intriguing because it highlights the extent of which many vulnerabilities go unnoticed, although it cannot be combined directly with the attacker for ROUGE. Intuitively, there could be various clever methods to attack BERTScore as well, such as adding a prefix to each string~\citep{wallace-etal-2019-universal,song-etal-2021-universal}. However, we here opt to develop a system that could output (one of) the most obviously bad strings (scrambled codes) to score high.

BERTScore is generally classified as a neural, untrained score~\citep{sai2022survey}. In other words, part of its forward computation (\emph{e.g.}, greedy matching) is rule-based, while the rest (\emph{e.g.}, getting every token embedded in the sequence) is not. Therefore, it is difficult to "design" an attack rationally. Gradient methods (white-box) or discrete optimization (black-box) are preferable. Likewise, while letting BERTScore generate soft predictions~\citep{jauregi-unanue-etal-2021-berttune} may allow attacks in a white-box setting, we found that black-box optimization is sufficient.

Inspired by the single-dot backdoor in BERTScore, we hypothesize that we can form longer catch-all emulators by using only non-alphanumeric tokens. Such an emulator has two benefits: first, it requires a small fitting set, which is important in targeted attacks on regression models. We will see that once an emulator is optimized to fit one natural sentence, it can also emulate almost any other natural sentence. The total number of natural sentences that need to be fitted before it can imitate decently is usually less than ten. Another benefit is that using non-alphanumeric tokens does not affect ROUGE.

Genetic Algorithm (GA, \citealp{holland1992genetic}) was used to discretely optimize the proposed non-alphanumeric strings. Genetic algorithm is a search-based optimization technique inspired by the natural selection process. GA starts by initializing a population of candidate solutions and iteratively making them progress towards better solutions. In each iteration, GA uses a fitness function to evaluate the quality of each candidate. High-quality candidates are likely to be selected and crossover-ed to produce the next set of candidates. New candidates are mutated to ensure search space diversity and better exploration. Applying GA to attacks has shown effectiveness and efficiency in maximizing the probability of a certain classification label~\citep{alzantot-etal-2018-generating} or the semantic similarity between two text sequences~\citep{maheshwary2021generating}. Our single fitness function is as follows,
\begin{equation}
    \mathbf{\hat{s}}_\text{emu} = \argmin_\mathbf{\hat{s}} -B(\hat{\mathbf{s}}, \mathbf{s}_\text{ref}),
\end{equation}
where $B$ stands for BERTScore. As for termination, we either use a threshold of -0.88, or maximum 2000 iterations.

To fit $\mathbf{\hat{s}}_\text{emu}$ to a set of natural sentences, we calculate BERTScore for each sentence in the set after each termination. We then select a proper $\mathbf{s}_\text{ref}$ to fit for the next round. We always select the natural sentence (in a finite set) that has the lowest BERTScore with the optimized $\mathbf{\hat{s}}_\text{emu}$ at the current stage. We then repeat this process till the average BERTScore achieved by this string is higher than many reputable summarizers.

Finally, to simultaneously game ROUGE and BERTScore, we concatenate $\mathbf{\hat{s}}_\text{emu}$ and the input-agnostic $\sigma_\text{ROUGE}(\mathbf{a})$. If we set the number of tokens in $\mathbf{\hat{s}}_\text{emu}$ greater than 512 (the max sequence length for BERT), $\sigma_\text{ROUGE}(\mathbf{a})$ would then not affect the effectiveness of $\mathbf{\hat{s}}_\text{emu}$, and we technically game them both. Additionally, this concatenated string games METEOR, too.

\subsection{Experiments}
We instantiate our evasion attack by conducting experiments on non-anonymized CNN/DailyMail (CNNDM, \citealp{nallapati2016abstractive,see-etal-2017-get}), a dataset that contains news articles and associated highlights as summaries. CNNDM includes 287,226 training pairs, 13,368 validation pairs and 11,490 test pairs. 

For $\sigma_\text{ROUGE}$ we use RoBERTa (base model, \citealp{liu2019roberta}) to instantiate $\phi$, which is an optimized pretrained encoding with a randomly initialized linear layer on top of the hidden states. Number of classes is set to three because we assume that each word appears at most twice in a summary. All 124,058,116 parameters are trained as a whole on CNNDM train split for one epoch. When the batch size is eight, the training time on an NVIDIA Tesla K80 graphics processing unit (GPU) is less than 14 hours. It then takes about 20 minutes to predict (including word ordering) all 11,490 samples in the CNNDM test split. Scripts and results are available at
\url{https://github.com} (URL will be made publicly available after paper acceptance).

For the universal trigger to BERTScore, we use the library from \citet{pymoo} for discrete optimizing, set population size at 10, and terminate at 2000 generations. $\mathbf{\hat{s}}_\text{emu}$ is a sequence of independent randomly initialized non-alphanumeric characters. For a reference $\mathbf{s}_\text{ref}$ from CNNDM, we start from randomly pick a summary text from train split and optimize for $\mathbf{\hat{s}}_{\text{emu}, i=0}$. We then pick the $\mathbf{s}_\text{ref}$ that is farthest away from $\mathbf{\hat{s}}_{\text{emu}, i=0}$ to optimize for $\mathbf{\hat{s}}_{\text{emu}, i=1}$, with $\mathbf{\hat{s}}_{\text{emu}, i=1}$ as initial population. Practically, we found that we can stop iterating when $i = 5$. Each iteration takes less two hours on a 2vCPU (Intel Xeon @ 2.30GHz).

\begin{landscape}
\begin{table*}[t]
\begin{tabular}{llllllll}
\hline
System                                        & ROUGE-1 & ROUGE-2 & ROUGE-L & ROUGE-A.M. & ROUGE-G.M. & METEOR & BERTScore \\
\hline
Pointer-generator(coverage)~\citeyear{see-etal-2017-get} & 39.53   & 17.28   & 36.38   & 31.06    & 29.18    & 33.1  & 86.44    \\
Bottom-Up~\citeyear{gehrmann-etal-2018-bottom}           & 41.22   & 18.68   & 38.34   & 32.75    & 30.91    & 34.2  & 87.71    \\
PNBERT~\citeyear{zhong-etal-2019-searching}             & 42.69   & 19.60   & 38.85   & 33.71    & 31.91    & \textbf{41.2}  & 87.73    \\
T5~\citeyear{raffel2019exploring}                 & 43.52   & 21.55   & 40.69   & 35.25    & 33.67    & 38.6  & \underline{88.66}    \\
BART~\citeyear{lewis-etal-2020-bart}                     & 44.16   & 21.28   & 40.90   & 35.45    & 33.75    & 40.5  & 88.62    \\
SimCLS~\citeyear{liu-liu-2021-simcls}                   & 46.67   & \textbf{22.15}   & 43.54   & \underline{37.45}    & \textbf{35.57}    & 40.5  & \textbf{88.85}    \\
\hline
Scrambled code + broken                       & \underline{46.71}   & 20.39   & \underline{43.56}   & 36.89    & 34.62    & 39.6  & 87.80     \\
Scrambled code + broken   (alter)             & \textbf{48.18}   & 19.84   & \textbf{45.35}   & \textbf{37.79}    & \underline{35.13}    & \underline{40.6}  & 87.80    \\
\hline
\end{tabular}
\caption{Results on CNNDM. Besides ROUGE-1/2/L, METEOR, and BERTScore, we also compute the arithmetic mean (A.M.) and geometric mean (G.M.) of ROUGE-1/2/L, which is commonly adopted~\citep{zhang-etal-2019-pretraining,bae-etal-2019-summary,chowdhery2022palm}. The best score in each column is in bold, the runner-up is underlined. Our attack system is compared with well-known summarizers from the past five years. The alternative version (last row) of our system changes $C$ in Algorithm~\ref{alg:b2s} from 3 to 2.}
\label{tab:result}
\end{table*}
\end{landscape}

\section{Results}
We compare ROUGE-1/2/L, METEOR, and BERTScore of our threat model with that achieved by the top summarizers in Table~\ref{tab:result}. We present two versions of threat models with a minor difference. As the results indicate, each version alone can exceed state-of-the-art summarizing algorithms on both ROUGE-1 and ROUGE-L. For METEOR, the threat model ranks second. As for ROUGE-2 and BERTScore, the threat model can score higher than other BERT-based summarizing algorithms\footnote{except MatchSum~\citep{zhong-etal-2020-extractive} and DiscoBERT~\citep{xu-etal-2020-discourse}, where our method is about 0.5 lower in ROUGE-2. We present the same results in tables with additional target thresholds in Appendix~\ref{sec:additional}}. Overall, we rank the systems by averaging their three relative ranking on ROUGE\footnote{Conservatively, We take geometric mean~\citep{chowdhery2022palm}. Combining metrics in other ways shows similar trends.}, METEOR, and BERTScore; our threat model gets runner-up (2.7), right behind SimCLS (1.7) and ahead of BART (3.3). This suggests that at the system level, even a combination of mainstream metrics is questionable in justifying the excellence of the summarizer.

These results reveal low robustness of popular metrics and how certain models can obtain high scores with inferior summaries. For example, our threat model is able to grasp the essence of ROUGE-1/2/L using a general but lightweight model, which requires less running time than summarizing algorithms. The training strategies for the model and word order are trivial. Not surprisingly, its output texts do not resemble human understandable "summaries" (Table~\ref{tab:example}).
\subsection{Related Work}

\subsection{Evasion Attacks in NLP}
In an evasion attack, the attacker modifies the input data so that the NLP model incorrectly identifies the input. The most widely studied evasion attack is the adversarial attack, in which insignificant changes are made to the input to make "adversarial examples" that greatly affect the model's output~\citep{szegedy2013intriguing}. There are other types of evasion attacks, and evasion attacks can be classified from at least three perspectives. (1) Targeted evasion attacks and untargeted evasion attacks~\citep{cao2017mitigating}. Targeted evasion attacks and untargeted evasion attacks. The former is intended for the model to predict a specific wrong output for that example. The latter is designed to mislead the model to predict any incorrect output.(2) Universal attacks and input-dependent attacks~\citep{wallace-etal-2019-universal,song-etal-2021-universal}. The former, also known as an "input-agnostic" attack, is a "unique model analysis tool". They are more threatening and expose more general input-output patterns learned by the model. The opposite is often referred to as an input-dependent attack, and is sometimes referred to as a local or typical attack. (3) Black-box attacks and white-box attacks. The difference is whether the attacker has access to the detailed computation of the victim model. The former does not, and the latter does. Often, targeted, universal, black-box attacks are more challenging. Evasion attacks have been used to expose vulnerabilities in sentiment analysis, natural language inference (NLI), automatic short answer grading (ASAG), and natural language generation (NLG)~\citep{alzantot-etal-2018-generating,wallace-etal-2019-universal,song-etal-2021-universal,filighera2020fooling,filighera2022cheating,zang-etal-2020-word,behjati2019universal}.

\subsection{Universal Triggers in Attacks on Classification}
A prefix can be a universal trigger. When a prefix is added to any input, it can cause the classifier to misclassify sentiment, textual entailment~\citep{wallace-etal-2019-universal}, or if a short answer is correct~\citep{filighera2020fooling}. These are usually untargeted attacks in a white-box setting\footnote{When the number of categories is small, the line between targeted and non-targeted attacks is blurred, especially when there are only two categories.}, where the gradients of neural models are computed during the trigger search phase.

\citeauthor{wallace-etal-2019-universal} also used prefixes to trigger a reading comprehension model to specifically choose an odd answer or an NLG model to generate something similar to an egregious set of targets. These two are universal, targeted attacks, but are mainly for classification tasks. Given that automatic scoring is a regression task, more research is needed.

\subsection{Adversarial Examples Search for Regression Models}
Compared with classification tasks in NLP, regression tasks (such as determining text similarity) are fewer and less frequently attacked. For example, the Universal Sentence Encoder (USE, \citep{cer-etal-2018-universal}) and BERTScore~\citep{zhang2019bertscore} are often taken as two constraints when searching adversarial examples for other tasks~\citep{alzantot-etal-2018-generating}. However, these regression models may also be flawed, vulnerable or not robust, which may invalidate the constraints~\citep{morris-2020-second}.

\citeauthor{morris-2020-second} shows that adversarial attacks could also threaten these regression models. For example, \citep{maheshwary2021generating} adopt a black-box setting to maximize the semantic similarity between the altered input text sequence and the original text. Similar attacks are mostly input dependent, probably because these regression models are mostly used as constraints. In contrast, universal attacks may better reveal the vulnerabilities of these regression models.

\subsection{Targeted Threshold for Attacks}
We use a threshold to determine whether a targeted attack on the regression model was successful. Intuitively, the threshold is given by the scores of the top summarizers, and we consider our attack to be successful if an attacker obtains a score higher than the threshold using clearly inferior summaries. We use representative systems that once achieved the state-of-the-art in the past five years: Pointer Generator~\citep{see-etal-2017-get}, Bottom-Up~\citep{gehrmann-etal-2018-bottom}, PNBERT~\citep{zhong-etal-2019-searching}, T5~\citep{raffel2019exploring},
BART~\citep{lewis-etal-2020-bart}, and SimCLS~\citep{liu-liu-2021-simcls}.

\section{Discussion}
\subsection{How does Shortcut Learning Come about?}
As suggested in the hypothetical story by \citeauthor{geirhos2020shortcut}, scoring draws students’ attention~\citep{filighera2022cheating} and Bob is thus considered a better student. Similarly, in automatic summarization, there are already works that are explicitly optimized for various scoring systems~\citep{jauregi-unanue-etal-2021-berttune,pasunuru-bansal-2018-multi}. Even in some cases, people subscribe more to automatic scoring than "aspects of good summarization". For example, \citet{pasunuru-bansal-2018-multi} employ reinforcement learning where entailment is one of the rewards, but in the end, ROUGE, not textual entailment, is the only justification for this summarizer.

We use a threat model to show that literally optimizing toward a flawed indicator does more harm than good. This is consistent with the findings by \citeauthor{paulus2017deep} but more often, not everyone scrutinizes the output like \citeauthor{paulus2017deep} do, and these damages can be overshadowed by a staggering increase in metrics, or made less visible by optimizing with other objectives. This is also because human evaluations are usually only used as a supplement, and it is only one per cent of the scale of automatic scoring, and how human evaluations are done also varies from group to group~\citep{van2021human}.

\subsection{Simple defence}
For score robustness, we believe that simply taking more scores as benchmark~\citep{gehrmann-etal-2021-gem} may not be enough. Instead, fixing the existing scoring system might be a better option. A well-defined attack leads to a well-defined defence. Our attacks can be detected, or neutralized through a few defences such as adversarial example detection~\citep{xu2017feature,metzen2017detecting,carlini2017adversarial}. During the model inference phase, detectors, determining if the sample is fluent/grammatical, can be applied before the input samples are scored. An even easier defence is to check whether there is a series of non-alphanumeric characters. Practically, grammar-based measures, like grammatical error correction (GEC\footnote{\url{https://github.com/PrithivirajDamodaran/Gramformer}}), could be promising~\citep{napoles-etal-2016-theres,novikova-etal-2017-need}, although they are also under development. To account for grammar in text, one can also try to parse predictions and references, and calculate F1-score of dependency triple overlap~\citep{riezler-etal-2003-statistical,clarke-lapata-2006-models}. Dependency triples compare grammatical relations of two texts. We found both useful to ensure input sanitization (Table~\ref{tab:defence}).

\begin{table}[t]
\centering
\scriptsize
\begin{tabularx}{0.5\textwidth}{Xll}
\hline
System                                           & Parse & GEC \\
\hline
Pointer-generator(coverage)~\citep{see-etal-2017-get}           & 0.131          & \underline{1.73}         \\ 
Bottom-Up~\citep{gehrmann-etal-2018-bottom}                     & 0.145          & 1.88        \\
PNBERT~\citep{zhong-etal-2020-extractive}                     & 0.179          & 2.15            \\
T5~\citep{raffel2019exploring}                        & \underline{0.198}          & \textbf{1.59} \\
BART~\citep{lewis-etal-2020-bart}                         & 0.170          & 2.07          \\
SimCLS~\citep{liu-liu-2021-simcls}                         & \textbf{0.202} & 2.17        \\
\hline
Scrambled code + broken                         & 0.168          & 2.64      \\
\hline

\end{tabularx}
\caption{\label{tab:defence}
Input sanitization checks, Parse and GEC, on the 100-sample CNNDM test split given by \citet{graham-2015-evaluating}. They penalize non-summary texts, but may introduce more disagreement with human evaluation, \emph{e.g.}, high-scoring Pointer-generator on GEC. Thus, their actual summary-evaluating capabilities on linguistic features (grammar, dependencies, or co-reference) require further investigation.
}
\end{table}

\subsection{Potential Objections on the Proposed Attacks}
\paragraph{The Flaw was Known.}
That many summarization scoring can be gamed is well known. For example, ROUGE grows when prediction length increases~\citep{sun-etal-2019-compare}. ROUGE-L is not reliable when output space is relatively large~\citep{krishna-etal-2021-hurdles}. That ROUGE correlates badly with human judgments at a system level has been revealed by findings of \citeauthor{paulus2017deep}. And, BERTScore does not improve upon the correlation of ROUGE~\citep{fabbri-etal-2021-summeval,gehrmann-etal-2021-gem}.

The current work goes beyond most conventional arguments and analyses against the metrics, and actually constructs a system that sets out to game ROUGE, METEOR, and BERTScore together. We believe that clearly showing the vulnerability is beneficial for scoring remediation efforts. From a behavioural viewpoint, each step of defence against an attack makes the scoring more robust. Compared with findings by \citeauthor{paulus2017deep}, we cover more metrics, and provide a more thorough overthrow of the monotonicity of the scoring systems, \emph{i.e.}, outputs from our threat model are significantly worse.

\paragraph{Shoddy Attack?}
The proposed attack is easy to detect, so its effectiveness may be questioned. In fact, since we are the first to see automatic scoring as a decent NLU task and attack the most widely used systems, evasion attacks are relatively easy. This just goes to show that even the crudest attack can work on these scoring systems. Certainly, as the scoring system becomes more robust, the attack has to be more crafted. For example, if the minimum accepted input to the scoring system is a "grammatically correct" sentence, an attacker may have to search for fluent but factually incorrect sentences. With contest like this, we may end up with a robust scoring system.

As for attack scope, we believe it is more urgent to explore popular metrics, as they currently have the greatest impact on summarization. Nonetheless, we will expand to a wider range of scoring and catch up with emerging ratings such as BLEURT~\citep{sellam-etal-2020-bleurt}.

\subsection{Potential Difficulties}
Performing evasion attacks with bad texts is easy, when texts are as bad as broken sentences or scrambled codes in Table~\ref{tab:example}. In this case, the output of the threat system does not need to be scrutinized by human evaluators. However, human evaluation of attack examples may be required to identify more complex flaws, such as untrue statements or those that the document does not entail. Therefore, more effort may be required when performing evasion attacks on more robust scoring systems.

\section{Packages}
For evaluation metrics, we used the following packages:
\begin{itemize}
    \item For ROUGE metrics~\citep{lin-hovy-2003-automatic}, we used the public \emph{rouge-score} package from Google Research: \\ \url{https://github.com/google-research/google-research/tree/master/rouge}
    \item For METEOR~\citep{lavie-agarwal-2007-meteor}, we used the public Natural Language Toolkit:\\
    \url{https://www.nltk.org/_modules/nltk/translate/meteor_score.html}
    \item For BERTScore~\citep{zhang2019bertscore}, we used the public \emph{datasets} package from Huggingface:\\
    \url{https://huggingface.co/metrics/bertscore}
\end{itemize}

\section{Additional Comparison with More Summarization Systems}
\label{sec:additional}
We present the same results in Table~\ref{tab:result} with additional systems in Table~\ref{tab:neighbours}. Table~\ref{tab:neighbours} also shows that about half of the listed works employ human evaluation to support the effectiveness of summarization systems.

\section{Applications}

Abstractive summarization creates a concise representation that retains most relevant information in an article. The technology is believed to have the potential to aid in legal contract analysis or financial research analysis. For example, investment banks analyse vast amounts of information. When a financial analyst reads market reports and news every day, efficient summaries can greatly improve her productivity. At present, the market believes that the summarization system can help analysts quickly obtain market signals from the content. Law and finance are both areas with a low tolerance for mistakes. In the presence of backdoors, especially when this backdoor is discovered after commercialisation, it is likely to cause irreparable economic losses.

We thus bridging the gap between the commercialisation and academic research.

The presented attack to regression model can also be used in testing the already deployed similarity checkers like Ithenticate\footnote{\url{www.ithenticate.com}} to uncover vulnerabilities, threats, risks in such software applications and prevent malicious text changes used to evade plagiarism detection.

\section{Conclusion}
We hereby answer the question: it is easy to create a threat system that simultaneously scores high on ROUGE, METEOR, and BERTScore using worse text. In this work, we treat automatic scoring as a regression machine learning task and conduct evasion attacks to probe its robustness or reliability. Our attacker, whose score competes with top-level summarizers, actually outputs non-summary strings. This further suggests that current mainstream scoring systems are not a sufficient condition to support the plausibility of summarizers, as they ignore the linguistic information required to compute sentence proximity. Intentionally or not, optimizing for flawed scores can prevent algorithms from summarizing well. The practical effectiveness of existing summarizing algorithms is not affected by this, since most of them optimize maximum likelihood estimation. Based on the exposed vulnerabilities, careful fixes to scoring systems that measure summary quality and sentence similarity are necessary.



\chapter{Linguistic Basis in Similarity Estimation} 

\label{Chapter6} 

\section{Bidirectional Implicature as Similarity}

Entailment is a phenomena between two linguistic expressions. It is sometimes called inference or implication, but overall stands for whether the truth value of one expression can be determined from another expression. Formally, an entailment is:

\begin{equation}\label{eq:entail}
    e: \mathcal{U}, \mathcal{U} \to \set{0, 1}
\end{equation}

The flip side of determining entailment is determining exclusivity/contrary, defined as:
\begin{equation}
\begin{aligned}
    c : &\ \mathcal{U}, \mathcal{U} \to \set{0, 1},\\
    \text{s.t. } & c(u_1, u_2) + e(u_1, u_2) \leq 1.   
\end{aligned}
\end{equation}

Determining the output value of $e$ is a task usually called \textit{recognizing textual entailment}, where the determination of output of $c$ is taken into account as well. Formula~\ref{eq:entail} looks identical to formula~\ref{eq:paraphrase_recognition}, except that entailment is not a symmetric phenomenon, i.e., $e(u_1, u_2) = e(u_2, u_1)$ is not necessary.

\citet{dagan1999contextual} then decomposed paraphrase recognition into a pair of entailment recognition as shown below:

\begin{equation}
    f(u_1, u_2) = e(u_1, u_2) \times e(u_2, u_1)
\end{equation}

From then on, a pair of paraphrases can be formally explained as a pair of expression with mutual entailment. However, if it was not for the extension of entailment determination to a formalisation of conditional probability~\citep{chen-etal-2020-uncertain}, we would still not be able to use formula from \citet{dagan1999contextual} to calculate similarity, as similarity is a regression task. \citet{chen-etal-2020-uncertain} offer an alternative version of formula~\ref{eq:entail},

\begin{equation}\label{eq:uncertain}
    e: \mathcal{U}, \mathcal{U} \to \mathbb{R} \cap [0, 1].
\end{equation}

The conditional probabilistic formalisation of entailment/inference considers how likely one expression is to hold if the other is true.

In formula 2.X, similarity between language units has been defined as a bivariate function that maps two units into a real number. Although in most cases a similarity scheme is a symmetric function where its value is the same no matter the order of its arguments, some schemes are not symmetric, such as METEOR~\citep{banerjee-lavie-2005-meteor} or Tversky index~\citep{tversky1977features}.

In this way, the conditional probability between two expressions can be regarded as an asymmetric measure of similarity. The problem of estimating similarity is successfully transformed into a problem of estimating (probabilistic) inference, where more studies can be found. This chapter studies the ways to determine of an entailment/inference relation.

\section{The Controversial Nature of Natural Language Inference}
\label{sec:nli}
Natural language inference has flourished over the past three decades, but there is no clear community consensus on its nature. Disagreements are reflected in deciding whether the inference is based on semantics or pragmatics, the strength of the inference, the form of the inference, and the relevance of the inference to the "real" world. Differing views on the nature of inferences can further lead to differing views on whether a particular inference holds. Therefore, a system may face difficulties in evaluating reasonably under different perspectives. We propose an interpretation of the inference that takes advantage of predicate logic and natural deduction to resolve seemingly inconsistencies between different viewpoints.

There is language, and then there are language-based inferences. There are often expressions that can be put together, but less often what constitutes an inference. Therefore, although there are reasonable inferences, expressions that do not constitute inferences are often mixed in.

Inferences between expressions often have stated and/or unstated assumptions. Although there is reasonable experimental accuracy in judging whether the inference is valid or not without knowing its assumption, it is difficult to make a stable judgment in real applications. The following passage is from the Analytical Writing section of the GRE General Test, which asks test-takers to identify possible loopholes in the argument.

\textit{"\textcolor{mypurple}{Most homes in the northeastern United States, where winters are typically cold, have traditionally used oil as their major fuel for heating. Last heating season that region experienced 90 days with below-normal temperatures, and climate forecasters predict that this weather pattern will continue for several more years. Furthermore, many new homes are being built in the region in response to recent population growth.} Because of these trends, we predict \textcolor{mymagenta}{an increased demand for heating oil} and ..."}

The argument seems convincing at first, and many high-ranking natural language inference (NLI) systems predict that the \textcolor{mypurple}{premise} supports, or so-called "entails", the \textcolor{mymagenta}{hypothesis}. However, many assumptions can be unwarranted. For instance, had people in the northeast United States not used the oil as their major fuel for heating during the last season when the region experienced below-normal temperature, the argument predicting increased demand for oil would become ill-founded.

We are not delusional to have an NLI system state clearly why an inference holds or not right now. Instead, we just need a simple answer. The misjudgment of the above case is just one of many evasion examples~\citep{cao2017mitigating}, and we have never been short of evasion attack examples ~\citep{alzantot-etal-2018-generating,behjati2019universal,wallace-etal-2019-universal,zang-etal-2020-word,song-etal-2021-universal} or stress test samples~\citep{saha-etal-2020-conjnli,hossain-etal-2020-analysis}. These examples highlights the current difficulties for NLI systems to make good judgments.

We believe that at least one of the reasons for the misjudgment of the NLI system is that there is no consensus in the community on the interpretation of inference~\citep{zaenen-etal-2005-local,manning2006local,poliak-2020-survey}. The rationale is to imagine an NLI system calibrated on test suite A and then run on test suite B; if test suites A and B are fundamentally different in labelling, the system may be considered poor in terms of robustness or even accuracy. In other words, B is actually A's out-of-distribution adversarial examples~\citep{sehwag2019analyzing,zhou-tan-2021-investigating}, and one of the causes of this distribution difference is the difference in the interpretation of inference. 

In this work, we analyze possible interpretations of inference from four perspectives: whether inference refers to a semantic or pragmatic level; the strength of the inference; the form of the inference; and the dependence of the inference on real-world knowledge. We then refer to natural deduction to come up with an interpretation that dissolves as much of the opposition between existing interpretations as possible. Relying on the proposed interpretation, we relabelled some existing data sample annotation, fine-tuned NLI systems, and compared the differences before and after the fine-tuning: we found that the adjusted systems were less likely to judge the validity of the inference through excessive imagination.

\subsection{NLI: Is It about Semantics or Pragmatics?}
The classical classification between semantics and pragmatics is that semantics deals with linguistic meaning and pragmatics with the use of language~\citep{morris1938foundations, kamp1978semantics}. A feature that can be derived is that expressions with different semantics may be pragmatically the same, while expressions with the same semantics, even the same expression, may be pragmatically different when used in different contexts.

Earlier, textual inference\footnote{Being called NLI, recognising textual entailment (RTE), or even recognising textual inference~\citep{poliak-2020-survey}, its name has always been a mystery.} is defined as a task that captures major \emph{semantic} inference needs across applications~\citep{dagan2005pascal}. \citet{zaenen-etal-2005-local} also follows the standard approach within linguistic semantics. Since the first large annotated corpus (SNLI, \citealp{bowman-etal-2015-large}), most works are about lexical and compositional semantics~\citep{williams-etal-2018-broad}.

While it appears that textual inference is only within the semantics, there is disagreement. The task defined by \citeauthor{dagan2005pascal} is hypothesized to be a "generic task for evaluating and comparing applied semantic inference models". Particularly, paraphrase acquisition (acquiring sets of lexical-syntactic expressions that convey largely equivalent or entailing meanings) can be cast in terms of textual inference. \citet{hirst2003paraphrasing} believes paraphrases "talk about the same situation in a different way" and argues that paraphrases are not fully synonymous~\citep{bhagat-hovy-2013-squibs}. In the following simple example~\citep{iyer2017qqp}:
\begin{example}
How can I be a good geologist? $\xleftrightarrow{?}$ What should I do to be a great geologist?
\end{example}

Literally, the two questions are different, and one of the many reasons is that a \textit{good geologist} may not be a \textit{great geologist}. Nevertheless, when both are asked on Quora, they count as the same question. Hence, the same dilemma about what is paraphrase also applies to what is textual inference.

Now that the meaning or use of a pair of texts can lead to different judgments, arguments against treating textual inference simply as a semantic task also include that standard theories of linguistic semantics are sometimes inappropriate for the task
~\citep{manning2006local}. \citeauthor{manning2006local} does not make it clear that textual inference is about pragmatics, either. "What a human would be happy to infer from a piece of text", rather than meaning or use, is considered the correct understanding of the text. 
This is probably the loosest definition of this task.

\subsection{Inference Strength}

A less loose workaround that seems to allow both semantic and pragmatic inference is to relax from demanding \emph{valid arguments} to \emph{strong arguments}. \citet{dolan2005microsoft} consider expressions that are "\textit{more or less} semantically equivalent" as paraphrases. Classifying valid arguments with invalid but strong arguments avoids excluding "all but the most trivial" inferential relationships. \citet{chen-etal-2020-uncertain} further refine the task by explicitly determining the probability that a hypothesis can be inferred from premises. Three benefits of this refinement are: first, the boundary between valid and invalid arguments/inferences is still clear; second, it circumvents the tricky question: with what probability must the hypothesis follow the premises for the inference to count as "strong"; third, it avoids making a slippery slope argument, if the probability estimates are correct. However, one question remains: how can we determine the probability? In the following example:

\begin{example}
90\% of humans are right-handed. Pat is human. $\xrightarrow{?}$ Pat is right-handed.
\end{example}

Pat is possibly one of the 10\% who are not \textit{right-handed}. Yet, the argument in this example provides good reasons to think that "Pat is right-handed", and "Pat" is 90\% likely to be "right-handed". Other probabilities are less obvious, as shown below.

\begin{example}\label{ex:boy}
A boy hits a ball, with a bat. $\xrightarrow{?}$  The kid is playing in a baseball game.
\end{example}

\citeauthor{chen-etal-2020-uncertain} estimate that this argument/inference is 78\% strong, and \citeauthor{bowman-etal-2015-large} think the inference hold true. In Example~\ref{ex:boy}, probabilities are explicit in text and the annotator has to guess, which actually creates more uncertainty and uninterpretability. In fact, many premise-hypothesis pairs in U-SNLI~\citep{chen-etal-2020-uncertain} with probability higher than 78\% are not considered valid/strong inferences in SNLI~\citep{bowman-etal-2015-large}.

\subsection{Relation to Knowledge}


\begin{example}\label{k1}
90\% of humans are \replace{right}{left}-handed. Pat is human. $\xrightarrow{?}$ Pat is \replace{right}{left}-handed.
\end{example}

\begin{example}\label{k2}
Pat is human. $\xrightarrow{?}$ Pat is right-handed.
\end{example}

\begin{example}\label{k3}
Pat is human. $\xrightarrow{?}$ Pat is \replace{right}{left}-handed.
\end{example}

The probability for premise in Example~\ref{k1} to infer its hypothesis is still 90\%. The premise is "factually" false~\citep{enwiki:1091909435} and the hypothesis is "factually" only 10\% likely to be true, but the argument is "strong". This is because the inference is 90\% true when the premise is plausible.

Without presupposed knowledge, neither Example~\ref{k2} nor \ref{k3} presents a strong argument, and the premises and hypotheses seem so independent that probabilities cannot be calculated. Ignoring knowledge is used to criticize the view that textual inferences are strictly logical entailment~\citep{manning2006local}. Presuppositions of knowledge may be necessary and unavoidable~\citep{zaenen-etal-2005-local}. With presupposed knowledge, these two examples have probabilities of 90\% and 10\%, respectively.

\subsection{Forms of Inference}
Standard form of textual inference is to determine how likely premises ($P$) infer a hypothesis ($H$). Both $P$ and $H$ are natural languages, and the likelihood could be in binary values (valid or invalid) or real numbers from 0 to 1~\citep{chen-etal-2020-uncertain}.

We may replace some premises with presupposed knowledge. If the premise ends up being empty, the inference task becomes a "fact" verification task (without looking for evidence). Besides textual bias~\citep{tsuchiya-2018-performance,gururangan-etal-2018-annotation}, this could be another explanation for the "hypothesis-only model"~\citep{poliak-etal-2018-hypothesis} performing twice as well as the "majority baseline" on SNLI. In addition to binarization, the likelihood of inference is sometimes 3-way~\citep{cooper1996using} or 4-way~\citep{valencia1991studies}.

\section{Natural Deduction Preliminary}

We then use the following natural deduction rules to search appropriate variants of $S$ or $N$:

\begin{align}
    \lnot\lnot P &\equiv P, \\
    P\lor Q &\equiv Q \lor P, \\
    P\land Q &\equiv Q\land P, \\
    (P\lor Q)\lor R &\equiv P\lor (Q\lor R), \\
    (P\land Q)\land R &\equiv P\land (Q\land R), \\
    P\lor (Q\land  R) &\equiv(P\lor Q)\land  (P\lor R),\\
    P\land (Q\lor R) &\equiv(P\land  Q)\lor (P\land  R), \\
    \lnot (P\lor Q) &\equiv\lnot P\land \lnot Q, \\
    \lnot (P\land Q) &\equiv\lnot P\lor \lnot Q,\\
\end{align}

\begin{align}
    \lnot \exists x\,P &\equiv \forall x\,(\lnot P),\\
    \lnot \forall x\,P &\equiv \exists x\,(\lnot P),\\
    \forall x\,(P\land Q) &\equiv \forall x\,P\land \forall x\,Q,\\
    \exists x\,(P\lor Q) &\equiv \exists x\,P\lor \exists x\,Q,\\
    \forall x(A(x)\lor B)&\equiv \forall x\,A(x)\lor B,\\
    \forall x(A(x)\land B)&\equiv \forall x\,A(x)\land B,\\
    \exists x(A(x)\lor B)&\equiv \exists x\,A(x)\lor B,\\
    \exists x(A(x)\land B)&\equiv \exists x\,A(x)\land B,\\
    \forall x\,A(x)\lor \forall x\,B(x)&\Rightarrow \forall x(A(x)\lor B(x)),\\
    \exists x(A(x)\land B(x))&\Rightarrow \exists x\,A(x)\land \exists x\,B(x)
\end{align}

\begin{align}
    P\lor (P\land Q) &\equiv P,\\
    P\land (P\lor Q) &\equiv P,\\
    P\lor \lnot P &\equiv \top,\\
    P\land \lnot P &\equiv \bot
\end{align} 

\begin{align}
    P\land Q &\Rightarrow P,\\
    P\land Q &\Rightarrow Q,\\
    P &\Rightarrow P\lor Q,\\
    Q &\Rightarrow P\lor Q,\\
    \lnot P, P\lor Q &\Rightarrow Q,\\
    P, P\to Q &\Rightarrow Q,\\
    \lnot Q, P\to Q &\Rightarrow \lnot P,\\
    P\to Q, Q\to R &\Rightarrow P\to R,\\
    P\lor Q, P\to R, Q\to R &\Rightarrow R
\end{align}

\section{Inference by Making Decisions about Assumptions}
\label{sec:assu}

We believe that the key to using knowledge while retaining strict semantic reasoning is to introduce assumptions. The process of reasoning is never about knowledge, but about assumptions. Explicitly stated assumptions are also called premises, and here we introduce unstated assumptions.

\begin{example}\label{shapiro}
About two weeks before the trial started, I was in Shapiro's office in Century City. $\xrightarrow{?}$ Shapiro works in Century City. 
\end{example}

In RTE2 Guidelines~\citep{haim2006second}, this example is very "probable" and judged as "YES". Its sample justification is that "although Shapiro's office is in Century City, he actually never arrives to his office, and works elsewhere. However, this interpretation is \textit{very unlikely}, and so the entailment holds with \textit{high probability}."

Is that "Shapiro works in Shapiro's office" really "very probable"? One may always \emph{assume} this to be true or false, but its truth value could change. Suppose Shapiro gets a job offer at the end of 2020 when people can work from home, taking that "Shapiro works in Shapiro's office" as granted can be ignorant. We do not mean to suggest here that Example~\ref{shapiro} \emph{should} be labelled as "No". Rather, we propose to transform it into two pairs:
\begin{enumerate}
    \item About ..., I was in Shapiro's office in Century City. \add{\textit{Most} people work in each of their offices.} $\xrightarrow{?}$ Shapiro works in Century City. 
    \item About ..., I was in Shapiro's office in Century City. \add{\textit{Some} people work in each of their offices.} $\xrightarrow{?}$ Shapiro works in Century City. 
\end{enumerate}

This way, the first inference may be correct most of the time, while the second inference is sometimes correct. On top of that, inference now only deals with semantics. The assumption likelihood is responsible for the hypothesis likelihood, and the process of inference can follow natural deduction. Although formalisms differ, an underlying idea of natural deduction is that one can "make an assumption $S$ and see that it leads to conclusion $X$", and then conclude that if the $S$ were true, then so would $X$ be~\citep{sep-natural-deduction}.

\begin{table*}[!t]
\centering
\tiny
\begin{tabular}{lll}
\hline
\textbf{\#} & \textbf{Expression} & \textbf{Actn.}\\
\hline
0 & \textbf{Premise} A turtle danced. \textbf{Hypothesis} A turtle moved.&  \\
1 & \textbf{P} $\exists\, e_1\, x_1.\, (Actor(e_1,x_1) \land v_\text{dance}(e_1) \land n_\text{turtle}(x_1))$.  \textbf{H} $\exists\, e_1\, x_1.\, (Actor(e_1,x_1) \land v_\text{move}(e_1) \land n_\text{turtle}(x_1))$. & Parse \\
2 & $S'$: $\lnot\exists\, e_1\, x_1.\, (Actor(e_1,x_1) \land v_\text{dance}(e_1) \land n_\text{turtle}(x_1)) \lor \exists\, e_1\, x_1.\, (Actor(e_1,x_1) \land v_\text{move}(e_1) \land n_\text{turtle}(x_1))$&  \\
3 & $S'$: $\exists\,e_1\,x_1.(v_\text{move}(e_1) \land n_\text{turtle}(x_1) \land Actor(e_1,x_1)) \lor \forall e_1\,x_1.(\lnot v_\text{dance}(e_1) \lor \lnot n_\text{turtle}(x_1) \lor \lnot Actor(e_1,x_1))$ & Sort \\
4 & $S''$: $\forall\,x. (v_\text{dance}(x) \to v_\text{move}(x))$ & Simp.\\
5 & \textbf{Assumption} Dancing is (a kind of) moving. & Text\\
\hline
\end{tabular}
\caption{\label{tab:exp}
Derive an assumption $S''$ from premise and hypothesis. First, parsing textual premises and hypotheses results in first-order logic (FOL) expressions. Next, we derive $S'$ from parsed premises and hypotheses. Then, there could be more than one way to simplify $S'$.
}
\end{table*}

The undetermined conclusion $X$ is ($P \to H$). We need to find an appropriate assumption ($S$), and if it is plausible ($\top(S)$), $(P \to H)$ is plausible. If $P$ is given, $S$ is a \emph{sufficient} assumption. Formally,

\begin{align}
    P \land S &\vdash H, \\
    P \land S &\nvdash \bot, \\
    S &\nvdash H.
\end{align}



We use natural deduction to obtain $S$.
\begin{equation}
\begin{aligned}
    P \land S &\vdash H, & \\
    P, S &\vdash H &\text{(Simplification)}, \\
    P &\vdash S \to H &\text{(Modus Ponens)}, \\
    P, \lnot H &\vdash \lnot S &\text{(Modus Tollens)}, \\
    P \land \lnot H &\vdash \lnot S&\text{(Conjunction)}.
\end{aligned}
\end{equation}

Hence, we know that any $S$ whose negation is entailed by $P \land \lnot H$ is a sufficient assumption. Moreover, any $S''$ satisfying $\lnot S' \vdash \lnot S''$ is a sufficient assumption, if $S'$ is a sufficient assumption. We can simply apply conjunction elimination on $S'$ to obtain $S''$ or more variants, as shown in Table~\ref{tab:exp}.

While the sample label in Table~\ref{tab:exp} is not affected by the proposed annotations, other samples may need to be re-annotated. That "there are two young ladies jogging, by the \textit{ocean side}" is thought to infer that "two women are jogging by the \textit{beach}"~\citep{bowman-etal-2015-large}. Even if we only ask for a strong argument and not a valid one, we will find it unreasonable. Assuming that 80\% of the world's ocean sides have beaches, we would happily infer "Peter cannot have three daughters" from "Peter has three children"~\citep{simonoff_2010}. The latter example is correct about 85\% of the time, but it would be a little odd to use it as an inference. Therefore, if we cannot assume "people must be at beaches if they are by the ocean side", the sample label will be "non-entailed".

\subsection{Semantic Parsing in Textual Inference}
\label{subsec:parse}

Semantic parsing, particularly discourse representation structure (DRS) parsing~\cite{bos-2008-wide}, has been used to translate natural languages into first-order logic expressions. Ideally, parsed premises and hypotheses can help derive all required assumptions. However, current DRS parsing systems may either make mistakes in negation parsing, or behave less stably when designed in neural models~\cite{liu-etal-2021-universal}. More refinement on the parsing system is needed. \citet{bos-markert-2005-recognising} also directly applied DRS parsing in textual inference, by manually generating about 20 different axioms. These works suggest that textual inference and semantic parsing closely related.

\subsection{Assumption Simplification}
\label{subsec:simp}
Simplifying $S'$ may result in several equivalent or non-equivalent expressions $S''$. The sample proof with NLTK Resolution Prover\footnote{Tableau Prover presents similar results, though its proof is longer. \url{https://www.nltk.org/howto/inference.html}} for the expression in Table~\ref{tab:exp} is shown below.

\begin{lstlisting}
[ 1] {-v_dance(z4), v_move(z4)}                        A 
[ 2] {v_dance(z9)}                                     A 
[ 3] {n_turtle(z10)}                                   A 
[ 4] {Actor(z12,z11)}                                  A 
[ 5] {-v_move(z13), -n_turtle(z14), -Actor(z13,z14)}   A 
[ 6] {v_move(z9)}                                      (1, 2) 
[ 7] {-n_turtle(z14), -Actor(z13,z14), -v_dance(z13)}  (1, 5) 
[ 8] {-n_turtle(z14), -Actor(z13,z14)}                 (2, 7) 
[ 9] {-Actor(z13,z14), -v_move(z13)}                   (3, 5) 
[10] {-Actor(z13,z14), -v_dance(z13)}                  (1, 9) 
[11] {-Actor(z13,z14)}                                 (2, 10) 
[12] {-Actor(z13,z14), -v_dance(z13)}                  (3, 7) 
[13] {-Actor(z13,z14)}                                 (2, 12) 
[14] {-Actor(z13,z14)}                                 (3, 8) 
[15] {-v_move(z13), -n_turtle(z14)}                    (4, 5) 
[16] {-n_turtle(z14), -v_dance(z13)}                   (1, 15) 
[17] {-n_turtle(z14)}                                  (2, 16) 
[18] {-v_move(z13)}                                    (3, 15) 
[19] {-v_dance(z13)}                                   (1, 18) 
[20] {}                                                (2, 19) 
\end{lstlisting}

\subsection{Stereotyped Ideas in Existing Annotation}
\label{subsec:stereo}

Besides the few examples stated in Section~\ref{sec:assu}, more stereotyped inferences have been observed in SNLI~\cite{bowman-etal-2015-large}. We show them in Table~\ref{tab:stereo} and \ref{tab:stereo2}.

\begin{table*}[!t]
\centering

\begin{tabularx}{\textwidth}{lXl}
\hline
\textbf{\#} & \textbf{Expression} & \textbf{Why stereotyped}\\
\hline
1 & \textbf{Premise} A car is loaded with items on the top.	 \textbf{Hypothesis} The car is a convertible. \textbf{SNLI label} Contradiction. \textbf{Assumption} Can convertible cars ever get loaded on their top? Yes, they can. & 
\begin{minipage}{.25\textwidth}
  \includegraphics[width=\linewidth]{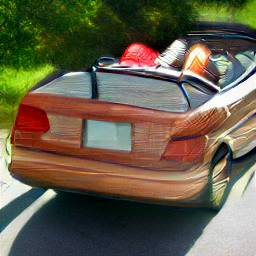}
\end{minipage}  \\
\hline
2 & \textbf{Premise} A family walking with a soldier.	 \textbf{Hypothesis} A group of people strolling. \textbf{SNLI label} Entailment. \textbf{Assumption} Do family members have to stroll, if they walk together with a soldier? No, they do not have to. & 
\begin{minipage}{.25\textwidth}
  \includegraphics[width=\linewidth]{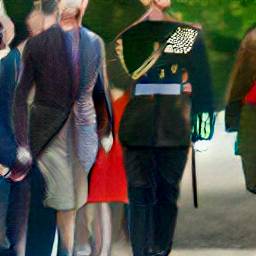}
\end{minipage}  \\
\hline
3 & \textbf{Premise} A man is wearing a blue and yellow racing uniform while holding a bottle. \textbf{Hypothesis} This guy is jumping rope. \textbf{SNLI label} Contradiction. \textbf{Assumption} Does one have to give up his bottle or stop jumping rope when he wears a blue and yellow racing uniform? No, he does not. & 
\begin{minipage}{.25\textwidth}
  \includegraphics[width=\linewidth]{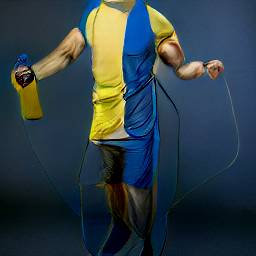}
\end{minipage}  \\
\hline
4 & \textbf{Premise} Young Asian girl is sitting on the ground in rubble.	 \textbf{Hypothesis} The young Asian girl is outside in the rubble. \textbf{SNLI label} Entailment. \textbf{Assumption} Must rubble be outside? No. & 
\begin{minipage}{.25\textwidth}
  \includegraphics[width=\linewidth]{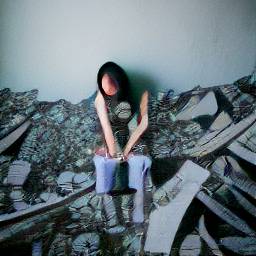}
\end{minipage}  \\
\hline
5 & \textbf{Premise} A woman wearing sunglasses is frowning.	 \textbf{Hypothesis} A woman wearing sunglasses is not smiling. \textbf{SNLI label} Contradiction. \textbf{Assumption} Have you ever seen someone smiles with frowning? Yes, try forced smile. & 
\begin{minipage}{.25\textwidth}
  \includegraphics[width=\linewidth]{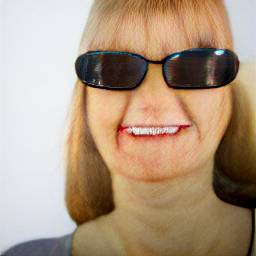}
\end{minipage}  \\
\hline
\end{tabularx}
\caption{\label{tab:stereo}
We ask with what assumption(s) would the given label hold. Then we assess the plausibility of these assumptions. Reasons or explanations are shown in the right-most column. The images are generated from Dalle-mini \url{https://huggingface.co/spaces/dalle-mini/dalle-mini}.
}
\end{table*}

\begin{table*}[!t]
\centering

\begin{tabularx}{\textwidth}{lXl}
\hline
\textbf{\#} & \textbf{Expression} & \textbf{Why stereotyped}\\
\hline
6 & \textbf{Premise} A statue at a museum that no seems to be looking at.  \textbf{Hypothesis} Tons of people are gathered around the statue. \textbf{SNLI label} Contradiction. \textbf{Assumption} Could tons of people around a statue doing something else? Yes, they could. & 
\begin{minipage}{.25\textwidth}
  \includegraphics[width=\linewidth]{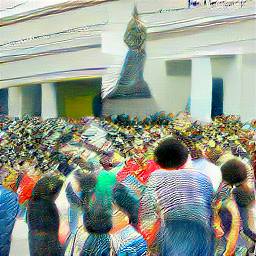}
\end{minipage}  \\
\hline
7 & \textbf{Premise} A blond-haired doctor and her African american assistant looking threw new medical manuals.  \textbf{Hypothesis} A man is eating pb and j. \textbf{SNLI label} Contradiction. \textbf{Assumption} Can a doctor eat pb and j while doing something else? Yes, they can. & 
\begin{minipage}{.25\textwidth}
  \includegraphics[width=\linewidth]{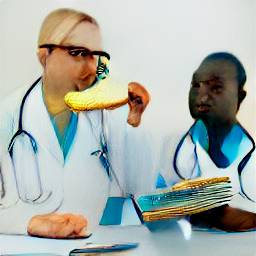}
\end{minipage}  \\
\hline
8 & \textbf{Premise} A young family enjoys feeling ocean waves lap at their feet.  \textbf{Hypothesis} A family is out at a restaurant. \textbf{SNLI label} Contradiction. \textbf{Assumption} Are there any beach restaurants where ocean waves are just around? Yes, there could be. & 
\begin{minipage}{.25\textwidth}
  \includegraphics[width=\linewidth]{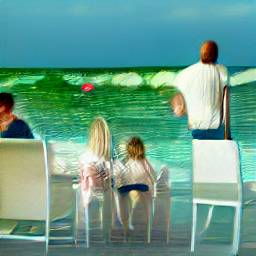}
\end{minipage}  \\
\hline
9 & \textbf{Premise} A person wearing a straw hat, standing outside working a steel apparatus with a pile of coconuts on the ground. 	 \textbf{Hypothesis} A person is burning a straw hat. \textbf{SNLI label} Contradiction. \textbf{Assumption} Can one wear a hat and burn another? Yes. & 
\begin{minipage}{.25\textwidth}
  \includegraphics[width=\linewidth]{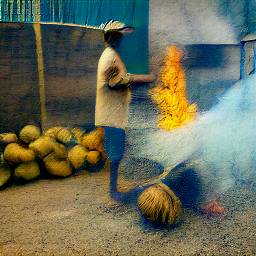}
\end{minipage}  \\
\hline
10 & \textbf{Premise} Man chopping wood with an axe.		 \textbf{Hypothesis} The man is outside. \textbf{SNLI label} Entailment. \textbf{Assumption} Must one chop wood outdoor/outside? Not really. & 
\begin{minipage}{.25\textwidth}
  \includegraphics[width=\linewidth]{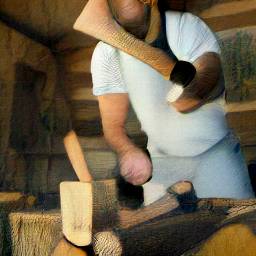}
\end{minipage}  \\
\hline
\end{tabularx}
\caption{\label{tab:stereo2}
(Continue Table~\ref{tab:stereo}) We ask with what assumption(s) would the given label hold. Then we assess the plausibility of these assumptions. Reasons or explanations are shown in the right-most column. The images are generated from Dalle-mini \url{https://huggingface.co/spaces/dalle-mini/dalle-mini}.
}
\end{table*}

\section{Experiment and Result}
We investigate how the proposed annotation scheme affects an NLI system by re-annotating a subset of SNLI~\citep{bowman-etal-2015-large}. This subset contains all test samples in U-SNLI~\citep{chen-etal-2020-uncertain}, and 210 samples from U-SNLI train split which are labelled as "contradiction" or "entailment" in SNLI. This is a similar amount compared with re-annotations by \citet{hossain-etal-2020-analysis} and \citet{kumar-talukdar-2020-nile}. In short, we relabel a sample that was initially "mistaken" as a strong argument as a weak argument. As a result, 20\% samples experience a label change. Comparing both original labels and new labels with the probabilities given by U-SNLI, we observe that the Spearman correlation coefficient ($\rho$, \citealp{zwillinger1999crc}) increases by 2.5\% from originally 0.8715 to 0.8933 now. We then fine-tuned existing RoBERTa~\citep{liu2019roberta} model\footnote{\url{https://huggingface.co/ynie/roberta-large-snli_mnli_fever_anli_R1_R2_R3-nli}} with the relabelled training split, and the accuracy increases from 0.8148 to 0.8185 on the relabelled test split (Similarly, the accuracy drops from 0.9185 to 0.9148 when tested on original labels). 


To test on real-world applications, we applied the model with and without fine-tuning on 27 GRE passages like the example in Section~\ref{sec:nli}. While all 27 samples should be "non-entailed", we observe that the model without fine-tuning predicts 14 out of 27 as "entailed", leading to an error rate much higher than 0.09 (the reported error rate on SNLI test split). After fine-tuning, the error rate drops from $\frac{14}{27}$ to $\frac{4}{9}$.

\section{Conclusion}

Ambiguity about "what is the right thing to infer from text" weakens robustness of NLI systems. In real-world applications, we observe that NLI models tend to "over-infer" from the premises. We propose separating the plausibility of the assumption from the validity of the argument, since the truth value of the assumption may vary from person to person. The proposed scheme simultaneously preserves strict semantic deduction and the participation of world knowledge. We verify its usage by fine-tuning an existing NLI system with the new annotations and observe an improvement in precision. We hope this work will facilitate construction of future NLI corpora and exploration of the robustness of NLI systems.




\chapter{Conclusions} 

\label{conclusions} 

\section{Conclusions}

In this thesis, I introduce a methodology for measuring the similarity between human language units. This methodology can be used to test and improve schemes that measure similarity between human language units.

In chapter 2, before describing these practical works, I systematically review about 70 existing schemes for measuring similarity from the perspective of mathematical form. In general, notwithstanding their deeper nature in abstract algebra, I classify their formalisation into four distinct types, the first three of which represent language units in some form and the last not. These are sets, sequences, and real vectors. Afterwards, standard similarity formulae such as the Jaccard index can be applied. As for the latter type, which accounts for a large part, the similarity between two linguistic units is simply a bivariate function without any abstract representation of the units. This is the most flexible, but can be difficult to interpret in many cases.

In chapter 3, I examine how existing schemes work in practice by studying their effectiveness in applications. I first choose to apply word embedding similarity in topic modelling on Twitter. The original ClusTop algorithm~\citep{lim2017clustop} is improved by using the word embedding similarity as edge weights to fill in the vertices of the network graph, and then applying community detection to partition the network graph. I empirically study the impact of multiple variants of the improved ClusTop algorithm with word embedding similarity as edge type, original variants with unigram or dyad co-occurrence as edge weights, and various LDA baselines. Empirical evaluations are based on metrics such as topic coherence, pointwise mutual information, precision, recall, and F-score. Experimental results based on three Twitter datasets with labelled topics (crises and events) show that the ClusTop algorithm with refined edges outperforms various LDA baselines and the original Clustop on these evaluation metrics.

In chapter 4, I look at the problem of whether we can to take these similarity values themselves very seriously if they are not to be rounded into zero or one. Existing similarity schemes, when used for topic modelling, have demonstrated their efficacy. However, my work in this chapter suggests that if our demand is similarity score itself rather than an external or indirect assessment like topic modelling. These similarity schemes are stress-tested using sentence pairs with high semantic similarity and low syntactic similarity. The results show that when the two sentences are more syntactically different, the scale reported by the semantic similarity scheme also tends to be lower. Therefore, current semantic similarity schemes still largely depend on the syntactic or lexical relationships between the language units being compared.

In chapter 5, I study the problem of how specific vulnerabilities of a similarity scheme can be uncovered such that we can refine the scheme. Stress tests can expose some of the shortcomings of similarity schemes, but merely providing such well-formed sentence pairs does little to improve the scheme. For example, fine-tuning a scheme with such sentence pairs only results in a general rise in all semantic similarity scores and a sink in all syntactic similarity scores. Therefore, I turn my attention to evasion attacks. Evasion attacks aim to manipulate the input to a model such that the model misjudges, whereas attacking a regression model is all about getting a high score on an input that should have gotten a low score or vice versa. We choose three of the most popular similarity schemes to attack, namely ROUGE~\citep{lin-2004-rouge}, METEOR~\citep{banerjee-lavie-2005-meteor}, and BERTScore~\citep{zhang2019bertscore}. In particular, we perform a universal attack, also known as an input-agnostic attack, where the manipulating policy is determined before the inputs to the regression model are known. Universal attacks are known to be better at illustrating the vulnerability of tested models. The flow of the whole attack is as follows. These regression models are used to score the output of the summarizers, and their work is to give a higher similarity judgment between the reference and the output given by the advanced summarizer, while predicting a lower score for the similarity between the reference and the output of the inferior summarizers. I take the state-of-the-art summarizer performance as a threshold, and try to create a worse summarizer to get a higher similarity than the threshold, where a worse summarizer means its output is clearly bad to a human. In practice, the systems I create either generate nonsense strings or generate scrambling codes, but the scores for these outputs are surprisingly high. These bad outputs are generated through token classification and word ordering, or black-box string optimisation.

For these discovered vulnerabilities, we provide corresponding defence methods, namely patching similarity schemes through dependency parsing, grammatical error correction, or similarity schemes other than the three attacked. Targeted defence and attack together constitute the proposed progressive-refinement framework for similarity schemes, and for each repair under this framework, we know exactly which aspect has been fixed.

In chapter 6, I look at what is the proper way for humans to determine the gold reference for this regression task. The continued refinement of the similarity scheme eventually leads to the question of how humans should determine similarities between language units. In the past, the gold reference value for similarity scheme was determined by one or a few manually labelled datasets. In fact, when it comes to each example, people's opinions can be very different. For example, there are controversies around the condition for an entailment to hold. Rather than pursue the philosophical question of what makes two units similar, as \citet{landauer1997solution} did, researchers tacitly acquiesced in transforming the definition of similarity into a somewhat guesswork~\citep{manning2006local}. To this end, I propose an annotating scheme for humans, that one must elaborate her assumption before judging the semantic similarity between two or more language units. Moreover, once the assumptions are given, natural deduction of first-order logic, and/or conditional probability can be applied to rigorously derive the similarity between language units.

In summary, I am gradually digging into the task of similarity measurement, from its application, testing, repair, and even reference determination. I appreciate the convenience and speedy application that existing similarity schemes can bring us, such as topic modelling, but also see their shortcomings in confirming details. I come up with a progressive approach to refining a vulnerable similarity scheme, but find that waiting at the end of the refined similarity measure was an equally imperfect reference scheme. Hence, the fundamental problem is finally revealed, we must first confirm how we humans intend to calculate similarity, before we can hand this task to the computer, otherwise, we can only stay on the approximate binary judgment of similarity. As a deeply philosophical, cognitive and computational science question, this task remains to be further explored.

While studying the similarity, I have three other contributions in this thesis. The first is a natural language generation task, revision for concision, designed to aid academic writing efforts, and I provide a baseline sequence-to-sequence model. The second is that, as far as I know, my attack on similarity scoring is the first attack on a regression model in NLP. The third is that the most immediate indication of these similarity schemes' vulnerability as summary raters is that summarizers who demonstrate validity based on these similarity scores lose (this part of) their justification.


\section{Applications}

The work in this thesis finds its realistic applications in the industry as follows. As mentioned in prior chapters, these works have great potential to contribute to bridging the gap between academia and industry, and even get commercialised to serve a larger population.

\paragraph{Applications of Topic Modelling}

In short, topic modelling is a text mining technique used to discover and categorise topics in documents. Recommendation systems are very popular these days and systems are algorithms aimed at suggesting relevant items to users. Topic modelling is a core part of this~\citep{mcgowan_2021}. Applying the presented topic modelling algorithm can effectively provide more accurate similar content recommendations (such as movie recommendations), and thus potentially benefit the service providers.

Topic modelling can also be used for spam detection. Due to the characteristics of spam itself, their subject similarity is very high, the use of subject modelling can effectively improve the capture rate of spam, reduce user confusion, and improve user experience.

\paragraph{Applications of Concise Paraphrases}
As mentioned in the Chapter~\ref{Chapter4}, the collected concise writing data is from institutional academic writing. Thus the intuitive commercial application of similar models can reduce teaching costs while improving teaching quality. In the future, the precise writing model has the potential to become a must-have learning tool for students like Edimension.

Our approach is not limited to helping academic writing but can be extended to other areas. Such as the detection of advanced plagiarism.

\paragraph{Applications of Abstractive Summarization}
Abstractive summarization has room for almost every field involving text. Potential future areas for abstracts include, but are not limited to, patent research, video content abstracts, medical case studies, and financial research.

Google has reportedly been working on projects trying to understand fiction. Abstracts can help consumers quickly understand what a book is about during the buying process. Current evaluation systems with shortcuts lead to models that learn shortcuts rather than true summaries. This may be an important reason to understand that the fiction system has not been pushed out.

Another often overlooked area of abstract abstraction is helping people with dyslexia. In the Internet age, a large amount of information is difficult for normal people to digest and process, and people with dyslexia are even more unable to obtain information effectively and are further out of touch with society. A Lupin summary model can help dyslexic people reduce their consumption to obtain information.


\section{Future Work}

The immediate next step based on this thesis is the progressive refinement of existing similarity schemes. When simple bugs are fixed, more realistic and complex vulnerabilities may surface. Besides, the refined similarity scheme can better supervise language generation tasks, and thus better generative models can be developed. After attacking the summarization system metric, we plan to propose a new metric that is robust to adversarial attacks including the attack discussed previously. Before that, we may propose some mitigating methods to patch this system.

After proposing the concise writing task, it would be necessary to provide strongly intuitive baselines and well-justified evaluating systems, which I will continue to follow up on.

In addition to working on practical tasks, the analysis of abstract representations of similarity can also deepen the understanding of this cognitive primitive.
 

\appendix 



\chapter{Dataset Concise-536 in Chapter~\ref{Chapter4}} 

\label{AppendixA} 

\section{All Sentence pairs of Concise-536}
Sentences are from college writing centres.\\

\textbf{\textcolor{red}{Wordy: }}The politician talked about several of the merits of after-school programs in his speech.

\textbf{\textcolor{blue}{Concise: }}The politician touted after-school programs in his speech. \cite{purdue_writing_lab2021}

\vspace{12pt}

\textbf{\textcolor{red}{Wordy: }}Suzie believed but could not confirm that Billy had feelings of affection for her.

\textbf{\textcolor{blue}{Concise: }}Suzie assumed that Billy adored her. \cite{purdue_writing_lab2021}

\vspace{12pt}

\textbf{\textcolor{red}{Wordy: }}Our Web site has made available many of the things you can use for making a decision on the best dentist.

\textbf{\textcolor{blue}{Concise: }}Our website presents criteria for determining the best dentist. \cite{purdue_writing_lab2021}

\vspace{12pt}

\textbf{\textcolor{red}{Wordy: }}Working as a pupil under someone who develops photos was an experience that really helped me learn a lot.

\textbf{\textcolor{blue}{Concise: }}Working as a photo technician's apprentice was an educational experience. \cite{purdue_writing_lab2021}

\vspace{12pt}

\textbf{\textcolor{red}{Wordy: }}The teacher demonstrated some of the various ways and methods for cutting words from my essay that I had written for class.

\textbf{\textcolor{blue}{Concise: }}The teacher demonstrated methods for cutting words from my essay. \cite{purdue_writing_lab2021}

\vspace{12pt}

\textbf{\textcolor{red}{Wordy: }}Eric Clapton and Steve Winwood formed a new band of musicians together in 1969, giving it the ironic name of Blind Faith because early speculation that was spreading everywhere about the band suggested that the new musical group would be good enough to rival the earlier bands that both men had been in, Cream and Traffic, which people had really liked and had been very popular.

\textbf{\textcolor{blue}{Concise: }}Eric Clapton and Steve Winwood formed a new band in 1969, ironically naming it Blind Faith because speculation suggested that the group would rival the musicians' previous popular bands, Cream and Traffic. \cite{purdue_writing_lab2021}

\vspace{12pt}

\textbf{\textcolor{red}{Wordy: }}Many have made the wise observation that when a stone is in motion rolling down a hill or incline that that moving stone is not as likely to be covered all over with the kind of thick green moss that grows on stationary unmoving things and becomes a nuisance and suggests that those things haven't moved in a long time and probably won't move any time soon.

\textbf{\textcolor{blue}{Concise: }}A rolling stone gathers no moss. \cite{purdue_writing_lab2021}

\vspace{12pt}

\textbf{\textcolor{red}{Wordy: }}For each and every book you purchase, you will receive a free bookmark.

\textbf{\textcolor{blue}{Concise: }}For every book you purchase, you will receive a free bookmark. \cite{north_carolina2021}

\vspace{12pt}

\textbf{\textcolor{red}{Wordy: }}Because a great many of the words in this sentence are basically unnecessary, it would really be a very good idea to edit somewhat for conciseness.

\textbf{\textcolor{blue}{Concise: }}Because many of the words in this sentence are unnecessary, we should edit it. \cite{north_carolina2021}

\vspace{12pt}

\textbf{\textcolor{red}{Wordy: }}The reason for the failure of the basketball team of the University of North Carolina in the Final Four game against the team from Kansas was that on that day and at that time, some players were frequently unable to rebound the ball.

\textbf{\textcolor{blue}{Concise: }}UNC's basketball team lost the Final Four game against Kansas because it could not consistently rebound the ball. \cite{north_carolina2021}

\vspace{12pt}

\textbf{\textcolor{red}{Wordy: }}Do not try to anticipate in advance those events that will completely revolutionize society.

\textbf{\textcolor{blue}{Concise: }}Do not try to anticipate revolutionary events. \cite{north_carolina2021}

\vspace{12pt}

\textbf{\textcolor{red}{Wordy: }}In the event that going out for the purpose of eating with them cannot be avoided, it is necessary that we first go to the ATM, in light of the fact that I am out of cash.

\textbf{\textcolor{blue}{Concise: }}If we must go out to eat with them, we should first go to the ATM because I am out of cash. \cite{north_carolina2021}

\vspace{12pt}

\textbf{\textcolor{red}{Wordy: }}If you do not have more than five years of experience, do not call for an interview if you have not already spoken to human resources.

\textbf{\textcolor{blue}{Concise: }}Applicants with more than five years of experience can bypass human resources and call for an interview. \cite{north_carolina2021}

\vspace{12pt}

\textbf{\textcolor{red}{Wordy: }}The 1780 constitution of Massachusetts was written by John Adams.

\textbf{\textcolor{blue}{Concise: }}John Adams wrote the 1780 Massachusetts Constitution. \cite{north_carolina2021}

\vspace{12pt}

\textbf{\textcolor{red}{Wordy: }}The letter was taken to the mailbox by Sally.

\textbf{\textcolor{blue}{Concise: }}Sally took the letter to the mailbox. \cite{north_carolina2021}

\vspace{12pt}

\textbf{\textcolor{red}{Wordy: }}The manager is deficient in interpersonal skills and invests minimal time in assisting the clerks to develop their expertise.

\textbf{\textcolor{blue}{Concise: }}The manager lacks interpersonal skills and spends little time helping the clerks develop their skills. \cite{monash2020}

\vspace{12pt}

\textbf{\textcolor{red}{Wordy: }}Coral reefs can be damaged by rapid and substantial climatic changes.

\textbf{\textcolor{blue}{Concise: }}Rapid and substantial climatic changes can damage coral reefs. \cite{monash2020}

\vspace{12pt}

\textbf{\textcolor{red}{Wordy: }}The current upsurge in stakeholder dissatisfaction with the outcomes of local government decision-making is at least partially a consequence of the predilection against long-term planning.

\textbf{\textcolor{blue}{Concise: }}The current rise in stakeholder dissatisfaction with local government decisions is at least partly due to the lack of long-term planning. \cite{monash2020}

\vspace{12pt}

\textbf{\textcolor{red}{Wordy: }}The next step will be to undertake a thorough analysis of the results.

\textbf{\textcolor{blue}{Concise: }}The next step will be to thoroughly analyse the results. \cite{monash2020}

\vspace{12pt}

\textbf{\textcolor{red}{Wordy: }}Regular reviews of online content should be scheduled.

\textbf{\textcolor{blue}{Concise: }}Online content should be reviewed regularly. \textbf{ OR } Companies should review online content regularly. \cite{monash2020}

\vspace{12pt}

\textbf{\textcolor{red}{Wordy: }}One important factor to consider is the age at which language instruction begins.

\textbf{\textcolor{blue}{Concise: }}One important consideration is the age at which language instruction begins. \cite{monash2020}

\vspace{12pt}

\textbf{\textcolor{red}{Wordy: }}The novel is preoccupied with matters such as post-colonial society and its mores, race, madness, and family relationships.

\textbf{\textcolor{blue}{Concise: }}The novel is preoccupied with post-colonial society and its mores, race, madness, and family relationships. \cite{monash2020}

\vspace{12pt}

\textbf{\textcolor{red}{Wordy: }}In the event that dividends continue to fall, it will be necessary to reduce staff numbers.

\textbf{\textcolor{blue}{Concise: }}If dividends continue to fall, it will be necessary to reduce staff numbers. \cite{monash2020}

\vspace{12pt}

\textbf{\textcolor{red}{Wordy: }}The opinion of the working group was that the budget for the project had been set too low.

\textbf{\textcolor{blue}{Concise: }}The working group's opinion was that the project budget had been set too low. \cite{monash2020}

\vspace{12pt}

\textbf{\textcolor{red}{Wordy: }}There are two goals for the project; to estimate the future energy needs of the South Pacific and to make recommendations as to the most viable renewable energy production options suitable for the region.

\textbf{\textcolor{blue}{Concise: }}The project goals are to estimate the future energy needs of the South Pacific and to recommend the most viable renewable energy production options for the region. \cite{monash2020}

\vspace{12pt}

\textbf{\textcolor{red}{Wordy: }}Measurement of scanner performance can be achieved through examination of four criteria; resolution, bit-depth, dynamic range and software.

\textbf{\textcolor{blue}{Concise: }}Scanner performance can be measured by examining four criteria: resolution, bit-depth, dynamic range and software. \cite{monash2020}

\vspace{12pt}

\textbf{\textcolor{red}{Wordy: }}In high school, where I had the opportunity for three years of working with the student government, I realized how significantly a person's enthusiasm can be destroyed merely by the attitudes of his superiors.

\textbf{\textcolor{blue}{Concise: }}In high school, during three years on the student council, I saw students' enthusiasm destroyed by insecure teachers and cynical administrators. \cite{gmu2021}

\vspace{12pt}

\textbf{\textcolor{red}{Wordy: }}The poverty of Anne Moody was also a crucial factor in the formation of her character.

\textbf{\textcolor{blue}{Concise: }}Anne Moody's poverty also helped form her character. \cite{gmu2021}

\vspace{12pt}

\textbf{\textcolor{red}{Wordy: }}Frequently, a chapter in a book reveals to the reader the main point that the author desires to bring out during the course of the chapter.

\textbf{\textcolor{blue}{Concise: }}A chapter's title often reveals its thesis. \cite{gmu2021}

\vspace{12pt}

\textbf{\textcolor{red}{Wordy: }}In both Orwell's and Baldwin's essays, the feeling of white supremacy is very important.

\textbf{\textcolor{blue}{Concise: }}Both Orwell and Baldwin trace the consequences of white supremacy. \cite{gmu2021}

\vspace{12pt}

\textbf{\textcolor{red}{Wordy: }}The scene is taking place at night, in front of the capitol building.

\textbf{\textcolor{blue}{Concise: }}The scene takes place at night, in front of the capitol building. \cite{gmu2021}

\vspace{12pt}

\textbf{\textcolor{red}{Wordy: }}The friar is knowledgeable about Juliet being alive.

\textbf{\textcolor{blue}{Concise: }}The friar knows that Juliet is alive. \cite{gmu2021}

\vspace{12pt}

\textbf{\textcolor{red}{Wordy: }}There are two pine trees which are growing behind this house.

\textbf{\textcolor{blue}{Concise: }}Two pine trees grow behind this house. \cite{gmu2021}

\vspace{12pt}

\textbf{\textcolor{red}{Wordy: }}Any student could randomly sit anywhere.

\textbf{\textcolor{blue}{Concise: }}Students could sit anywhere. \cite{gmu2021}

\vspace{12pt}

\textbf{\textcolor{red}{Wordy: }}Housing for married students is not unworthy of consideration.

\textbf{\textcolor{blue}{Concise: }}Housing for married students is worthy of consideration. \cite{gmu2021}

\vspace{12pt}

\textbf{\textcolor{red}{Wordy: }}This is a quote from Black Elk's autobiography that discloses his prophetic powers.

\textbf{\textcolor{blue}{Concise: }}This quote from Black Elk's autobiography discloses his prophetic powers. \cite{gmu2021}

\vspace{12pt}

\textbf{\textcolor{red}{Wordy: }}It is frequently considered that Hamlet is Shakespeare's most puzzling play.

\textbf{\textcolor{blue}{Concise: }}Hamlet is frequently considered Shakespeare's most puzzling play. \cite{gmu2021}

\vspace{12pt}

\textbf{\textcolor{red}{Wordy: }}It has been argued by Stargill that there is no topic in education on which there is greater agreement and consensus among educators than on the need for parental involvement in classrooms (Stargill, 2009).

\textbf{\textcolor{blue}{Concise: }}Educators agree on the need for parental involvement in classrooms (Stargill, 2009). \cite{waldenu2021}

\vspace{12pt}

\textbf{\textcolor{red}{Wordy: }}Scholars, researchers, and writers have recommended and promoted student assessment as a means by which to address the achievement gap.

\textbf{\textcolor{blue}{Concise: }}Researchers have recommended student assessment to address the achievement gap. \cite{waldenu2021}

\vspace{12pt}

\textbf{\textcolor{red}{Wordy: }}There are 30 participants who volunteered for the study.

\textbf{\textcolor{blue}{Concise: }}Thirty participants volunteered for the study. \cite{waldenu2021}

\vspace{12pt}

\textbf{\textcolor{red}{Wordy: }}I feel that the study is very significant to scholars in psychology.

\textbf{\textcolor{blue}{Concise: }}The study is significant to psychology scholars. \cite{waldenu2021}

\vspace{12pt}

\textbf{\textcolor{red}{Wordy: }}We will be home in a period of ten days.

\textbf{\textcolor{blue}{Concise: }}We will be home in ten days. \cite{uagc2021}

\vspace{12pt}

\textbf{\textcolor{red}{Wordy: }}It has come to my attention that there is a vast proliferation of undesirable vegetation surrounding the periphery of this facility.

\textbf{\textcolor{blue}{Concise: }}I have noticed many weeds growing around the building. \cite{uagc2021}

\vspace{12pt}

\textbf{\textcolor{red}{Wordy: }}A decision to vote on the issue was made by the committee this week.

\textbf{\textcolor{blue}{Concise: }}This week, the committee decided to vote on the issue. \cite{uagc2021}

\vspace{12pt}

\textbf{\textcolor{red}{Wordy: }}Jon will let me know in the event that he can get away and make the trip.

\textbf{\textcolor{blue}{Concise: }}Jon will call me if he can go. \cite{uagc2021}

\vspace{12pt}

\textbf{\textcolor{red}{Wordy: }}There are four officers who report to the captain.

\textbf{\textcolor{blue}{Concise: }}Four officers report to the captain. \cite{uagc2021}

\vspace{12pt}

\textbf{\textcolor{red}{Wordy: }}Luis was interested in the data processing field.

\textbf{\textcolor{blue}{Concise: }}Luis was interested in data processing. \cite{uagc2021}

\vspace{12pt}

\textbf{\textcolor{red}{Wordy: }}I said that I was tired.

\textbf{\textcolor{blue}{Concise: }}I said I was tired. \cite{uagc2021}

\vspace{12pt}

\textbf{\textcolor{red}{Wordy: }}I stepped off of the curb.

\textbf{\textcolor{blue}{Concise: }}I stepped off the curb. \cite{uagc2021}

\vspace{12pt}

\textbf{\textcolor{red}{Wordy: }}I got up on the ladder.

\textbf{\textcolor{blue}{Concise: }}I got on the ladder. \cite{uagc2021}

\vspace{12pt}

\textbf{\textcolor{red}{Wordy: }}Many people prefer Marvel Comics  (to DC Comics), due to the fact that they explore more characters in their canon.

\textbf{\textcolor{blue}{Concise: }}Many people prefer Marvel Comics (to DC Comics) because they explore more characters in their canon. \cite{uis2021}

\vspace{12pt}

\textbf{\textcolor{red}{Wordy: }}Marvel fans watch Agents of S.H.I.E.L.D. in order to expand their knowledge of the Marvel canon.

\textbf{\textcolor{blue}{Concise: }}Marvel fans watch Agents of S.H.I.E.L.D. to expand their knowledge of the Marvel canon. \cite{uis2021}

\vspace{12pt}

\textbf{\textcolor{red}{Wordy: }}There are many examples why Marvel is the stronger comic.

\textbf{\textcolor{blue}{Concise: }}Many examples support why Marvel is the stronger comic. \cite{uis2021}

\vspace{12pt}

\textbf{\textcolor{red}{Wordy: }}It is clear that DC has superior superheroes in its canon.

\textbf{\textcolor{blue}{Concise: }}DC has superior superheroes in its canon. \cite{uis2021}

\vspace{12pt}

\textbf{\textcolor{red}{Wordy: }}I believe that Channing Tatum may not make the best Gambit in the upcoming movie.

\textbf{\textcolor{blue}{Concise: }}Channing Tatum may not make the best Gambit in the upcoming movie. \cite{uis2021}

\vspace{12pt}

\textbf{\textcolor{red}{Wordy: }}My favorite superheroes are both Batman and Iron Man.

\textbf{\textcolor{blue}{Concise: }}My favorite superheroes are Batman and Iron Man. \cite{uis2021}

\vspace{12pt}

\textbf{\textcolor{red}{Wordy: }}Man of Steel and Batman vs. Superman disappointed at the box office, and the reason why is because people are tired of origin stories.

\textbf{\textcolor{blue}{Concise: }}Man of Steel and Batman vs. Superman disappointed at the box office because people are tired of origin stories. \cite{uis2021}

\vspace{12pt}

\textbf{\textcolor{red}{Wordy: }}She was in an elated state of mind.

\textbf{\textcolor{blue}{Concise: }}She was elated. \cite{uis2021}

\vspace{12pt}

\textbf{\textcolor{red}{Wordy: }}They are connected together.

\textbf{\textcolor{blue}{Concise: }}They are connected. \cite{uis2021}

\vspace{12pt}

\textbf{\textcolor{red}{Wordy: }}They have many traits in common to both.

\textbf{\textcolor{blue}{Concise: }}They share many traits. \cite{uis2021}

\vspace{12pt}

\textbf{\textcolor{red}{Wordy: }}The houses are in close proximity.

\textbf{\textcolor{blue}{Concise: }}The houses are nearby. \cite{uis2021}

\vspace{12pt}

\textbf{\textcolor{red}{Wordy: }}He has no emotional feelings.

\textbf{\textcolor{blue}{Concise: }}He has no feelings. \cite{uis2021}

\vspace{12pt}

\textbf{\textcolor{red}{Wordy: }}Each and every person should come.

\textbf{\textcolor{blue}{Concise: }}Everyone should come. \cite{uis2021}

\vspace{12pt}

\textbf{\textcolor{red}{Wordy: }}The Acme Corporation is developing a new consumer device that allows users to communicate vocally in real time.

\textbf{\textcolor{blue}{Concise: }}The Acme Corporation is developing a new cell phone. \cite{stanford2021}

\vspace{12pt}

\textbf{\textcolor{red}{Wordy: }}Basically, the first widget pretty much surpassed the second one in overall performance.

\textbf{\textcolor{blue}{Concise: }}The first widget performed better than the second. \cite{stanford2021}

\vspace{12pt}

\textbf{\textcolor{red}{Wordy: }}The engineer considered the second monitor an unneeded luxury.

\textbf{\textcolor{blue}{Concise: }}The engineer considered the second monitor a luxury. \cite{stanford2021}

\vspace{12pt}

\textbf{\textcolor{red}{Wordy: }}The test revealed conduction activity that was peculiar in nature.

\textbf{\textcolor{blue}{Concise: }}The test revealed peculiar conduction activity. \cite{stanford2021}

\vspace{12pt}

\textbf{\textcolor{red}{Wordy: }}We redid the experiment due to the fact that our initial method was incorrect.

\textbf{\textcolor{blue}{Concise: }}We redid the experiment because our initial method was incorrect. \cite{stanford2021}

\vspace{12pt}

\textbf{\textcolor{red}{Wordy: }}Your audience will not appreciate the details that lack relevance.

\textbf{\textcolor{blue}{Concise: }}Your audience will appreciate relevant details. \cite{stanford2021}

\vspace{12pt}

\textbf{\textcolor{red}{Wordy: }}The Acme Corporation continues to work on the cell phone case configuration revision project.

\textbf{\textcolor{blue}{Concise: }}The Acme Corporation is developing a redesigned cell phone case. \cite{stanford2021}

\vspace{12pt}

\textbf{\textcolor{red}{Wordy: }}The design was completed by us.

\textbf{\textcolor{blue}{Concise: }}We completed the design. \cite{stanford2021}

\vspace{12pt}

\textbf{\textcolor{red}{Wordy: }}Our lack of data prevented evaluation of the areas in most need of assistance.

\textbf{\textcolor{blue}{Concise: }}Because we lacked data, we could not evaluate the areas that most needed assistance. \cite{stanford2021}

\vspace{12pt}

\textbf{\textcolor{red}{Wordy: }}The discovery of a method for the manufacture of artificial skin will have the result of an increase in the survival of patients with radical burns.

\textbf{\textcolor{blue}{Concise: }}If researchers discover how to manufacture artificial skin, more patients will survive radical burns. \cite{stanford2021}

\vspace{12pt}

\textbf{\textcolor{red}{Wordy: }}The outcome is dependent on the data.

\textbf{\textcolor{blue}{Concise: }}The outcome depends on the data. \cite{stanford2021}

\vspace{12pt}

\textbf{\textcolor{red}{Wordy: }}There is the possibility of approval of the study ahead of time.

\textbf{\textcolor{blue}{Concise: }}Robin may approve of the study ahead of time. \cite{stanford2021}

\vspace{12pt}

\textbf{\textcolor{red}{Wordy: }}The lack of any knowledge on the part of the investigators about local conditions precluded determination of committee action effectiveness in fund allocation to those areas in greatest need of assistance.

\textbf{\textcolor{blue}{Concise: }}Because the investigators did not know anything about local conditions, they could not determine how effectively the committee had allocated funds to the areas that most needed assistance. \cite{stanford2021}

\vspace{12pt}

\textbf{\textcolor{red}{Wordy: }}It was instructed by the professor that the assignments must be submitted by the due date.

\textbf{\textcolor{blue}{Concise: }}The professor instructed that the assignment be submitted by the due date. \cite{uri2019}

\vspace{12pt}

\textbf{\textcolor{red}{Wordy: }}It was earlier demonstrated that climate change can be caused by air pollution.

\textbf{\textcolor{blue}{Concise: }}Early studies demonstrated that air pollution can cause climate change. \cite{uri2019}

\vspace{12pt}

\textbf{\textcolor{red}{Wordy: }}An evaluation of the methods needs to be done.

\textbf{\textcolor{blue}{Concise: }}The methods need to be evaluated. \textbf{ OR } We need to evaluate the methods. \cite{uri2019}

\vspace{12pt}

\textbf{\textcolor{red}{Wordy: }}It was the final result that finally persuaded me.

\textbf{\textcolor{blue}{Concise: }}The final result finally persuaded me. \cite{uri2019}

\vspace{12pt}

\textbf{\textcolor{red}{Wordy: }}There are likely to be many researchers that raise questions about this methodological approach.

\textbf{\textcolor{blue}{Concise: }}Many researchers are likely to question this methodology. \cite{uri2019}

\vspace{12pt}

\textbf{\textcolor{red}{Wordy: }}Research is increasing in the field of nutrition and food science.

\textbf{\textcolor{blue}{Concise: }}Research is increasing in nutrition and food science. \textbf{ OR } Research within nutrition and food science is increasing. \cite{uri2019}

\vspace{12pt}

\textbf{\textcolor{red}{Wordy: }}The focus of this project was to study the effects of sea level rise on coastal habitats.

\textbf{\textcolor{blue}{Concise: }}This project examined how sea level rise affects coastal habitats. \cite{uri2019}

\vspace{12pt}

\textbf{\textcolor{red}{Wordy: }}The students were researching climate change impacts.

\textbf{\textcolor{blue}{Concise: }}The students researched climate change impacts. \cite{uri2019}

\vspace{12pt}

\textbf{\textcolor{red}{Wordy: }}These results are in agreement with prior findings.

\textbf{\textcolor{blue}{Concise: }}These results agree with prior findings. \cite{boras2021}

\vspace{12pt}

\textbf{\textcolor{red}{Wordy: }}We performed an analysis of several factors.

\textbf{\textcolor{blue}{Concise: }}We analysed several factors. \cite{boras2021}

\vspace{12pt}

\textbf{\textcolor{red}{Wordy: }}Our results are in opposition to Johnson's study.

\textbf{\textcolor{blue}{Concise: }}Our results contradict Johnson's study. \cite{boras2021}

\vspace{12pt}

\textbf{\textcolor{red}{Wordy: }}Our method is an improvement over prior systems.

\textbf{\textcolor{blue}{Concise: }}Our method improves on prior systems. \cite{boras2021}

\vspace{12pt}

\textbf{\textcolor{red}{Wordy: }}AjeA was found to be present in the nucleus.

\textbf{\textcolor{blue}{Concise: }}AjeA localised to the nucleus. \cite{boras2021}

\vspace{12pt}

\textbf{\textcolor{red}{Wordy: }}Table 1 presents a summary of the patient data.

\textbf{\textcolor{blue}{Concise: }}Table 1 summarises the patient data. \cite{boras2021}

\vspace{12pt}

\textbf{\textcolor{red}{Wordy: }}An analysis of learning outcomes was made on the basis of the findings.

\textbf{\textcolor{blue}{Concise: }}Learning outcomes were analysed on the basis of the findings. \cite{boras2021}

\vspace{12pt}

\textbf{\textcolor{red}{Wordy: }}There was considerable waste after production.

\textbf{\textcolor{blue}{Concise: }}The production created considerable waste. \cite{boras2021}

\vspace{12pt}

\textbf{\textcolor{red}{Wordy: }}They made the decision to adjust.

\textbf{\textcolor{blue}{Concise: }}They decided to adjust. \cite{boras2021}

\vspace{12pt}

\textbf{\textcolor{red}{Wordy: }}The reduction in the census was caused by the lack of a favourable response by officials.

\textbf{\textcolor{blue}{Concise: }}The lack of a favourable official response reduced the census. \cite{boras2021}

\vspace{12pt}

\textbf{\textcolor{red}{Wordy: }}The cornerstone to the government providing essential emergency services to its citizens is ensuring there has been effective planning to anticipate a variety of challenges to deliver these critical services.

\textbf{\textcolor{blue}{Concise: }}To provide essential emergency services to its citizens, the government must plan effectively. \cite{naval2021}

\vspace{12pt}

\textbf{\textcolor{red}{Wordy: }}The most authoritative account of the Great Umbrella Tariffs of 1810-13 was written by Philippa von Phrasewitz in the years immediately following those storied events.

\textbf{\textcolor{blue}{Concise: }}Philippa von Phrasewitz wrote the most authoritative account of the Great Umbrella Tariffs of 1810-13 in the years immediately following those storied events. \cite{naval2021}

\vspace{12pt}

\textbf{\textcolor{red}{Wordy: }}The survey was designed by the research assistant.

\textbf{\textcolor{blue}{Concise: }}The research assistant designed the survey. \cite{leeds2017}

\vspace{12pt}

\textbf{\textcolor{red}{Wordy: }}First and foremost, the pieces must be aligned in an accurate manner.

\textbf{\textcolor{blue}{Concise: }}First, the pieces must be aligned accurately. \cite{tamu2021}

\vspace{12pt}

\textbf{\textcolor{red}{Wordy: }}Whatever results are found during the trials, and they may or may not be favorable, we will stand behind the science.

\textbf{\textcolor{blue}{Concise: }}Whatever results are found during the trials, we will stand behind the science. \cite{tamu2021}

\vspace{12pt}

\textbf{\textcolor{red}{Wordy: }}We must make, if we want to be respected, sound choices.

\textbf{\textcolor{blue}{Concise: }}If we want to be respected, we must make sound choices. \textbf{ OR } We must make sound choices if we want to be respected. \cite{tamu2021}

\vspace{12pt}

\textbf{\textcolor{red}{Wordy: }}There are several syntactic devices that let you manage where you locate units of new information in a sentence.

\textbf{\textcolor{blue}{Concise: }}Several syntactic devices let you manage where you locate units of new information in a sentence. \cite{tamu2021}

\vspace{12pt}

\textbf{\textcolor{red}{Wordy: }}The use of this method would eliminate the problem.

\textbf{\textcolor{blue}{Concise: }}This method would eliminate the problem. \cite{tamu2021}

\vspace{12pt}

\textbf{\textcolor{red}{Wordy: }}Scientists have learned that their observations are as subjective as those in any other field in recent years.

\textbf{\textcolor{blue}{Concise: }}In recent years, scientists have learned that that their observations are as subjective as those in any other field. \cite{tamu2021}

\vspace{12pt}

\textbf{\textcolor{red}{Wordy: }}The group of science students sat their exams.

\textbf{\textcolor{blue}{Concise: }}The science students sat their exams. \cite{massey2021}

\vspace{12pt}

\textbf{\textcolor{red}{Wordy: }}Smith (2006) also believed this to be true, but took into consideration the fact that some managers also preferred to have long meetings that took all day.

\textbf{\textcolor{blue}{Concise: }}Smith(2006) agreed, but considered the fact that some managers preferred to have longer, all-day meetings. \cite{massey2021}

\vspace{12pt}

\textbf{\textcolor{red}{Wordy: }}The farmer sheared the sheep and removed all their wool.

\textbf{\textcolor{blue}{Concise: }}The farmer sheared the sheep. \cite{massey2021}

\vspace{12pt}

\textbf{\textcolor{red}{Wordy: }}There are several of the soldiers, each with their guns and ammunition, who gathered at the gates of the camp before dawn.

\textbf{\textcolor{blue}{Concise: }}Several of the soldiers, each with their own guns and ammunition, gathered at the camp gates before dawn. \cite{massey2021}

\vspace{12pt}

\textbf{\textcolor{red}{Wordy: }}As far as my professor is concerned, the problem of wordiness is the thing she'd really like to see us involved with actually eliminating.

\textbf{\textcolor{blue}{Concise: }}My professor wants us to focus on eliminating wordiness. \cite{ualr2021}

\vspace{12pt}

\textbf{\textcolor{red}{Wordy: }}There are many women who never marry.

\textbf{\textcolor{blue}{Concise: }}Many women never marry. \cite{ualr2021}

\vspace{12pt}

\textbf{\textcolor{red}{Wordy: }}It is his last book that shows his genius best.

\textbf{\textcolor{blue}{Concise: }}His last book shows his genius best. \cite{ualr2021}

\vspace{12pt}

\textbf{\textcolor{red}{Wordy: }}Many uneducated citizens who have never attended school continue to vote for better schools.

\textbf{\textcolor{blue}{Concise: }}Many uneducated citizens who have never attended school continue to vote for better schools. \cite{commnet2021}

\vspace{12pt}

\textbf{\textcolor{red}{Wordy: }}All things considered, Connecticut's woodlands are in better shape now than ever before.

\textbf{\textcolor{blue}{Concise: }}Connecticut's woodlands are in better shape now than ever before. \cite{commnet2021}

\vspace{12pt}

\textbf{\textcolor{red}{Wordy: }}As a matter of fact, there are more woodlands in Connecticut now than there were in 1898.

\textbf{\textcolor{blue}{Concise: }}There are more woodlands in Connecticut now than there were in 1898. \cite{commnet2021}

\vspace{12pt}

\textbf{\textcolor{red}{Wordy: }}As far as I'm concerned, there is no need for further protection of woodlands.

\textbf{\textcolor{blue}{Concise: }}Further protection of woodlands is not needed. \cite{commnet2021}

\vspace{12pt}

\textbf{\textcolor{red}{Wordy: }}This is because there are fewer farmers at the present time.

\textbf{\textcolor{blue}{Concise: }}This is because there are fewer farmers now. \cite{commnet2021}

\vspace{12pt}

\textbf{\textcolor{red}{Wordy: }}Woodlands have grown in area because of the fact that farmers have abandoned their fields.

\textbf{\textcolor{blue}{Concise: }}Woodlands have grown in area because farmers have abandoned their fields. \cite{commnet2021}

\vspace{12pt}

\textbf{\textcolor{red}{Wordy: }}Major forest areas are coming back by means of natural processes.

\textbf{\textcolor{blue}{Concise: }}Major forest areas are coming back naturally. \cite{commnet2021}

\vspace{12pt}

\textbf{\textcolor{red}{Wordy: }}Our woodlands are coming back by virtue of the fact that our economy has shifted its emphasis.

\textbf{\textcolor{blue}{Concise: }}Our woodlands are coming back because our economy has shifted its emphasis. \cite{commnet2021}

\vspace{12pt}

\textbf{\textcolor{red}{Wordy: }}Due to the fact that their habitats are being restored, forest creatures are also re-establishing their population bases.

\textbf{\textcolor{blue}{Concise: }}Because their habitats are being restored, forest creatures are also re-establishing their population bases. \cite{commnet2021}

\vspace{12pt}

\textbf{\textcolor{red}{Wordy: }}The fear that exists among many people that we are losing our woodlands is uncalled for.

\textbf{\textcolor{blue}{Concise: }}The fear among many people that we are losing our woodlands is uncalled for. \cite{commnet2021}

\vspace{12pt}

\textbf{\textcolor{red}{Wordy: }}The era in which we must aggressively defend our woodlands has, for all intents and purposes, passed.

\textbf{\textcolor{blue}{Concise: }}The era in which we must aggressively defend our woodlands has passed. \cite{commnet2021}

\vspace{12pt}

\textbf{\textcolor{red}{Wordy: }}For the most part, people's suspicions are based on a misunderstanding of the facts.

\textbf{\textcolor{blue}{Concise: }}People's suspicions are based on a misunderstanding of the facts. \cite{commnet2021}

\vspace{12pt}

\textbf{\textcolor{red}{Wordy: }}Many woodlands, in fact, have been purchased for the purpose of creating public parks.

\textbf{\textcolor{blue}{Concise: }}Many woodlands, in fact, have been purchased as public parks. \cite{commnet2021}

\vspace{12pt}

\textbf{\textcolor{red}{Wordy: }}This policy has a tendency to isolate some communities.

\textbf{\textcolor{blue}{Concise: }}This policy tends to isolate some communities. \cite{commnet2021}

\vspace{12pt}

\textbf{\textcolor{red}{Wordy: }}The policy has, in a manner of speaking, begun to Balkanize the more rural parts of our state.

\textbf{\textcolor{blue}{Concise: }}The policy has begun to Balkanize the more rural parts of our state. \cite{commnet2021}

\vspace{12pt}

\textbf{\textcolor{red}{Wordy: }}In a very real sense, this policy works to the detriment of those it is supposed to help.

\textbf{\textcolor{blue}{Concise: }}This policy works to the detriment of those it is supposed to help. \cite{commnet2021}

\vspace{12pt}

\textbf{\textcolor{red}{Wordy: }}In my opinion, this wasteful policy ought to be revoked.

\textbf{\textcolor{blue}{Concise: }}This wasteful policy ought to be revoked. \cite{commnet2021}

\vspace{12pt}

\textbf{\textcolor{red}{Wordy: }}In the case of this particular policy, citizens of northeast Connecticut became very upset.

\textbf{\textcolor{blue}{Concise: }}Citizens of northeast Connecticut became very upset about his policy. \cite{commnet2021}

\vspace{12pt}

\textbf{\textcolor{red}{Wordy: }}In the final analysis, the state would have been better off without such a policy.

\textbf{\textcolor{blue}{Concise: }}The state would have been better off without such a policy. \cite{commnet2021}

\vspace{12pt}

\textbf{\textcolor{red}{Wordy: }}In the event that enough people protest, it will probably be revoked.

\textbf{\textcolor{blue}{Concise: }}If enough people protest, it will probably be revoked. \cite{commnet2021}

\vspace{12pt}

\textbf{\textcolor{red}{Wordy: }}Something in the nature of a repeal may soon take place.

\textbf{\textcolor{blue}{Concise: }}Something like a repeal may soon take place. \cite{commnet2021}

\vspace{12pt}

\textbf{\textcolor{red}{Wordy: }}Legislators are already in the process of reviewing the statutes.

\textbf{\textcolor{blue}{Concise: }}Legislators are already reviewing the statutes. \cite{commnet2021}

\vspace{12pt}

\textbf{\textcolor{red}{Wordy: }}It seems that they can't wait to get rid of this one.

\textbf{\textcolor{blue}{Concise: }}They can't wait to get rid of this one. \cite{commnet2021}

\vspace{12pt}

\textbf{\textcolor{red}{Wordy: }}They have monitored the activities of conservationists in a cautious manner.

\textbf{\textcolor{blue}{Concise: }}They have cautiously monitored the activities of conservationists. \cite{commnet2021}

\vspace{12pt}

\textbf{\textcolor{red}{Wordy: }}The point I am trying to make is that sometimes public policy doesn't accomplish what it set out to achieve.

\textbf{\textcolor{blue}{Concise: }}Sometimes public policy doesn't accomplish what it set out to achieve. \cite{commnet2021}

\vspace{12pt}

\textbf{\textcolor{red}{Wordy: }}Legislators need to be more careful of the type of policy they propose.

\textbf{\textcolor{blue}{Concise: }}Legislators need to be more careful of the policy they propose. \cite{commnet2021}

\vspace{12pt}

\textbf{\textcolor{red}{Wordy: }}What I mean to say is that well intentioned lawmakers sometimes make fools of themselves.

\textbf{\textcolor{blue}{Concise: }}Well intentioned lawmakers sometimes make fools of themselves. \cite{commnet2021}

\vspace{12pt}

\textbf{\textcolor{red}{Wordy: }}Smith College, which was founded in 1871, is the premier all-women's college in the United States.

\textbf{\textcolor{blue}{Concise: }}Founded in 1871, Smith College is the premier all-women's college in the United States. \cite{commnet2021}

\vspace{12pt}

\textbf{\textcolor{red}{Wordy: }}Citizens who knew what was going on voted him out of office.

\textbf{\textcolor{blue}{Concise: }}Knowledgeable citizens voted him out of office. \cite{commnet2021}

\vspace{12pt}

\textbf{\textcolor{red}{Wordy: }}Recommending that a student copy from another student's paper is not something he would recommend.

\textbf{\textcolor{blue}{Concise: }}He would never tell a student to copy from another student's paper. \cite{commnet2021}

\vspace{12pt}

\textbf{\textcolor{red}{Wordy: }}Unencumbered by a sense of responsibility, Jasion left his wife with forty-nine kids and a can of beans.

\textbf{\textcolor{blue}{Concise: }}Jasion irresponsibly left his wife with forty-nine kids and a can of beans. \cite{commnet2021}

\vspace{12pt}

\textbf{\textcolor{red}{Wordy: }}At this point in time we can't ascertain the reason as to why the screen door was left open.

\textbf{\textcolor{blue}{Concise: }}We don't know why the screen door was left open. \cite{commnet2021}

\vspace{12pt}

\textbf{\textcolor{red}{Wordy: }}My sister, who is employed as a nutritionist at the University of Michigan, recommends the daily intake of megadoses of Vitamin C.

\textbf{\textcolor{blue}{Concise: }}My sister, a nutritionist at the University of Michigan, recommends daily megadoses of Vitamin C. \cite{commnet2021}

\vspace{12pt}

\textbf{\textcolor{red}{Wordy: }}Basically, in light of the fact that Congressman Fuenches was totally exhausted by his last campaign, there was an expectation on the part of the voters that he would not reduplicate his effort to achieve office in government again.

\textbf{\textcolor{blue}{Concise: }}Because Congressman Fuenches was exhausted by his last campaign, voters expected he would not seek re-election. \textbf{ OR } Voters thought that Congressman Fuenches was so exhausted by his last campaign that he wouldn't seek re-election. \cite{commnet2021}

\vspace{12pt}

\textbf{\textcolor{red}{Wordy: }}It is to be hoped that we discover a means to create an absolutely proper and fitting tribute to Professor Espinoza.

\textbf{\textcolor{blue}{Concise: }}We hope for an appropriate tribute to Professor Espinoza. \cite{commnet2021}

\vspace{12pt}

\textbf{\textcolor{red}{Wordy: }}There is a desire on the part of many of us to maintain a spring recess for the purpose of getting away from the demands of our studies.

\textbf{\textcolor{blue}{Concise: }}We want a spring recess so we can get away from our studies. \textbf{ OR } We want a spring recess to escape our studies. \cite{commnet2021}

\vspace{12pt}

\textbf{\textcolor{red}{Wordy: }}What is your basic understanding of predestination?

\textbf{\textcolor{blue}{Concise: }}Explain predestination. \cite{commnet2021}

\vspace{12pt}

\textbf{\textcolor{red}{Wordy: }}At what point in time will a downturn in the stock market have a really serious effect on the social life of people as a whole?

\textbf{\textcolor{blue}{Concise: }}When will a downturn in the stock market affect society? \cite{commnet2021}

\vspace{12pt}

\textbf{\textcolor{red}{Wordy: }}I would call your attention to the fact that our President, who was formerly the Governor of Arkansas, is basically a Southerner.

\textbf{\textcolor{blue}{Concise: }}Our President, formerly the Governor of Arkansas, is a Southerner. \cite{commnet2021}

\vspace{12pt}

\textbf{\textcolor{red}{Wordy: }}There are millions of fans who desperately want the Hartford Whalers to stay in the city.

\textbf{\textcolor{blue}{Concise: }}Millions of fans desperately want the Hartford Whalers to stay in the city. \cite{commnet2021}

\vspace{12pt}

\textbf{\textcolor{red}{Wordy: }}Bothered by allergies, a condition that made them sneeze, some of the preschool children had sinus troubles that caused them to miss several days in nursery school this spring.

\textbf{\textcolor{blue}{Concise: }}Bothered by allergies, some children missed several days in nursery school this spring. \cite{commnet2021}

\vspace{12pt}

\textbf{\textcolor{red}{Wordy: }}The nursery school teacher education training sessions involve active interfacing with preschool children of the appropriate age as well as intensive peer interaction in the form of role playing.

\textbf{\textcolor{blue}{Concise: }}Training for nursery school teachers involves interacting with preschoolers and role playing with peers. \cite{commnet2021}

\vspace{12pt}

\textbf{\textcolor{red}{Wordy: }}He found his neighbor who lived next door to be attractive in appearance.

\textbf{\textcolor{blue}{Concise: }}He found his neighbor attractive. \cite{commnet2021}

\vspace{12pt}

\textbf{\textcolor{red}{Wordy: }}He was really late to his English class due to the fact that he had to finish his math test.

\textbf{\textcolor{blue}{Concise: }}He was late for English because he had to finish his math test. \cite{commnet2021}

\vspace{12pt}

\textbf{\textcolor{red}{Wordy: }}Although they were several in number, the street gang feared the police.

\textbf{\textcolor{blue}{Concise: }}Although they were a large group, the street gang feared the police. \cite{commnet2021}

\vspace{12pt}

\textbf{\textcolor{red}{Wordy: }}Bob provided an explanation of the computer to his grandmother.

\textbf{\textcolor{blue}{Concise: }}Bob explained the computer to his grandmother. \cite{commnet2021}

\vspace{12pt}

\textbf{\textcolor{red}{Wordy: }}During the time when I lived in South Carolina, it was my intention to go to college in Florida.

\textbf{\textcolor{blue}{Concise: }}When I lived in South Carolina, I planned to attend college in Florida. \cite{commnet2021}

\vspace{12pt}

\textbf{\textcolor{red}{Wordy: }}In order to prove that he could hold his own on the track team, Gordo had to train hard like the old runners.

\textbf{\textcolor{blue}{Concise: }}To prove that he could hold his own on the track team, Gordo had to train hard like the older runners. \cite{commnet2021}

\vspace{12pt}

\textbf{\textcolor{red}{Wordy: }}If you go to the store, you will see that the store is closed on Sundays because the storeowner likes to go to church.

\textbf{\textcolor{blue}{Concise: }}If you go to the store, you will see that it is closed on Sunday because the owner likes to go to church. \textbf{ OR } The store is closed on Sundays because the owner attends church. \cite{commnet2021}

\vspace{12pt}

\textbf{\textcolor{red}{Wordy: }}Due to the fact that Jim liked chocolate, he was very upset when the candy company canceled production of his favorite chocolate bar.

\textbf{\textcolor{blue}{Concise: }}Because Jim liked chocolate, he was very upset when the candy company canceled his favorite chocolate bar. \textbf{ OR } Jim was upset when the candy company canceled production of his favorite chocolate bar. \cite{commnet2021}

\vspace{12pt}

\textbf{\textcolor{red}{Wordy: }}One time when I went to the park, my friend, whose name is Jake, went with me and we had fun due to the fact that it was a nice day out.

\textbf{\textcolor{blue}{Concise: }}When I went to the park with my friend Jake, we had fun because it was such a nice day. \cite{commnet2021}

\vspace{12pt}

\textbf{\textcolor{red}{Wordy: }}In this report I will conduct a study of ants and the setup of their colonies.

\textbf{\textcolor{blue}{Concise: }}I will study ants and their colonies in this report. \textbf{ OR } This report studies ants. \cite{commnet2021}

\vspace{12pt}

\textbf{\textcolor{red}{Wordy: }}An investigation of the symbolic overtones in Melville's "Billy Budd, Sailor" reveals the opposing personas of Billy Budd's naivete and Claggart's vituperative demeanor.

\textbf{\textcolor{blue}{Concise: }}The powerful symbols in Melville's "Billy Budd, Sailor" develop Budd's innocence in contrast to the corrupt Claggart. \cite{commnet2021}

\vspace{12pt}

\textbf{\textcolor{red}{Wordy: }}In hopes of rejuvenating their romantic liaison, the couple went on a pilgrimage to become one with nature.

\textbf{\textcolor{blue}{Concise: }}In hopes of rekindling their romance, the couple went camping. \cite{commnet2021}

\vspace{12pt}

\textbf{\textcolor{red}{Wordy: }}A mathematical mindset best suits the making of decisions that profoundly affect this company's future business relations.

\textbf{\textcolor{blue}{Concise: }}Executive decisions require precise calculations. \cite{commnet2021}

\vspace{12pt}

\textbf{\textcolor{red}{Wordy: }}Joe found that the fictional novel by Alcott, Little Women, conveyed a sense of emotion and feeling.

\textbf{\textcolor{blue}{Concise: }}Joe found Alcott's book, Little Women, both delightful and tragic. \cite{commnet2021}

\vspace{12pt}

\textbf{\textcolor{red}{Wordy: }}The existence of computers and computer technology has greatly influenced commercial enterprise and information exchange.

\textbf{\textcolor{blue}{Concise: }}Computers have improved business communication. \cite{commnet2021}

\vspace{12pt}

\textbf{\textcolor{red}{Wordy: }}Attentive focusing on the professor's informational lecture was not always completely possible.

\textbf{\textcolor{blue}{Concise: }}Focusing on professor's boring lectures was a problem for most students. \cite{commnet2021}

\vspace{12pt}

\textbf{\textcolor{red}{Wordy: }}After extensive nightly labors on an academic assignment, Sally needed a lengthy period of somnolence.

\textbf{\textcolor{blue}{Concise: }}After staying up all night finishing a paper, Sally needed a long nap. \cite{commnet2021}

\vspace{12pt}

\textbf{\textcolor{red}{Wordy: }}My first visit to Boston will always be remembered by me.

\textbf{\textcolor{blue}{Concise: }}I shall always remember my first visit to Boston. \cite{strunk_1918}

\vspace{12pt}

\textbf{\textcolor{red}{Wordy: }}There were a great number of dead leaves lying on the ground.

\textbf{\textcolor{blue}{Concise: }}Dead leaves covered the ground. \cite{strunk_1918}

\vspace{12pt}

\textbf{\textcolor{red}{Wordy: }}The sound of the falls could still be heard.

\textbf{\textcolor{blue}{Concise: }}The sound of the falls still reached our ears. \cite{strunk_1918}

\vspace{12pt}

\textbf{\textcolor{red}{Wordy: }}The reason that he left college was that his health became impaired.

\textbf{\textcolor{blue}{Concise: }}Failing health compelled him to leave college. \cite{strunk_1918}

\vspace{12pt}

\textbf{\textcolor{red}{Wordy: }}It was not long before he was very sorry that he had said what he had.

\textbf{\textcolor{blue}{Concise: }}He soon repented his words. \cite{strunk_1918}

\vspace{12pt}

\textbf{\textcolor{red}{Wordy: }}Gold was not allowed to be exported.

\textbf{\textcolor{blue}{Concise: }}It was forbidden to export gold. \cite{strunk_1918}

\vspace{12pt}

\textbf{\textcolor{red}{Wordy: }}He has been proved to have been seen entering the building.

\textbf{\textcolor{blue}{Concise: }}It has been proved that he was seen to enter the building. \cite{strunk_1918}

\vspace{12pt}

\textbf{\textcolor{red}{Wordy: }}A survey of this region was made in 1900.

\textbf{\textcolor{blue}{Concise: }}This region was surveyed in 1900. \cite{strunk_1918}

\vspace{12pt}

\textbf{\textcolor{red}{Wordy: }}Mobilization of the army was rapidly carried out.

\textbf{\textcolor{blue}{Concise: }}The army was rapidly mobilized. \cite{strunk_1918}

\vspace{12pt}

\textbf{\textcolor{red}{Wordy: }}Confirmation of these reports cannot be obtained.

\textbf{\textcolor{blue}{Concise: }}These reports cannot be confirmed. \cite{strunk_1918}

\vspace{12pt}

\textbf{\textcolor{red}{Wordy: }}He was not very often on time.

\textbf{\textcolor{blue}{Concise: }}He usually came late. \cite{strunk_1918}

\vspace{12pt}

\textbf{\textcolor{red}{Wordy: }}He did not think that studying Latin was much use.

\textbf{\textcolor{blue}{Concise: }}He thought the study of Latin useless. \cite{strunk_1918}

\vspace{12pt}

\textbf{\textcolor{red}{Wordy: }}Formerly, science was taught by the textbook method, while now the laboratory method is employed.

\textbf{\textcolor{blue}{Concise: }}Formerly, science was taught by the textbook method; now it is taught by the laboratory method. \cite{strunk_1918}

\vspace{12pt}

\textbf{\textcolor{red}{Wordy: }}It was both a long ceremony and very tedious.

\textbf{\textcolor{blue}{Concise: }}The ceremony was both long and tedious. \cite{strunk_1918}

\vspace{12pt}

\textbf{\textcolor{red}{Wordy: }}Wordsworth, in the fifth book of The Excursion, gives a minute description of this church.

\textbf{\textcolor{blue}{Concise: }}In the fifth book of The Excursion, Wordsworth gives a minute description of this church. \cite{strunk_1918}

\vspace{12pt}

\textbf{\textcolor{red}{Wordy: }}Cast iron, when treated in a Bessemer converter, is changed into steel.

\textbf{\textcolor{blue}{Concise: }}By treatment in a Bessemer converter, cast iron is changed into steel. \cite{strunk_1918}

\vspace{12pt}

\textbf{\textcolor{red}{Wordy: }}There was a look in his eye that boded mischief.

\textbf{\textcolor{blue}{Concise: }}In his eye was a look that boded mischief. \cite{strunk_1918}

\vspace{12pt}

\textbf{\textcolor{red}{Wordy: }}He wrote three articles about his adventures in Spain, which were published in Harper's Magazine.

\textbf{\textcolor{blue}{Concise: }}He published in Harper's Magazine three articles about his adventures in Spain. \cite{strunk_1918}

\vspace{12pt}

\textbf{\textcolor{red}{Wordy: }}I received your inquiry that you wrote about tennis rackets yesterday, and read it thoroughly.

\textbf{\textcolor{blue}{Concise: }}I received your inquiry about tennis rackets yesterday. \cite{purdue_writing_lab2021}

\vspace{12pt}

\textbf{\textcolor{red}{Wordy: }}It goes without saying that we are acquainted with your policy on filing tax returns, and we have every intention of complying with the regulations that you have mentioned.

\textbf{\textcolor{blue}{Concise: }}We intend to comply with the tax-return regulations that you have mentioned. \cite{purdue_writing_lab2021}

\vspace{12pt}

\textbf{\textcolor{red}{Wordy: }}Imagine a mental picture of someone engaged in the intellectual activity of trying to learn what the rules are for how to play the game of chess.

\textbf{\textcolor{blue}{Concise: }}Imagine someone trying to learn the rules of chess. \cite{purdue_writing_lab2021}

\vspace{12pt}

\textbf{\textcolor{red}{Wordy: }}Baseball, one of our oldest and most popular outdoor summer sports in terms of total attendance at ball parks and viewing on television, has the kind of rhythm of play on the field that alternates between times when players passively wait with no action taking place between the pitches to the batter and then times when they explode into action as the batter hits a pitched ball to one of the players and the player fields it.

\textbf{\textcolor{blue}{Concise: }}Baseball has a rhythm that alternates between waiting and explosive action. \cite{purdue_writing_lab2021}

\vspace{12pt}

\textbf{\textcolor{red}{Wordy: }}Any particular type of dessert is fine with me.

\textbf{\textcolor{blue}{Concise: }}Any dessert is fine with me. \cite{purdue_writing_lab2021}

\vspace{12pt}

\textbf{\textcolor{red}{Wordy: }}Balancing the budget by Friday is an impossibility without some kind of extra help.

\textbf{\textcolor{blue}{Concise: }}Balancing the budget by Friday is impossible without extra help. \cite{purdue_writing_lab2021}

\vspace{12pt}

\textbf{\textcolor{red}{Wordy: }}For all intents and purposes, American industrial productivity generally depends on certain factors that are really more psychological in kind than of any given technological aspect.

\textbf{\textcolor{blue}{Concise: }}American industrial productivity depends more on psychological than on technological factors. \cite{purdue_writing_lab2021}

\vspace{12pt}

\textbf{\textcolor{red}{Wordy: }}The supply manager considered the correcting typewriter an unneeded luxury.

\textbf{\textcolor{blue}{Concise: }}The supply manager considered the correcting typewriter a luxury. \cite{purdue_writing_lab2021}

\vspace{12pt}

\textbf{\textcolor{red}{Wordy: }}Before the travel agent was completely able to finish explaining the various differences among all of the many very unique vacation packages his travel agency was offering, the customer changed her future plans.

\textbf{\textcolor{blue}{Concise: }}Before the travel agent finished explaining the differences among the unique vacation packages his travel agency was offering, the customer changed her plans. \cite{purdue_writing_lab2021}

\vspace{12pt}

\textbf{\textcolor{red}{Wordy: }}During that time period, many car buyers preferred cars that were pink in color and shiny in appearance.

\textbf{\textcolor{blue}{Concise: }}During that period, many car buyers preferred pink, shiny cars. \cite{purdue_writing_lab2021}

\vspace{12pt}

\textbf{\textcolor{red}{Wordy: }}The microscope revealed a group of organisms that were round in shape and peculiar in nature.

\textbf{\textcolor{blue}{Concise: }}The microscope revealed a group of peculiar, round organisms. \cite{purdue_writing_lab2021}

\vspace{12pt}

\textbf{\textcolor{red}{Wordy: }}Jeff Converse, our chief of consulting, suggested at our last board meeting the installation of microfilm equipment in the department of data processing.

\textbf{\textcolor{blue}{Concise: }}At our last board meeting, Chief Consultant Jeff Converse suggested that we install microfilm equipment in the data processing department. \cite{purdue_writing_lab2021}

\vspace{12pt}

\textbf{\textcolor{red}{Wordy: }}We read the letter we received yesterday and reviewed it thoroughly.

\textbf{\textcolor{blue}{Concise: }}We thoroughly read the letter we received yesterday. \cite{purdue_writing_lab2021}

\vspace{12pt}

\textbf{\textcolor{red}{Wordy: }}As you carefully read what you have written to improve your wording and catch small errors of spelling, punctuation, and so on, the thing to do before you do anything else is to try to see where a series of words expressing action could replace the ideas found in nouns rather than verbs.

\textbf{\textcolor{blue}{Concise: }}As you edit, first find nominalizations that you can replace with verb phrases. \cite{purdue_writing_lab2021}

\vspace{12pt}

\textbf{\textcolor{red}{Wordy: }}An account was opened by Mrs. Simms.

\textbf{\textcolor{blue}{Concise: }}Mrs. Simms opened an account. \cite{purdue_writing_lab2021}

\vspace{12pt}

\textbf{\textcolor{red}{Wordy: }}Your figures were checked by the research department.

\textbf{\textcolor{blue}{Concise: }}The research department checked your figures. \cite{purdue_writing_lab2021}

\vspace{12pt}

\textbf{\textcolor{red}{Wordy: }}It is the governor who signs or vetoes bills.

\textbf{\textcolor{blue}{Concise: }}The governor signs or vetoes bills. \cite{purdue_writing_lab2021}

\vspace{12pt}

\textbf{\textcolor{red}{Wordy: }}There are four rules that should be observed.

\textbf{\textcolor{blue}{Concise: }}Four rules should be observed. \cite{purdue_writing_lab2021}

\vspace{12pt}

\textbf{\textcolor{red}{Wordy: }}There was a big explosion, which shook the windows, and people ran into the street.

\textbf{\textcolor{blue}{Concise: }}A big explosion shook the windows, and people ran into the street. \cite{purdue_writing_lab2021}

\vspace{12pt}

\textbf{\textcolor{red}{Wordy: }}The function of this department is the collection of accounts.

\textbf{\textcolor{blue}{Concise: }}This department collects accounts. \cite{purdue_writing_lab2021}

\vspace{12pt}

\textbf{\textcolor{red}{Wordy: }}The current focus of the medical profession is disease prevention.

\textbf{\textcolor{blue}{Concise: }}The medical profession currently focuses on disease prevention. \cite{purdue_writing_lab2021}

\vspace{12pt}

\textbf{\textcolor{red}{Wordy: }}The duty of a clerk is to check all incoming mail and to record it.

\textbf{\textcolor{blue}{Concise: }}A clerk checks and records all incoming mail. \cite{purdue_writing_lab2021}

\vspace{12pt}

\textbf{\textcolor{red}{Wordy: }}A shortage of tellers at our branch office on Friday and Saturday during rush hours has caused customers to become dissatisfied with service.

\textbf{\textcolor{blue}{Concise: }}A teller shortage at our branch office on Friday and Saturday during rush hours has caused customer dissatisfaction. \cite{purdue_writing_lab2021}

\vspace{12pt}

\textbf{\textcolor{red}{Wordy: }}It is possible that nothing will come of these preparations.

\textbf{\textcolor{blue}{Concise: }}Nothing may come of these preparations. \cite{purdue_writing_lab2021}

\vspace{12pt}

\textbf{\textcolor{red}{Wordy: }}She has the ability to influence the outcome.

\textbf{\textcolor{blue}{Concise: }}She can influence the outcome. \cite{purdue_writing_lab2021}

\vspace{12pt}

\textbf{\textcolor{red}{Wordy: }}It is necessary that we take a stand on this pressing issue.

\textbf{\textcolor{blue}{Concise: }}We must take a stand on this pressing issue. \cite{purdue_writing_lab2021}

\vspace{12pt}

\textbf{\textcolor{red}{Wordy: }}The point I wish to make is that the employees working at this company are in need of a much better manager of their money.

\textbf{\textcolor{blue}{Concise: }}Employees at this company need a better money manager. \cite{purdue_writing_lab2021}

\vspace{12pt}

\textbf{\textcolor{red}{Wordy: }}It is widely known that the engineers at Sandia Labs have become active participants in the Search and Rescue operations in most years.

\textbf{\textcolor{blue}{Concise: }}In recent years, engineers at Sandia Labs have participated in the Search and Rescue operations. \cite{purdue_writing_lab2021}

\vspace{12pt}

\textbf{\textcolor{red}{Wordy: }}After reviewing the results of your previous research, and in light of the relevant information found within the context of the study, there is ample evidence for making important, significant changes to our operating procedures.

\textbf{\textcolor{blue}{Concise: }}After reviewing the results of your research, and within the context of the study, we find evidence supporting significant changes in our operating procedures. \cite{purdue_writing_lab2021}

\vspace{12pt}

\textbf{\textcolor{red}{Wordy: }}The experiment was conducted at 1330 GMT and was conducted with an increased basal rate with a double bolus.

\textbf{\textcolor{blue}{Concise: }}The experiment was conducted at 1330 GMT using an increased basal rate and a double bolus. \cite{purdue_writing_lab2021}

\vspace{12pt}

\textbf{\textcolor{red}{Wordy: }}The baseball game was attended by Morgan Latour.

\textbf{\textcolor{blue}{Concise: }}Morgan Latour attended the baseball game. \cite{purdue_writing_lab2021}

\vspace{12pt}

\textbf{\textcolor{red}{Wordy: }}In the following experiment, we used the feline cell line, W9, to evaluate cell growth in the presence of the growth factor.

\textbf{\textcolor{blue}{Concise: }}The feline cell line W9 was used to evaluate cell growth in the presence of growth factor. \cite{purdue_writing_lab2021}

\vspace{12pt}

\textbf{\textcolor{red}{Wordy: }}During the procedure, we cultured the cells for 48 hours in media that we modified with 78, 80, 90, and 110 ng/mL BMP.

\textbf{\textcolor{blue}{Concise: }}Cells were cultured for 48 hours in media modified with 78, 80, 90, and 110 ng/mL BMP. \cite{purdue_writing_lab2021}

\vspace{12pt}

\textbf{\textcolor{red}{Wordy: }}At 48 hours, we harvested cells from the cell culture dish and counted; we used a hemocytometer.

\textbf{\textcolor{blue}{Concise: }}At 48 hours, cells were harvested from the culture dish and counted using hemocytometer. \cite{purdue_writing_lab2021}

\vspace{12pt}

\textbf{\textcolor{red}{Wordy: }}Usner's work is an attempt at revision of orthodox historiography.

\textbf{\textcolor{blue}{Concise: }}Usner attempts to revise orthodox historiography. \cite{hamilton2021}

\vspace{12pt}

\textbf{\textcolor{red}{Wordy: }}An example of thorough analysis is when Tierney cites Wiseman's extensive data.

\textbf{\textcolor{blue}{Concise: }}Tierney's citing of Wiseman's extensive data exemplifies thorough analysis. \cite{hamilton2021}

\vspace{12pt}

\textbf{\textcolor{red}{Wordy: }}The process of modernization in any society is seen as a positive change.

\textbf{\textcolor{blue}{Concise: }}Most people see modernization in a society as a positive change. \cite{hamilton2021}

\vspace{12pt}

\textbf{\textcolor{red}{Wordy: }}Donner's misuse of information is exposed by the facts.

\textbf{\textcolor{blue}{Concise: }}The facts expose Donner's misuse of information. \cite{hamilton2021}

\vspace{12pt}

\textbf{\textcolor{red}{Wordy: }}Successful globalization depends on factors that involve culture more than economics.

\textbf{\textcolor{blue}{Concise: }}Successful globalization depends more on culture than economics. \cite{hamilton2021}

\vspace{12pt}

\textbf{\textcolor{red}{Wordy: }}Tom Jones is a novel that comically portrays English society in the eighteenth century.

\textbf{\textcolor{blue}{Concise: }}The novel Tom Jones comically portrays eighteenth century English society. \cite{hamilton2021}

\vspace{12pt}

\textbf{\textcolor{red}{Wordy: }}There are indications of Privo's misunderstanding of natural selection in her argument.

\textbf{\textcolor{blue}{Concise: }}Privo's argument demonstrates her misunderstanding of natural selection. \cite{hamilton2021}

\vspace{12pt}

\textbf{\textcolor{red}{Wordy: }}Modern society is in need of a recalibration of its moral values.

\textbf{\textcolor{blue}{Concise: }}Modern society needs to recalibrate its moral values. \cite{hamilton2021}

\vspace{12pt}

\textbf{\textcolor{red}{Wordy: }}Attempts by economists at defining full employment have been met with failure.

\textbf{\textcolor{blue}{Concise: }}Economists' attempts at defining full employment have failed. \cite{hamilton2021}

\vspace{12pt}

\textbf{\textcolor{red}{Wordy: }}One of the most important indications of the sensorimotor period is the development of object permanence.

\textbf{\textcolor{blue}{Concise: }}The development of object permanence is a key indicator of the sensorimotor period. \cite{hamilton2021}

\vspace{12pt}

\textbf{\textcolor{red}{Wordy: }}Bill is kicked by Jim.

\textbf{\textcolor{blue}{Concise: }}Jim kicks Bill. \cite{ioyno2021}

\vspace{12pt}

\textbf{\textcolor{red}{Wordy: }}Bill screamed loudly as he was being kicked very forcefully by Jim.

\textbf{\textcolor{blue}{Concise: }}Bill screamed as Jim kicked him forcefully. \cite{ioyno2021}

\vspace{12pt}

\textbf{\textcolor{red}{Wordy: }}This manual of instructions was prepared by our company to aid and assist our dealers in being helpful to their customers.

\textbf{\textcolor{blue}{Concise: }}Our company prepared this instruction manual to help our dealers serve their customers. \cite{ioyno2021}

\vspace{12pt}

\textbf{\textcolor{red}{Wordy: }}As a result of an infraction of a minor nature, Jonathan was kicked in the shin by Matthew, which caused Jonathan pain of a medium degree.

\textbf{\textcolor{blue}{Concise: }}After Jonathan committed a minor infraction, Mathew kicked Jonathan in the shin, causing Jonathan pain. \cite{rcsj2021}

\vspace{12pt}

\textbf{\textcolor{red}{Wordy: }}The author's intention of the article is to argue for the elimination of the passivity of the learner in college through using recent studies that show the value of collaborative learning in the classroom environment.

\textbf{\textcolor{blue}{Concise: }}The author cites recent research and argues for collaboration as a way to eliminate passivity and stimulate classroom active learning. \cite{antioch2021}

\vspace{12pt}

\textbf{\textcolor{red}{Wordy: }}The new interpretations on canine aggressive behavior researched by Cesar Milano are actually conceived by using a method that originally was used by Skinner and the early scholars in behaviorism.

\textbf{\textcolor{blue}{Concise: }}Cesar Millan uses Behaviorism in researching canine aggression. \cite{antioch2021}

\vspace{12pt}

\textbf{\textcolor{red}{Wordy: }}In this paragraph is a demonstration of the use of good style in the writing of a report.

\textbf{\textcolor{blue}{Concise: }}This paragraph demonstrates good style in report writing. \cite{mcdaniel_college2021}

\vspace{12pt}

\textbf{\textcolor{red}{Wordy: }}The point that I wish to make is that the employees working at this company are in need of a much better manager of their money.

\textbf{\textcolor{blue}{Concise: }}This company's employees need a much better money manager. \cite{mcdaniel_college2021}

\vspace{12pt}

\textbf{\textcolor{red}{Wordy: }}If the two materials have drastically different coefficients of expansion, the difference in the rates of expansion will cause stress on the coupling and will cause the point where the two materials are joined to become weak and less secure.

\textbf{\textcolor{blue}{Concise: }}Joining two materials with significantly different coefficients of expansion can cause stress in the coupling due to the different rates of expansion, weakening the joint. \cite{uiowa2021}

\vspace{12pt}

\textbf{\textcolor{red}{Wordy: }}Before we ran any of the tests, we measured the cross-sectional width and thickness of the sample as well as the initial length of the sample.

\textbf{\textcolor{blue}{Concise: }}The initial length, thickness, and cross-sectional width of the sample were measured before running the tests. \cite{uiowa2021}

\vspace{12pt}

\textbf{\textcolor{red}{Wordy: }}I think that all I can usefully say on this point is that in the normal course of their professional activities social anthropologists are usually concerned with the third of these alternatives, while the other two levels are treated as raw data for analysis.

\textbf{\textcolor{blue}{Concise: }}Social anthropologists usually concentrate on the third alternative, treating the other two as raw data. \cite{ucf2021}

\vspace{12pt}

\textbf{\textcolor{red}{Wordy: }}The best teachers help each student become a better student both academically and emotionally.

\textbf{\textcolor{blue}{Concise: }}The best teachers help each student grow both academically and emotionally. \cite{csbsju2021}

\vspace{12pt}

\textbf{\textcolor{red}{Wordy: }}It is imperative that all night managers follow strict procedures when locking the safe.

\textbf{\textcolor{blue}{Concise: }}All night managers must follow strict procedures when locking the safe. \cite{csbsju2021}

\vspace{12pt}

\textbf{\textcolor{red}{Wordy: }}The migration away from Java to Singapore by the British Royal Navy, which was a westward shifting of their base of power in Southeast Asia, was also the dominant factor in ensuring the Dutch were able to control Indonesia for decades.

\textbf{\textcolor{blue}{Concise: }}The British Royal Navy migrated from Java to Singapore, shifting their Southeast Asian power base westward and ensuring the Dutch controlled Indonesia for decades. \cite{cmu2021}

\vspace{12pt}

\textbf{\textcolor{red}{Wordy: }}There are problems with the lease.

\textbf{\textcolor{blue}{Concise: }}The lease has problems. \cite{butte2021}

\vspace{12pt}

\textbf{\textcolor{red}{Wordy: }}There are several good reasons to delay making this decision.

\textbf{\textcolor{blue}{Concise: }}We should delay making this decision for several reasons. \cite{butte2021}

\vspace{12pt}

\textbf{\textcolor{red}{Wordy: }}There is a natural desire among adolescents to experience freedom from authority.

\textbf{\textcolor{blue}{Concise: }}Adolescents naturally crave freedom from authority. \cite{butte2021}

\vspace{12pt}

\textbf{\textcolor{red}{Wordy: }}There is a requirement that all students have an evaluation of their transcripts for placement purposes or to meet a prerequisite.

\textbf{\textcolor{blue}{Concise: }}The college requires that the admissions office evaluate all student transcripts for placement and prerequisites. \cite{butte2021}

\vspace{12pt}

\textbf{\textcolor{red}{Wordy: }}That sofa is large in size.

\textbf{\textcolor{blue}{Concise: }}That sofa is large. \cite{uark2021}

\vspace{12pt}

\textbf{\textcolor{red}{Wordy: }}Compulsory attendance at social events is required.

\textbf{\textcolor{blue}{Concise: }}Attendance at social events is required. \cite{uark2021}

\vspace{12pt}

\textbf{\textcolor{red}{Wordy: }}The housing situation can have a big impact on the social aspect of a student's life.

\textbf{\textcolor{blue}{Concise: }}Housing can influence a student's life. \cite{uark2021}

\vspace{12pt}

\textbf{\textcolor{red}{Wordy: }}I think that there are too many issues with this computer.

\textbf{\textcolor{blue}{Concise: }}There are too many issues with this computer. \cite{uark2021}

\vspace{12pt}

\textbf{\textcolor{red}{Wordy: }}The point that I wish to make is that Google is a very effective and innovative company.

\textbf{\textcolor{blue}{Concise: }}Google is an effective and innovative company. \cite{uark2021}

\vspace{12pt}

\textbf{\textcolor{red}{Wordy: }}The fact of the matter is that culture is a complicated term to define.

\textbf{\textcolor{blue}{Concise: }}Culture is a complicated term to define. \cite{uark2021}

\vspace{12pt}

\textbf{\textcolor{red}{Wordy: }}There is general agreement among climate scientists as to the elevation of global temperatures as a result of carbon dioxide in the atmosphere.

\textbf{\textcolor{blue}{Concise: }}Most climate scientists agree that carbon dioxide in the atmosphere is elevating global temperatures. \cite{ding2021}

\vspace{12pt}

\textbf{\textcolor{red}{Wordy: }}Engineers cannot avoid utilising complex and difficult technical terms in order to clearly analyse requirements and describe them in a precise and meaningful way.

\textbf{\textcolor{blue}{Concise: }}Engineers must use technical terms when writing clear requirements specifications. \cite{ding2021}

\vspace{12pt}

\textbf{\textcolor{red}{Wordy: }}Attempts by researchers at identification of the AIDS virus have met with success; however, attempts at development of a vaccine for the immunisation of those at risk have failed.

\textbf{\textcolor{blue}{Concise: }}AIDS researchers have identified the AIDS virus but have failed to develop a vaccine that will immunise those at risk. \cite{ding2021}

\vspace{12pt}

\textbf{\textcolor{red}{Wordy: }}Any departures by the members from established procedures may cause termination of membership by the Board.

\textbf{\textcolor{blue}{Concise: }}If members depart from established procedures, their membership may be terminated by the Board. \cite{ding2021}

\vspace{12pt}

\textbf{\textcolor{red}{Wordy: }}Employability actually depends on certain factors that basically involve generic skills, such as ability to communicate, as much as any particular technical disciplinary knowledge.

\textbf{\textcolor{blue}{Concise: }}Employability depends on generic skills, such as ability to communicate, as much as disciplinary knowledge. \cite{ding2021}

\vspace{12pt}

\textbf{\textcolor{red}{Wordy: }}Going forward, it is recommended that such kinds of issues be considered in the future planning process.

\textbf{\textcolor{blue}{Concise: }}It is recommended that such issues be considered in the planning process. \cite{ding2021}

\vspace{12pt}

\textbf{\textcolor{red}{Wordy: }}It is interesting to note that the fascinating creature called the common poorwill is actually known to hibernate in the wintertime, going into a deep sleep during the coldest months of the year.

\textbf{\textcolor{blue}{Concise: }}Common poorwills hibernate in the winter. \cite{middlebury2021}

\vspace{12pt}

\textbf{\textcolor{red}{Wordy: }}Both of the two samples had very similar concentrations.

\textbf{\textcolor{blue}{Concise: }}The two samples had similar concentrations. \cite{middlebury2021}

\vspace{12pt}

\textbf{\textcolor{red}{Wordy: }}The results suggest that it may be possible to start to show how lasers may be used in such a way.

\textbf{\textcolor{blue}{Concise: }}The results suggest how lasers may be used in this way. \cite{middlebury2021}

\vspace{12pt}

\textbf{\textcolor{red}{Wordy: }}An experiment helps elucidate the transition of charge carriers from polaronic metal states to insulating ordered states.

\textbf{\textcolor{blue}{Concise: }}Charge carriers can transition from polaronic metal states to insulating ordered states. \cite{middlebury2021}

\vspace{12pt}

\textbf{\textcolor{red}{Wordy: }}Unfortunately, inferred mass changes using the gravity field solutions obtained by the groups we previously mentioned do not agree to a level that is considered truly acceptable.

\textbf{\textcolor{blue}{Concise: }}Inferred mass changes using the gravity field solutions obtained by aforementioned groups do not agree to an acceptable level. \cite{middlebury2021}

\vspace{12pt}

\textbf{\textcolor{red}{Wordy: }}At some point in the future, the incredible field of research known as molecular dynamics simulations could be used in this way for use with plant DNA and nanoparticles to potentially provide an insight into changes observed here in plant health and output.

\textbf{\textcolor{blue}{Concise: }}Molecular dynamics simulations of plant DNA and nanoparticles could provide an insight into changes observed in plant health and output. \cite{middlebury2021}

\vspace{12pt}

\textbf{\textcolor{red}{Wordy: }}It is interesting to note that our results show that the topological states are remarkably robust and that their dispersion has the capacity to be tuned in the range that we explored here (m=0-3; n=1) as well as in bulk Sb.

\textbf{\textcolor{blue}{Concise: }}Our results show that the topological states are remarkably robust and that their dispersion can be tuned in the explored range (m=0-3; n=1) and in bulk Sb. \cite{middlebury2021}

\vspace{12pt}

\textbf{\textcolor{red}{Wordy: }}Experiments performed using reverse transcriptase-polymerase chain reaction obviously demonstrated that the hTRT mRNA originated from the cDNA that had been transfected and not the endogenous gene.

\textbf{\textcolor{blue}{Concise: }}Reverse transcriptase-polymerase chain reaction experiments demonstrated that the hTRT mRNA originated from the transfected cDNA and not the endogenous gene. \cite{middlebury2021}

\vspace{12pt}

\textbf{\textcolor{red}{Wordy: }}The stepmother's house was cleaned by Cinderella.

\textbf{\textcolor{blue}{Concise: }}Cinderella cleaned the stepmother's house. \cite{berkeley2021}

\vspace{12pt}

\textbf{\textcolor{red}{Wordy: }}The stepsisters were jealous and envious.

\textbf{\textcolor{blue}{Concise: }}The stepsisters were jealous. \textbf{ OR } The stepsisters were envious. \cite{berkeley2021}

\vspace{12pt}

\textbf{\textcolor{red}{Wordy: }}The mystery lady was one who every eligible man at the ball admired.

\textbf{\textcolor{blue}{Concise: }}Every eligible man at the ball admired the mystery lady. \cite{berkeley2021}

\vspace{12pt}

\textbf{\textcolor{red}{Wordy: }}The research of Yuan et al. (2007) on sustainable architecture in Singapore is considered to be the cream of the crop.

\textbf{\textcolor{blue}{Concise: }}The research of Yuan et al. (2007) on sustainable architecture in Singapore is considered to be the best. \cite{nus2021}

\vspace{12pt}

\textbf{\textcolor{red}{Wordy: }}The veteran researcher has churned out many articles in recent years.

\textbf{\textcolor{blue}{Concise: }}The veteran researcher has produced many articles in recent years. \cite{nus2021}

\vspace{12pt}

\textbf{\textcolor{red}{Wordy: }}The team that was hurriedly put together has not been productive because the members do not share common objectives.

\textbf{\textcolor{blue}{Concise: }}The team that was hurriedly assembled has not been productive because the members do not share common objectives. \cite{nus2021}

\vspace{12pt}

\textbf{\textcolor{red}{Wordy: }}In his attempt to establish absolute control, the dictator sought to wipe out all who were opposed to his rule.

\textbf{\textcolor{blue}{Concise: }}In his attempt to establish absolute control, the dictator sought to eliminate all who were opposed to his rule. \cite{nus2021}

\vspace{12pt}

\textbf{\textcolor{red}{Wordy: }}The auditors' report suggests that the treasurer had tried to cover up the fi nancial irregularities.

\textbf{\textcolor{blue}{Concise: }}The auditors' report suggests that the treasurer had tried to hide the fi nancial irregularities. \cite{nus2021}

\vspace{12pt}

\textbf{\textcolor{red}{Wordy: }}We must be prepared for discomfort in various sectors if we want to bring about change in the system.

\textbf{\textcolor{blue}{Concise: }}We must be prepared for discomfort in various sectors if we want to effect change in the system. \cite{nus2021}

\vspace{12pt}

\textbf{\textcolor{red}{Wordy: }}In the following, I will argue that there is a need for alternative punishments.

\textbf{\textcolor{blue}{Concise: }}The following will argue that there is a need for alternative punishments. \cite{kuleuven2021}

\vspace{12pt}

\textbf{\textcolor{red}{Wordy: }}I think the potential mismatch is particularly important.

\textbf{\textcolor{blue}{Concise: }}The potential mismatch is particularly important. \cite{kuleuven2021}

\vspace{12pt}

\textbf{\textcolor{red}{Wordy: }}I find it surprising that these reasons were completely neglected.

\textbf{\textcolor{blue}{Concise: }}Surprisingly, these reasons were completely neglected. \cite{kuleuven2021}

\vspace{12pt}

\textbf{\textcolor{red}{Wordy: }}This is the basic background on which he bases his thinking on popular music.

\textbf{\textcolor{blue}{Concise: }}He based his thinking on popular music on the aforementioned theories. \cite{kuleuven2021}

\vspace{12pt}

\textbf{\textcolor{red}{Wordy: }}It is believed by the candidate that a ceiling must be placed on the budget by Congress.

\textbf{\textcolor{blue}{Concise: }}The candidate believes that Congress must place a ceiling on the budget. \cite{wisc2021}

\vspace{12pt}

\textbf{\textcolor{red}{Wordy: }}The man was bitten by the dog.

\textbf{\textcolor{blue}{Concise: }}The dog bit the man. \cite{wisc2021}

\vspace{12pt}

\textbf{\textcolor{red}{Wordy: }}The establishment of a different approach on the part of the committee has become a necessity.

\textbf{\textcolor{blue}{Concise: }}The committee has to approach it differently. \cite{wisc2021}

\vspace{12pt}

\textbf{\textcolor{red}{Wordy: }}An evaluation of the procedures needs to be done.

\textbf{\textcolor{blue}{Concise: }}We need to evaluate the procedures. \cite{wisc2021}

\vspace{12pt}

\textbf{\textcolor{red}{Wordy: }}The procedures need to be evaluated.

\textbf{\textcolor{blue}{Concise: }}We need to evaluate the procedures. \cite{wisc2021}

\vspace{12pt}

\textbf{\textcolor{red}{Wordy: }}The stability and quality of our financial performance will be developed through the profitable execution of our existing business, as well as the acquisition or development of new businesses.

\textbf{\textcolor{blue}{Concise: }}We will improve our financial performance not only by executing our existing business more profitably but by acquiring or developing new businesses. \cite{wisc2021}

\vspace{12pt}

\textbf{\textcolor{red}{Wordy: }}In a situation in which a class is overenrolled, you may request that the instructor force-add you.

\textbf{\textcolor{blue}{Concise: }}When a class is overenrolled, you may ask the instructor to force-add you. \cite{wisc2021}

\vspace{12pt}

\textbf{\textcolor{red}{Wordy: }}I will now make a few observations concerning the matter of contingency funds.

\textbf{\textcolor{blue}{Concise: }}I will now make a few observations about contingency funds. \cite{wisc2021}

\vspace{12pt}

\textbf{\textcolor{red}{Wordy: }}The obvious effect of such a range of reference is to assure the audience of the author's range of learning and intellect.

\textbf{\textcolor{blue}{Concise: }}The wide-ranging references in this talk assure the audience that the author is intelligent and well-read. \cite{wisc2021}

\vspace{12pt}

\textbf{\textcolor{red}{Wordy: }}It is a matter of the gravest possible importance to the health of anyone with a history of a problem with disease of the heart that he or she should avoid the sort of foods with a high percentage of saturated fats.

\textbf{\textcolor{blue}{Concise: }}Anyone with a history of heart disease should avoid saturated fats. \cite{wisc2021}

\vspace{12pt}

\textbf{\textcolor{red}{Wordy: }}It was her last argument that finally persuaded me.

\textbf{\textcolor{blue}{Concise: }}Her last argument finally persuaded me. \cite{wisc2021}

\vspace{12pt}

\textbf{\textcolor{red}{Wordy: }}There are likely to be many researchers raising questions about this methodological approach.

\textbf{\textcolor{blue}{Concise: }}Many researchers are likely to raise questions about this methodological approach. \cite{wisc2021}

\vspace{12pt}

\textbf{\textcolor{red}{Wordy: }}It is inevitable that oil prices will rise.

\textbf{\textcolor{blue}{Concise: }}Oil prices will inevitably rise. \cite{wisc2021}

\vspace{12pt}

\textbf{\textcolor{red}{Wordy: }}Consumer demand is rising in the area of services.

\textbf{\textcolor{blue}{Concise: }}Consumers are demanding more services. \cite{wisc2021}

\vspace{12pt}

\textbf{\textcolor{red}{Wordy: }}Strong reading skills are an important factor in students' success in college.

\textbf{\textcolor{blue}{Concise: }}Students' success in college depends on their reading skills. \cite{wisc2021}

\vspace{12pt}

\textbf{\textcolor{red}{Wordy: }}MHS has a hospital employee relations improvement program.

\textbf{\textcolor{blue}{Concise: }}MHS has a program to improve relations among employees. \cite{wisc2021}

\vspace{12pt}

\textbf{\textcolor{red}{Wordy: }}The magazines that are on the table belong to me.

\textbf{\textcolor{blue}{Concise: }}The magazines on the table belong to me. \cite{gmu2020}

\vspace{12pt}

\textbf{\textcolor{red}{Wordy: }}Students who are living on campus will receive a refund.

\textbf{\textcolor{blue}{Concise: }}Students living on campus will receive a refund. \cite{gmu2020}

\vspace{12pt}

\textbf{\textcolor{red}{Wordy: }}The scholar who had been nominated for the award published another article.

\textbf{\textcolor{blue}{Concise: }}The scholar nominated for the award published another article. \cite{gmu2020}

\vspace{12pt}

\textbf{\textcolor{red}{Wordy: }}On domestic flights that last at least three hours, a meal is served.

\textbf{\textcolor{blue}{Concise: }}On domestic flights lasting at least three hours, a meal is served. \cite{gmu2020}

\vspace{12pt}

\textbf{\textcolor{red}{Wordy: }}The project is one of complexity.

\textbf{\textcolor{blue}{Concise: }}It is a complex project. \cite{cityu2021}

\vspace{12pt}

\textbf{\textcolor{red}{Wordy: }}IT assisted in the maintenance of the database.

\textbf{\textcolor{blue}{Concise: }}IT helped to maintain the database. \cite{cityu2021}

\vspace{12pt}

\textbf{\textcolor{red}{Wordy: }}Everyone's co-operation is of vital importance.

\textbf{\textcolor{blue}{Concise: }}It's very important for everyone to co-operate. \cite{cityu2021}

\vspace{12pt}

\textbf{\textcolor{red}{Wordy: }}The Equal Opportunities Commission (EOC) is responsible for the implementation of anti-discrimination legislation.

\textbf{\textcolor{blue}{Concise: }}The Equal Opportunities Commission (EOC) is responsible for implementing anti-discrimination legislation. \cite{cityu2021}

\vspace{12pt}

\textbf{\textcolor{red}{Wordy: }}We have conducted an investigation and arrived at the conclusion.

\textbf{\textcolor{blue}{Concise: }}We have investigated and concluded. \cite{cityu2021}

\vspace{12pt}

\textbf{\textcolor{red}{Wordy: }}I have made some enquiries.

\textbf{\textcolor{blue}{Concise: }}I have enquired. \cite{cityu2021}

\vspace{12pt}

\textbf{\textcolor{red}{Wordy: }}We are making arrangements.

\textbf{\textcolor{blue}{Concise: }}We are arranging. \cite{cityu2021}

\vspace{12pt}

\textbf{\textcolor{red}{Wordy: }}We are in receipt of your proposal dated 26 January.

\textbf{\textcolor{blue}{Concise: }}We have received your proposal of 26 January. \cite{cityu2021}

\vspace{12pt}

\textbf{\textcolor{red}{Wordy: }}Upon completion of the tendering process, the project can be awarded to the successful bidder.

\textbf{\textcolor{blue}{Concise: }}When the tendering process is completed, we will award the project to the successful bidder. \cite{cityu2021}

\vspace{12pt}

\textbf{\textcolor{red}{Wordy: }}Arrangements have been made to make the payment to you next Monday.

\textbf{\textcolor{blue}{Concise: }}We will pay you next Monday. \cite{cityu2021}

\vspace{12pt}

\textbf{\textcolor{red}{Wordy: }}In order to facilitate the implementation of the new strategic plan, we require the assistance of either one or two experienced specialists.

\textbf{\textcolor{blue}{Concise: }}To implement the new plan, we need two specialist staff. \cite{cityu2021}

\vspace{12pt}

\textbf{\textcolor{red}{Wordy: }}Consuello Calvin, our Chief Consultant, suggested at our last board meeting the installation of new software to enhance security protection.

\textbf{\textcolor{blue}{Concise: }}At our last board meeting, Chief Consultant Consuello Calvin suggested that we install new security protection software. \cite{cityu2021}

\vspace{12pt}

\textbf{\textcolor{red}{Wordy: }}I am currently engaged in a study of the target market.

\textbf{\textcolor{blue}{Concise: }}I am studying the target market at the moment. \cite{cityu2021}

\vspace{12pt}

\textbf{\textcolor{red}{Wordy: }}We didn't renew the contract with our supplier on account of the fact that it was necessary for us to find a cheaper supplier.

\textbf{\textcolor{blue}{Concise: }}We didn't renew the contract with our supplier because we had to find a cheaper one. \cite{cityu2021}

\vspace{12pt}

\textbf{\textcolor{red}{Wordy: }}It is important that you read the notes, advice and information detailed opposite then complete the form overleaf (all sections) prior to its immediate return to the bank by way of the envelope provided.

\textbf{\textcolor{blue}{Concise: }}Please read the attached information and complete and return the form. \cite{cityu2021}

\vspace{12pt}

\textbf{\textcolor{red}{Wordy: }}In the event that there is a fire on the premises, it is vitally important that you shall leave the premises as soon as you hear the alarm bell ringing.

\textbf{\textcolor{blue}{Concise: }}If you hear the fire alarm, please leave the building. \cite{cityu2021}

\vspace{12pt}

\textbf{\textcolor{red}{Wordy: }}My team of staff and I are committed to maintaining and the promotion of Hong Kong as a major international financial centre.

\textbf{\textcolor{blue}{Concise: }}My team of staff and I are committed to maintaining and promoting Hong Kong as a major international financial centre. \cite{cityu2021}

\vspace{12pt}

\textbf{\textcolor{red}{Wordy: }}Thomas Jefferson's support of the new Constitution was documented in a letter to James Madison.

\textbf{\textcolor{blue}{Concise: }}Thomas Jefferson documented his support of the new Constitution in a letter to James Madison. \cite{yale_nus2021}

\vspace{12pt}

\textbf{\textcolor{red}{Wordy: }}It is the combination of these two elements that makes the argument weak.

\textbf{\textcolor{blue}{Concise: }}The combination of these two elements weakens the argument. \cite{yale_nus2021}

\vspace{12pt}

\textbf{\textcolor{red}{Wordy: }}The reason that General Lee invaded Pennsylvania in June 1863 was to draw the Army of the Potomac away from Richmond.

\textbf{\textcolor{blue}{Concise: }}General Lee invaded Pennsylvania in June 1863 to draw the Army of the Potomac away from Richmond. \cite{yale_nus2021}

\vspace{12pt}

\textbf{\textcolor{red}{Wordy: }}Tom Jones is a novel by Fielding that comically portrays English society in the eighteenth century.

\textbf{\textcolor{blue}{Concise: }}Fielding's novel Tom Jones comically portrays English society in the eighteenth century. \cite{yale_nus2021}

\vspace{12pt}

\textbf{\textcolor{red}{Wordy: }}In Crew's argument there are many indications of her misunderstanding of natural selection.

\textbf{\textcolor{blue}{Concise: }}Crew's argument repeatedly demonstrates misunderstanding of natural selection. \cite{yale_nus2021}

\vspace{12pt}

\textbf{\textcolor{red}{Wordy: }}In the original state constitution, they allowed polygamy.

\textbf{\textcolor{blue}{Concise: }}The original state constitution allowed polygamy. \cite{yale_nus2021}

\vspace{12pt}

\textbf{\textcolor{red}{Wordy: }}Each student must meet his or her advisor.

\textbf{\textcolor{blue}{Concise: }}Students must meet with their advisors. \cite{yale_nus2021}

\vspace{12pt}

\textbf{\textcolor{red}{Wordy: }}The majority opinion contains a discussion of legislative history.

\textbf{\textcolor{blue}{Concise: }}The majority opinion discusses legislative history. \cite{uvalawyer2021}

\vspace{12pt}

\textbf{\textcolor{red}{Wordy: }}If you find yourself swimming in the ocean, be wary of sharks.

\textbf{\textcolor{blue}{Concise: }}Be wary of sharks when swimming in the ocean. \cite{hotaling2020simple}

\vspace{12pt}

\textbf{\textcolor{red}{Wordy: }}We conducted an investigation of the accident.

\textbf{\textcolor{blue}{Concise: }}We investigated the accident. \cite{nps2021}

\vspace{12pt}

\textbf{\textcolor{red}{Wordy: }}In each picture, the responses are shown.

\textbf{\textcolor{blue}{Concise: }}Each picture shows the responses. \cite{nps2021}

\vspace{12pt}

\textbf{\textcolor{red}{Wordy: }}It has been found experimentally that genetically altered strawberries are frost-resistant.

\textbf{\textcolor{blue}{Concise: }}In this experiment, we found that genetically altered strawberries are frost-resistant. \cite{nps2021}

\vspace{12pt}

\textbf{\textcolor{red}{Wordy: }}Indirect predator control is the exclusion of predators through means.

\textbf{\textcolor{blue}{Concise: }}Indirect predator control excludes predators through means. \cite{nps2021}

\vspace{12pt}

\textbf{\textcolor{red}{Wordy: }}The increasing electrical output requires the addition of extra modules.

\textbf{\textcolor{blue}{Concise: }}The increasing electrical output requires adding extra modules. \cite{nps2021}

\vspace{12pt}

\textbf{\textcolor{red}{Wordy: }}Early childhood thought disorders misdiagnosis often occurs as a result of unfamiliarity with recent research literature describing such conditions.

\textbf{\textcolor{blue}{Concise: }}Physicians unfamiliar with the literature on recent research often misdiagnose disordered thought in young children. \cite{nps2021}

\vspace{12pt}

\textbf{\textcolor{red}{Wordy: }}There are several advantages to genetic screening.

\textbf{\textcolor{blue}{Concise: }}Genetic screening presents several advantages. \cite{nps2021}

\vspace{12pt}

\textbf{\textcolor{red}{Wordy: }}There has been a lot of research done to determine its causes.

\textbf{\textcolor{blue}{Concise: }}Much research has been done to determine its causes. \cite{nps2021}

\vspace{12pt}

\textbf{\textcolor{red}{Wordy: }}There were many factors that affected this situation.

\textbf{\textcolor{blue}{Concise: }}Many factors affected this situation. \cite{nps2021}

\vspace{12pt}

\textbf{\textcolor{red}{Wordy: }}It is recommended that we complete more research.

\textbf{\textcolor{blue}{Concise: }}We should complete more research. \cite{nps2021}

\vspace{12pt}

\textbf{\textcolor{red}{Wordy: }}It is our goal to serve the university community.

\textbf{\textcolor{blue}{Concise: }}Our goal is to serve the university community. \cite{nps2021}

\vspace{12pt}

\textbf{\textcolor{red}{Wordy: }}The new regulations could cause problems for both the winners and for those who lose.

\textbf{\textcolor{blue}{Concise: }}The new regulations could cause problems for both winners and losers. \cite{nps2021}

\vspace{12pt}

\textbf{\textcolor{red}{Wordy: }}Everyone should be cognizant of the danger of explosion.

\textbf{\textcolor{blue}{Concise: }}Everyone should be aware of the danger of explosion. \cite{nps2021}

\vspace{12pt}

\textbf{\textcolor{red}{Wordy: }}Utilization of crystal clear goals and objectives will optimize our capacity to prioritize our concerns so that we will impact upon the major thrust of our company's future plans and prospects.

\textbf{\textcolor{blue}{Concise: }}If we clarify our goals and objectives, we will be better able to concentrate on what is most important for our company's future. \cite{nps2021}

\vspace{12pt}

\textbf{\textcolor{red}{Wordy: }}After a time interval of one to two minutes, the tone usually stops.

\textbf{\textcolor{blue}{Concise: }}After one to two minutes, the tone usually stops. \cite{nps2021}

\vspace{12pt}

\textbf{\textcolor{red}{Wordy: }}Another activity that I would use my computer for is the wordprocessing software package.

\textbf{\textcolor{blue}{Concise: }}I would also use my computer for wordprocessing. \cite{nps2021}

\vspace{12pt}

\textbf{\textcolor{red}{Wordy: }}In response to the issue of equality for educational and occupational mobility, it is my belief that a system of gender inequality exists in the school system.

\textbf{\textcolor{blue}{Concise: }}I believe that gender inequality exists in the schools. \cite{umass2021}

\vspace{12pt}

\textbf{\textcolor{red}{Wordy: }}The human immune system is responsible not only for the identification of foreign molecules, but also for actions leading to their immobilization, neutralization, and destruction.

\textbf{\textcolor{blue}{Concise: }}The human immune system not only identifies foreign molecules, but also immobilizes, neutralizes, and destroys them. \cite{celia2021}

\vspace{12pt}

\textbf{\textcolor{red}{Wordy: }}A numerical approach was devised that enables the fast and efficient determination of the ternary diagrams associated with our Gibbs free energy.

\textbf{\textcolor{blue}{Concise: }}A new numerical approach quickly and efficiently determines the ternary diagrams associated with our Gibbs free energy. \cite{celia2021}

\vspace{12pt}

\textbf{\textcolor{red}{Wordy: }}Generally, a storage ring is used not only to store charged particles, but also to define their energy and trajectory.

\textbf{\textcolor{blue}{Concise: }}Generally, a storage ring not only stores charged particles but also defines their energy and trajectory. \cite{celia2021}

\vspace{12pt}

\textbf{\textcolor{red}{Wordy: }}There were several methods used to produce the thin metal substrates - hot stamping, cold rolling, and cleaving.

\textbf{\textcolor{blue}{Concise: }}Thin metal substrates were produced by several methods - hot stamping, cold rolling, and cleaving. \cite{celia2021}

\vspace{12pt}

\textbf{\textcolor{red}{Wordy: }}In the future, please turn in assignments on time, by or on the deadline.

\textbf{\textcolor{blue}{Concise: }}Please turn in assignments on time. \cite{boras2021}

\vspace{12pt}

\textbf{\textcolor{red}{Wordy: }}I finished my math and science homework while I was in study hall.

\textbf{\textcolor{blue}{Concise: }}I finished my math and science homework in study hall. \cite{boras2021}

\vspace{12pt}

\textbf{\textcolor{red}{Wordy: }}We outlined the letters for the banner in a careful way.

\textbf{\textcolor{blue}{Concise: }}We outlined the letters for the banner carefully. \cite{boras2021}

\vspace{12pt}

\textbf{\textcolor{red}{Wordy: }}Use a diminutive word, please.

\textbf{\textcolor{blue}{Concise: }}Use a small word, please. \cite{boras2021}

\vspace{12pt}

\textbf{\textcolor{red}{Wordy: }}The man was seen by the cat resting on the window.

\textbf{\textcolor{blue}{Concise: }}The cat resting on the window saw the man. \cite{arkansas_university2021}

\vspace{12pt}

\textbf{\textcolor{red}{Wordy: }}When I went over to the meadow, I picked up some flowers and then placed the flowers inside of a vase at my grandmother's house.

\textbf{\textcolor{blue}{Concise: }}I placed flowers in a vase at my grandmother's house that I picked from a meadow. \cite{arkansas_university2021}

\vspace{12pt}

\textbf{\textcolor{red}{Wordy: }}He went to lie down on the bed.

\textbf{\textcolor{blue}{Concise: }}He lay on the bed. \cite{arkansas_university2021}

\vspace{12pt}

\textbf{\textcolor{red}{Wordy: }}The fact that the cat had a sweet personality factored into why I decided to adopt her.

\textbf{\textcolor{blue}{Concise: }}I adopted the cat because she had a sweet personality. \cite{arkansas_university2021}

\vspace{12pt}

\textbf{\textcolor{red}{Wordy: }}She used things like examples to convey her message.

\textbf{\textcolor{blue}{Concise: }}She used examples to convey her message. \cite{arkansas_university2021}

\vspace{12pt}

\textbf{\textcolor{red}{Wordy: }}It was found that after the main shock, several smaller movements continued to occur.

\textbf{\textcolor{blue}{Concise: }}Several smaller movements continued to occur after the main shock. \cite{adelaide2021}

\vspace{12pt}

\textbf{\textcolor{red}{Wordy: }}It was not until after the last batch of votes was counted, that victory could be declared by the Senator.

\textbf{\textcolor{blue}{Concise: }}The Senator declared victory after the last batch of votes was counted. \cite{adelaide2021}

\vspace{12pt}

\textbf{\textcolor{red}{Wordy: }}The warning system promulgated its message just minutes before the tsunami struck.

\textbf{\textcolor{blue}{Concise: }}The warning system was activated just minutes before the tsunami struck. \cite{adelaide2021}

\vspace{12pt}

\textbf{\textcolor{red}{Wordy: }}This essay discusses the successes of Japan's Disaster Risk Management (DRM) system, as well as the ways in which that system could be improved.

\textbf{\textcolor{blue}{Concise: }}This essay discusses the successes of Japan's Disaster Risk Management (DRM) system, and how that system could be improved. \cite{adelaide2021}

\vspace{12pt}

\textbf{\textcolor{red}{Wordy: }}As mentioned earlier, Japan's DRM system could be improved.

\textbf{\textcolor{blue}{Concise: }}Japan's DRM system could be improved. \cite{adelaide2021}

\vspace{12pt}

\textbf{\textcolor{red}{Wordy: }}In my opinion, it is important that all college students participate in the electoral process by casting their vote.

\textbf{\textcolor{blue}{Concise: }}All college students should vote. \cite{san_jose2021}

\vspace{12pt}

\textbf{\textcolor{red}{Wordy: }}Actually, Mary kind of glanced at Bob when she realized they had basically lost the battle.

\textbf{\textcolor{blue}{Concise: }}Mary glanced at Bob when she realized they had lost the battle. \cite{san_jose2021}

\vspace{12pt}

\textbf{\textcolor{red}{Wordy: }}The future plan Congress will propose to completely overhaul the healthcare system could negatively frustrate constituents.

\textbf{\textcolor{blue}{Concise: }}The plan Congress will propose to overhaul the healthcare system could frustrate constituents. \cite{san_jose2021}

\vspace{12pt}

\textbf{\textcolor{red}{Wordy: }}When I asked her about the new job, she sort of looked at me with anger and basically did not reply.

\textbf{\textcolor{blue}{Concise: }}When I asked her about the new job, she looked at me with anger and did not reply. \cite{san_jose2021}

\vspace{12pt}

\textbf{\textcolor{red}{Wordy: }}Now that Bob has truly been fired, appealing to the CEO is kind of useless.

\textbf{\textcolor{blue}{Concise: }}Now that Bob has been fired, appealing to the CEO is useless. \cite{san_jose2021}

\vspace{12pt}

\textbf{\textcolor{red}{Wordy: }}Each individual will receive a free gift during the ceremony.

\textbf{\textcolor{blue}{Concise: }}Each individual will receive a gift during the ceremony. \cite{san_jose2021}

\vspace{12pt}

\textbf{\textcolor{red}{Wordy: }}Committee members first began by discussing the new marketing plan; then they presented the true facts about foreign imports from Belgium.

\textbf{\textcolor{blue}{Concise: }}Committee members first discussed the new marketing plan; then they presented the facts about imports from Belgium. \cite{san_jose2021}

\vspace{12pt}

\textbf{\textcolor{red}{Wordy: }}Anthropologists really believe these old fossils are absolutely crucial to uncovering our past history.

\textbf{\textcolor{blue}{Concise: }}Anthropologists believe these fossils are crucial to uncovering our history. \cite{san_jose2021}

\vspace{12pt}

\textbf{\textcolor{red}{Wordy: }}The zombies' feast of the survivors occurred.

\textbf{\textcolor{blue}{Concise: }}The zombies feasted on the survivors. \cite{nevada2021}

\vspace{12pt}

\textbf{\textcolor{red}{Wordy: }}In this passage is an example of the use of the rule of justice in argumentation.

\textbf{\textcolor{blue}{Concise: }}This passage exemplifies argumentation using the rule of justice. \cite{richmond2021}

\vspace{12pt}

\textbf{\textcolor{red}{Wordy: }}The point I wish to make is that fish sleep with their eyes open.

\textbf{\textcolor{blue}{Concise: }}Fish sleep with their eyes open. \cite{richmond2021}

\vspace{12pt}

\textbf{\textcolor{red}{Wordy: }}Burning books is considered censorship by some people.

\textbf{\textcolor{blue}{Concise: }}Some people consider burning books censorship. \cite{richmond2021}

\vspace{12pt}

\textbf{\textcolor{red}{Wordy: }}The theory of relativity isn't demonstrated by this experiment.

\textbf{\textcolor{blue}{Concise: }}This experiment does not demonstrate the theory of relativity. \cite{richmond2021}

\vspace{12pt}

\textbf{\textcolor{red}{Wordy: }}For the most part, individuals in graduate classes are appreciative of conciseness.

\textbf{\textcolor{blue}{Concise: }}Graduate students generally appreciate conciseness. \cite{une2020}

\vspace{12pt}

\textbf{\textcolor{red}{Wordy: }}Superintendents will be interviewed by the teachers they work for.

\textbf{\textcolor{blue}{Concise: }}Teachers will interview their superintendents. \cite{une2020}

\vspace{12pt}

\textbf{\textcolor{red}{Wordy: }}The success of the pilot program is reflected in the number of children in elementary school who are healthy.

\textbf{\textcolor{blue}{Concise: }}The number of healthy elementary students reflects the pilot program's success. \cite{une2020}

\vspace{12pt}

\textbf{\textcolor{red}{Wordy: }}Improvements in your writing will be seen if you remove prepositional phrases that aren't necessary.

\textbf{\textcolor{blue}{Concise: }}Removing unnecessary prepositional phrases will improve your writing. \cite{une2020}

\vspace{12pt}

\textbf{\textcolor{red}{Wordy: }}Write a biographical essay of your life.

\textbf{\textcolor{blue}{Concise: }}Write a biography. \cite{une2020}

\vspace{12pt}

\textbf{\textcolor{red}{Wordy: }}Reach a consensus of opinion with your group members.

\textbf{\textcolor{blue}{Concise: }}Reach a consensus with your group members. \cite{une2020}

\vspace{12pt}

\textbf{\textcolor{red}{Wordy: }}To get started with your reflective writing assignment, imagine a mental picture of a time when you found learning to be extremely difficult.

\textbf{\textcolor{blue}{Concise: }}As you begin your reflection, remember a time when learning was difficult. \cite{une2020}

\vspace{12pt}

\textbf{\textcolor{red}{Wordy: }}As you complete the analysis, be keep in mind the methods and procedures that direct the processes of our judicial system.

\textbf{\textcolor{blue}{Concise: }}As you complete the analysis, consider the common processes of the judiciary. \cite{une2020}

\vspace{12pt}

\textbf{\textcolor{red}{Wordy: }}Your assigned reading for the course will describe many of the problems and mistakes that can make projects more difficult for students than they need to be.

\textbf{\textcolor{blue}{Concise: }}Your assigned course reading describes many project pitfalls students may encounter. \cite{une2020}

\vspace{12pt}

\textbf{\textcolor{red}{Wordy: }}The brief was read by us.

\textbf{\textcolor{blue}{Concise: }}We read the brief. \cite{georgetown_law2021}

\vspace{12pt}

\textbf{\textcolor{red}{Wordy: }}The evidence was suppressed by the court.

\textbf{\textcolor{blue}{Concise: }}The court suppressed the evidence. \cite{georgetown_law2021}

\vspace{12pt}

\textbf{\textcolor{red}{Wordy: }}The holding was reached by the court.

\textbf{\textcolor{blue}{Concise: }}The court held. \cite{georgetown_law2021}

\vspace{12pt}

\textbf{\textcolor{red}{Wordy: }}The argument was presented by the plaintiff.

\textbf{\textcolor{blue}{Concise: }}The plaintiff argued. \cite{georgetown_law2021}

\vspace{12pt}

\textbf{\textcolor{red}{Wordy: }}A complaint was filed by the union.

\textbf{\textcolor{blue}{Concise: }}The union filed the complaint. \cite{georgetown_law2021}

\vspace{12pt}

\textbf{\textcolor{red}{Wordy: }}Our conclusion is supported by legislative history.

\textbf{\textcolor{blue}{Concise: }}Legislative history supports our conclusion. \cite{georgetown_law2021}

\vspace{12pt}

\textbf{\textcolor{red}{Wordy: }}It is possible for the court to modify the judgment.

\textbf{\textcolor{blue}{Concise: }}The court can modify the judgment. \cite{georgetown_law2021}

\vspace{12pt}

\textbf{\textcolor{red}{Wordy: }}The implementation of the plan by the team was successful.

\textbf{\textcolor{blue}{Concise: }}The team successfully implemented the plan. \cite{georgetown_law2021}

\vspace{12pt}

\textbf{\textcolor{red}{Wordy: }}There was an affirmative decision for expansion by the administration.

\textbf{\textcolor{blue}{Concise: }}The administration decided to expand the program. \cite{georgetown_law2021}

\vspace{12pt}

\textbf{\textcolor{red}{Wordy: }}A revision of the program will result in increases of efficiency to our clients.

\textbf{\textcolor{blue}{Concise: }}If we revise the program, we can serve clients more efficiently. \cite{georgetown_law2021}

\vspace{12pt}

\textbf{\textcolor{red}{Wordy: }}The smoke that comes from factories that are situated in the valley pollutes the air.

\textbf{\textcolor{blue}{Concise: }}The smoke from factories in the valley pollutes the air. \cite{byu2021}

\vspace{12pt}

\textbf{\textcolor{red}{Wordy: }}University students are required by the university to make payments of their tuition fees before the time of their registration.

\textbf{\textcolor{blue}{Concise: }}University students are required to pay tuition before registering. \cite{byu2021}

\vspace{12pt}

\textbf{\textcolor{red}{Wordy: }}The match had been won by the world champion shortly after it started.

\textbf{\textcolor{blue}{Concise: }}The world champion won the match shortly after it started. \cite{byu2021}

\vspace{12pt}

\textbf{\textcolor{red}{Wordy: }}The accident occurred due to the fact that there was nothing to prevent it.

\textbf{\textcolor{blue}{Concise: }}The accident occurred because there was nothing to prevent it. \cite{byu2021}

\vspace{12pt}

\textbf{\textcolor{red}{Wordy: }}The nature of the crisis situation was such that it called for our immediate attention.

\textbf{\textcolor{blue}{Concise: }}The crisis called for our immediate attention. \cite{byu2021}

\vspace{12pt}

\textbf{\textcolor{red}{Wordy: }}Upon reflection, I remembered eating bread and cheese at the reception.

\textbf{\textcolor{blue}{Concise: }}I remembered eating bread and cheese at the reception. \cite{byu2021}

\vspace{12pt}

\textbf{\textcolor{red}{Wordy: }}In my opinion, in response to the issue of wordy sentences in college writing, it is my belief that a tool for revision is needed.

\textbf{\textcolor{blue}{Concise: }}I believe college students need sentence revision strategies. \cite{jmu2021}

\vspace{12pt}

\textbf{\textcolor{red}{Wordy: }}What I wish to show here is that James Madison University is a good representation of a place with many backgrounds and interests of the students.

\textbf{\textcolor{blue}{Concise: }}James Madison University represents a diverse university. \cite{jmu2021}

\vspace{12pt}

\textbf{\textcolor{red}{Wordy: }}The report is waiting for your approval.

\textbf{\textcolor{blue}{Concise: }}The report awaits your approval. \cite{ieee2021}

\vspace{12pt}

\textbf{\textcolor{red}{Wordy: }}It is a matter of the gravest importance to the health of anyone who uses a microwave and has a heart condition to avoid standing in front of the microwave while it is running.

\textbf{\textcolor{blue}{Concise: }}Anyone with a heart condition should avoid standing in front of an operating microwave oven. \cite{ieee2021}

\vspace{12pt}

\textbf{\textcolor{red}{Wordy: }}My sister is loud when she is telling her children to do their homework.

\textbf{\textcolor{blue}{Concise: }}My sister bellows when telling her children to do their homework. \cite{stlcc2021}

\vspace{12pt}

\textbf{\textcolor{red}{Wordy: }}The restaurant's food is excellent.

\textbf{\textcolor{blue}{Concise: }}The restaurant serves excellent food. \cite{stlcc2021}

\vspace{12pt}

\textbf{\textcolor{red}{Wordy: }}Jennifer Lawrence is the star of The Hunger Games.

\textbf{\textcolor{blue}{Concise: }}Jennifer Lawrence stars in The Hunger Games. \cite{stlcc2021}

\vspace{12pt}

\textbf{\textcolor{red}{Wordy: }}Our mother was the driver of the bus.

\textbf{\textcolor{blue}{Concise: }}Our mother drove the bus. \cite{stlcc2021}

\vspace{12pt}

\textbf{\textcolor{red}{Wordy: }}It is my intention to transfer to a four-year university.

\textbf{\textcolor{blue}{Concise: }}I intend to transfer to a four-year university. \cite{stlcc2021}

\vspace{12pt}

\textbf{\textcolor{red}{Wordy: }}There were some important findings resulting from this experiment.

\textbf{\textcolor{blue}{Concise: }}This experiment resulted in some important findings. \cite{stlcc2021}

\vspace{12pt}

\textbf{\textcolor{red}{Wordy: }}He struggled with the paper that was assigned by the professor.

\textbf{\textcolor{blue}{Concise: }}He struggled with the paper assigned by the professor. \cite{stlcc2021}

\vspace{12pt}

\textbf{\textcolor{red}{Wordy: }}Electric car technology was in existence as early as 1830.

\textbf{\textcolor{blue}{Concise: }}Electric car technology existed as early as 1830. \cite{stlcc2021}

\vspace{12pt}

\textbf{\textcolor{red}{Wordy: }}Negative advertising is influential on voters' perceptions of candidates.

\textbf{\textcolor{blue}{Concise: }}Negative advertising influences voters' perceptions of candidates. \cite{stlcc2021}

\vspace{12pt}

\textbf{\textcolor{red}{Wordy: }}I am in receipt of your letter.

\textbf{\textcolor{blue}{Concise: }}I received your letter. \cite{stlcc2021}

\vspace{12pt}

\textbf{\textcolor{red}{Wordy: }}The stock market was fluctuating wildly before the crash.

\textbf{\textcolor{blue}{Concise: }}The stock market fluctuated wildly before the crash. \cite{stlcc2021}

\vspace{12pt}

\textbf{\textcolor{red}{Wordy: }}The concept of immortality is intriguing to me.

\textbf{\textcolor{blue}{Concise: }}The concept of immortality intrigues me. \cite{stlcc2021}

\vspace{12pt}

\textbf{\textcolor{red}{Wordy: }}In reviewing the notes of his lecture, he referred back to what he had written before class.

\textbf{\textcolor{blue}{Concise: }}He reviewed his lecture notes. \cite{ndsu2021}

\vspace{12pt}

\textbf{\textcolor{red}{Wordy: }}There are four employees who have filed grievances.

\textbf{\textcolor{blue}{Concise: }}Four employees have filed grievances. \cite{ndsu2021}

\vspace{12pt}

\textbf{\textcolor{red}{Wordy: }}It is our suspicion that the basement was substandard.

\textbf{\textcolor{blue}{Concise: }}We suspect the basement was substandard. \cite{ndsu2021}

\vspace{12pt}

\textbf{\textcolor{red}{Wordy: }}What we need to do next is simplify our sign-off procedure.

\textbf{\textcolor{blue}{Concise: }}Next, we need to simplify our sign-off procedure. \cite{ndsu2021}

\vspace{12pt}

\textbf{\textcolor{red}{Wordy: }}Make a revision of this sentence.

\textbf{\textcolor{blue}{Concise: }}Revise this sentence. \cite{ndsu2021}

\vspace{12pt}

\textbf{\textcolor{red}{Wordy: }}We would like to call your attention and to the attention of others in your office to the fact that your report is overdue because the deadline was last Friday.

\textbf{\textcolor{blue}{Concise: }}We remind you that your report is overdue. \cite{palomar2021}

\vspace{12pt}

\textbf{\textcolor{red}{Wordy: }}The document that you refer back to is not included in the sample documents that go to make up the enclosure.

\textbf{\textcolor{blue}{Concise: }}The document is not included in the samples. \cite{palomar2021}

\vspace{12pt}

\textbf{\textcolor{red}{Wordy: }}The purpose of this memo is to provide a reference for all your staff members of the formats of various documents that go forward by transmission to headquarters.

\textbf{\textcolor{blue}{Concise: }}This memo describes the formats for documents going to headquarters. \cite{palomar2021}

\vspace{12pt}

\textbf{\textcolor{red}{Wordy: }}The need for reports that have logic and are relevant is still great.

\textbf{\textcolor{blue}{Concise: }}We need logical, relevant reports. \cite{palomar2021}

\vspace{12pt}

\textbf{\textcolor{red}{Wordy: }}Your statement in your letter that has the contention that the information submitted and sent to you by us contained certain inaccuracies and errors has prompted us to embark on a careful and thorough reevaluation of the information submitted, with the result that the original informative data has been determined to be accurate and correct in all instances and aspects of the information.

\textbf{\textcolor{blue}{Concise: }}As you suggested, we have checked our information and confirmed its accuracy. \cite{palomar2021}

\vspace{12pt}

\textbf{\textcolor{red}{Wordy: }}In the event that some information concerning Mr. Smith should be brought to your attention, it should be forwarded via mail or courier or telephone to us in view of the possibility that the information may reveal any attempt on the part of Mr. Smith to depart from the United States.

\textbf{\textcolor{blue}{Concise: }}If you get any information about Mr. Smith, please contact us in case he tries to leave the country. \cite{palomar2021}

\vspace{12pt}

\textbf{\textcolor{red}{Wordy: }}Sometimes it is not easy to avoid the dummy subject.

\textbf{\textcolor{blue}{Concise: }}Sometimes the dummy subject is not easily avoided. \cite{loyola2021}

\vspace{12pt}

\textbf{\textcolor{red}{Wordy: }}She fell down due to the fact that she hurried.

\textbf{\textcolor{blue}{Concise: }}She fell because she hurried. \cite{loyola2021}

\vspace{12pt}

\textbf{\textcolor{red}{Wordy: }}The resolution to the problem can be seen in author's attempt.

\textbf{\textcolor{blue}{Concise: }}The author resolves the problem. \cite{loyola2021}

\vspace{12pt}

\textbf{\textcolor{red}{Wordy: }}His objective was to win, but playing fair also mattered to him.

\textbf{\textcolor{blue}{Concise: }}His objective was not only to win, but also to play fair. \cite{loyola2021}

\vspace{12pt}

\textbf{\textcolor{red}{Wordy: }}The author shows the reader the path to being virtuous rather than to vice.

\textbf{\textcolor{blue}{Concise: }}The author shows the reader the path to virtue rather than to vice. \cite{loyola2021}

\vspace{12pt}

\textbf{\textcolor{red}{Wordy: }}Both Smith and Jones took different views of the war.

\textbf{\textcolor{blue}{Concise: }}Smith and Jones took different views of the war. \cite{loyola2021}

\vspace{12pt}

\textbf{\textcolor{red}{Wordy: }}Both Smith and Jones took the same view of the war.

\textbf{\textcolor{blue}{Concise: }}Smith and Jones took the same view of the war. \cite{loyola2021}

\vspace{12pt}

\textbf{\textcolor{red}{Wordy: }}For his young readers, the author must avoid intimidating them by taking too much for granted.

\textbf{\textcolor{blue}{Concise: }}The author must avoid intimidating young readers by taking too much for granted. \cite{loyola2021}

\vspace{12pt}

\textbf{\textcolor{red}{Wordy: }}It was earlier demonstrated that heart attacks can be caused by high stress.

\textbf{\textcolor{blue}{Concise: }}Researchers earlier showed that high stress can cause heart attacks. \cite{wisc2021}

\vspace{12pt}

\textbf{\textcolor{red}{Wordy: }}It is imperative that we find a solution.

\textbf{\textcolor{blue}{Concise: }}We must find a solution. \cite{hccs2021}

\vspace{12pt}

\textbf{\textcolor{red}{Wordy: }}As far as I am concerned, discrimination against certain ethnic groups continues to exist for all intents and purposes.

\textbf{\textcolor{blue}{Concise: }}Discrimination against certain ethnic groups continues to exist. \cite{sanmateo2021}

\vspace{12pt}

\textbf{\textcolor{red}{Wordy: }}At this point in time, the latest iPad is expensive due to the fact that it has no competition.

\textbf{\textcolor{blue}{Concise: }}Now, the new iPad is expensive because it has no competition. \cite{sanmateo2021}

\vspace{12pt}

\textbf{\textcolor{red}{Wordy: }}The reason why Jhumpa Lahiri's novel is so great is because it has vivid imagery.

\textbf{\textcolor{blue}{Concise: }}Jhumpa Lahiri 's novel is great because it has vivid imagery. \cite{sanmateo2021}

\vspace{12pt}

\textbf{\textcolor{red}{Wordy: }}A healthy lifestyle enhances your ability both to live life to its fullest and to live to a ripe old age.

\textbf{\textcolor{blue}{Concise: }}A healthy lifestyle helps you live life fully and longer. \cite{sanmateo2021}

\vspace{12pt}

\textbf{\textcolor{red}{Wordy: }}Past history shows that the students who transfer are actually very few in number.

\textbf{\textcolor{blue}{Concise: }}History shows that the students who transfer are actually very few. \cite{sanmateo2021}

\vspace{12pt}

\textbf{\textcolor{red}{Wordy: }}President George W. Bush made serious errors in responding to the sudden crisis that followed the terrible tragedy of September 11.

\textbf{\textcolor{blue}{Concise: }}President George W. Bush made serious errors in responding to the crisis that followed the tragedy of September 11. \cite{sanmateo2021}

\vspace{12pt}

\textbf{\textcolor{red}{Wordy: }}I think that basically we need to use fewer words, or our papers will really be much too long.

\textbf{\textcolor{blue}{Concise: }}We need to use fewer words, or our papers will be too long. \cite{ursinus2021}

\vspace{12pt}

\textbf{\textcolor{red}{Wordy: }}Due to the fact that Ursula babysat every Thursday evening, she could not attend Michael's recital.

\textbf{\textcolor{blue}{Concise: }}Ursula babysat every Thursday evening, she could not attend Michael's recital. \cite{lakeforest2021}

\vspace{12pt}

\textbf{\textcolor{red}{Wordy: }}The committee's first and foremost goal is to strategize ways to eliminate excess spending.

\textbf{\textcolor{blue}{Concise: }}The committee's foremost goal is to strategize ways to eliminate excess spending. \cite{lakeforest2021}

\vspace{12pt}

\textbf{\textcolor{red}{Wordy: }}Do not write in the negative.

\textbf{\textcolor{blue}{Concise: }}Write in the affirmative. \cite{lakeforest2021}

\vspace{12pt}

\textbf{\textcolor{red}{Wordy: }}Not many parishioners commented that Father Donnelly does not have the same charisma as Father Marks.

\textbf{\textcolor{blue}{Concise: }}Few parishioners commented that Father Donnelly lacks Father Marks' charisma. \cite{lakeforest2021}

\vspace{12pt}

\textbf{\textcolor{red}{Wordy: }}There were many students at the lecture.

\textbf{\textcolor{blue}{Concise: }}Many students attended the lecture. \cite{ucalgary2021}

\vspace{12pt}

\textbf{\textcolor{red}{Wordy: }}The patient's condition was very unstable.

\textbf{\textcolor{blue}{Concise: }}The patient's condition was unstable. \cite{ucalgary2021}

\vspace{12pt}

\textbf{\textcolor{red}{Wordy: }}The author does an analysis of the problem.

\textbf{\textcolor{blue}{Concise: }}The author analyzes the problem. \cite{ucalgary2021}

\vspace{12pt}

\textbf{\textcolor{red}{Wordy: }}In my opinion, one of the most important steps of the learning process is understanding how the learning process happens.

\textbf{\textcolor{blue}{Concise: }}One of the most important steps of the learning process is understanding how it happens. \cite{ucalgary2021}

\vspace{12pt}

\textbf{\textcolor{red}{Wordy: }}In my opinion, the role of nurse practitioner (NP) should be introduced to more clinics since NPs can perform many of the tasks assumed by physicians.

\textbf{\textcolor{blue}{Concise: }}The role of nurse practitioner (NP) should be introduced to more clinics since NPs can perform many of the tasks assumed by physicians. \cite{ucalgary2021}

\vspace{12pt}

\textbf{\textcolor{red}{Wordy: }}They did not succeed in reviving the patient.

\textbf{\textcolor{blue}{Concise: }}They failed to revive the patient. \cite{ucalgary2021}

\vspace{12pt}

\textbf{\textcolor{red}{Wordy: }}They were not successful in reviving the patient.

\textbf{\textcolor{blue}{Concise: }}They failed to revive the patient. \cite{ucalgary2021}

\vspace{12pt}

\textbf{\textcolor{red}{Wordy: }}Unencumbered by a sense of responsibility, Jason left his new leather jacket on the bus.

\textbf{\textcolor{blue}{Concise: }}Jason irresponsibly left his new leather jacket on the bus. \cite{college2021}

\vspace{12pt}

\textbf{\textcolor{red}{Wordy: }}There is a unique, eccentric, and opposite way to accomplish this burdensome task.

\textbf{\textcolor{blue}{Concise: }}There is an unusual way to complete this task. \cite{indianatech2021}

\vspace{12pt}

\textbf{\textcolor{red}{Wordy: }}I really enjoy sourdough bread.

\textbf{\textcolor{blue}{Concise: }}I enjoy sourdough bread. \cite{indianatech2021}

\vspace{12pt}

\textbf{\textcolor{red}{Wordy: }}She basically said no to me.

\textbf{\textcolor{blue}{Concise: }}She said no to me. \cite{indianatech2021}

\vspace{12pt}

\textbf{\textcolor{red}{Wordy: }}The gift was received by James.

\textbf{\textcolor{blue}{Concise: }}James received the gift. \cite{indianatech2021}

\vspace{12pt}

\textbf{\textcolor{red}{Wordy: }}Productivity actually depends on certain factors that basically involve psychology more than any particular technology.

\textbf{\textcolor{blue}{Concise: }}Productivity depends on psychology more than on technology. \cite{ualberta2021}

\vspace{12pt}

\textbf{\textcolor{red}{Wordy: }}She proposed various and sundry explanations for the procedure.

\textbf{\textcolor{blue}{Concise: }}She proposed various explanations for the procedure. \cite{ualberta2021}

\vspace{12pt}

\textbf{\textcolor{red}{Wordy: }}Their software programs demonstrated and showed that many companies and businesses could benefit and profit from eliminating unnecessary overhead and expenses.

\textbf{\textcolor{blue}{Concise: }}Their software showed that companies could profit from eliminating unnecessary overhead. \cite{ualberta2021}

\vspace{12pt}

\textbf{\textcolor{red}{Wordy: }}Do not try to predict those future events that will completely revolutionize society, because past history shows that it is the final outcome of minor events that unexpectedly surprises us more.

\textbf{\textcolor{blue}{Concise: }}Do not try to predict revolutionary events, because history shows that the outcome of minor events surprises us more. \cite{ualberta2021}

\vspace{12pt}

\textbf{\textcolor{red}{Wordy: }}In the event that you finish early, contact this office.

\textbf{\textcolor{blue}{Concise: }}If you finish early, contact the office. \cite{ualberta2021}

\vspace{12pt}

\textbf{\textcolor{red}{Wordy: }}There is a need for more careful inspections of all welds.

\textbf{\textcolor{blue}{Concise: }}You must inspect all welds more carefully. \cite{ualberta2021}

\vspace{12pt}

\textbf{\textcolor{red}{Wordy: }}By the time she got home, Merdine was very tired.

\textbf{\textcolor{blue}{Concise: }}By the time she got home, Merdine was exhausted. \cite{ualberta2021}

\vspace{12pt}

\textbf{\textcolor{red}{Wordy: }}She was also really hungry.

\textbf{\textcolor{blue}{Concise: }}She was also hungry. \cite{ualberta2021}

\vspace{12pt}

\textbf{\textcolor{red}{Wordy: }}The last point I would like to make is that in regard to men-women relationships, it is important to keep in mind that the greatest changes have occurred in how they work together.

\textbf{\textcolor{blue}{Concise: }}Men and women have changed their relationships most in how they work together. \cite{ualberta2021}

\vspace{12pt}

\textbf{\textcolor{red}{Wordy: }}High divorce rates have been observed to occur in areas that have been determined to have low population density.

\textbf{\textcolor{blue}{Concise: }}High divorce rates occur in areas with low population density. \cite{ualberta2021}

\vspace{12pt}

\textbf{\textcolor{red}{Wordy: }}This section introduces another problem, that of noise pollution.

\textbf{\textcolor{blue}{Concise: }}Another problem is noise pollution. \cite{ualberta2021}

\vspace{12pt}

\textbf{\textcolor{red}{Wordy: }}The church, which was completed in 1875, dominated the skyline.

\textbf{\textcolor{blue}{Concise: }}Completed in 1875, the church dominated the skyline. \cite{evansville2021}

\vspace{12pt}

\textbf{\textcolor{red}{Wordy: }}The man, who was intelligent, built a time machine.

\textbf{\textcolor{blue}{Concise: }}The intelligent man built a time machine. \cite{evansville2021}

\vspace{12pt}

\textbf{\textcolor{red}{Wordy: }}There are twelve children who would like ice cream.

\textbf{\textcolor{blue}{Concise: }}Twelve children would like ice cream. \cite{evansville2021}

\vspace{12pt}

\textbf{\textcolor{red}{Wordy: }}It is he who stole the car.

\textbf{\textcolor{blue}{Concise: }}He stole the car. \cite{evansville2021}

\vspace{12pt}

\textbf{\textcolor{red}{Wordy: }}Brother Edward is now employed at a private rehabilitation center working as a certified physical therapist.

\textbf{\textcolor{blue}{Concise: }}Brother Edward works at a private rehabilitation center as a certified physical therapist. \cite{uliberty2021}

\vspace{12pt}

\textbf{\textcolor{red}{Wordy: }}Sally was determined in her mind to lose weight.

\textbf{\textcolor{blue}{Concise: }}Sally was determined to lose weight. \cite{uliberty2021}

\vspace{12pt}

\textbf{\textcolor{red}{Wordy: }}Thirty-somethings are often thought of or stereotyped as apathetic.

\textbf{\textcolor{blue}{Concise: }}Thirty-somethings are often stereotyped as apathetic. \cite{uliberty2021}

\vspace{12pt}

\textbf{\textcolor{red}{Wordy: }}Our fifth patient, in room six, is a mentally ill patient.

\textbf{\textcolor{blue}{Concise: }}Our fifth patient, in room six, is mentally ill. \cite{uliberty2021}

\vspace{12pt}

\textbf{\textcolor{red}{Wordy: }}The best instructors help each student to become a better student both academically and emotionally.

\textbf{\textcolor{blue}{Concise: }}The best instructors help each student grow both academically and emotionally. \cite{uliberty2021}

\vspace{12pt}

\textbf{\textcolor{red}{Wordy: }}In my opinion, our current tax policy is misguided.

\textbf{\textcolor{blue}{Concise: }}Our current tax policy is misguided. \cite{uliberty2021}

\vspace{12pt}

\textbf{\textcolor{red}{Wordy: }}It certainly seems that The Telltale Heart is Edgar Allen Poe's most macabre story.

\textbf{\textcolor{blue}{Concise: }}The Telltale Heart is Edgar Allen Poe's most macabre story. \cite{uliberty2021}

\vspace{12pt}

\textbf{\textcolor{red}{Wordy: }}I will file the proper papers in the event that I am unable to meet the deadline.

\textbf{\textcolor{blue}{Concise: }}I will file the proper papers if I cannot meet the deadline. \cite{uliberty2021}

\vspace{12pt}

\textbf{\textcolor{red}{Wordy: }}The investment analyst claimed that because of volatile market conditions, he could not make an estimate of the company's future profits.

\textbf{\textcolor{blue}{Concise: }}The investment analyst claimed that because of volatile market conditions, he could not estimate the company's future profits. \cite{uliberty2021}

\vspace{12pt}

\textbf{\textcolor{red}{Wordy: }}Mervina is responsible for monitoring and balancing the budgets for travel, contract services, and personnel.

\textbf{\textcolor{blue}{Concise: }}Mervina monitors and balances the budgets for travel, contract services, and personnel. \cite{uliberty2021}

\vspace{12pt}

\textbf{\textcolor{red}{Wordy: }}There is another book that tells the story of Karl Marx and explains his theory of socialism.

\textbf{\textcolor{blue}{Concise: }}Another book tells the story of Karl Marx and explains his theory of socialism. \cite{uliberty2021}

\vspace{12pt}

\textbf{\textcolor{red}{Wordy: }}All too often, athletes with minimal academic skills have been recruited by our coaches.

\textbf{\textcolor{blue}{Concise: }}All too often, our coaches have recruited athletes with minimal academic skills. \cite{uliberty2021}

\vspace{12pt}

\textbf{\textcolor{red}{Wordy: }}My family took a side trip to Monticello, which was the home of Thomas Jefferson.

\textbf{\textcolor{blue}{Concise: }}My family took a side trip to Monticello, the home of Thomas Jefferson. \cite{uliberty2021}

\vspace{12pt}

\textbf{\textcolor{red}{Wordy: }}For her birthday we gave Harriet a stylish vest made out of silk.

\textbf{\textcolor{blue}{Concise: }}For her birthday we gave Harriet a stylish silk vest. \cite{uliberty2021}

\vspace{12pt}

\textbf{\textcolor{red}{Wordy: }}The pizza was eaten by the students.

\textbf{\textcolor{blue}{Concise: }}The students ate the pizza. \cite{ukentucky2021}

\vspace{12pt}

\textbf{\textcolor{red}{Wordy: }}The attorney made an investigation of the case.

\textbf{\textcolor{blue}{Concise: }}The attorney investigated the case. \cite{ukentucky2021}

\vspace{12pt}

\textbf{\textcolor{red}{Wordy: }}The cause of our schools' failure at teaching basic skills is not understanding the influence of cultural background on learning.

\textbf{\textcolor{blue}{Concise: }}Our schools have failed to teach basic skills because educators do not understand how cultural backgrounds influence learning. \cite{duke_university2021}

\vspace{12pt}

\textbf{\textcolor{red}{Wordy: }}A revision of the program will result in increases in our efficiency in the servicing of our customers.

\textbf{\textcolor{blue}{Concise: }}If we revise the program, we can serve our customers more efficiently. \cite{duke_university2021}

\vspace{12pt}

\textbf{\textcolor{red}{Wordy: }}It was decided by the Director to expand the program.

\textbf{\textcolor{blue}{Concise: }}The Director decided to expand the program. \cite{duke_university2021}

\vspace{12pt}

\textbf{\textcolor{red}{Wordy: }}It is vital that we delete the word "absolutely."

\textbf{\textcolor{blue}{Concise: }}We must delete the word "absolutely." \cite{duke_university2021}

\vspace{12pt}

\textbf{\textcolor{red}{Wordy: }}There are five car alarms that are blaring in the parking lot.

\textbf{\textcolor{blue}{Concise: }}Five car alarms are blaring in the parking lot. \cite{duke_university2021}

\vspace{12pt}

\textbf{\textcolor{red}{Wordy: }}There are many engineers who like to write.

\textbf{\textcolor{blue}{Concise: }}Many engineers like to write. \cite{iqbal2021}

\vspace{12pt}

\textbf{\textcolor{red}{Wordy: }}There are many ways in which we can arrange the pulleys.

\textbf{\textcolor{blue}{Concise: }}We can arrange the pulleys in many ways. \cite{iqbal2021}

\vspace{12pt}

\textbf{\textcolor{red}{Wordy: }}The meeting happened on Monday.

\textbf{\textcolor{blue}{Concise: }}The meeting happened Monday. \cite{iqbal2021}

\vspace{12pt}

\textbf{\textcolor{red}{Wordy: }}They agreed that it was true.

\textbf{\textcolor{blue}{Concise: }}They agreed it was true. \cite{iqbal2021}

\vspace{12pt}

\textbf{\textcolor{red}{Wordy: }}George's wife is a woman who is unhappy because of the fact that George ignores her.

\textbf{\textcolor{blue}{Concise: }}George's wife is unhappy because George ignores her. \cite{rambo2019}

\vspace{12pt}

\textbf{\textcolor{red}{Wordy: }}In this day and age, people are under the impression that it is important to express ideas in a concise manner.

\textbf{\textcolor{blue}{Concise: }}Today, people think it is important to express ideas concisely. \cite{rambo2019}

\vspace{12pt}

\textbf{\textcolor{red}{Wordy: }}Hemingway conveys the idea that the setting of the story reflects the troubled relationship between George and his wife.

\textbf{\textcolor{blue}{Concise: }}The setting of the story reflects the troubled relationship between George and his wife. \cite{rambo2019}

\vspace{12pt}

\textbf{\textcolor{red}{Wordy: }}If one looks at the painting carefully, one notices that there are dark clouds gathering over the mountains.

\textbf{\textcolor{blue}{Concise: }}Dark clouds gather over the mountains. \textbf{ OR } There are dark clouds gathering over the mountains. \cite{rambo2019}

\vspace{12pt}

\textbf{\textcolor{red}{Wordy: }}There are children that are playing near the base of the mountain.

\textbf{\textcolor{blue}{Concise: }}Children are playing near the base of the mountain. \cite{rambo2019}

\vspace{12pt}

\textbf{\textcolor{red}{Wordy: }}There is a snow-capped mountain that appears on the left, and there are storm clouds that are gathering in the background.

\textbf{\textcolor{blue}{Concise: }}A snow-capped mountain appears on the left, and storm clouds are gathering in the background. \cite{rambo2019}

\vspace{12pt}

\textbf{\textcolor{red}{Wordy: }}In the story "Cat in the Rain," the setting of the story reflects the troubled relationship between George and his wife in the story.

\textbf{\textcolor{blue}{Concise: }}In the story "Cat in the Rain," the setting reflects the troubled relationship between George and his wife. \cite{rambo2019}

\vspace{12pt}

\textbf{\textcolor{red}{Wordy: }}The gathering storm clouds appear threatening and ominous.

\textbf{\textcolor{blue}{Concise: }}The gathering storm clouds appear threatening. \textbf{ OR } The gathering storm clouds appear ominous. \cite{rambo2019}

\vspace{12pt}

\textbf{\textcolor{red}{Wordy: }}The gathering clouds are emphasized by the artist, but the approaching storm does not seem to be noticed by the children.

\textbf{\textcolor{blue}{Concise: }}The artist emphasizes the gathering clouds, but the children do not seem to notice the approaching storm. \cite{rambo2019}

\vspace{12pt}

\textbf{\textcolor{red}{Wordy: }}The elephant is killed by Orwell even though the villagers are no longer threatened by it.

\textbf{\textcolor{blue}{Concise: }}Orwell kills the elephant even though it no longer threatens the villagers. \cite{rambo2019}

\vspace{12pt}

\textbf{\textcolor{red}{Wordy: }}I think that George's wife is unhappy because George ignores her.

\textbf{\textcolor{blue}{Concise: }}George's wife is unhappy because George ignores her. \cite{rambo2019}

\vspace{12pt}

\textbf{\textcolor{red}{Wordy: }}I believe that the storm clouds suggest the danger that, in my opinion, children sometimes ignore.

\textbf{\textcolor{blue}{Concise: }}The storm clouds suggest the danger that children sometimes ignore. \cite{rambo2019}

\vspace{12pt}

\textbf{\textcolor{red}{Wordy: }}George's wife appears to be unhappy.

\textbf{\textcolor{blue}{Concise: }}George's wife appears unhappy. \cite{rambo2019}

\vspace{12pt}

\textbf{\textcolor{red}{Wordy: }}The storm clouds appear to be dark and ominous, but the children seem to be unaware of the approaching storm.

\textbf{\textcolor{blue}{Concise: }}The storm clouds appear dark and ominous, but the children seem unaware of the approaching storm. \cite{rambo2019}

\vspace{12pt}

\textbf{\textcolor{red}{Wordy: }}George is reading his book while his wife is looking out the window.

\textbf{\textcolor{blue}{Concise: }}George reads his book while his wife looks out the window. \cite{rambo2019}

\vspace{12pt}

\textbf{\textcolor{red}{Wordy: }}Storm clouds are gathering over the mountains.

\textbf{\textcolor{blue}{Concise: }}Storm clouds gather over the mountains. \cite{rambo2019}

\vspace{12pt}

\textbf{\textcolor{red}{Wordy: }}Rather than taking the bull by the horns, she was quiet as a church mouse.

\textbf{\textcolor{blue}{Concise: }}She avoided confrontation by remaining silent. \cite{uwaterloo2018}

\vspace{12pt}

\textbf{\textcolor{red}{Wordy: }}The bridge is unstable due to the fact that it was constructed with inferior material.

\textbf{\textcolor{blue}{Concise: }}The bridge is unstable because it was constructed with inferior material. \cite{uwaterloo2018}

\vspace{12pt}

\textbf{\textcolor{red}{Wordy: }}All things considered, climate change should be given more attention, in my opinion.

\textbf{\textcolor{blue}{Concise: }}Climate change should be given more attention. \cite{uwaterloo2018}

\vspace{12pt}

\textbf{\textcolor{red}{Wordy: }}Last but not least, researchers found several connections between the subjects.

\textbf{\textcolor{blue}{Concise: }}Lastly, researchers found several connections between the subjects. \cite{uwaterloo2018}

\vspace{12pt}

\textbf{\textcolor{red}{Wordy: }}Historical context is an important factor to consider while writing literature reviews.

\textbf{\textcolor{blue}{Concise: }}Historical context must be considered while writing literature reviews. \cite{uwaterloo2018}

\vspace{12pt}

\textbf{\textcolor{red}{Wordy: }}He received a wound from the clock while he wound it.

\textbf{\textcolor{blue}{Concise: }}The clock injured him while he wound it. \cite{uwaterloo2018}

\vspace{12pt}

\textbf{\textcolor{red}{Wordy: }}He was right to assume his subjects are right-handed.

\textbf{\textcolor{blue}{Concise: }}He correctly assumed his subjects are right-handed. \cite{uwaterloo2018}

\vspace{12pt}

\textbf{\textcolor{red}{Wordy: }}Subjects with little technical training tend to perform poorly due to their lack of technical experience.

\textbf{\textcolor{blue}{Concise: }}Some subjects' lack of technical experience resulted in poor performance. \cite{uwaterloo2018}

\vspace{12pt}

\textbf{\textcolor{red}{Wordy: }}The reason she moved is because she was offered a better position.

\textbf{\textcolor{blue}{Concise: }}She moved because she was offered a better position. \cite{uwaterloo2018}

\vspace{12pt}

\textbf{\textcolor{red}{Wordy: }}As already stated above, beluga whales use sounds and echolocation to hunt in dark or turbid waters.

\textbf{\textcolor{blue}{Concise: }}As stated above, beluga whales use sounds and echolocation to hunt in dark or turbid waters. \cite{uwaterloo2018}

\vspace{12pt}

\textbf{\textcolor{red}{Wordy: }}It is challenging to read Shakespeare.

\textbf{\textcolor{blue}{Concise: }}Reading Shakespeare is challenging. \cite{uwaterloo2018}

\vspace{12pt}

\textbf{\textcolor{red}{Wordy: }}It is significant that a study of ethics complaints against social workers found that half of these involved violation of professional boundaries.

\textbf{\textcolor{blue}{Concise: }}Significantly, a study of ethics complaints against social workers found that half of these involved violations of professional boundaries. \cite{uwaterloo2018}

\vspace{12pt}

\textbf{\textcolor{red}{Wordy: }}The researchers conducted an investigation of the effects of caffeine on students writing timed examinations.

\textbf{\textcolor{blue}{Concise: }}The researchers investigated the effects of caffeine on students writing timed examinations. \cite{uwaterloo2018}

\vspace{12pt}

\textbf{\textcolor{red}{Wordy: }}The teacher could understand why her students failed the test.

\textbf{\textcolor{blue}{Concise: }}The teacher understood why her students failed the test. \cite{uwaterloo2018}

\vspace{12pt}

\textbf{\textcolor{red}{Wordy: }}The reaction was catalyzed by the introduction of light.

\textbf{\textcolor{blue}{Concise: }}The introduction of light catalyzed the reaction. \cite{uwaterloo2018}

\vspace{12pt}

\textbf{\textcolor{red}{Wordy: }}The author's expostulation impugns litterateurs of yore.

\textbf{\textcolor{blue}{Concise: }}The author's argument disproves earlier scholars. \cite{uwaterloo2018}

\vspace{12pt}

\textbf{\textcolor{red}{Wordy: }}Measurements were taken of variations in pressure as function of temperature.

\textbf{\textcolor{blue}{Concise: }}Pressure was measured as it varied with temperature. \cite{celia2021}

\vspace{12pt}

\textbf{\textcolor{red}{Wordy: }}There is a prize in every box of Quacko cereal.

\textbf{\textcolor{blue}{Concise: }}A prize is in every box of Quacko cereal. \cite{ualberta2021}

\vspace{12pt}

\textbf{\textcolor{red}{Wordy: }}There are two security guards standing at the gate.

\textbf{\textcolor{blue}{Concise: }}Two security guards stand at the gate. \cite{ualberta2021}

\vspace{12pt}

\section{More sentences from other sources}
\textbf{\textcolor{red}{Wordy: }}Due to the fact that the measure was unavailable, I selected another.

\textbf{\textcolor{blue}{Concise: }}Because the measure was unavailable, I selected another. \cite{lee2015}

\vspace{12pt}

\textbf{\textcolor{red}{Wordy: }}In spite of the fact that half the participants dropped out the study, we still conducted Phase 2.

\textbf{\textcolor{blue}{Concise: }}Although half the participants dropped out of the study, we still conducted Phase 2. \cite{lee2015}

\vspace{12pt}

\textbf{\textcolor{red}{Wordy: }}We administered surveys for the purpose of assessing motivation.

\textbf{\textcolor{blue}{Concise: }}We administered surveys to assess motivation. \cite{lee2015}

\vspace{12pt}

\textbf{\textcolor{red}{Wordy: }}It is important to note that our study has implications for counseling practice.

\textbf{\textcolor{blue}{Concise: }}Our study has important implications for counseling practice. \cite{lee2015}

\vspace{12pt}

\textbf{\textcolor{red}{Wordy: }}The boring textbook was being read by the students.

\textbf{\textcolor{blue}{Concise: }}The students read the boring textbook . \cite{sheehan2018}

\vspace{12pt}

\textbf{\textcolor{red}{Wordy: }}It is better to become a nurse instead of a teacher.

\textbf{\textcolor{blue}{Concise: }}Nursing pays better than teaching does. \cite{sheehan2018}

\vspace{12pt}

\textbf{\textcolor{red}{Wordy: }}The paper was written by Roseanne.

\textbf{\textcolor{blue}{Concise: }}Roseanne wrote the paper. \cite{sheehan2018}

\vspace{12pt}

\textbf{\textcolor{red}{Wordy: }}The Old Man and the Sea was written by Hemingway.

\textbf{\textcolor{blue}{Concise: }}Hemingway wrote The Old Man and the Sea. \cite{sheehan2018}

\vspace{12pt}

\textbf{\textcolor{red}{Wordy: }}It is believed by some critics that Psycho is Hitchcock's greatest film.

\textbf{\textcolor{blue}{Concise: }}Some critics believe that Psycho is Hitchcock's greatest film. \cite{sheehan2018}

\vspace{12pt}

\textbf{\textcolor{red}{Wordy: }}Rob decided to retake the class at a later date in time.

\textbf{\textcolor{blue}{Concise: }}Rob decided to retake the class later. \cite{sheehan2018}

\vspace{12pt}

\textbf{\textcolor{red}{Wordy: }}Many local farmers plan to attend next Friday's meeting.

\textbf{\textcolor{blue}{Concise: }}Many local farmers plan to attend next Friday's meeting. \cite{sheehan2018}

\vspace{12pt}

\textbf{\textcolor{red}{Wordy: }}Although Bradley Hall is regularly populated by students, close study of the building as a structure is seldom undertaken by them.

\textbf{\textcolor{blue}{Concise: }}Bradley Hall is usually filled with students who do not study the building as a structure. \cite{sheehan2018}

\vspace{12pt}

\textbf{\textcolor{red}{Wordy: }}He dropped out of school on account of the fact that it was necessary for him to help support his family.

\textbf{\textcolor{blue}{Concise: }}He dropped out of school to support his family \cite{sheehan2018}

\vspace{12pt}

\textbf{\textcolor{red}{Wordy: }}It is expected that the new schedule will be announced by the bus company within the next few days.

\textbf{\textcolor{blue}{Concise: }}The bus company will probably announce its schedule during the next few days. \cite{sheehan2018}

\vspace{12pt}

\textbf{\textcolor{red}{Wordy: }}There are many ways in which a student who is interested in meeting foreign students may come to know one.

\textbf{\textcolor{blue}{Concise: }}Any student who wants to meet foreign students can do so in many ways. \cite{sheehan2018}

\vspace{12pt}

\textbf{\textcolor{red}{Wordy: }}It is very unusual to find someone who has never told a deliberate lie on purpose.

\textbf{\textcolor{blue}{Concise: }}Rarely will you find someone who has never told a deliberate lie. \cite{sheehan2018}

\vspace{12pt}

\textbf{\textcolor{red}{Wordy: }}The subjects that are considered most important by students are those that have been shown to be useful to them after graduation.

\textbf{\textcolor{blue}{Concise: }}. Students think that the most important subjects are those that will be useful after graduation. \cite{sheehan2018}

\vspace{12pt}

\textbf{\textcolor{red}{Wordy: }}In our company there are wide-open opportunities for professional growth with a company that enjoys an enviable record for stability in the dynamic atmosphere of aerospace technology.

\textbf{\textcolor{blue}{Concise: }}Our company provides opportunities for professional growth and stability in the dynamic field of aerospace technology. \cite{sheehan2018}

\vspace{12pt}

\textbf{\textcolor{red}{Wordy: }}Some people believe in capital punishment, while other people are against it; there are many opinions on this subject.

\textbf{\textcolor{blue}{Concise: }}There are people who are for and people who are against capital punishment. \cite{sheehan2018}

\vspace{12pt}

\textbf{\textcolor{red}{Wordy: }}Even after the teacher entered the classroom, the boys went on playing.

\textbf{\textcolor{blue}{Concise: }}Even after the teacher entered the classroom, the boys continued playing. \cite{grammar2015}

\vspace{12pt}

\textbf{\textcolor{red}{Wordy: }}He is in debt because of his habit of spending money wastefully.

\textbf{\textcolor{blue}{Concise: }}He is in debt because of his extravagance. \cite{grammar2015}

\vspace{12pt}

\textbf{\textcolor{red}{Wordy: }}The celebrations went on for a whole month.

\textbf{\textcolor{blue}{Concise: }}The celebrations lasted for a whole month. // The celebrations continued for a whole month \cite{grammar2015}

\vspace{12pt}

\textbf{\textcolor{red}{Wordy: }}His uncle is a famous designer of buildings.

\textbf{\textcolor{blue}{Concise: }}His uncle is a famous architect. \cite{grammar2015}

\vspace{12pt}

\textbf{\textcolor{red}{Wordy: }}The drunkard abused people without discrimination.

\textbf{\textcolor{blue}{Concise: }}The drunkard abused people indiscriminately \cite{grammar2015}

\vspace{12pt}

\textbf{\textcolor{red}{Wordy: }}He spent most of his life abroad as one sent out of his country.

\textbf{\textcolor{blue}{Concise: }}He spent most of his life abroad as an exile. \cite{grammar2015}

\vspace{12pt}

\textbf{\textcolor{red}{Wordy: }}The country is passing through a critical phase

\textbf{\textcolor{blue}{Concise: }}The country is passing through a crisis. \cite{grammar2015}

\vspace{12pt}

\textbf{\textcolor{red}{Wordy: }}Bacon's essays are full of terse, witty, pointed statements

\textbf{\textcolor{blue}{Concise: }}Bacon's essays are full of epigrams. \cite{grammar2015}

\vspace{12pt}

\textbf{\textcolor{red}{Wordy: }}If you lose your good name, it is not easy to get it back.

\textbf{\textcolor{blue}{Concise: }}If you lose your reputation, it is not easy to get it back. \cite{grammar2015}

\vspace{12pt}

\textbf{\textcolor{red}{Wordy: }}There was no famine in our country during the last twenty years.

\textbf{\textcolor{blue}{Concise: }}There was no famine in our country during the last two decades \cite{grammar2015}

\vspace{12pt}

\textbf{\textcolor{red}{Wordy: }}The Mona Lisa is the best among the works of Leonardo da Vinci.

\textbf{\textcolor{blue}{Concise: }}The Mona Lisa is the masterpiece of Leonardo da Vinci. \cite{grammar2015}

\vspace{12pt}

\textbf{\textcolor{red}{Wordy: }}My friend had the special right or advantage of visiting England as the official guest of the Queen.

\textbf{\textcolor{blue}{Concise: }}My friend had the privilege of visiting England as the official guest of the Queen. \cite{grammar2015}

\vspace{12pt}

\textbf{\textcolor{red}{Wordy: }}In the cellar, there are four wooden-type crates with nothing in them that might perhaps be used by us for storing paint cans inside of.

\textbf{\textcolor{blue}{Concise: }}We could store the paint cans in the four wooden crates in the cellar. \cite{nordquist2020}

\vspace{12pt}

\textbf{\textcolor{red}{Wordy: }}This morning at 6:30 a.m., I woke up out of sleep to hear my alarm go off, but the alarm was turned off by me, and I returned back to a sleeping state.

\textbf{\textcolor{blue}{Concise: }}I awoke this morning at 6:30 but then turned off the alarm and went back to sleep. \cite{nordquist2020}

\vspace{12pt}

\textbf{\textcolor{red}{Wordy: }}The reason that Merdine was not able to be in attendance at the hockey game was because she had jury duty.

\textbf{\textcolor{blue}{Concise: }}Because she had jury duty, Merdine was not at the hockey game. \cite{nordquist2020}

\vspace{12pt}

\textbf{\textcolor{red}{Wordy: }}Omar and I, we returned back to the hometown where we both grew up to attend a reunion of the people that we went to high school with ten years ago in the past.

\textbf{\textcolor{blue}{Concise: }}Omar and I returned to our hometown to attend our ten-year high school \cite{nordquist2020}

\vspace{12pt}

\textbf{\textcolor{red}{Wordy: }}Melba has designed a very unique kind of shirt that is made out of a polyester type of material that never creases into wrinkles when it rains and the shirt gets wet.

\textbf{\textcolor{blue}{Concise: }}Melba has designed a polyester shirt that never creases when wet. \cite{nordquist2020}

\vspace{12pt}

\textbf{\textcolor{red}{Wordy: }}She used her money to purchase a large-type desk made of mahogany wood that is dark brown in color and handsome to look at.

\textbf{\textcolor{blue}{Concise: }}She purchased a large, handsome-looking mahogany desk. \cite{nordquist2020}

\vspace{12pt}

\textbf{\textcolor{red}{Wordy: }}In view of the fact that it was raining down, orders were given that the game be canceled.

\textbf{\textcolor{blue}{Concise: }}The game was canceled because of rain. \cite{nordquist2020}

\vspace{12pt}

\textbf{\textcolor{red}{Wordy: }}At that point in time when Marie was a teenager the basic fundamentals of how to dance were first learned by her.

\textbf{\textcolor{blue}{Concise: }}Marie learned how to dance when she was a teenager. \cite{nordquist2020}

\vspace{12pt}

\textbf{\textcolor{red}{Wordy: }}Some sort of identification that would show how old we were was requested of us by the man that collects tickets from people at the movie theater.

\textbf{\textcolor{blue}{Concise: }}The ticket collector at the movie theater asked us for identification. \cite{nordquist2020}

\vspace{12pt}

\textbf{\textcolor{red}{Wordy: }}There is a possibility that one of the causes of so many teenagers running away from home is the fact that many of them have indifferent parents who don't really care about them.

\textbf{\textcolor{blue}{Concise: }}Perhaps one reason that so many teenagers run away from home is that their parents don't care about them. \cite{nordquist2020}

\vspace{12pt}

\textbf{\textcolor{red}{Wordy: }}The Senator explained the ways in which his electoral victory were unique.

\textbf{\textcolor{blue}{Concise: }}The Senator explained how his electoral victory was unique. \cite{firth2015}

\vspace{12pt}

\textbf{\textcolor{red}{Wordy: }}As I have previously argued, it was not until after the last batch of votes was counted, that the Senator was able to declare victory.

\textbf{\textcolor{blue}{Concise: }}The Senator declared victory after the last batch of votes was counted. \cite{firth2015}

\vspace{12pt}

\textbf{\textcolor{red}{Wordy: }}A majority of respondents were single parents.

\textbf{\textcolor{blue}{Concise: }}Most respondents were single parents. \cite{vinz2020}

\vspace{12pt}

\textbf{\textcolor{red}{Wordy: }}A sufficient number of cases were selected.

\textbf{\textcolor{blue}{Concise: }}Enough cases were selected \cite{vinz2020}

\vspace{12pt}

\textbf{\textcolor{red}{Wordy: }}The interview was cancelled as a result of illness.

\textbf{\textcolor{blue}{Concise: }}The interview was cancelled due to illness. \cite{vinz2020}

\vspace{12pt}

\textbf{\textcolor{red}{Wordy: }}The laptop was kept in a locked office at all times.

\textbf{\textcolor{blue}{Concise: }}The laptop was always kept in a locked office.* \cite{vinz2020}

\vspace{12pt}

\textbf{\textcolor{red}{Wordy: }}The IT sector is expanding at the present time.

\textbf{\textcolor{blue}{Concise: }}The IT sector is currently expanding. \cite{vinz2020}

\vspace{12pt}

\textbf{\textcolor{red}{Wordy: }}The surveys were distributed by means of email.

\textbf{\textcolor{blue}{Concise: }}The surveys were distributed by email. \cite{vinz2020}

\vspace{12pt}

\textbf{\textcolor{red}{Wordy: }}Articles often draw attention to the most problematic cases.

\textbf{\textcolor{blue}{Concise: }}Articles often point to the most problematic cases. \cite{vinz2020}

\vspace{12pt}

\textbf{\textcolor{red}{Wordy: }}This definition cannot be used due to the fact that it is too limiting.

\textbf{\textcolor{blue}{Concise: }}This definition cannot be used because it is too limiting. \cite{vinz2020}

\vspace{12pt}

\textbf{\textcolor{red}{Wordy: }}A spreadsheet was used for the purpose of recording the data.

\textbf{\textcolor{blue}{Concise: }}A spreadsheet was used for recording the data. \cite{vinz2020}

\vspace{12pt}

\textbf{\textcolor{red}{Wordy: }}Consultants were excluded for the reason that they are not regular staff.

\textbf{\textcolor{blue}{Concise: }}Consultants were excluded because they are not regular staff. \cite{vinz2020}

\vspace{12pt}

\textbf{\textcolor{red}{Wordy: }}Economists have a tendency to favor policy reform.

\textbf{\textcolor{blue}{Concise: }}Economists tend to favor policy reform. \cite{vinz2020}

\vspace{12pt}

\textbf{\textcolor{red}{Wordy: }}Age appears to have an impact on confidence.

\textbf{\textcolor{blue}{Concise: }}Age appears to affect confidence. \cite{vinz2020}

\vspace{12pt}

\textbf{\textcolor{red}{Wordy: }}The scale has the ability to measure to the microgram.

\textbf{\textcolor{blue}{Concise: }}The scale can measure to the microgram. \cite{vinz2020}

\vspace{12pt}

\textbf{\textcolor{red}{Wordy: }}The data showed that CEOs earn more in comparison to CFOs.

\textbf{\textcolor{blue}{Concise: }}The results showed that CEOs earn more than CFOs. \cite{vinz2020}

\vspace{12pt}

\textbf{\textcolor{red}{Wordy: }}Saad's theory is chosen in light of the fact that it is most relevant.

\textbf{\textcolor{blue}{Concise: }}Saad's theory is chosen because it is most relevant. \cite{vinz2020}

\vspace{12pt}

\textbf{\textcolor{red}{Wordy: }}More research is needed in order to fill this gap.

\textbf{\textcolor{blue}{Concise: }}More research is needed to fill this gap. \cite{vinz2020}

\vspace{12pt}

\textbf{\textcolor{red}{Wordy: }}The factory produces in the neighborhood of 5,000 cars a week.

\textbf{\textcolor{blue}{Concise: }}The factory produces about 5,000 cars a week. \cite{vinz2020}

\vspace{12pt}

\textbf{\textcolor{red}{Wordy: }}Sales peaked in the year 2001.

\textbf{\textcolor{blue}{Concise: }}Sales peaked in 2001. \cite{vinz2020}

\vspace{12pt}

\textbf{\textcolor{red}{Wordy: }}Wang (2009) is of the opinion that the model must be expanded.

\textbf{\textcolor{blue}{Concise: }}Wang (2009) believes that the model must be expanded. \cite{vinz2020}

\vspace{12pt}

\textbf{\textcolor{red}{Wordy: }}We made a calculation of the average IQ.

\textbf{\textcolor{blue}{Concise: }}We calculated the average IQ. \cite{vinz2020}

\vspace{12pt}

\textbf{\textcolor{red}{Wordy: }}Marketers must make decisions about their target audience.

\textbf{\textcolor{blue}{Concise: }}Marketers must decide on their target audience. \cite{vinz2020}

\vspace{12pt}

\textbf{\textcolor{red}{Wordy: }}The patients were tested on two occasions.

\textbf{\textcolor{blue}{Concise: }}The patients were tested twice. \cite{vinz2020}

\vspace{12pt}

\textbf{\textcolor{red}{Wordy: }}The control group is relatively small in size.

\textbf{\textcolor{blue}{Concise: }}The control group is relatively small. \cite{vinz2020}

\vspace{12pt}

\textbf{\textcolor{red}{Wordy: }}The people who are located in rural areas had fewer symptoms.

\textbf{\textcolor{blue}{Concise: }}The people in rural areas had fewer symptoms. \cite{vinz2020}

\vspace{12pt}

\textbf{\textcolor{red}{Wordy: }}The reason why the population decreased is unknown.

\textbf{\textcolor{blue}{Concise: }}The reason the population decreased is unknown. \cite{vinz2020}

\vspace{12pt}

\textbf{\textcolor{red}{Wordy: }}Smartphones will be used until such time as a new technology is developed.

\textbf{\textcolor{blue}{Concise: }}Smartphones will be used until a new technology is developed. \cite{vinz2020}

\vspace{12pt}

\textbf{\textcolor{red}{Wordy: }}The goal was to identify whether or not gender made a difference.

\textbf{\textcolor{blue}{Concise: }}The goal was to identify whether gender made a difference. \cite{vinz2020}

\vspace{12pt}

\textbf{\textcolor{red}{Wordy: }}The creditor must first establish that the debtor is undoubtedly bankrupt.

\textbf{\textcolor{blue}{Concise: }}The creditor must establish that the debtor is bankrupt. \cite{syllables2021}

\vspace{12pt}

\textbf{\textcolor{red}{Wordy: }}Professor Smith was picked by each and every person on the committee.

\textbf{\textcolor{blue}{Concise: }}Professor Smith was picked by each person on the committee. \cite{syllables2021}

\vspace{12pt}

\textbf{\textcolor{red}{Wordy: }}As a matter of fact, Pleasantville has a strong position in the forestry industry.

\textbf{\textcolor{blue}{Concise: }}Pleasantville has a strong position in the forestry industry. \cite{syllables2021}

\vspace{12pt}

\textbf{\textcolor{red}{Wordy: }}The student needs to obtain high marks in science in order to study medicine.

\textbf{\textcolor{blue}{Concise: }}The student needs high marks in science to study medicine. \cite{syllables2021}

\vspace{12pt}

\textbf{\textcolor{red}{Wordy: }}The manager will, insofar as is possible, make sure that the information is true and accurate.

\textbf{\textcolor{blue}{Concise: }}The manager will make sure that the information is accurate. \cite{syllables2021}

\vspace{12pt}

\textbf{\textcolor{red}{Wordy: }}We will send out brochures to the general public.

\textbf{\textcolor{blue}{Concise: }}We will send brochures to the public. \cite{syllables2021}

\vspace{12pt}

\textbf{\textcolor{red}{Wordy: }}You must be willing to challenge yourself in order to get the most out of your education at university.

\textbf{\textcolor{blue}{Concise: }}You must be willing to challenge yourself to get the most out of university. \cite{syllables2021}

\vspace{12pt}

\textbf{\textcolor{red}{Wordy: }}We are working hard so the poor are given the basic essentials to lift themselves out of poverty.

\textbf{\textcolor{blue}{Concise: }}We are working hard to give the poor the basics to lift themselves out of poverty. \cite{syllables2021}

\vspace{12pt}

\textbf{\textcolor{red}{Wordy: }}Students should make contact with us for help with their future plans prior to the start of semester.

\textbf{\textcolor{blue}{Concise: }}Before semester starts, students should contact us for help with their plans. \cite{syllables2021}

\vspace{12pt}

\textbf{\textcolor{red}{Wordy: }}It is apparent that the tragedy could have been avoided if the company had talked to workers regarding hazards which existed in the workplace.

\textbf{\textcolor{blue}{Concise: }}It appears the tragedy could have been avoided if the company had talked to workers about workplace dangers. \cite{syllables2021}

\vspace{12pt}

\textbf{\textcolor{red}{Wordy: }}The total population of the island is 12,046 with the future possibility of 5

\textbf{\textcolor{blue}{Concise: }}The population of the island is 12,046 with the possibility of 5

\vspace{12pt}

\textbf{\textcolor{red}{Wordy: }}A commerce graduate can leverage knowledge they have acquired in their coursework and apply it to the real world.

\textbf{\textcolor{blue}{Concise: }}A commerce graduate can apply knowledge from their course to their work. \cite{syllables2021}

\vspace{12pt}

\textbf{\textcolor{red}{Wordy: }}The clerk's job is to check all incoming mail and to record it.

\textbf{\textcolor{blue}{Concise: }}The clerk's job is to check and record all incoming mail. \cite{edgecase2021}

\vspace{12pt}

\textbf{\textcolor{red}{Wordy: }}The method will be tested by Paul.

\textbf{\textcolor{blue}{Concise: }}Paul will test the method. \cite{edgecase2021}

\vspace{12pt}

\textbf{\textcolor{red}{Wordy: }}There are five women interested in the course.

\textbf{\textcolor{blue}{Concise: }}Five women want to take the course. \cite{edgecase2021}

\vspace{12pt}

\textbf{\textcolor{red}{Wordy: }}The teacher talked about many of the different benefits of the after-school homework club when she addressed the class.

\textbf{\textcolor{blue}{Concise: }}The teacher hyped the homework club to the class. \cite{aresearchguide2018}

\vspace{12pt}

\textbf{\textcolor{red}{Wordy: }}Tara believed but could not prove that Cara stole her tuna sandwich.

\textbf{\textcolor{blue}{Concise: }}Tara assumed that Cara stole her lunch. \cite{aresearchguide2018}

\vspace{12pt}

\textbf{\textcolor{red}{Wordy: }}Our website has available a list of social services in the city that can help residents find a hot meal.

\textbf{\textcolor{blue}{Concise: }}A list of hot meal programs are available on our website. \cite{aresearchguide2018}

\vspace{12pt}

\textbf{\textcolor{red}{Wordy: }}The coach demonstrated some of the various techniques that players could use to improve their stick handling and defence skills.

\textbf{\textcolor{blue}{Concise: }}The coach showed the players how to improve their defence. \cite{aresearchguide2018}

\vspace{12pt}

\textbf{\textcolor{red}{Wordy: }}Kensington College, which was first opened in 1902, is the leading Liberal Arts college in London.

\textbf{\textcolor{blue}{Concise: }}Founded in 1902, Kensington College is the leading Liberal Arts college in London. \cite{aresearchguide2018}

\vspace{12pt}

\textbf{\textcolor{red}{Wordy: }}The grant proposals were reviewed by the students.

\textbf{\textcolor{blue}{Concise: }}The students reviewed the grant proposals. \cite{nordquist2019}

\vspace{12pt}

\textbf{\textcolor{red}{Wordy: }}At this moment in time, students who are matriculating through high school should be empowered to participate in the voting process.

\textbf{\textcolor{blue}{Concise: }}High school students should have the right to vote. \cite{nordquist2019}

\vspace{12pt}

\textbf{\textcolor{red}{Wordy: }}All things being equal, what I am trying to say is that in my opinion all students should, in the final analysis, have the right to vote for all intents and purposes.

\textbf{\textcolor{blue}{Concise: }}Students should have the right to vote. \cite{nordquist2019}

\vspace{12pt}

\textbf{\textcolor{red}{Wordy: }}The presentation of the arguments by the students was convincing.

\textbf{\textcolor{blue}{Concise: }}The students argued convincingly. \cite{nordquist2019}

\vspace{12pt}

\textbf{\textcolor{red}{Wordy: }}After reading several things in the area of psychology-type subjects, I decided to put myself in a situation where I might change my major.

\textbf{\textcolor{blue}{Concise: }}After reading several psychology books, I decided to change my major. \cite{nordquist2019}

\vspace{12pt}

\textbf{\textcolor{red}{Wordy: }}Children's literature has the ability to entertain both the children for whom it has been written and the adults who may be reading it.

\textbf{\textcolor{blue}{Concise: }}Children's literature provides entertainment for children and adults alike. \cite{natureofwriting2021}

\vspace{12pt}

\textbf{\textcolor{red}{Wordy: }}The concept or idea that led to the talent show was the notion that even babies and infants are able to communicate and interact with others.

\textbf{\textcolor{blue}{Concise: }}The talent show was inspired by the fact that even babies can communicate. \cite{natureofwriting2021}

\vspace{12pt}

\textbf{\textcolor{red}{Wordy: }}Jamal has a chance to be able to win his badminton match.

\textbf{\textcolor{blue}{Concise: }}Jamal has a chance to win his badminton match. \cite{natureofwriting2021}

\vspace{12pt}

\textbf{\textcolor{red}{Wordy: }}Multiple instances of red herrings being seen in the lake water were reported several times all day yesterday.

\textbf{\textcolor{blue}{Concise: }}Red herrings were seen in the lake several times yesterday. \cite{inc2021}

\vspace{12pt}

\textbf{\textcolor{red}{Wordy: }}We made such a grandiloquent verbal exodus from the gathering that everyone in our immediate proximity was agog, their mouths fluctuating and trilling in surprise.

\textbf{\textcolor{blue}{Concise: }}Our parting comments left everyone at the party speechless and surprised. \cite{inc2021}

\vspace{12pt}

\textbf{\textcolor{red}{Wordy: }}Consumer demand is rising in the area of services

\textbf{\textcolor{blue}{Concise: }}Consumers are demanding more services \cite{sssi2021}

\vspace{12pt}

\textbf{\textcolor{red}{Wordy: }}Gardner's earlier work describes complete economic stability as something that is impossible to achieve.

\textbf{\textcolor{blue}{Concise: }}Gardner's earlier work describes complete economic stability as impossible. \cite{mitchellmentor2015}

\vspace{12pt}

\textbf{\textcolor{red}{Wordy: }}The admissions committee will read many personal statements, and they want to gain the information in as efficient a manner as possible.

\textbf{\textcolor{blue}{Concise: }}The admissions committee will read many personal statements, and they want to gain the information as efficiently as possible. \cite{mitchellmentor2015}

\vspace{12pt}

\textbf{\textcolor{red}{Wordy: }}These are situations in which there is not very much social interaction.

\textbf{\textcolor{blue}{Concise: }}These situations lack social interaction. \cite{mitchellmentor2015}

\vspace{12pt}

\textbf{\textcolor{red}{Wordy: }}I think the risks of hydraulic fracturing clearly outweigh the benefits.

\textbf{\textcolor{blue}{Concise: }}The risks of hydraulic fracturing clearly outweigh the benefits. \cite{mitchellmentor2015}

\vspace{12pt}

\textbf{\textcolor{red}{Wordy: }}Nowadays, the price of oil is higher than it has ever been before.

\textbf{\textcolor{blue}{Concise: }}The price of oil is higher than it has ever been. \cite{mitchellmentor2015}

\vspace{12pt}

\textbf{\textcolor{red}{Wordy: }}Professor Li does research on productivity in the workplace.

\textbf{\textcolor{blue}{Concise: }}Professor Li researches workplace productivity. \cite{mitchellmentor2015}

\vspace{12pt}

\textbf{\textcolor{red}{Wordy: }}She demonstrated a great deal of empathy for others.

\textbf{\textcolor{blue}{Concise: }}She demonstrated remarkable empathy. \cite{mitchellmentor2015}

\vspace{12pt}

\textbf{\textcolor{red}{Wordy: }}Hundreds of different people attended the conference.

\textbf{\textcolor{blue}{Concise: }}Hundreds of people attended the conference. \cite{mitchellmentor2015}

\vspace{12pt}

\textbf{\textcolor{red}{Wordy: }}Participants sat in a circular formation during the meeting.

\textbf{\textcolor{blue}{Concise: }}Participants sat in a circle during the meeting. \cite{mitchellmentor2015}

\vspace{12pt}

\textbf{\textcolor{red}{Wordy: }}Smith draws the conclusion that global warming is a threat to 125,000 species of insects.

\textbf{\textcolor{blue}{Concise: }}Smith concludes that global warming threatens 125,000 insect species. \cite{mitchellmentor2015}

\vspace{12pt}

\textbf{\textcolor{red}{Wordy: }}In my honest opinion, I deeply feel that a really firm comprehensive understanding of the basic fundamentals of English is an essential prerequisite to a person's success in college or university, the business world, and the community at large.

\textbf{\textcolor{blue}{Concise: }}An understanding of English fundamentals is a prerequisite to success in college or university, business, and the community. \cite{langara2021}

\vspace{12pt}

\textbf{\textcolor{red}{Wordy: }}In my own personal opinion, I think that baseball is a very boring sport.

\textbf{\textcolor{blue}{Concise: }}I think baseball is boring. \cite{langara2021}

\vspace{12pt}

\textbf{\textcolor{red}{Wordy: }}Mr. Smith is far from perfect and has many faults.

\textbf{\textcolor{blue}{Concise: }}Mr. Smith has many faults. \cite{langara2021}

\vspace{12pt}

\textbf{\textcolor{red}{Wordy: }}The dog was more exuberant in emitting threatening sounds than in attempting to engage in physical assault.

\textbf{\textcolor{blue}{Concise: }}The dog's bark was worse than its bite. \cite{langara2021}

\vspace{12pt}

\textbf{\textcolor{red}{Wordy: }}If people look at the flowers, they'll see that the color of the flowers is quite bright.

\textbf{\textcolor{blue}{Concise: }}The flowers are brightly coloured. \cite{langara2021}

\vspace{12pt}

\textbf{\textcolor{red}{Wordy: }}The president was supportive of the council's plans.

\textbf{\textcolor{blue}{Concise: }}The president supported the council's plans. \cite{langara2021}

\vspace{12pt}

\textbf{\textcolor{red}{Wordy: }}She really is trying very hard to quit smoking completely.

\textbf{\textcolor{blue}{Concise: }}She is trying hard to quit smoking. \cite{langara2021}

\vspace{12pt}

\textbf{\textcolor{red}{Wordy: }}I am writing you this email to let you know that I will not be coming into work on Friday, September 23.

\textbf{\textcolor{blue}{Concise: }}I will not be coming into work on Friday, September 23. \cite{langara2021}

\vspace{12pt}

\textbf{\textcolor{red}{Wordy: }}It is because Bob likes bananas that he eats them every day.

\textbf{\textcolor{blue}{Concise: }}Bob eats bananas every day because he likes them \cite{langara2021}

\vspace{12pt}



\printbibliography[heading=bibintoc]


\end{document}